\documentclass[twocolumn]{svjour3}

\usepackage[latin1]{inputenc}
\usepackage{times}

\usepackage{psfrag, amsmath, amssymb, amscd, amsfonts, latexsym, epsf, graphicx}

\def\bbbr{{\mathbb R}}

\journalname{In Journal of
  Mathematical Imaging and Vision doi:10.1007/s10851-016-0691-3 
  Jan 2017}

\begin{document}

\title{\bf Temporal scale selection in time-causal scale space%
\thanks{The support from the Swedish Research Council 
              (Contract No.\ 2014-4083) and Stiftelsen Olle Engkvist
              Byggm{\"a}stare (Contract No.\ 2015/465) is gratefully acknowledged.}}

\titlerunning{Temporal scale selection in time-causal scale space}

\author{Tony Lindeberg}

\institute{Tony Lindeberg,
                Computational Brain Science Lab,
                Department of Computational Science and Technology,
                School of Computer Science and Communication,
                KTH Royal Institute of Technology,
                SE-100 44 Stockholm, Sweden. 
                \email{tony@kth.se}}

\date{Received: date / Accepted: date}

\maketitle

\begin{abstract}
\noindent
When designing and developing scale selection mechanisms for
generating hypotheses about characteristic scales in signals, it is
essential that the selected scale levels reflect the extent of the
underlying structures in the signal.

This paper presents a theory and in-depth theoretical analysis about
the scale selection properties of methods for automatically selecting
local temporal scales in time-dependent signals based on local extrema
over temporal scales of scale-normalized temporal derivative
responses. Specifically, this paper develops a novel theoretical
framework for performing such temporal scale selection over a
time-causal and time-recursive temporal domain as is necessary 
when processing continuous video or audio streams in real time
or when modelling biological perception.

For a recently developed time-causal and time-recursive scale-space 
concept defined by convolution with a scale-invariant limit kernel, we
show that it is possible to transfer a large number of the desirable
scale selection properties that hold for the Gaussian scale-space
concept over a non-causal temporal domain to this temporal scale-space
concept over a truly time-causal domain. Specifically, we show that
for this temporal scale-space concept, it is possible to achieve true
temporal scale invariance although the temporal scale levels have to be
discrete, which is a novel theoretical construction.

The analysis starts from a detailed comparison of different temporal scale-space concepts
and their relative advantages and disadvantages,
leading the focus to a class of recently extended time-causal and time-recursive
temporal scale-space concepts based on first-order integrators or
equivalently truncated exponential kernels coupled in cascade.
Specifically, by the discrete nature of the temporal scale levels in
this class of time-causal scale-space concepts, we study two special cases of distributing the
intermediate temporal scale levels, by using either a uniform
distribution in terms of the variance of the composed temporal scale-space
kernel or a logarithmic distribution.

In the case of a uniform distribution of the temporal scale levels, we
show that scale selection based on local extrema of scale-normalized
derivatives over temporal scales makes it possible to estimate the
temporal duration of sparse local features defined in terms of temporal extrema
of first- or second-order temporal derivative responses. For dense
features modelled as a sine wave, the lack of temporal scale
invariance does, however, constitute a major limitation for handling
dense temporal structures of different temporal duration in a uniform
manner.

In the case of a logarithmic distribution of the temporal scale
levels, specifically taken to the limit of a time-causal
limit kernel with an infinitely dense distribution of the temporal
scale levels towards zero temporal scale, we show that it is possible to
achieve true temporal scale invariance to handle dense features
modelled as a sine wave in a uniform manner over different temporal
durations of the temporal structures as well to achieve more general temporal scale
invariance for any signal over any temporal scaling transformation with a 
scaling factor that is an integer power of the distribution parameter
of the time-causal limit kernel.

It is shown how these temporal scale
selection properties developed for a pure temporal domain carry over
to feature detectors defined over time-causal
spatio-temporal and spectro-temporal domains.


\end{abstract}

\section{Introduction}

When processing sensory data by automatic methods in areas of signal
processing such as computer vision or audio processing or in
computational modelling of biological perception, the notion of
receptive field constitutes an essential concept
(Hubel and Wiesel \cite{HubWie59-Phys,HubWie05-book};
 Aertsen and Johannesma \cite{AerJoh81-BICY};
 DeAngelis {\em et al.}
 \cite{DeAngOhzFre95-TINS,deAngAnz04-VisNeuroSci};
 Miller {\em et al} \cite{MilEscReaSch01-JNeuroPhys}).

For sensory data as obtained from vision or hearing, or their
counterparts in artificial perception, the measurement from a single
light sensor in a video camera or on the retina, or the instantaneous sound pressure
registered by a microphone is hardly meaningful at all, since
any such measurement
is strongly dependent on external factors such as the illumination of
a visual scene regarding vision or the distance between the sound source
and the microphone regarding hearing.
Instead, the essential information is carried by the relative
relations between local measurements at different points and temporal
moments regarding vision or local measurements over different
frequencies and temporal moments regarding hearing.
Following this paradigm, sensory measurements should be performed over local
neighbourhoods over space-time regarding vision and over local
neighbourhoods in the time-frequency domain regarding hearing, leading to the notions of
spatio-temporal and spectro-temporal receptive fields.

Specifically, spatio-temporal receptive fields constitute a
main class of primitives for expressing methods for video
analysis
(Zelnik-Manor and Irani \cite{ZelIra01-CVPR},
 Laptev and Lindeberg \cite{LapLin04-ECCVWS,LapCapSchLin07-CVIU};
 Jhuang {\em et al.\/}\ \cite{JhuSerWolPog07-ICCV};
 Kl{\"a}ser {\em et al.\/}\ \cite{KlaMarSch08-BMVC};
 Niebles {\em et al.\/}\ \cite{NieWanFei08-IJCV};
 Wang {\em et al.\/}\ \cite{WanUllKlaLapSch09-BMVC};
 Poppe {\em et al.\/}\ \cite{Pop09-IVC};
 Shao and Mattivi \cite{ShaMatt10-CIVR};
 Weinland {\em et al.\/}\ \cite{WeiRonBoy11-CVIU};
 Wang {\em et al.\/}\ \cite{WanQiaTan15-CVPR}),
whereas spectro-temporal receptive fields constitute a main class of
primitives for expressing methods for machine hearing
(Patterson {\em et al.\/}\ \cite{PatRobHolMcKeoZhaAll92-AudPhysPerc,PatAllGig95-JASA};
 Kleinschmidt \cite{Kle02-ActAcust};
 Ezzat {\em et al.\/}\ \cite{EzzBouPog07-InterSpeech};
 Meyer and Kollmeier \cite{MeyKol08-InterSpeech};
 Schlute {\em et al.\/}\ \cite{SchBezWagNey07-ICASSP};
 Heckmann {\em et al.\/}\ \cite{HecDomJouGoe11-SpeechComm};
 Wu {\em et al.\/}\ \cite{WuZhaShi11-ASLP};
 Alias {\em et al.\/}\ \cite{AliSocJoaSev16-ApplSci}).

A general problem when applying the notion of receptive fields in
practice, however, is that the types of responses that are obtained
in a specific situation can be strongly dependent on the scale levels at which they are
computed. A spatio-temporal receptive field is determined by at least a
spatial scale parameter and a temporal scale parameter, whereas a
spectro-temporal receptive field is determined by at least a spectral
and a temporal scale parameter. 
Beyond ensuring that local sensory measurements at different spatial, temporal and
spectral scales are treated in a consistent manner, which by itself
provides strong contraints on the shapes of the receptive fields
(Lindeberg \cite{Lin13-BICY,Lin16-JMIV}; Lindeberg and Friberg
\cite{LinFri15-PONE,LinFri15-SSVM}), it is necessary for computer vision or machine
hearing algorithms to decide what responses within the families of receptive fields
over different spatial, temporal and spectral scales they should base
their analysis on.

Over the spatial domain, theoretically well-founded methods have been
developed for choosing spatial scale levels among receptive
field responses over multiple spatial scales (Lindeberg
\cite{Lin97-IJCV,Lin98-IJCV,Lin99-CVHB,Lin12-JMIV,Lin14-EncCompVis}) leading to 
{\em e.g.\/} robust methods
for image-based matching and recognition
(Lowe \cite{Low04-IJCV}; 
 Mikolajczyk and Schmid \cite{MikSch04-IJCV};
 Tuytelaars and van Gool \cite{TuyGoo04-IJCV};
 Bay {\em et al.\/}\ \cite{BayEssTuyGoo08-CVIU};
 Tuytelaars and Mikolajczyk \cite{TuyMik08-Book};
  van de Sande {\em et al.\/}\ \cite{SanGevSno10-PAMI};
 Larsen {\em et al.\/}\ \cite{LarDarDahPed12-ECCV})
that are able to handle large variations of the size of the objects in
the image domain and
with numerous applications regarding object recognition, object categorization,
multi-view geometry, construction of 3-D models from visual input, human-computer
interaction, biometrics and robotics.

Much less research has, however, been performed regarding the topic of
choosing local appropriate scales in temporal data. While some methods
for temporal scale selection have been developed
(Lindeberg \cite{Lin97-AFPAC};
 Laptev and Lindeberg \cite{LapLin03-ICCV};
 Willems {\em et al.\/}\ \cite{WilTuyGoo08-ECCV}),
these methods suffer from either theoretical or practical 
limitations.

A main subject of this paper is present a theory for how to compare filter
responses in terms of temporal derivatives that have been computed at different temporal scales,
specifically with a detailed theoretical analysis of
the possibilities of having temporal scale estimates as obtained from
a temporal scale selection mechanism reflect the temporal duration of
the underlying temporal structures that gave rise to the feature
responses.
Another main subject of this paper is to present a theoretical framework
for temporal scale selection that leads to temporal scale invariance
and enables the computation of scale covariant temporal scale
estimates.
While these topics can for a non-causal temporal domain be addressed
by the non-causal Gaussian scale-space concept 
(Iijima \cite{Iij62}; Witkin \cite{Wit83}; Koenderink \cite{Koe84-BC};
 Koenderink and van Doorn \cite{KoeDoo92-PAMI};
 Lindeberg \cite{Lin93-Dis,Lin94-SI,Lin10-JMIV};
 Florack \cite{Flo97-book}; 
 ter Haar Romeny \cite{Haa04-book}), the development of
such a theory has been missing regarding a time-causal temporal domain.

\subsection{Temporal scale selection}

When processing time-dependent signals in video or audio or more
generally any temporal signal, special attention has to be put to the facts that:
\begin{itemize}
\item
  the physical phenomena that generate the temporal signals may
  occur at different speed --- faster or slower, and
\item
  the temporal signals may contain qualitatively different types of
temporal structures at different temporal scales.
\end{itemize}
In certain controlled situations where the physical system that
generates the temporal signals that is to be processed is sufficiently
well known and if the
variability of the temporal scales over time in the domain
is sufficiently constrained, suitable temporal scales for processing the 
signals may in some situations be chosen manually and then be verified experimentally.
If the sources that generate the temporal signals are sufficiently
complex and/or if the temporal structures in the signals vary
substantially in temporal duration by the underlying physical processes 
occurring significantly faster or slower,  it is on the other hand natural to (i)~include
a mechanism for processing the temporal data at multiple temporal
scales and 
(ii)~try to detect in a bottom-up manner at what
 temporal scales the interesting  temporal phenomena are likely to occur.

The subject of this article is to develop a theory for temporal scale
selection in a time-causal temporal scale space as an extension of a previously
developed theory for spatial scale selection in a spatial scale space
(Lindeberg
\cite{Lin97-IJCV,Lin98-IJCV,Lin99-CVHB,Lin12-JMIV,Lin14-EncCompVis}), 
to generate bottom-up hypotheses about characteristic temporal scales
in time-dependent signals, intended to serve as estimates of the
temporal duration of local temporal structures in time-dependent signals.
Special focus will be on developing mechanisms analogous to scale
selection in non-causal Gaussian scale-space, based on local extrema
over scales of scale-normalized derivatives, while expressed within the
framework of a time-causal and time-recursive temporal scale space
in which the future cannot be accessed and the signal processing 
operations are thereby only allowed to make use of information from the
present moment and a compact buffer of what has occurred in the past.

When designing and developing such scale selection mechanisms, it is
essential that the computed scale estimates reflect the temporal
duration of the corresponding temporal structures that gave rise to
the feature responses. 
To understand the pre-requisites for developing such temporal scale
selection methods, we will in this paper perform an in-depth
theoretical analysis of the scale selection properties that such temporal scale selection
mechanisms give rise to for different temporal scale-space concepts
and for different ways of defining scale-normalized temporal derivatives.

Specifically, after an examination of the theoretical properties of
different types of temporal scale-space concepts, we will focus on a
class of recently extended time-causal temporal
scale-space concepts obtained by convolution with truncated exponential kernels 
coupled in cascade 
(Lindeberg \cite{Lin90-PAMI,Lin15-SSVM,Lin16-JMIV}; 
 Lindeberg and Fagerstr{\"o}m \cite{LF96-ECCV}).
For two natural ways of distributing the discrete temporal scale
levels in such a representation, in terms of either a uniform
distribution over the scale parameter $\tau$ corresponding to the
variance of the composed scale-space kernel or a logarithmic
distribution, we will study the scale selection properties that result
from detecting local temporal scale levels from local extrema over
scale of scale-normalized temporal derivatives.
The motivation for studying a logarithmic distribution of the temporal
scale levels, is that it corresponds to a uniform distribution in units of effective
scale $\tau_{eff} = A + B \log \tau$ for
some constants $A$ and $B$, which has been shown to constitute the natural metric for measuring
the scale levels in a spatial scale space (Koenderink \cite{Koe84-BC};
Lindeberg \cite{Lin92-PAMI}).

As we shall see from the detailed theoretical analysis that will
follow, this will imply certain differences in scale selection properties 
of a temporally asymmetric time-causal scale space compared to scale 
selection in a spatially mirror symmetric Gaussian scale space.
These differences in theoretical properties are in turn essential to
take into explicit account when formulating algorithms for temporal scale
selection in {\em e.g.\/} video analysis or audio analysis applications.

For the temporal scale-space concept based on a uniform distribution
of the temporal scale levels in units of the variance of the composed
scale-space kernel, it will be shown that temporal scale selection
from local extrema over temporal scales will make it possible to
estimate the temporal duration of local temporal structures modelled
as local temporal peaks and local temporal ramps.
For a dense temporal structure modelled as a temporal sine wave, the
lack of true scale invariance for this concept will, however, imply
that the temporal scale estimates will not be directly proportional to the
wavelength of the temporal sine wave. Instead, the scale estimates are
affected by a bias, which is not a desirable property.

For the temporal scale-space concept based on a logarithmic
distribution of the temporal scale levels, and taken to the limit to
scale-invariant time-causal limit kernel (Lindeberg \cite{Lin16-JMIV})
corresponding to an infinite number of temporal scale levels that cluster infinitely
close near the temporal scale level zero, it will on the other hand be
shown that the temporal scale estimates of a dense temporal sine wave
will be truly proportional to the wavelength of the signal.
By a general proof, it will be shown this scale invariant property of temporal scale
estimates can also be extended to any sufficiently regular signal,
which constitutes a general foundation for expressing scale invariant
temporal scale selection mechanisms for time-dependent video and audio
and more generally also other classes of time-dependent measurement signals.

As complement to this proposed overall framework for temporal scale
selection, we will also present a set of general theoretical results
regarding time-causal scale-space representations:
(i)~showing that previous application of the assumption of a semi-group
property for time-causal scale-space concepts leads to undesirable
temporal dynamics, which however can be remedied by replacing the
assumption of a semi-group structure be a weaker assumption of a cascade
property in turn based on a transitivity property, (ii)~formulations of scale-normalized temporal
derivatives for Koenderink's time-causal scale-time model \cite{Koe88-BC}
and
(iii)~ways of translating the temporal scale estimates from local
extrema over temporal scales in the temporal scale-space
representation based on the scale-invariant time-causal limit kernel into
quantitative measures of the temporal duration of the corresponding
underlying temporal structures and in turn based on a scale-time
approximation of the limit kernel.

In these ways, this paper is intended to provide a theoretical
foundation for expressing theoretically well-founded temporal scale
selection methods for selecting local temporal scales over time-causal
temporal domains, such as video and audio
with specific focus on real-time image or sound streams.
Applications of this scale selection methodology for detecting both sparse
and dense spatio-temporal features in video are presented in a companion paper 
\cite{Lin16-spattempscsel}.

\subsection{Structure of this article}

As a conceptual background to the theoretical developments that will
be performed, we will start in Section~\ref{sec-related-work} with
an overview of different approaches to handling temporal data within 
the scale-space framework including a comparison of relative
advantages and disadvantages of different types of temporal scale-space concepts.

As a theoretical baseline for the later developments of methods for
temporal scale selection in a time-causal scale space, we shall then in
Section~\ref{sec-scsel-prop-gauss-temp-scsp} give an overall description
of basic temporal scale selection properties that will hold if the
non-causal Gaussian scale-space concept with its corresponding
selection methodology for a spatial image domain is applied to a
one-dimensional non-causal temporal domain,
{\em e.g.\/}\ for the purpose of handling the temporal domain when analysing pre-recorded video or audio
in an offline setting.

In
Sections~\ref{sec-scsel-prop-time-caus-trunc-exp-uni-distr}--\ref{sec-scsel-prop-time-caus-trunc-exp-log-distr}
we will then continue with a theoretical analysis of the consequences of
performing temporal scale selection in the time-causal scale space
obtained by convolution with truncated exponential kernels 
coupled in cascade 
(Lindeberg \cite{Lin90-PAMI,Lin15-SSVM,Lin16-JMIV}; 
 Lindeberg and Fagerstr{\"o}m \cite{LF96-ECCV}). 
By selecting local temporal scales from the scales at which
scale-normalized temporal derivatives assume local extrema over
temporal scales, we will analyze the resulting temporal scale selection
properties for two ways of defining scale-normalized temporal
derivatives, by either variance-based normalization as determined by a
scale normalization parameter $\gamma$ or $L_p$-normalization for
different values of the scale normalization power $p$.

With the temporal scale levels required to be discrete because of the
very nature of this temporal scale-space concept, we will specifically
study two ways of distributing the temporal scale levels over scale, using
either a uniform distribution relative to the temporal scale parameter
$\tau$ corresponding to the variance of the composed temporal
scale-space kernel in Section~\ref{sec-scsel-prop-time-caus-trunc-exp-uni-distr} or a
logarithmic distribution of the temporal scale levels in 
Section~\ref{sec-scsel-prop-time-caus-trunc-exp-log-distr}.

Because of the analytically simpler form for the time-causal
scale-space kernels corresponding to a {\em uniform distribution of the
temporal scale levels\/}, some theoretical scale-space properties will
turn out to be easier to study in closed form for this temporal
scale-space concept. We will specifically show that for a temporal peak
modelled as the impulse response to a set of truncated exponential
kernels coupled in cascade, the selected temporal scale level will
serve as a good approximation of the temporal duration of the peak or
be proportional to this measure depending on the value of the scale
normalization parameter $\gamma$ used for scale-normalized temporal
derivatives based on variance-based normalization or the scale
normalization power $p$ for
scale-norm\-alized temporal derivatives based on $L_p$-normalization.
For a temporal onset ramp, the selected temporal scale level will on
the other hand be either a good approximation of the time constant of
the onset ramp or proportional to this measure of the temporal duration
of the ramp. For a temporal sine wave, the selected temporal scale
level will, however, not be directly proportional to the wavelength of
the signal, but instead affected by a systematic bias. Furthermore,
the corresponding scale-normalized magnitude measures will not be
independent of the wavelength of the sine wave but instead show 
systematic wavelength dependent deviations. A main reason for this is that this temporal scale-space
concept does not guarantee temporal scale invariance if the temporal
scale levels are distributed uniformly in terms of the temporal scale
parameter $\tau$ corresponding to the temporal variance of the
temporal scale-space kernel.

With a {\em logarithmic distribution of the temporal scale levels\/}, 
we will on the other hand show that for the temporal scale-space
concept defined by convolution with {\em the time-causal limit kernel\/} (Lindeberg \cite{Lin16-JMIV}) 
corresponding to an infinitely dense distribution of the temporal
scale levels towards zero temporal scale, the temporal scale estimates will
be perfectly proportional to the wavelength of a sine wave for this
temporal scale-space concept. It will also be shown that this temporal
scale-space concept leads to perfect scale invariance in the sense
that
(i)~local extrema over temporal scales are preserved under temporal
scaling factors corresponding to integer powers of the distribution
parameter $c$ of the time-causal limit kernel underlying this temporal
scale-space concept and are transformed in a scale-covariant way 
for any temporal input signal and
(ii)~if the scale normalization parameter $\gamma = 1$ or equivalently
if the scale normalization power $p = 1$, the magnitude values at the local
extrema over scale will be equal under corresponding temporal scaling
transformations.
For this temporal scale-space concept we can therefore fulfil basic
requirements to {\em achieve temporal scale invariance also over a
time-causal and time-recursive temporal domain\/}.

To simplify the theoretical analysis we will in some cases temporarily
extend the definitions of temporal scale-space representations over
discrete temporal scale levels to a continuous scale variable, to make
it possible to compute local extrema over temporal scales from
differentiation with respect to the temporal scale parameter.
Section~\ref{sec-infl-disc-temp-scale-levels} discusses the influence
that this approximation has on the overall theoretical analysis.

Section~\ref{sec-temp-scsel-1D-temp-signal} then illustrates how the
proposed theory for temporal scale selection can be used for
computing local scale estimates from 1-D signals with substantial
variabilities in the characteristic temporal duration of the
underlying structures in the temporal signal.

In Section~\ref{sec-sc-sel-video-data}, we analyse how the derived
scale selection properties carry over to a set of spatio-temporal feature detectors defined over
both multiple spatial scales and multiple temporal scales in a
time-causal spatio-temporal scale-space representation for video analysis.
Section~\ref{sec-sc-sel-audio-data} then outlines how corresponding
selection of local temporal and logspectral scales can be expressed
for audio analysis operations over a time-causal spectro-temporal domain.
Finally, Section~\ref{sec-sum-disc} concludes with a summary and discussion.

To simplify the presentation, we have put some derivations and
theoretical analysis in the appendix.
Appendix~\ref{app-undesired-temp-dyn-temp-semi-group} presents a
general theoretical argument of why a requirement about a
semi-group property over temporal scales will lead to undesirable
temporal dynamics for a time-causal scale space and argue that the
essential structure of non-creation of new image structures from any
finer to any coarser temporal scale can instead nevertheless be achieved with the less restrictive
assumption about a cascade smoothing property over temporal scales,
which then allows for better temporal dynamics in terms of
{\em e.g.\/}\ shorter temporal delays.

In relation to Koenderink's scale-time model \cite{Koe88-BC},
Appendix~\ref{sec-sc-norm-koe-scale-time} shows how corresponding
notions of scale-normal\-ized temporal derivatives based on either
variance-based normalization or $L_p$-normalization can be defined also for
this time-causal temporal scale-space concept.

Appendix~\ref{sec-scaletime-approx-limit-kernel} shows how the
temporal duration of the time-causal limit kernel proposed in
(Lindeberg \cite{Lin16-JMIV}) can be estimated by a
scale-time approximation of the limit kernel via Koenderink's
scale-time model leading to estimates of how a selected temporal
scale level $\hat{\tau}$ from local extrema over temporal scale can be
translated into a estimates of the temporal duration of temporal
structures in the temporal scale-space representation obtained by
convolution with the time-causal limit kernel. Specifically, explicit
expressions are given for such temporal duration estimates based on first- and
second-order temporal derivatives.


\begin{figure*}[hbtp]
  \begin{center}
    \begin{tabular}{ccc}
       {\small $g(t;\; \tau)$} 
      & {\small $g_{t}(t;\; \tau)$} 
      & {\small $g_{tt}(t;\; \tau)$} \\
      \includegraphics[width=0.25\textwidth]{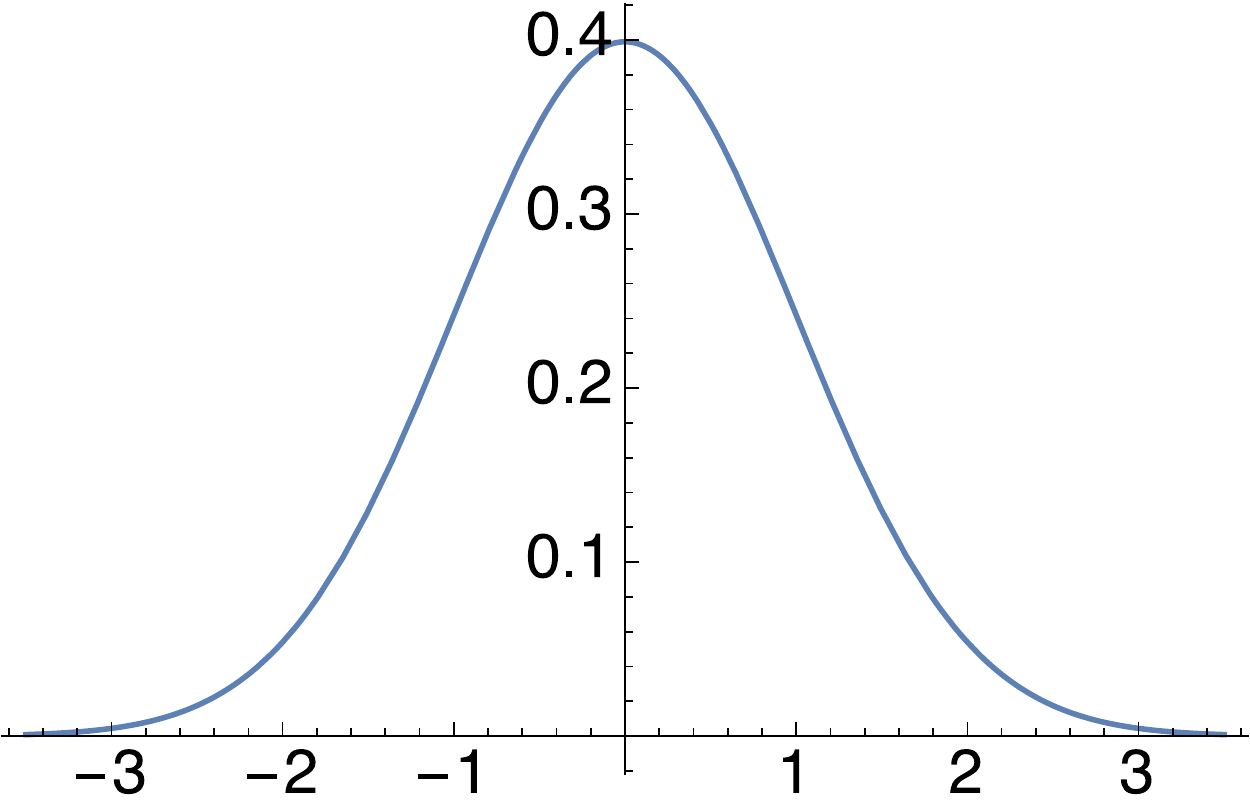} &
      \includegraphics[width=0.25\textwidth]{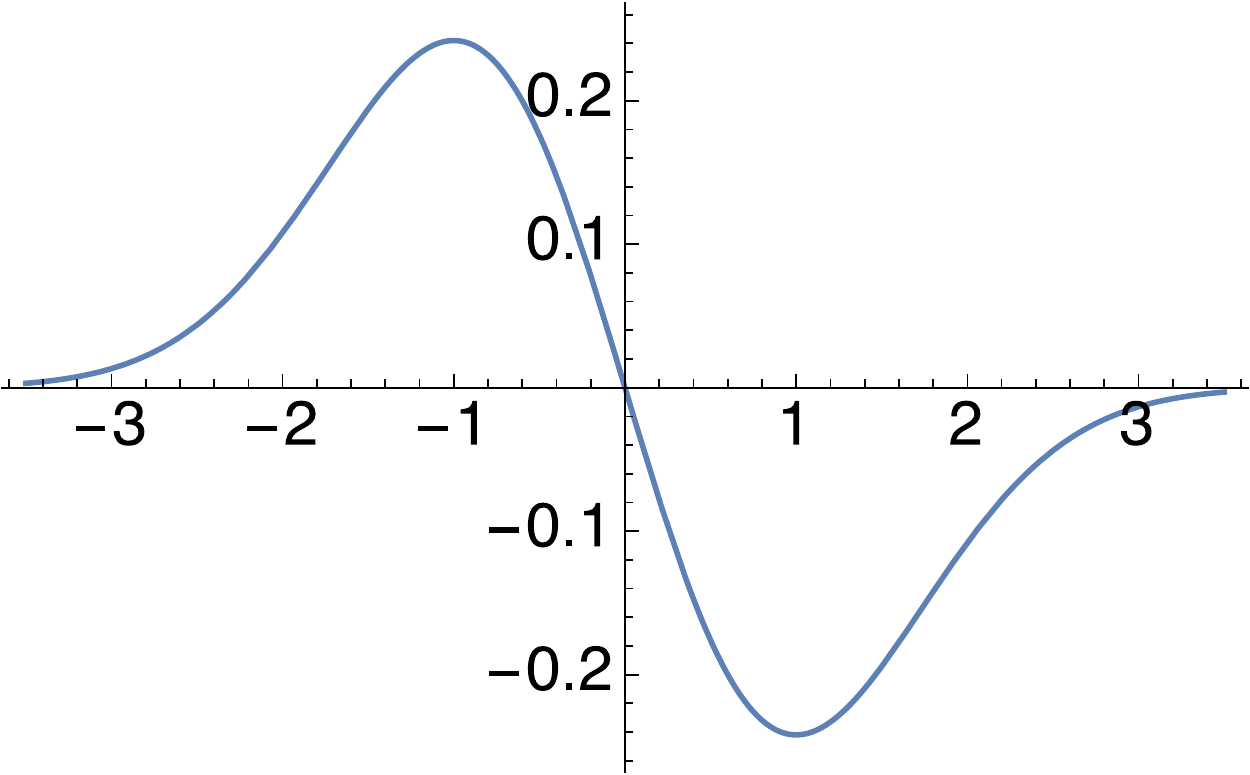} &
      \includegraphics[width=0.25\textwidth]{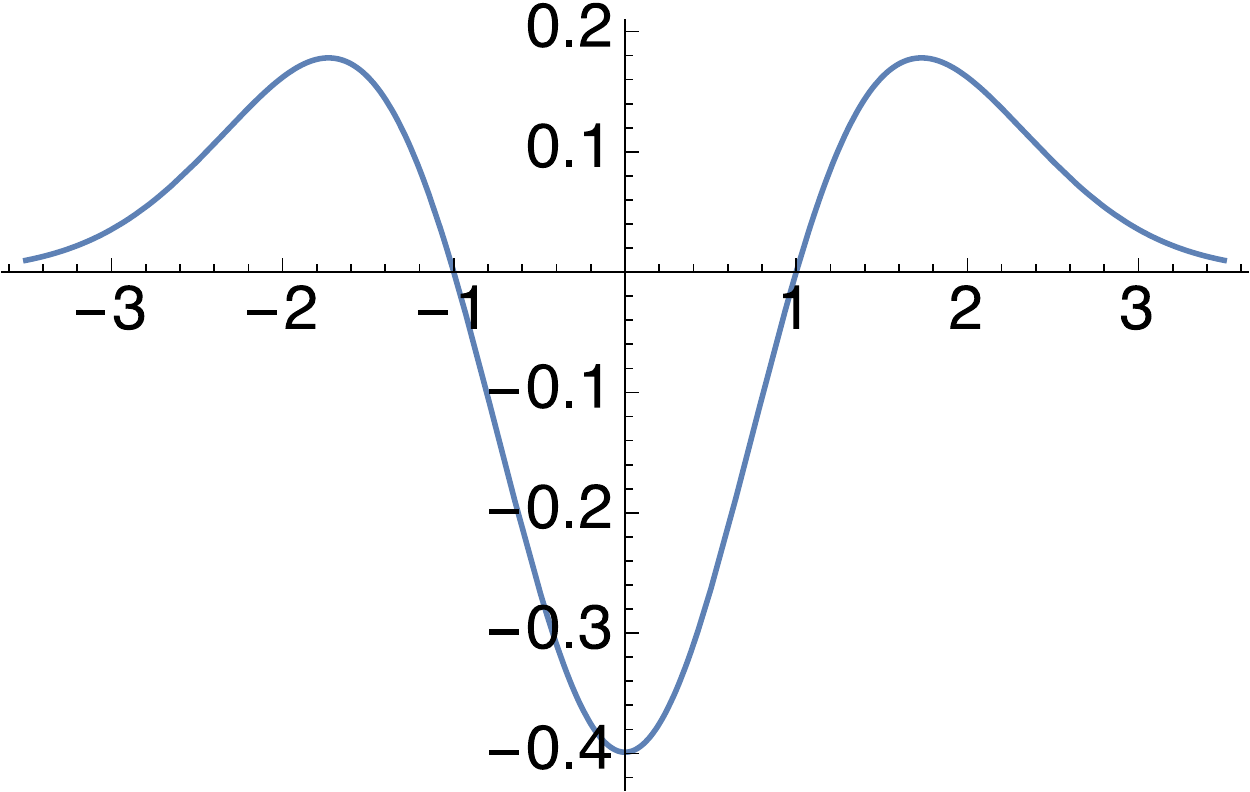} \\
  \\
     {\small $h(t;\; \mu, K=10)$} 
      & {\small $h_{t}(t;\; \mu, K=10)$} 
      & {\small $h_{tt}(t;\; \mu, K=10)$} \\
      \includegraphics[width=0.25\textwidth]{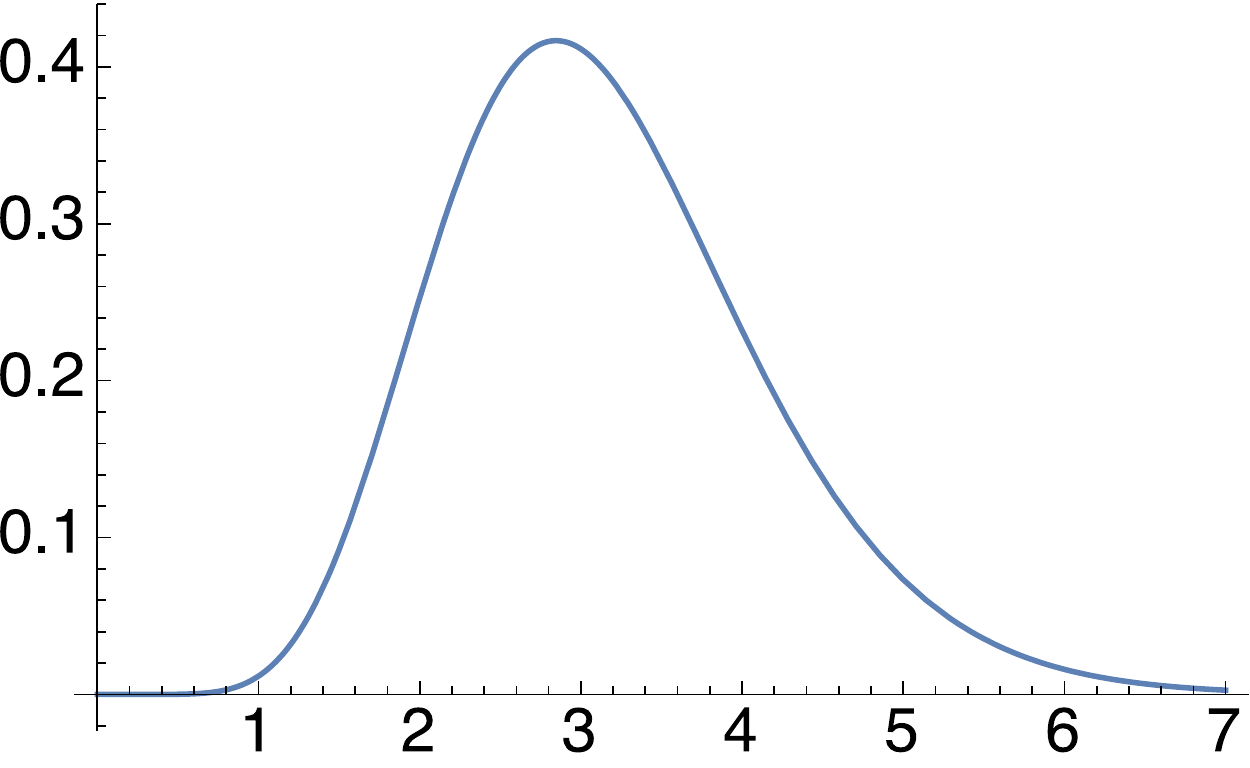} &
      \includegraphics[width=0.25\textwidth]{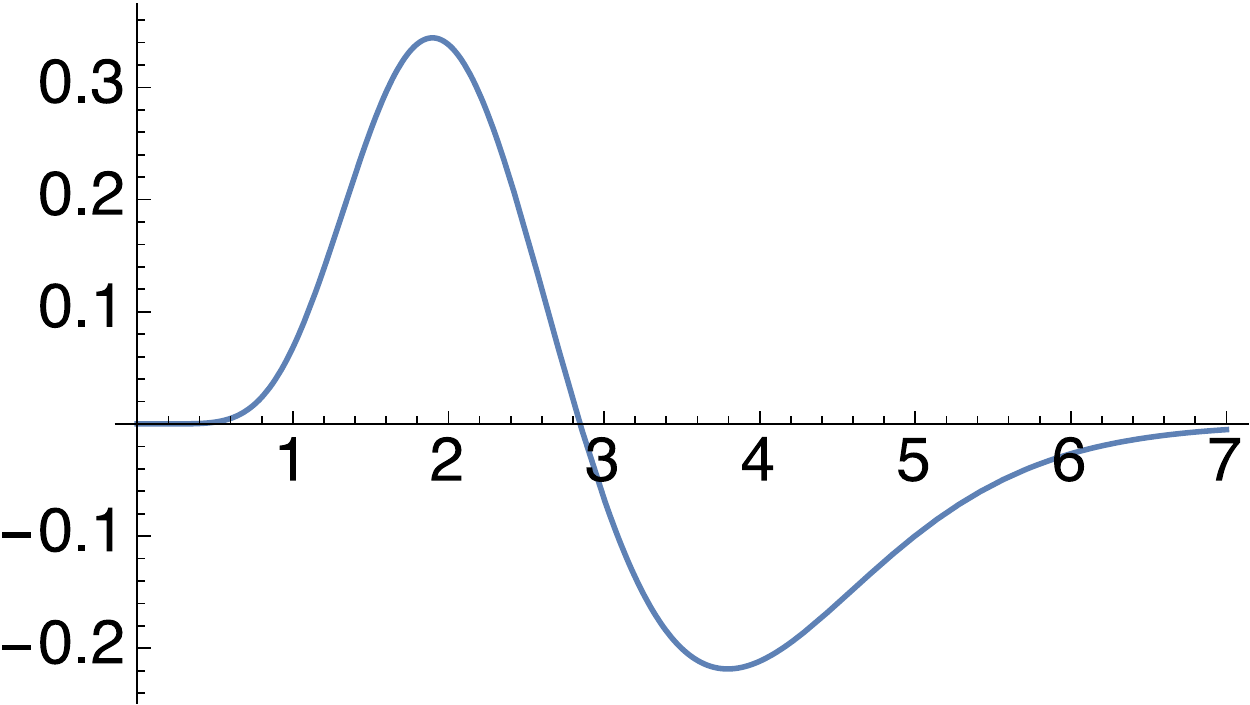} &
      \includegraphics[width=0.25\textwidth]{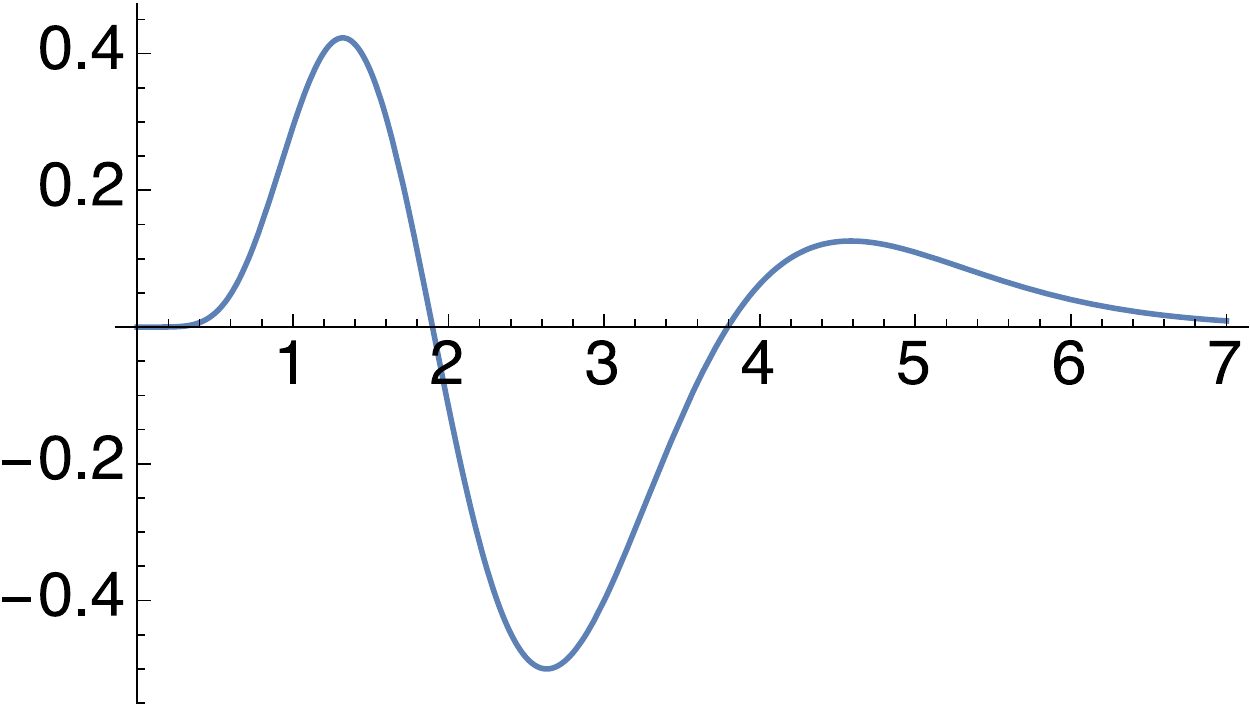} \\
  \\
      {\small $h(t;\; K=10, c = \sqrt{2})$} 
      & {\small $h_{t}(t;\; K=10, c = \sqrt{2})$} 
      & {\small $h_{tt}(t;\; K=10, c = \sqrt{2})$} \\
      \includegraphics[width=0.25\textwidth]{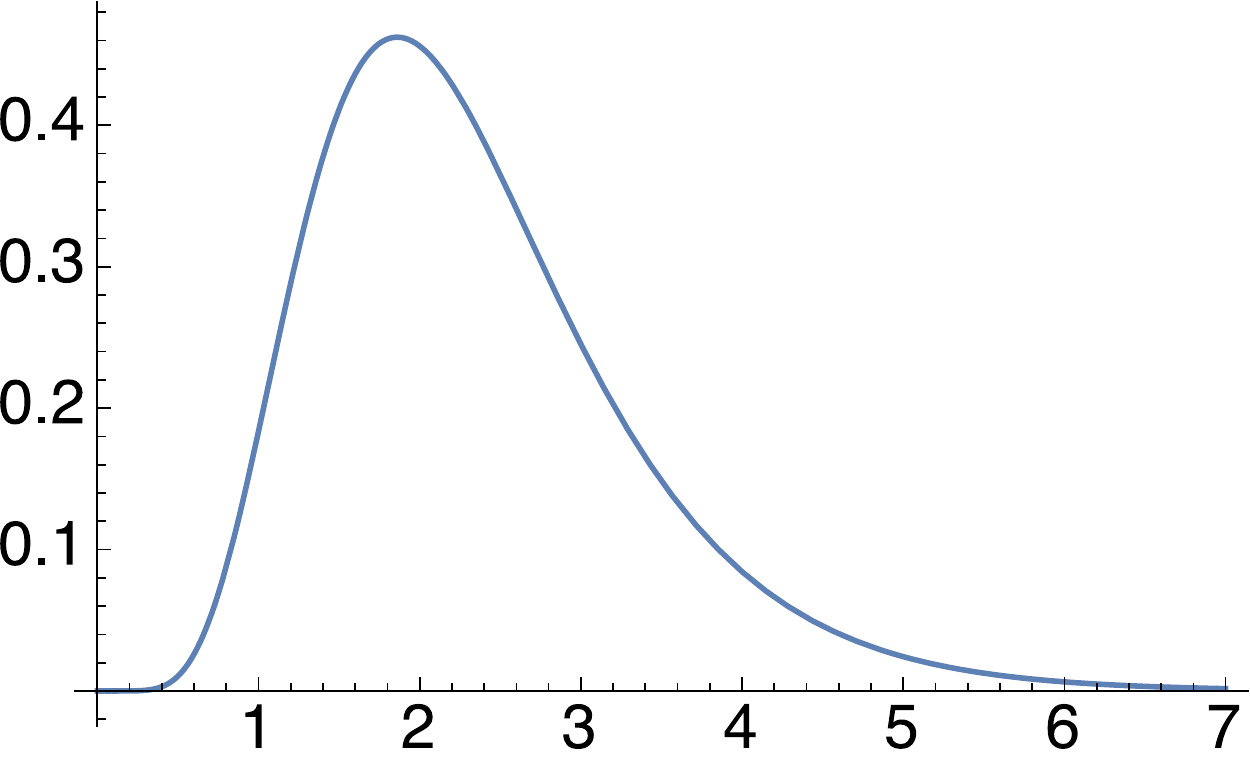} &
      \includegraphics[width=0.25\textwidth]{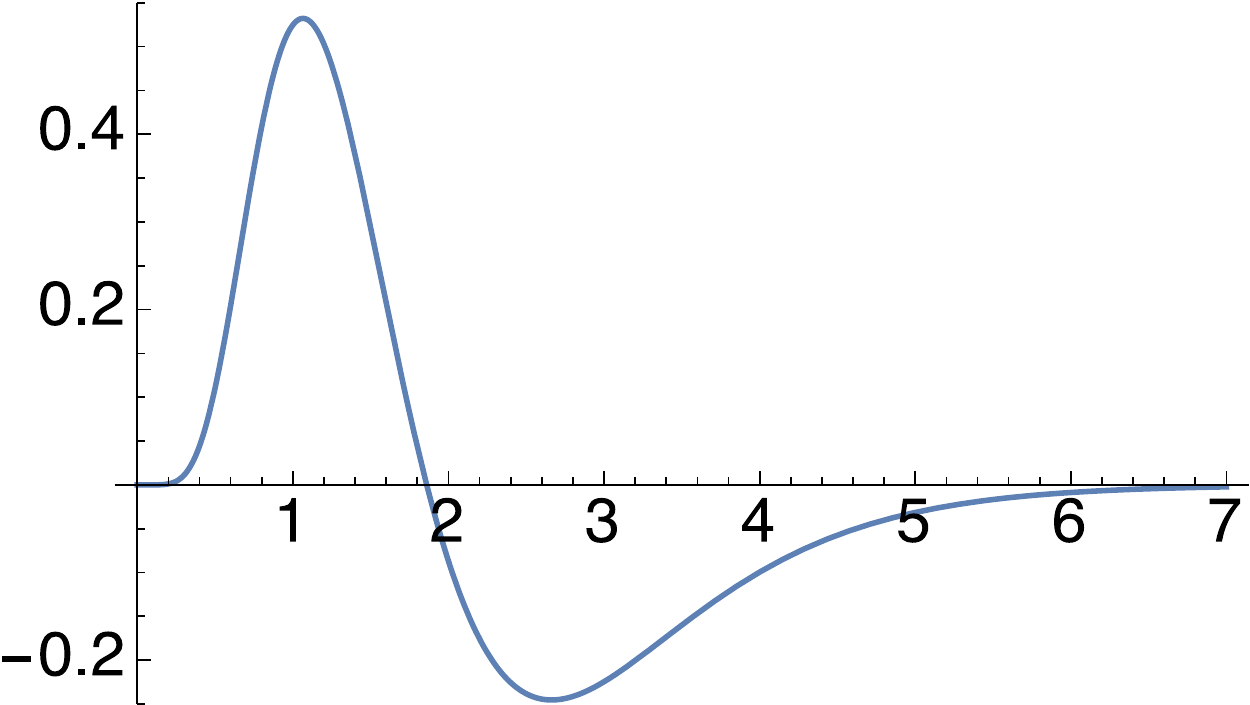} &
      \includegraphics[width=0.25\textwidth]{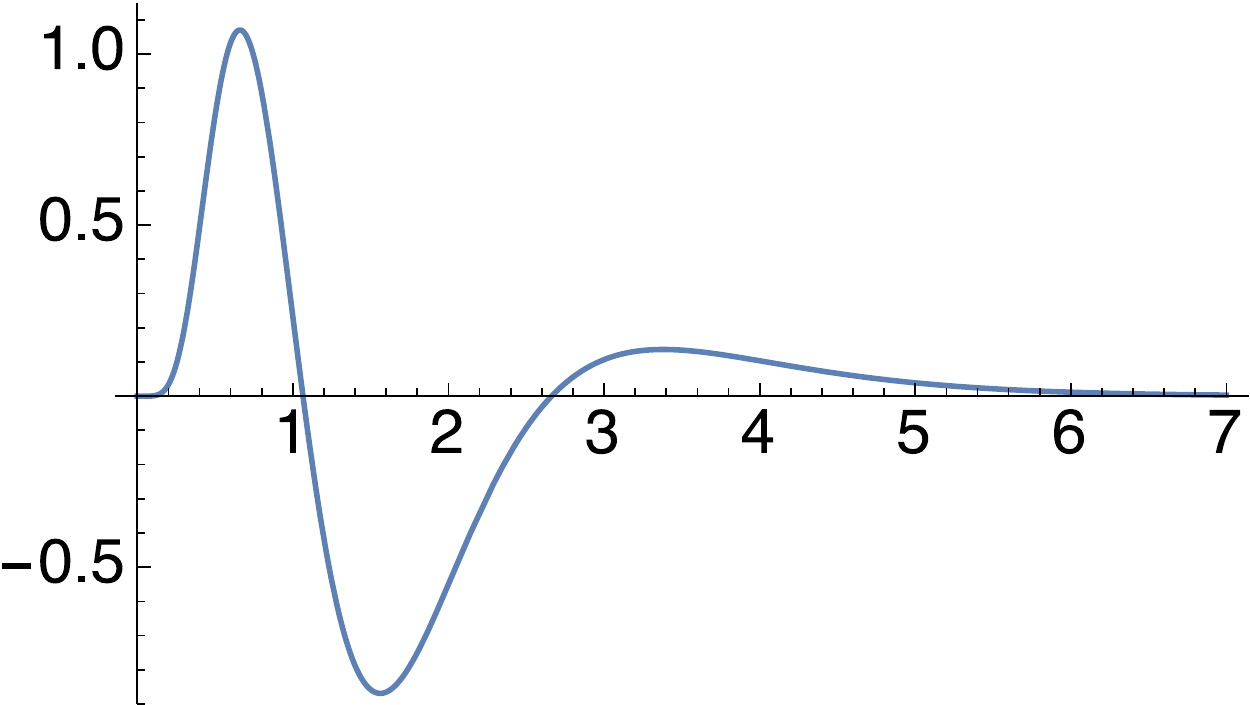} \\
  \\
      {\small $h(t;\; K=10, c = 2)$} 
      & {\small $h_{t}(t;\; K=10, c = 2)$} 
      & {\small $h_{tt}(t;\; K=10, c = 2)$} \\
      \includegraphics[width=0.25\textwidth]{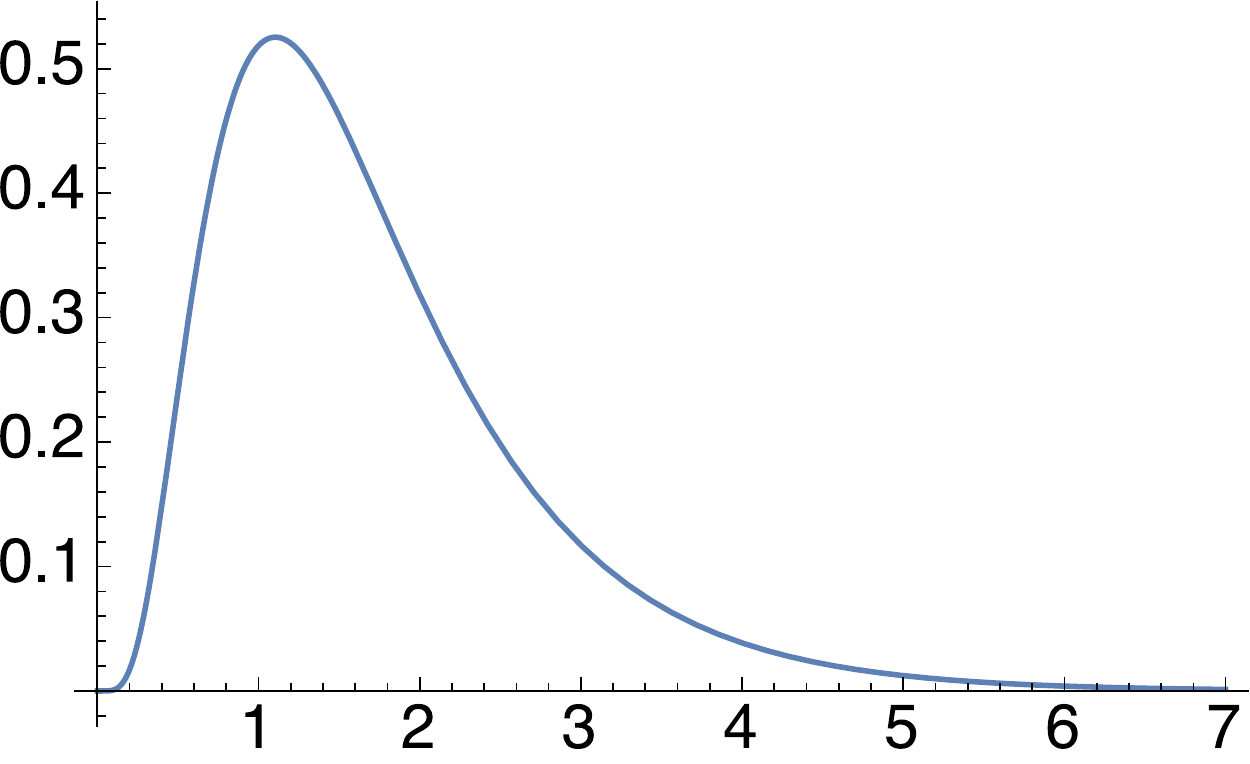} &
      \includegraphics[width=0.25\textwidth]{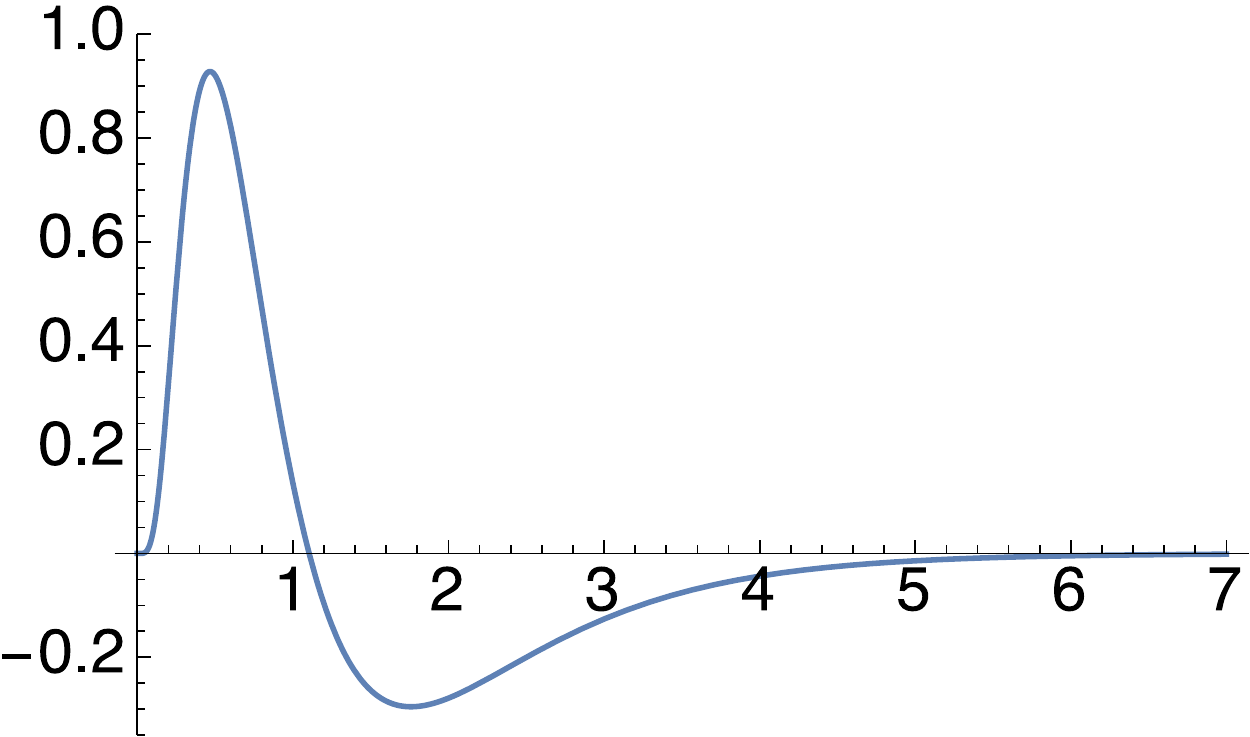} &
      \includegraphics[width=0.25\textwidth]{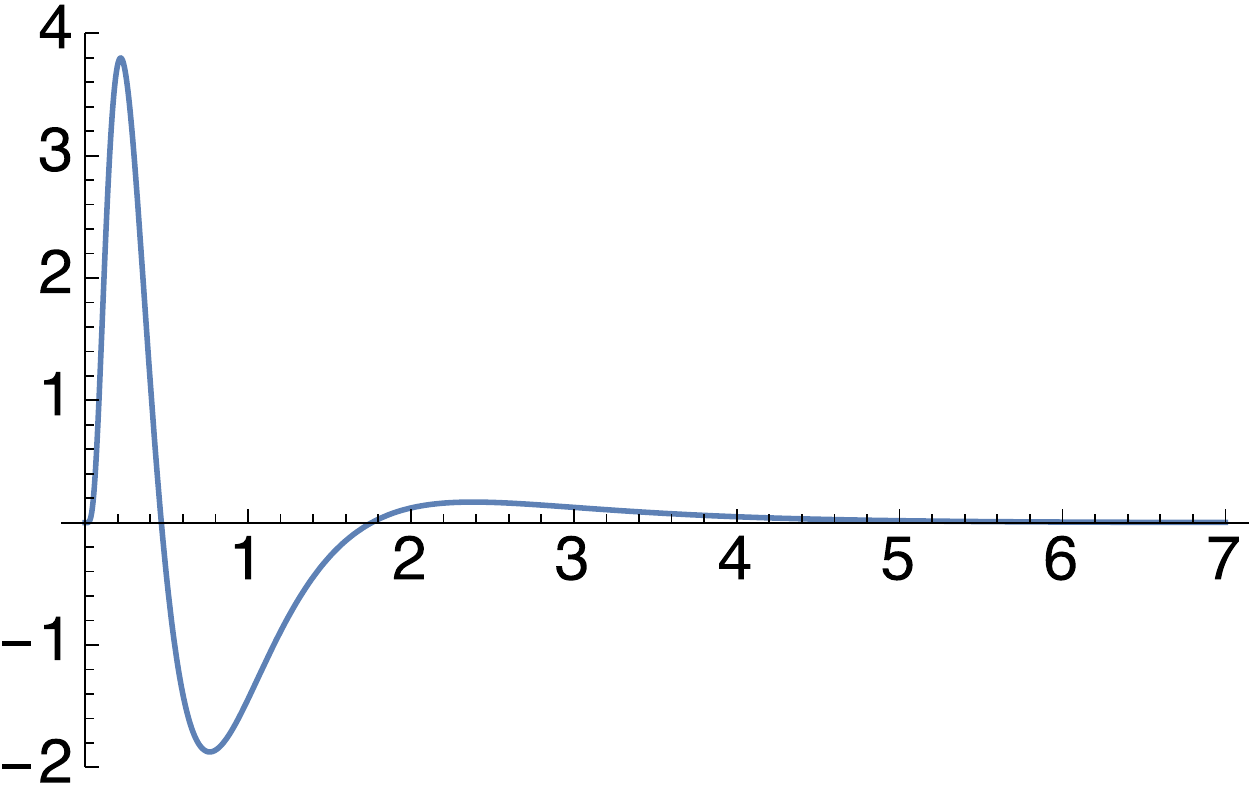} \\
  \\
      {\small $h_{Koe}(t;\; c = \sqrt{2})$} 
      & {\small $h_{Koe,t}(t;\; c = \sqrt{2})$} 
      & {\small $h_{Koe,tt}(t;\; c = \sqrt{2})$} \\
      \includegraphics[width=0.25\textwidth]{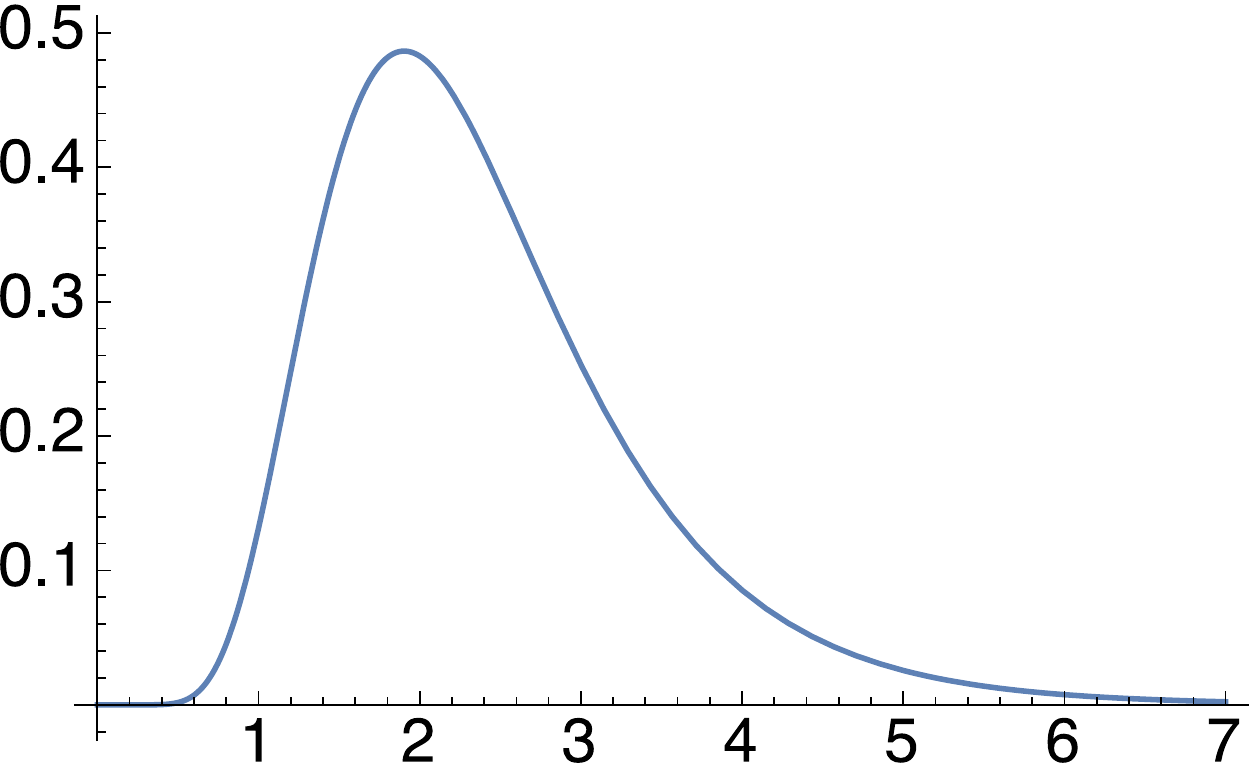} &
      \includegraphics[width=0.25\textwidth]{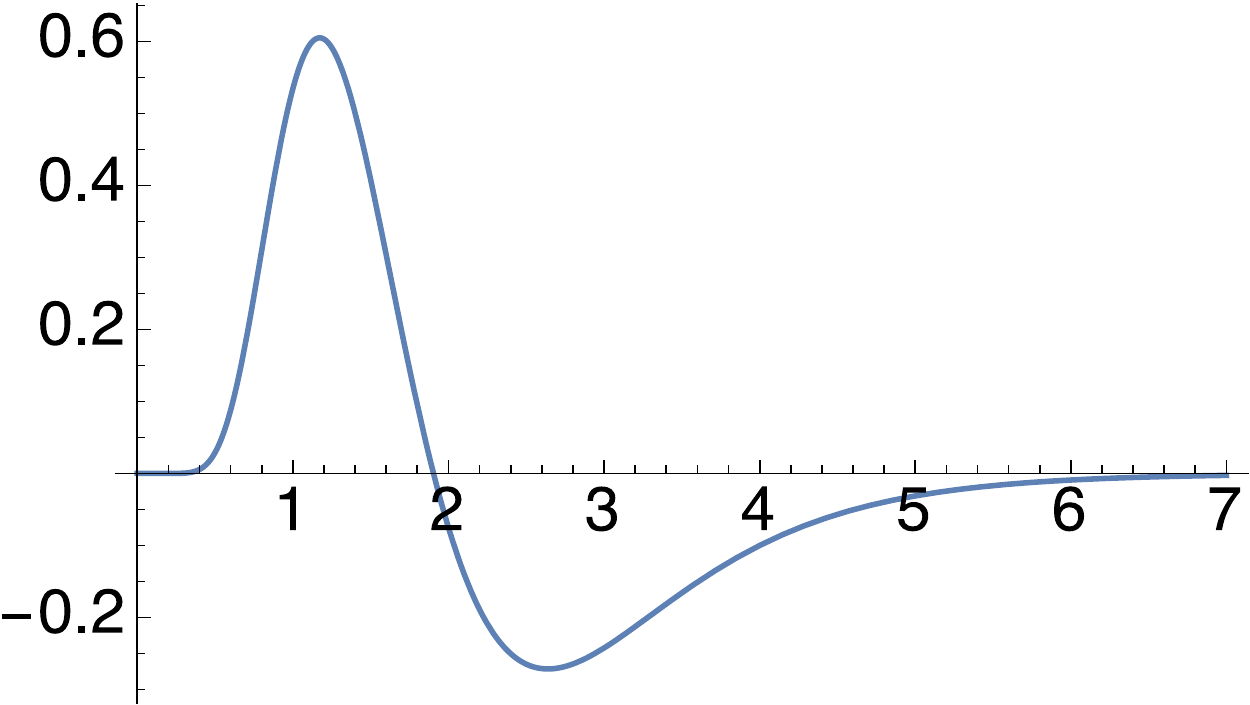} &
      \includegraphics[width=0.25\textwidth]{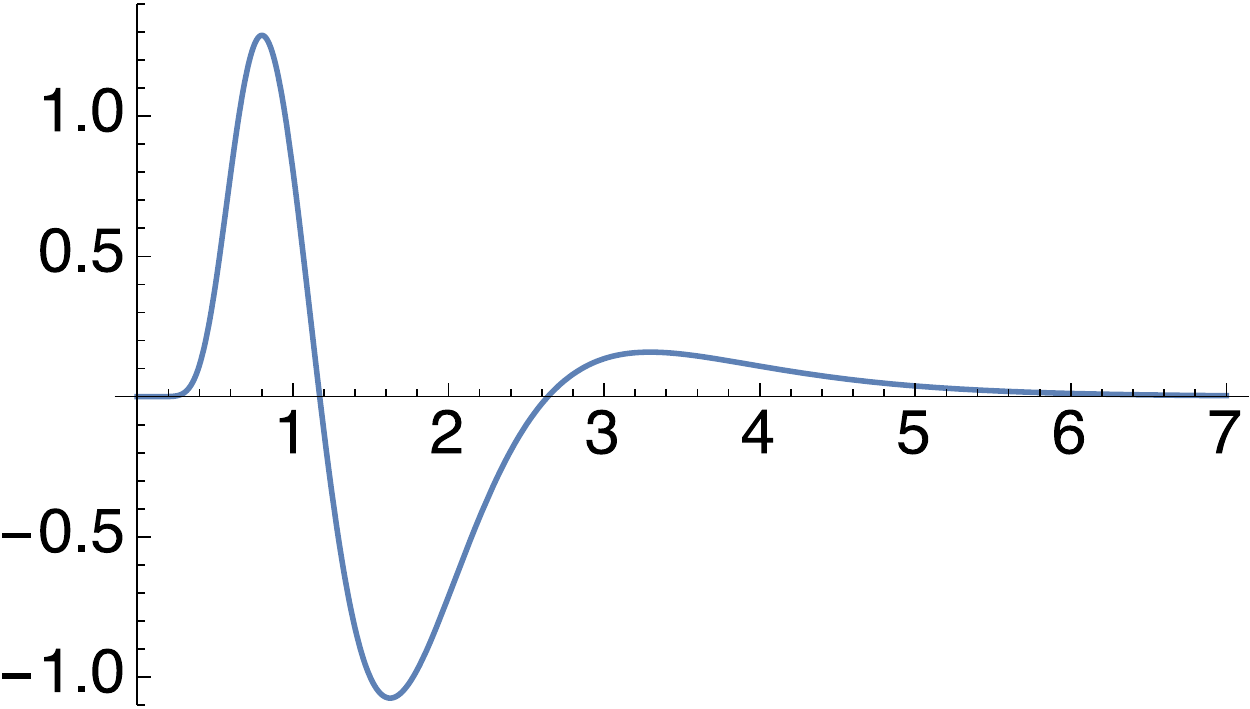} \\
  \\
      {\small $h_{Koe}(t;\; c = 2)$} 
      & {\small $h_{Koe,t}(t;\; c = 2)$} 
      & {\small $h_{Koe,tt}(t;\; c = 2)$} \\
      \includegraphics[width=0.25\textwidth]{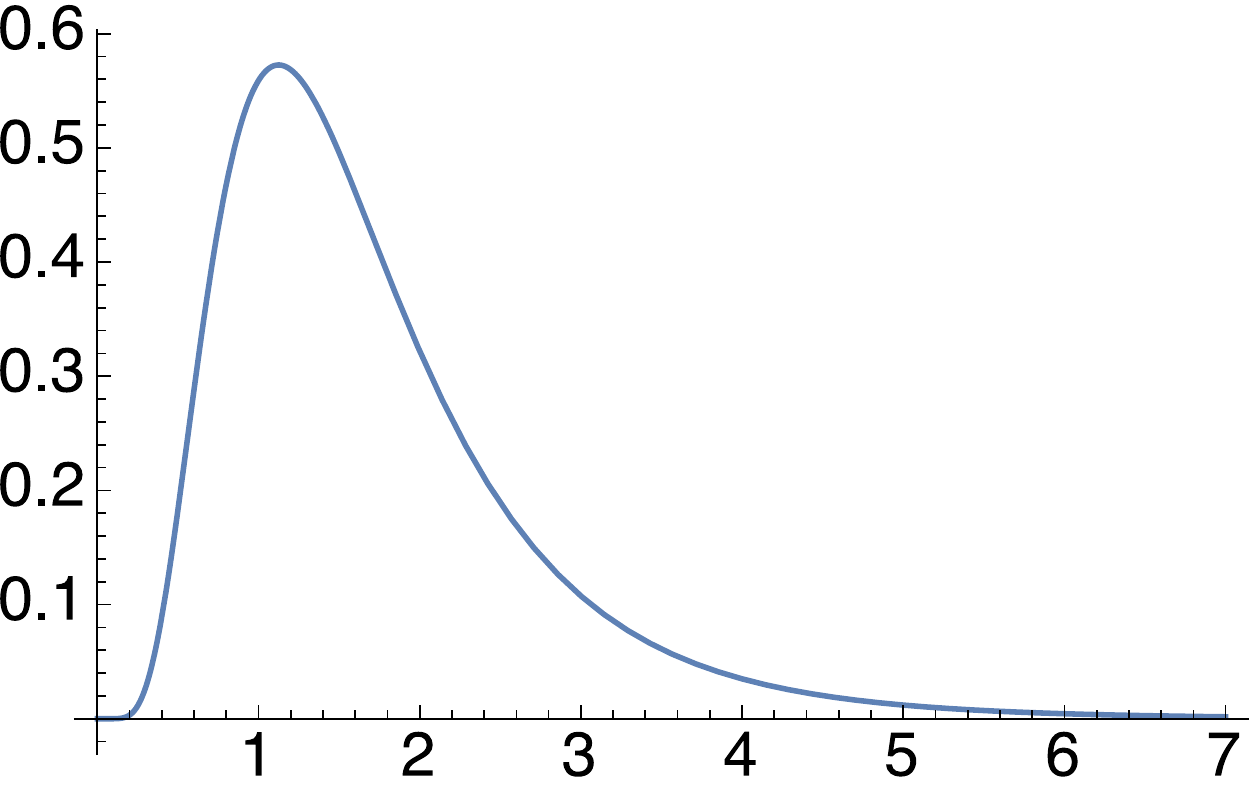} &
      \includegraphics[width=0.25\textwidth]{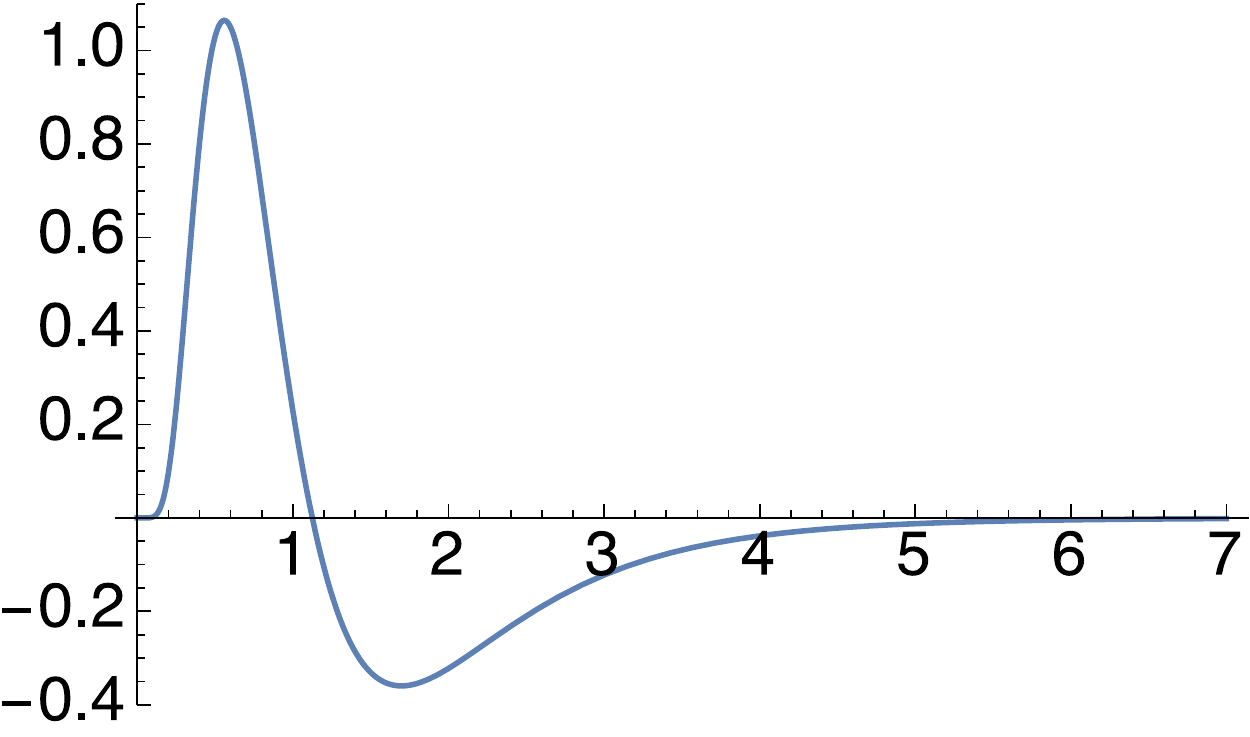} &
      \includegraphics[width=0.25\textwidth]{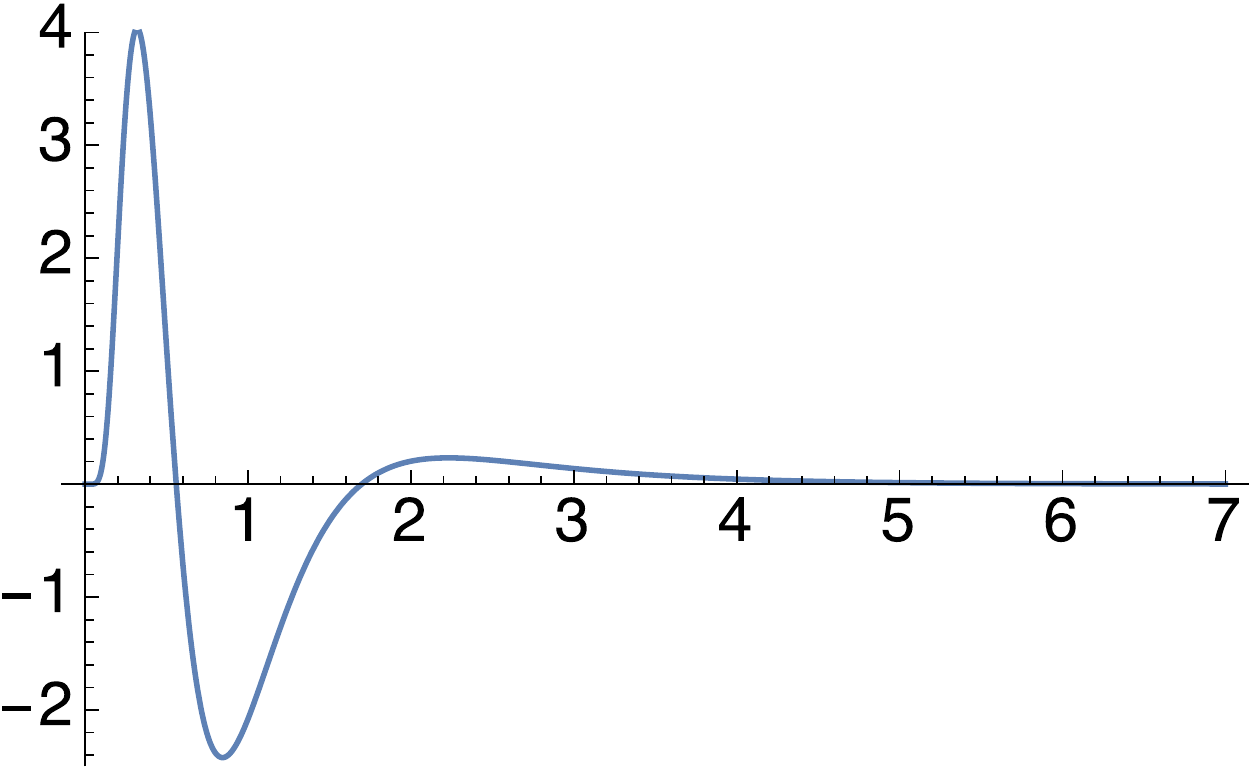} \\
      \end{tabular} 
  \end{center}
   \caption{Temporal scale-space kernels with composed temporal variance $\tau =
     1$ for the main types of temporal scale-space concepts considered in this
     paper and with their first- and second-order temporal derivatives:
     (top row) the non-causal Gaussian kernel $g(t;\; \tau)$,
     (second row) the composition $h(t;\; \mu, K=10)$ of $K = 10$ truncated exponential
     kernels with equal time constants,
     (third row) the composition $h(t;\; K=10, c = \sqrt{2})$ of $K = 10$ truncated exponential
     kernels with logarithmic distribution of the temporal scale
     levels for $c = \sqrt{2}$,
     (fourth row) corresponding kernels $h(t;\; K=10, c = 2)$ for $c = 2$,
     (fifth row) Koenderink's scale-time kernels $h_{Koe}(t;\; c = \sqrt{2})$ corresponding to
     Gaussian convolution over a logarithmically transformed temporal
     axis with the parameters
     determined to match the time-causal limit kernel
     corresponding to truncated exponential kernels with an
     infinite number of logarithmically distributed temporal scale levels according to
     (\protect\ref{eq-var-transf-par-tau-c-sigma-delta-explogdistr-scaletime-models})
     for $c = \sqrt{2}$,
     (bottom row) corresponding scale-time kernels $h_{Koe}(t;\; c = 2)$ for $c = 2$.
     (Horizontal axis: time $t$)}
  \label{fig-temp-kernels-1D}
\end{figure*}

\section{Theoretical background and related work}
\label{sec-related-work}

\subsection{Temporal scale-space concepts}

For processing temporal signals at multiple temporal scales, different
types of temporal scale-space concepts have been developed in the
computer vision literature
(see Figure~\ref{fig-temp-kernels-1D}):

For off-line processing of pre-recorded signals, a {\em non-causal Gaussian
temporal scale-space\/} concept may in many situations be sufficient.
A Gaussian temporal scale-space concept is constructed over the 1-D
temporal domain in a similar manner as a Gaussian spatial scale-space
concept is constructed over a D-dimensional spatial domain
(Iijima \cite{Iij62}; Witkin \cite{Wit83}; Koenderink \cite{Koe84-BC};
 Koenderink and van Doorn \cite{KoeDoo92-PAMI};
 Lindeberg \cite{Lin93-Dis,Lin94-SI,Lin10-JMIV};
 Florack \cite{Flo97-book}; 
 ter Haar Romeny \cite{Haa04-book}),
with or without the difference that a model for temporal delays may or may not
be additionally included (Lindeberg \cite{Lin10-JMIV}).

When processing temporal signals in real time, or when modelling
sensory processes in biological perception computationally, it is on
the other hand necessary to base the temporal analysis on time-causal
operations.

The first time-causal temporal scale-space concept was developed by 
Koenderink \cite{Koe88-BC}, who proposed to apply Gaussian smoothing
on a logarithmically transformed time axis with the present
moment mapped to the unreachable infinity.
This temporal scale-space concept does, however, not have any known
time-recursive formulation. Formally, it requires an infinite memory
of the past and has therefore not been extensively applied in computational applications.

Lindeberg \cite{Lin90-PAMI,Lin15-SSVM,Lin16-JMIV} and 
Lindeberg and Fagerstr{\"o}m \cite{LF96-ECCV}
proposed a time-causal temporal scale-space concept based on truncated exponential
kernels or equivalently first-order integrators coupled in cascade,
based on theoretical results by Schoenberg \cite{Sch50} 
(see also Schoenberg \cite{Sch88-book} and Karlin \cite{Kar68}) 
implying that
such kernels are the only variation-dimini\-shing kernels over a 1-D
temporal domain that guarantee non-creation of new local extrema or
equivalently zero-cross\-ings with increasing temporal scale.
This temporal scale-space concept is additionally time-recursive and can be
implemented in terms of computationally highly efficient first-order
integrators or recursive filters over time.
This theory has been recently extended into a scale-invariant
time-causal limit kernel (Lindeberg \cite{Lin16-JMIV}), which  allows for
scale invariance over the temporal scaling transformations that
correspond to exact mappings between the temporal scale levels in the
temporal scale-space representation based on a discrete set of
logarithmically distributed temporal scale levels.

Based on semi-groups that guarantee either self-similarity over
temporal scales or non-enhancement of local extrema with increasing
temporal scales, Fagerstr{\"o}m \cite{Fag05-IJCV} and 
Lindeberg \cite{Lin10-JMIV} have derived time-causal semi-groups that
allow for a continuous temporal scale parameter and studied theoretical
properties of these kernels.

Concerning temporal processing over discrete time, 
Fleet and Langley \cite{FleLan95-PAMI} performed temporal filtering
for optic flow computations based on recursive filters over time.
Lindeberg \cite{Lin90-PAMI,Lin15-SSVM,Lin16-JMIV} and
Lindeberg and Fagerstr{\"o}m \cite{LF96-ECCV} showed that
first-order recursive filters coupled in cascade constitutes a natural
time-causal scale-space concept over discrete time, based on the
requirement that the temporal filtering over a 1-D temporal signal
must not increase the number of local extrema or equivalently the
number of zero-crossings in the signal.
In the specific case when all the time constants in this model are
equal and tend to zero while simultaneously increasing the number of
temporal smoothing steps in such a way that the composed temporal
variance is held constant, these kernels can be shown to approach the temporal Poisson
kernel \cite{LF96-ECCV}. If on the other hand the time constants of
the first-order integrators are
chosen so that the temporal scale levels become logarithmically
distributed, these temporal smoothing kernels approach a discrete
approximation of the time-causal limit kernel \cite{Lin16-JMIV}.

Applications of using these linear temporal scale-space concepts for modelling
the temporal smoothing step in visual and auditory receptive fields
have been presented by
Lindeberg \cite{Lin97-ICSSTCV,CVAP257,Lin10-JMIV,Lin13-BICY,Lin13-PONE,Lin15-SSVM,Lin16-JMIV},
ter~Haar Romeny {\em et al.\/}\ \cite{RomFloNie01-SCSP},
Lindeberg and Friberg \cite{LinFri15-PONE,LinFri15-SSVM}
and Mahmoudi \cite{Mah16-JMIV}.
Non-linear spatio-temporal scale-space concepts have been proposed by
Guichard \cite{Gui98-TIP}.
Applications of the non-causal Gaussian temporal scale-space concept for
computing spatio-temporal features have been presented by
Laptev and Lindeberg
\cite{LapLin03-ICCV,LapLin04-ECCVWS,LapCapSchLin07-CVIU},
Kl{\"a}ser {\em et al.\/}\ \cite{KlaMarSch08-BMVC},
Willems {\em et al.\/}\ \cite{WilTuyGoo08-ECCV},
Wang {\em et al.\/}\ \cite{WanUllKlaLapSch09-BMVC},
Shao and Mattivi \cite{ShaMatt10-CIVR} and others,
see specifically Poppe \cite{Pop09-IVC} for a survey of early approaches to
vision-based human human action recognition,
Jhuang {\em et al.\/}\ \cite{JhuSerWolPog07-ICCV} and
Niebles {\em et al.\/}\ \cite{NieWanFei08-IJCV} for 
conceptually related non-causal Gabor approaches,
Adelson and Bergen \cite{AdeBer85-JOSA} and
Derpanis and Wildes \cite{DerWil12-PAMI} for closely related 
spatio-temporal orientation models
and Han {\em et al.\/}\ \cite{HanXuZhu15-JMIV} for a related mid-level
temporal representation termed the video primal sketch.

Applications of the temporal scale-space model based on truncated exponential
kernels with equal time constants coupled in cascade and corresponding
to Laguerre functions (Laguerre polynomials multiplied by a truncated
exponential kernel) for computing
spatio-temporal features have presented by 
Rivero-Moreno and Bres \cite{RivBre04-ImAnalRec},
Shabani {\em et al.\/}\ \cite{ShaClaZel12-BMVC} and
Berg {\em et al.\/}\ \cite{BerReyRid14-SensMEMSElOptSyst}
as well as for handling time scales in video surveillance 
(Jacob and Pless \cite{JacPle08-CircSystVidTech}), for performing
edge preserving smoothing in video streams
(Paris \cite{Par08-ECCV}) and is closely related to Tikhonov
regularization as used for image restoration by {\em e.g.\/} Surya {\em et al.\/}\ \cite{SurVorPelJosSeePal15-JMIV}.
A general framework for performing spatio-temporal feature detection based on the temporal scale-space
model based on truncated exponential kernels coupled in cascade with
specifically the both theoretical and practical advantages of using logarithmic
distribution of the intermediated temporal scale levels in terms of
temporal scale invariance and better temporal dynamics (shorter
temporal delays) has been presented in Lindeberg \cite{Lin16-JMIV}.

\subsection{Relative advantages of different temporal scale spaces}

When developing a temporal scale selection mechanism over a time-causal
temporal domain, a first problem concerns what time-causal scale-space concept
to base the multi-scale temporal analysis upon.
The above reviewed temporal scale-space concepts have different
relative advantages from a theoretical and computational viewpoint.
In this section, we will perform an in-depth examination of the
different temporal scale-space concepts that have been developed in
the literature, which will lead us to a class of time-causal scale-space
concepts that we argue is particularly suitable with respect to the
set of desirable properties we aim at.

The non-causal Gaussian temporal scale space is in many cases
the conceptually easiest temporal scale-space concept to handle
and to study analytically (Lindeberg \cite{Lin10-JMIV}).
The corresponding temporal kernels are scale invariant, have compact closed-form
expressions over both the temporal and frequency domains and obey a
semi-group property over temporal scales.
When applied to pre-recorded signals, temporal delays can if desirable
be disregarded, which eliminates any need for temporal delay
compensation.
This scale-space concept is, however, not time-causal and not
time-recursive, which implies fundamental limitations with regard to 
real-time applications and realistic modelling of biological perception.

Koenderink's scale-time kernels \cite{Koe88-BC} are truly time-causal,
allow for a continuous temporal scale parameter, have good
temporal dynamics and have a compact explicit expression over the temporal domain.
These kernels are, however, not time-recursive, which implies that
they in principle require an infinite memory of the past
(or at least extended temporal buffers corresponding to the temporal extent to
which the infinite support temporal kernels are truncated at the tail).
Thereby, the application of Koenderink's scale-time model to video analysis
implies that substantial temporal buffers are needed when
implementing this non-recursive temporal scale-space in practice.
Similar problems with substantial need for extended temporal buffers
arise when applying the non-causal Gaussian temporal scale-space
concept to offline analysis of extended video sequences.
The algebraic expressions for the temporal kernels in the scale-time
model are furthermore not always straightforward to
handle and there is no known simple expression for the Fourier
transform of these kernels or no known simple explicit cascade
smoothing property over temporal scales with respect to the
regular (untransformed) temporal domain.
Thereby, certain algebraic calculations with the scale-time kernels
may become quite
complicated.

The temporal scale-space kernels obtained by coupling truncated
exponential kernels or equivalently first-order integrators in cascade
are both truly time-causal and truly time-recursive
(Lindeberg \cite{Lin90-PAMI,Lin15-SSVM,Lin16-JMIV}; 
Lindeberg and Fagerstr{\"o}m \cite{LF96-ECCV}).  The temporal
scale levels are on the other hand required to be discrete.
If the goal is to construct a real-time signal
processing system that analyses continuous streams of signal data in
real time, one can however argue that a restriction of the theory to a
discrete set of temporal scale levels is less of a contraint, since
the signal processing system anyway has to be based on a finite amount
of sensors and hardware/wetware for sampling and processing the
continuous stream of signal data.

In the special case when all the time constants are equal, the
corresponding temporal kernels in the temporal scale-space model based
on truncated exponential kernels coupled in cascade have compact 
explicit expressions that are easy to handle both
in the temporal domain and in the frequency domain, which simplifies
theoretical analysis.
These kernels obey a semi-group property over
temporal scales, but are not scale invariant and lead to slower
temporal dynamics when a larger number of primitive temporal filters
are coupled in cascade  (Lindeberg \cite{Lin15-SSVM,Lin16-JMIV}).

In the special case when the temporal scale levels in this scale-space
model are logarithmically distributed, these kernels have a manageable explicit
expression over the Fourier domain that enables some closed-form
theoretical calculations. Deriving an explicit expression over the
temporal domain is, however, harder, since the explicit expression
then corresponds to a linear combination of truncated exponential
filters for all the time constants, with the coefficients determined
from a partial fraction expansion of the Fourier transform, which
may lead to rather complex closed-form expressions.
Thereby certain analytical calculations may become harder to handle.
As shown in \cite{Lin16-JMIV} and
Appendix~\ref{sec-scaletime-approx-limit-kernel}, some such
calculations can on the other hand be well approximated via a
scale-time approximation of the time-causal temporal scale-space kernels.
When using a logarithmic distribution of the temporal scales, the
composed temporal kernels do however have very good temporal dynamics and much
better temporal dynamics compared to corresponding kernels obtained by
using truncated exponential kernels with equal time constants coupled
in cascade.
Moreover, these kernels lead to a
computationally very efficient numerical implementation.
Specifically, these kernels allow for the formulation of a time-causal
limit kernel that obeys scale invariance under temporal scaling
transformations, 
which cannot be achieved if using a uniform
distribution of the temporal scale levels
 (Lindeberg \cite{Lin15-SSVM,Lin16-JMIV}).

The temporal scale-space representations obtained from the
self-similar time-causal semi-groups have a continuous scale parameter
and obey temporal scale invariance
(Fagerstr{\"o}m \cite{Fag05-IJCV}; Lindeberg \cite{Lin10-JMIV}).
These kernels do, however, have
less desirable temporal dynamics (see
Appendix~\ref{app-undesired-temp-dyn-temp-semi-group} for a general
theoretical argument about undesirable consequences of imposing a
temporal semi-group property on temporal kernels with temporal delays) and/or lead to pseudodifferential
equations that are harder to handle both theoretically and in terms of
computational implementation.
For these reasons, we shall not consider those time-causal semi-groups
further in this treatment.

\subsection{Previous work on methods for scale selection}

A general framework for performing scale selection for local
differential operations was proposed in Lindeberg
\cite{Lin93-SCIA,Lin93-Dis} 
based on the detection of local extrema over scale of scale-normalized derivative 
expressions and then refined in Lindeberg \cite{Lin97-IJCV,Lin98-IJCV}
--- see 
Lindeberg \cite{Lin99-CVHB,Lin14-EncCompVis} for tutorial overviews.

This scale selection approach has been applied to a large number of 
feature detection tasks over spatial 
image domains including
detection of scale-invariant interest points
(Lindeberg \cite{Lin97-IJCV,Lin12-JMIV},
Mikolajczyk and Schmid \cite{MikSch04-IJCV};
Tuytelaars and Mikolajczyk \cite{TuyMik08-Book}),
performing feature tracking (Bretzner and Lindeberg \cite{BL97-CVIU}),
computing shape from texture and disparity gradients
(Lindeberg and G{\aa}rding \cite{LG93-ICCV}; G{\aa}rding and Lindeberg \cite{GL94-IJCV}),
detecting 2-D and 3-D ridges 
(Lindeberg \cite{Lin98-IJCV}; Sato {\em et al.\/}\ \cite{SatNakShiAtsYouKolGerKik98-MIA};
Frangi {\em et al.\/}\ \cite{FraNieHooWalVie00-MED};
Krissian {\em et al.\/}\ \cite{KriMalAyaValTro00-CVIU}),
computing receptive field responses for object recognition
(Chomat {\em et al.\/}\ \cite{ChoVerHalCro00-ECCV};
Hall {\em et al.\/}\ \cite{HalVerCro00-ECCV}),
performing hand tracking and hand gesture recognition
(Bretzner {\em et al.\/}\  \cite{BreLapLin02-FG}) and
computing time-to-collision
(Negre {\em et al.\/}\ \cite{NegBraCroLau08-ExpRob}).

Specifically, very successful applications have been achi\-eved in
the area of image-based matching and recognition 
(Lowe \cite{Low04-IJCV}; 
Bay {\em et al.\/}\ \cite{BayEssTuyGoo08-CVIU};
Lindeberg \cite{Lin12-Scholarpedia,Lin15-JMIV}).
The combination of local scale selection from local extrema of
scale-normalized derivatives over scales (Lindeberg \cite{Lin93-Dis,Lin97-IJCV})  
with affine shape adaptation (Lindeberg and Garding \cite{LG96-IVC})
has made it possible to perform multi-view image matching over large
variations in viewing distances and viewing directions
(Mikolajczyk and Schmid \cite{MikSch04-IJCV};
Tuytelaars and van Gool \cite{TuyGoo04-IJCV};
Lazebnik {\em et al.\/}\ \cite{LazSchPon05-PAMI};
Mikolajczyk {\em et al.\/}\ \cite{MikTuySchZisMatSchKadGoo05-IJCV};
Rothganger {\em et al.\/}\ \cite{RotLazSchPon06-IJCV}).
The combination of interest point detection from scale-space extrema
of scale-normalized differential invariants (Lindeberg \cite{Lin93-Dis,Lin97-IJCV})  
with local image descriptors 
(Lowe \cite{Low04-IJCV}; Bay {\em et al.\/}\ \cite{BayEssTuyGoo08-CVIU}) 
has made it possible to design robust methods for performing object
recognition of natural objects in natural environments 
with numerous applications to object recognition (Lowe \cite{Low04-IJCV}; Bay {\em et al.\/}\ \cite{BayEssTuyGoo08-CVIU}),
object category classification (Bosch {\em et al.\/}\ \cite{BosZisMun07-ICCV}; Mutch and Lowe \cite{MutLow08-IJCV}), 
multi-view geometry (Hartley and Zisserman \cite{HarZis04-Book}),
panorama stitching (Brown and Lowe \cite{BroLow07-IJCV}), 
automated construction of 3-D object and scene models from visual input
(Brown and Lowe \cite{BroLow05-3DIM}; Agarwal {\em et al.\/}\ \cite{AgaSnaSimSeiSze09-ICCV}),
synthesis of novel views from previous views of the same object (Liu \cite{LiuYueTor11-PAMI}),
visual search in image databases 
(Lew {\em et al.\/}\ \cite{LewSebDjeJai06-ACM-Multi}; Datta {\em et  al.\/}\ \cite{DatJosLiWan08-CompSurv}), 
human computer interaction based on visual input 
(Porta \cite{Por02-HumCompStud}; Jaimes and Sebe \cite{JaiSeb07-CVIU}),
biometrics (Bicego {\em et al.\/}\ \cite{BicLagGroTis06-CVPRW}; Li \cite{Li09-EncBiometr}) and 
robotics (Se {\em et al.\/}\ \cite{SeLowLit05-TROB}; Siciliano and
Khatib \cite{SicKha08-HandBookRob}).

Alternative approaches for performing scale selection over spatial
image domains have also been proposed in terms of
(i)~detecting peaks of weighted entropy measures
(Kadir and Brady \cite{KadBra01-IJCV}) or 
Lyaponov functionals (Sporring {\em et al.\/}
\cite{SpoCoilTra00-ICIP}) over scales,
(ii)~minimising normalized error measures over scale 
(Lindeberg \cite{Lin97-IVC}),
(iii)~determining minimum reliable scales for edge detection based on
a noise suppression model (Elder and Zucker \cite{EldZuc98-PAMI}),
(iv)~determining at what scale levels to stop in non-linear diffusion-based image restoration methods
based on similarity measurements relative to the original
image data (Mr{\'a}zek and Navara \cite{MraNav03-IJCV}),
(v)~by comparing reliability measures from statistical classifiers for texture analysis at multiple
scales (Kang {\em et al.\/}\ \cite{KanMorNag05-ScSp}),
(vi)~by computing image segmentations from the scales at which a
supervised classifier delivers class labels with the highest reliability measure
(Loog {\em et al.\/}\ \cite{LooLiTax09-LNCS}; Li {\em et al.\/}
\cite{LiTaxLoo11-ScSp}), 
(vii) selecting scales for edge detection by estimating the saliency of
elongated edge segments (Liu {\em et al.\/}\ \cite{LiuWanYaoZha12-CVPR}) or
(viii) considering subspaces generated by local image descriptors
computed over multiple scales (Hassner {\em et al.\/}
\cite{HasMayZel12-CVPR}).

More generally, spatial scale selection can be seen as a specific
instance of computing invariant receptive field responses under
natural image transformations, to (i)~handle objects in the world of
different physical size and to account for scaling transformations
caused by the perspective mapping, and with extensions to (ii)~affine image
deformations to account for variations in the viewing direction and
(iii)~Galilean transformations to account for relative motions between
objects in the world and the observer as well as to (iv)~illumination variations
(Lindeberg \cite{Lin13-PONE}).

Early theoretical work on temporal scale selection in a time-causal
scale space was presented in 
Lindeberg \cite{Lin97-AFPAC} with primary focus on the temporal 
Poisson scale-space, which possesses a temporal
semi-group structure over a discrete time-causal temporal domain 
while leading to long temporal delays 
(see Appendix~\ref{app-undesired-temp-dyn-temp-semi-group} for a
general theoretical argument).
Temporal scale selection in non-causal Gaussian spatio-temp\-oral scale space has
been used by Laptev and Lindeberg \cite{LapLin03-ICCV} and
Willems {\em et al.\/}\ \cite{WilTuyGoo08-ECCV} for computing
spatio-temporal interest points, however, with certain theoretical
limitations that are explained in a companion paper
\cite{Lin16-spattempscsel}.%
\footnote{The spatio-temporal scale selection method in (Laptev and Lindeberg
   \cite{LapLin03-ICCV}) is based on a spatio-temporal Laplacian
   operator that is not scale covariant under independent relative
   scaling transformations of the spatial {\em vs.\/} the temporal
   domains \cite{Lin16-spattempscsel}, which implies that the
   spatial and temporal scale estimate will not be robust under
   independent variabilities of the spatial and temporal scales in video data.
   The spatio-temporal scale selection method applied to the determinant of
   the spatio-temporal Hessian in (Willems {\em et al.\/}
   \cite{WilTuyGoo08-ECCV}) does not make use of the full flexibility
   of the notion of $\gamma$-normalized derivative operators \cite{Lin16-spattempscsel} and has
   not previously been developed over a time-causal spatio-temporal domain.}
The purpose of this article is to present a much further developed
and more general theory for temporal scale selection in time-causal scale spaces over 
continuous temporal domains and to analyse the theoretical scale 
selection properties for different types of model signals.

\section{Scale selection properties for the non-causal Gaussian
  temporal scale space concept}
\label{sec-scsel-prop-gauss-temp-scsp}

In this section, we will present an overview of theoretical properties
that will hold if the Gaussian temporal scale-space concept is
applied to a non-causal temporal domain, if additionally the scale
selection mechanism that has been developed for a non-causal spatial
domain is directly transferred to a non-causal temporal domain.
The set of temporal scale-space properties that we will arrive at will
then be used as a theoretical base-line for developing temporal
scale-space properties over a time-causal temporal domain.

\subsection{Non-causal Gaussian temporal scale-space}

Over a one-dimensional temporal domain, axiomatic derivations of a
temporal scale-space representation based on the assumptions of
(i)~linearity, (ii)~temporal shift invariance, 
(iii)~semi-group property over temporal scale,
(iv)~sufficient regularity properties over time and temporal scale and 
(v)~non-enhancement of local extrema imply 
that the temporal scale-space representation
\begin{equation}
  \label{eq-non-caus-gauss-temp-scsp}
  L(\cdot;\; \tau, \delta) = g(\cdot;\; \tau, \delta) * f(\cdot)
\end{equation}
should be generated by convolution with possibly
time-delayed temporal kernels of the form (Lindeberg \cite{Lin10-JMIV})
\begin{equation}
  \label{eq-gauss-time-delay}
  g(t;\; \tau, \delta) = \frac{1}{\sqrt{2 \pi \tau}} e^{-\frac{(t - \delta)^2}{2\tau}}
\end{equation}
where $\tau$ is a temporal scale parameter corresponding to the
variance of the Gaussian kernel and $\delta$ is a temporal delay.
Differentiating the kernel with respect to time gives
\begin{align}
  \begin{split}
     g_t(t;\; \tau, \delta) = - \frac{(t - \delta)}{\tau} \, g(t;\; \tau, \delta)
  \end{split}\\
  \begin{split}
     g_{tt}(t;\; \tau, \delta) = \frac{((t - \delta)^2 - \tau)}{\tau^2} \, g(t;\; \tau, \delta)
  \end{split}
\end{align}
see the top row in Figure~\ref{fig-temp-kernels-1D} for graphs.
When analyzing pre-recorded temporal signals, it can be preferable to
set the temporal delay to zero, leading to temporal scale-space
kernels having a similar form as spatial Gaussian kernels:
\begin{equation}
  g(t;\; \tau) = \frac{1}{\sqrt{2 \pi \tau}} e^{-\frac{t^2}{2\tau}}.
\end{equation}

\subsection{Temporal scale selection from scale-normalized 
  derivatives}

As a conceptual background to the treatments that we shall later
develop regarding temporal scale selection in
time-causal temporal scale spaces, we will in this section describe
the theoretical structure that arises by transferring the theory for
scale selection in a Gaussian scale space over a spatial domain to the
non-causal Gaussian temporal scale space:

Given the temporal scale-space representation $L(t;\; \tau)$ 
of a temporal signal $f(t)$ obtained by convolution with the Gaussian
kernel $g(t;\; \tau)$ according to (\ref{eq-non-caus-gauss-temp-scsp}),
temporal scale selection can be performed by detecting
{\em local extrema over temporal scales\/} of differential expressions
expressed in terms of {\em scale-normalized temporal derivatives\/} 
at any scale $\tau$ according to (Lindeberg \cite{Lin97-IJCV,Lin98-IJCV,Lin99-CVHB,Lin14-EncCompVis})
\begin{equation}
  \partial_{\zeta^n} = \tau^{n \gamma/2} \, \partial_{t^n},
\end{equation}
where $\zeta = t/\tau^{\gamma/2}$ is the scale-normalized temporal variable,
$n$ is the order of temporal differentiation and
$\gamma$ is a free parameter. It can be shown 
\cite[Section~9.1]{Lin97-IJCV} that
this notion of $\gamma$-normalized derivatives corresponds to 
normalizing the $n$th order Gaussian derivatives 
$g_{\zeta^n}(t;\; \tau)$ over a one-dimensional domain
to constant $L_p$-norms over scale $\tau$
\begin{equation}
  \label{eq-Lp-norm-gauss-ders}
  \| g_{\zeta^n}(\cdot;\; \tau) \|_p 
  = \left( 
        \int_{t \in \bbbr} |g_{\zeta^n}(t;\; \tau)|^p \, dt
      \right)^{1/p} 
   = G_{n,\gamma}
\end{equation}
with
\begin{equation}
 \label{eq-sc-norm-p-from-gamma}
  p = \frac{1}{1 + n(1 - \gamma)}
\end{equation}
where the perfectly scale invariant case $\gamma = 1$ corresponds to
$L_1$-normalization for all orders $n$ of temporal differentiation.

\paragraph{Temporal scale invariance.}

A general and very useful scale invariant property that results from this
construction of the notion of scale-normalized temporal derivatives
can be stated as follows: 
Consider two signals $f$ and $f'$ that are related by a temporal scaling transformation
\begin{equation}
 f'(t') = f(t) 
    \quad \mbox{with} 
    \quad  t' = S \, t,
\end{equation}
and assume that there is a local extremum over scales at $(t_0;\; \tau_0)$
in a differential expression ${\cal D}_{\gamma-norm} L$ defined as a
homogeneous polynomial of Gaussian derivatives computed from
the scale-space representation $L$ of the original signal $f$.
Then, there will be a corresponding local extremum over scales at
$(t_0';\; \tau_0') = (S \, t_0;\; S^2 \tau_0)$ in the corresponding differential
expression  ${\cal D}_{\gamma-norm} L'$  computed from
the scale-space representation $L'$ of the rescaled signal $f'$
\cite[Section~4.1]{Lin97-IJCV}.

This scaling result holds for all homogeneous polynomial 
differential expression and implies that local extrema over 
scales of $\gamma$-normalized derivatives are
preserved under scaling transformations.
Specifically, this scale invariant property implies that if a local scale temporal level
level in dimension of time $\sigma = \tau$
is selected to be proportional to the temporal scale
estimate $\hat{\sigma} = \sqrt{\hat{\tau}}$ such that $\sigma = C \, \hat{\sigma}$, 
then if the temporal signal $f$ is transformed by a temporal scale factor $S$, the temporal
scale estimate and therefore also the selected temporal scale level 
will be transformed by a similar temporal factor $\hat{\sigma}' = S \, \hat{\sigma}$,
implying that the selected temporal scale levels will automatically
adapt to variations in the characteristic temporal scale of the signal.
Thereby, such local extrema over temporal scale provide a theoretically 
well-founded way to automatically adapt the scale levels to local
scale variations.

Specifically, scale-normalized scale-space derivatives of order $n$ at
corresponding temporal moments will
be related according to
\begin{equation}
  \label{eq-sc-inv-temp-scaling-gauss-temp-scsp}
  L'_{\zeta'^n} (t';\; \tau') =   S^{n(\gamma-1)} L_{\zeta^n} (t;\; \tau)
\end{equation}
which means that $\gamma = 1$ implies perfect scale-invariance
in the sense that the $\gamma$-normalized derivatives at corresponding
points will be equal. If $\gamma \neq 1$, the difference in magnitude
can on the other hand be easily compensated for using the scale values
of the corresponding scale-adaptive image features (see below).

\subsection{Temporal peak}

For a temporal peak modelled as a Gaussian function with variance $\tau_0$
\begin{equation}
  g(t;\; \tau_0) = \frac{1}{\sqrt{2 \pi \tau_0}} e^{-\frac{t^2}{2\tau_0}}.
\end{equation}
it can be shown that scale selection from local extrema over
scale of second-order scale-normalized temporal derivatives 
\begin{equation}
  L_{\zeta\zeta} = \tau^{\gamma} L_{tt}
\end{equation}
implies that the scale estimate at the position $t = 0$ of the peak 
will be given by
(Lindeberg \cite[Equation~(56)]{Lin98-IJCV} \cite[Equation~(212)]{Lin12-JMIV})
\begin{equation}
  \label{eq-sc-sel-peak-1D-general-gamma}
  \hat{\tau} = \frac{2\gamma}{3 - 2\gamma} \, \tau_0.
\end{equation}
If we require the scale estimate to reflect the temporal duration of the peak such
that 
\begin{equation}
  \label{eq-sc-est-temp-peak}
  \hat{\tau} = q^2 \tau_0,
\end{equation} 
then this implies
\begin{equation}
  \label{eq-gamma-fcn-of-q-2nd-der-gauss-temp-scsp}
   \gamma = \frac{3 q^2}{2 \left(q^2+1\right)}
\end{equation}
which in the specific case of $q = 1$ corresponds to
\cite[Section~5.6.1]{Lin98-IJCV} 
\begin{equation}
  \label{eq-gamma-0p75-2nd-der-gauss-temp-scsp}
  \gamma = \gamma_2 = \frac{3}{4}
\end{equation}
and in turn corresponding to $L_p$-normalization for $p = 2/3$ according to (\ref{eq-sc-norm-p-from-gamma}).

If we additionally renormalize the original Gaussian peak to having
maximum value equal to one 
\begin{equation}
  p(t;\; t_0) = \sqrt{2 \pi \tau_0} \, g(t;\; \tau_0) = e^{-\frac{t^2}{2\tau_0}},
\end{equation}
then if using the same value of $\gamma$ for computing the magnitude
response as for selecting the temporal scale, the maximum magnitude
value over scales will be given by
\begin{equation}
  L_{\zeta\zeta,maxmagn} 
  = \frac{2^{\gamma } (2 \gamma -3) }{3 \tau_0}
     \left(\frac{\gamma \, \tau_0}{3-2 \gamma}\right)^{\gamma}
\end{equation}
and will not be independent of the temporal scale $\tau_0$ of the original peak
unless $\gamma = 1$.
If on the other hand using $\gamma = 3/4$ as motivated by requirements of scale
calibration (\ref{eq-sc-est-temp-peak}) for $q = 1$, the scale
dependency will for a Gaussian peak be of the form
\begin{equation}
  \left. L_{\zeta\zeta,maxmagn} \right|_{\gamma=3/4} 
   = \frac{1}{2 \tau_0^{1/4}}.
\end{equation}
To get a scale-invariant magnitude measure for comparing the responses of
second-order temporal derivative responses at different temporal
scales for the purpose of scale calibration, we should therefore consider a scale-invariant magnitude
measure for peak detection of the form
\begin{equation}
  \label{eq-sc-sel-post-norm-temp-peak-scale-calib}
  \left. L_{\zeta\zeta,maxmagn,postnorm} \right|_{\gamma=1} = \tau^{1/4}  \left. L_{\zeta\zeta,maxmagn} \right|_{\gamma=3/4} 
\end{equation}
which for a Gaussian temporal peak will assume the value
\begin{equation}
  \label{eq-sc-sel-post-norm-temp-peak-const-magn}
  \left. L_{\zeta\zeta,maxmagn,postnorm} \right|_{\gamma=1} = \frac{1}{2}
\end{equation}
Specifically, this form of post-normalization corresponds to computing the
scale-normalized derivatives for $\gamma = 1$ at the selected scale
(\ref{eq-sc-est-temp-peak}) of the temporal peak, which according to
(\ref{eq-sc-norm-p-from-gamma}) corresponds to
$L_1$-normalization of the second-order temporal derivative kernels.

\subsection{Temporal onset ramp}

If we model a temporal onset ramp with temporal duration $\tau_0$ as 
the primitive function of the Gaussian kernel with variance $\tau_0$
\begin{equation}
  \Phi(t;\; \tau_0) = \int_{u = - \infty}^{t} g(u;\; \tau_0) \, du,
\end{equation}
it can be shown that scale selection from local extrema over
scale of first-order scale-normalized temporal derivatives 
\begin{equation}
  L_{\zeta} = \tau^{\gamma/2} L_t
\end{equation}
implies that the scale estimate at the central position $t = 0$ will be given by
\cite[Equation~(23)]{Lin98-IJCV} 
\begin{equation}
  \label{eq-sc-sel-ramp-1D-general-gamma}
  \hat{\tau} = \frac{\gamma}{1 - \gamma} \, \tau_0.
\end{equation}
If we require this scale estimate to reflect the temporal duration of the ramp such
that 
\begin{equation}
  \label{eq-sc-est-onset-ramp}
  \hat{\tau} = q^2 \tau_0, 
\end{equation}
then this implies
\begin{equation}
  \label{eq-gamma-fcn-of-q-1st-der-gauss-temp-scsp}
  \gamma = \frac{q^2}{q^2+1}
\end{equation}
which in the specific case of $q = 1$ corresponds to
\cite[Section~4.5.1]{Lin98-IJCV} 
\begin{equation}
  \label{eq-gamma-0p5-1st-der-gauss-temp-scsp}
  \gamma = \gamma_1 = \frac{1}{2}
\end{equation}
and in turn corresponding to $L_p$-normalization for $p = 2/3$ according to
(\ref{eq-sc-norm-p-from-gamma}).

If using the same value of $\gamma$ for computing the magnitude
response as for selecting the temporal scale, the maximum magnitude
value over scales will be given by
\begin{equation}
  L_{\zeta,maxmagn} 
  = \frac{\gamma ^{\gamma /2}}{\sqrt{2 \pi}}
      \left(\frac{1-\gamma}{\tau_0}\right)^{\frac{1}{2}-\frac{\gamma }{2}},
\end{equation}
which is not independent of the temporal scale $\tau_0$ of the original onset
ramp unless $\gamma = 1$.
If using $\gamma = 1$ for temporal scale selection, the selected temporal
scale according to (\ref{eq-sc-sel-ramp-1D-general-gamma}) would,
however, become infinite.
If on the other hand using $\gamma = 1/2$ as motivated by requirements of scale
calibration (\ref{eq-sc-est-onset-ramp}) for $q = 1$, the scale
dependency will for a Gaussian onset ramp be of the form
\begin{equation}
  \left. L_{\zeta,maxmagn} \right|_{\gamma=1/2} 
  = \frac{1}{2 \sqrt{\pi } \sqrt[4]{\tau_0}}.
\end{equation}
To get a scale-invariant magnitude measure for comparing the responses of
first-order temporal derivative responses at different temporal
scales, we should therefore consider a scale-invariant magnitude
measure for ramp detection of the form
\begin{equation}
  \left.  L_{\zeta,maxmagn,postnorm} \right|_{\gamma=1} 
  = \tau^{1/4}  \left. L_{\zeta,maxmagn} \right|_{\gamma=1/2} 
\end{equation}
which for a Gaussian onset ramp will assume the value
\begin{equation}
  \label{eq-sc-sel-onset-ramp-const-magn-after-post-norm}
  \left.  L_{\zeta,maxmagn,postnorm} \right|_{\gamma=1} 
  = \frac{1}{2 \sqrt{\pi}} \approx 0.282
\end{equation}
Specifically, this form of post-normalization corresponds to computing the
scale-normalized derivatives for $\gamma = 1$ at the selected scale
(\ref{eq-sc-est-onset-ramp}) of the onset ramp and thus also to
$L_p$-normalization of the first-order temporal derivative kernels for
$p = 1$.

\subsection{Temporal sine wave}

For a signal defined as a temporal sine wave 
\begin{equation}
  \label{eq-temp-sine-wave}
  f(t) = \sin (\omega_0 t),
\end{equation}
it can be shown that there will be a peak over temporal scales in the magnitude of the
$n$th order temporal derivative $L_{\zeta^n} = \tau^{n\gamma/2} L_{t^n}$ 
at temporal scale \cite[Section~3]{Lin97-IJCV} 
\begin{equation}
  \label{eq-scsel-sine-wave-gauss-temp-scsp-omega}
   \tau_{max} = \frac{n\gamma}{\omega_0^2}.
\end{equation}
If we define a temporal scale parameter $\sigma$ of dimension 
$[\mbox{time}]$ according to $\sigma = \sqrt{\tau}$, then this implies
that the scale estimate is proportional to the wavelength 
$\lambda_0 = 2\pi/\omega_0$ of the sine wave according to
\cite[Equation~(9)]{Lin97-IJCV} 
\begin{equation}
  \label{eq-scsel-sine-wave-gauss-temp-scsp-lambda}
  \sigma_{max} = \frac{\sqrt{\gamma n}}{2 \pi} \lambda_0
\end{equation}
and does in this respect reflect a characteristic time constant over which the
temporal phenomena occur.
Specifically, the maximum magnitude measure over scale \cite[Equation~(10)]{Lin97-IJCV} 
\begin{equation}
  \label{eq-scsel-sine-wave-gauss-temp-scsp-max-magn}
  L_{\zeta^n,max} = \frac{(\gamma n)^{\gamma n/2}}{e^{\gamma n/2}}  \omega_0^{(1-\gamma)n}
\end{equation}
is for $\gamma = 1$ independent of the angular frequency $\omega_0$ of
the sine wave and thereby scale invariant.

In the following, we shall investigate how these
scale selection properties can be transferred to two types of 
time-causal temporal scale-space concepts.

\section{Scale selection properties for the time-causal temporal scale
  space concept based on
  first-order integrators with equal time constants}
\label{sec-scsel-prop-time-caus-trunc-exp-uni-distr}

In this section, we will present a theoretical analysis of the scale selection
properties that are obtained in the time-causal scale-space based on
truncated exponential kernels coupled in cascade, for the specific
case of a uniform distribution of the temporal scale levels in units
of the composed variance of the composed temporal scale-space kernels,
and corresponding to the time-constants of all the primitive
truncated exponential kernels being equal.

We will study three types of idealized
model signals for which closed-form theoretical analysis is possible: 
(i)~a temporal peak modelled as a set of $K_0$ truncated exponential
kernels with equal time constants coupled in cascade,
(ii)~a temporal onset ramp modelled as the primitive function of the
temporal peak model and
(iii)~a temporal sine wave.
Specifically, we will analyse how the selected scale levels $\hat{K}$
obtained from local extrema of temporal derivatives over scale relate
to the temporal duration of a temporal peak or a temporal onset
ramp alternatively how the selected scale levels $\hat{K}$ depends on
the the wavelength of a sine wave. 

We will also study how good
approximation the scale-normalized magnitude measure at the maximum
over temporal scales is compared to the corresponding fully scale-invariant
magnitude measures that are obtained from the non-causal temporal scale concept
as listed in Section~\ref{sec-scsel-prop-gauss-temp-scsp}.

\subsection{Time-causal scale space based on truncated exponential
  kernels with equal time constants coupled in cascade}

Given the requirements that the temporal smoothing operation in a
temporal scale-space representation should obey
(i)~linearity, (ii)~temporal shift invariance,
(iii)~temporal causality and (iv)~guarantee non-creation of new local
extrema or equivalently new zero-crossings with increasing temporal
scale for any one-dimensional temporal signal,
it can be shown 
(Lindeberg \cite{Lin90-PAMI,Lin15-SSVM,Lin16-JMIV};
Lindeberg and Fagerstr{\"o}m \cite{LF96-ECCV})
that the temporal scale-space kernels should be constructed as a cascade of
truncated exponential kernels of the form
 \begin{equation}
  \label{eq-trunc-exp-kern-prim}
    h_{exp}(t;\; \mu_k) 
    = \left\{
        \begin{array}{ll}
          \frac{1}{\mu_k} e^{-t/\mu_k} & t \geq 0, \\
          0         & t < 0.
        \end{array}
      \right.
  \end{equation}
If we additionally require the time constants of all such primitive
kernels that are coupled in cascade to be equal, then this leads to a
composed temporal scale-space kernel of the form
\begin{equation}
 \label{eq-temp-scsp-kernel-uni-distr}
  h_{composed}(t;\; \mu, K) 
  = \frac{t^{K-1} \, e^{-t/\mu}}{\mu^K \, \Gamma(K)} 
  = U(t;\; \mu, K)
\end{equation}
corresponding to Laguerre functions (Laguerre polynomials multiplied by a truncated
exponential kernel)  and also equal to the probability density function of the Gamma distribution
having a Laplace transform of the form
  \begin{align}
    \begin{split}
       H_{composed}(q;\; \mu) 
       & = \int_{t = - \infty}^{\infty} (*_{k=1}^{K} h_{exp}(t;\; \mu_k)) \, e^{-qt} \, dt
    \end{split}\nonumber\\
    \begin{split}
       \label{eq-Laplace-temp-scsp-kernel-uni-distr}
        & =  \frac{1}{(1 + \mu q)^K}
           = \bar{U}(q;\; \mu, K).
    \end{split}
  \end{align}
Differentiating the temporal scale-space kernel with respect to time 
$t$ gives
\begin{align}
  \begin{split}
      U_t(t;\; \mu, K) 
      &  = -\frac{(t -(K-1) \mu)}{\mu  t} \, U(t;\; \mu, K)
  \end{split}\\
  \begin{split}
      U_{tt}(t;\; \mu, K) 
      & = \frac{\left( \left(K^2-3 K+2\right) \mu ^2-2 (K-1) \mu t+t^2\right)}{\mu ^2 t^2} \times
  \end{split}\nonumber\\
  \begin{split}
       & \phantom{=}  \quad U(t;\; \mu, K),
  \end{split}
\end{align}
see the second row in Figure~\ref{fig-temp-kernels-1D} for graphs.
The $L_1$-norms of these kernels are given by
\begin{align}
  \begin{split}
     \label{eq-L1norm-temp1der-uni-distr}
      & \| U_t(\cdot;\; \mu, K) \|_1
      = \frac{2 e^{1-K} (K-1)^{K-1}}{\mu  \, \Gamma (K)},
  \end{split}\\
  \begin{split}
     \label{eq-L1norm-temp2der-uni-distr}
      & \| U_{tt}(\cdot;\; \mu, K) \|_1
       = 2 e^{-K-\sqrt{K-1}+1} \times
 \end{split}\nonumber\\
 \begin{split}
        &  \quad   \left(
                e^{2 \sqrt{K-1}} \left(K+2 \sqrt{K-1}\right) \left(K-\sqrt{K-1}-1\right)^K +
             \right.
  \end{split}\nonumber\\
 \begin{split}
     &  \quad\quad \left.
                 \left(K-2 \sqrt{K-1}\right) \left(K+\sqrt{K-1}-1\right)^K
              \right)/
 \end{split}\nonumber\\
 \begin{split}
     & \quad \left( (K-2)^2 \sqrt{K-1} \mu ^2 \, \Gamma (K) \right).
  \end{split}
\end{align}
The temporal scale level at level $K$ corresponds to temporal variance
$\tau = K \mu^2$ and temporal standard deviation $\sigma = \sqrt{\tau} = \mu \sqrt{K}$.

\subsection{Temporal peak}
\label{sec-scsel-temp-blob-uni-distr}

Consider an input signal defined as a time-causal temporal peak corresponding
to filtering a delta function with $K_0$ first-order integrators with
time constants $\mu$ coupled in cascade:
\begin{equation}
 \label{eq-peak-model-uni-distr}
  f(t)
  =  \frac{t^{K_0-1} \, e^{-t/\mu}}{\mu^{K_0} \, \Gamma(K_0)} 
  = U(t;\; \mu, K_0).
\end{equation}
With regard to the application area of vision, this signal can be seen
as an idealized model of an object with
temporal duration $\tau_0 = K_0 \, \mu^2$ that
first appears and then disappears from the field of view, and modelled on a
form to be algebraically compatible with the algebra of the
temporal receptive fields.
With respect to the application area of hearing, this signal can be
seen as an idealized model of a beat sound over some frequency range
of the spectrogram, also modelled on a form to be compatible with the
algebra of the temporal receptive fields.

\begin{table*}[!hbt]
  \addtolength{\tabcolsep}{1pt}
  \begin{center}
   \footnotesize
  \begin{tabular}{ccccc}
  \hline
   \multicolumn{5}{c}{Scale estimate $\hat{K}$ and maximum magnitude
    $L_{\zeta\zeta,max}$ from temporal peak (uniform distr)} \\
  \hline
    $K_0$ 
       & $\hat{K}$ (var, $\gamma=3/4$) 
       & $\left. L_{\zeta\zeta,maxmagn,postnorm} \right|_{\gamma=1}$ (var, $\gamma=3/4$) 
       & $\hat{K}$ (var, $\gamma=1$) 
       & $\hat{K}$ ($L_p$, $p=1$) \\
  \hline
     4 &   3.1 & 0.504 &     6.1 &   10.3 \\
     8 &   7.1 & 0.502 &   14.1 &   18.3 \\
   16 & 15.1 & 0.501 &  30.1 &   34.3 \\
   32 & 31.1 & 0.500 &  62.1 &   66.3 \\
   64 & 63.1 & 0.500 & 126.1 & 130.3\\
\hline
  \end{tabular}
\end{center}
\caption{Numerical estimates of the value of $\hat{K}$ at which the
  scale-normalized second-order temporal derivative
  assumes its maximum over temporal scale for a {\em temporal peak\/} (with the discrete
  expression over discrete temporal scales extended to a continuous
  variation) as function of $K_0$ and for either 
  (i)~variance-based normalization with $\gamma = 3/4$,
  (iii)~variance-based normalization with $\gamma = 1$ and
 (iv)~$L_p$-normalization with $p = 1$.
 For the case of variance-based normalization with $\gamma = 3/4$, (ii)~the
post-normalized magnitude measure $\left. L_{\zeta\zeta,maxmagn,postnorm} \right|_{\gamma=1}$ according to
(\protect\ref{eq-sc-sel-post-norm-temp-peak-scale-calib}) and at the corresponding
scale (i) is also shown. Note that the temporal scale estimates
$\hat{K}$ do for $\gamma = 3/4$ constitute a good approximation of the temporal scale
$\hat{K_0}$ of the underlying structure and that the maximum magnitude
estimates obtained at this temporal scale do for $\gamma = 1$
constitute a good approximation to a scale-invariant constant
maximum magnitude measure over temporal scales.}
  \label{tab-sc-est-temp-peak-2nd-der-uni-distr}

\medskip

  \addtolength{\tabcolsep}{1pt}
  \begin{center}
   \footnotesize
  \begin{tabular}{ccccc}
  \hline
   \multicolumn{5}{c}{Scale estimate $\hat{K}$ and maximum magnitude
    $L_{\zeta,max}$ from temporal ramp (uniform distr)} \\
  \hline
    $K_0$ 
       & $\hat{K}$ (var, $\gamma=1/2$) 
       & $\hat{K}$ ($L_p$, $p=2/3$) 
       & $L_{\zeta,max}$ (var, $\gamma=1$) 
       & $L_{\zeta,max}$ ($L_p$, $p=1$) \\
  \hline
     4 &   3.2 &  3.6 & 0.282  & 0.254 \\
     8 &   7.2 &  7.7 & 0.282  & 0.272 \\
   16 & 15.2 & 15.8 & 0.282 & 0.277 \\
   32 & 31.2 & 31.8 & 0.282 & 0.279 \\
   64 & 63.2 & 64.0 & 0.282 & 0.281 \\
\hline
  \end{tabular}
\end{center}
\caption{(columns 2-3) Numerical estimates of the value of $\hat{K}$ at which the
  scale-normalized first-order temporal derivative
  assumes its maximum over temporal scale for a {\em temporal onset ramp\/} (with the discrete
  expression over discrete temporal scales extended to a continuous
  variation) as function of $K_0$ and for 
  (i)~variance-based normalization for $\gamma = 1/2$ and
  (ii)~$L_p$-normalization for $p = 2/3$.
  (columns 4-5) Maximum magnitude values $L_{\zeta,max}$ at the
  corresponding temporal scales, with
the magnitude values defined by
  (iii)~variance-based normalization for $\gamma = 1$ and
  (iv)~$L_p$-normalization for $p = ~1$. Note that for $\gamma = 1/2$
  as well as for $p = 2/3$ the temporal scale estimates $\hat{K}$
  constitute a good approximation of the temporal scale $K_0$ of
  the underlying onset ramp as well as that the scale-normalized
  maximum magnitude estimates $L_{\zeta,max}$ computed for $\gamma =
  1$ and $p = 1$ constitute a good approximation to a scale-invariant
  constant magnitude measure over temporal scales.}
  \label{tab-sc-est-temp-ramp-1st-der-uni-distr}
\end{table*}

Define the temporal scale-space representation by convolving this
signal with the temporal scale-space kernel
(\ref{eq-peak-model-uni-distr}) corresponding to $K$ first-order
integrators having the same time constants $\mu$
\begin{align}
   \begin{split}
      L(t;\; \mu, K) 
      & = (U(\cdot;\; \mu, K) * f(\cdot))(t;\; \mu, K)
  \end{split}\nonumber\\
  \begin{split}
    \label{eq-temp-scsp-peak-uni-distr}
     & = \frac{e^{-\frac{t}{\mu }} \mu ^{-K-K_0} t^{K+K_0-1}}{\Gamma (K+K_0)} 
        = U(t;\; \mu, K_0 + K) 
  \end{split}
\end{align}
where we have applied the semi-group property that follows immediately
from the corresponding Laplace transforms
\begin{align}
   \begin{split}
      \bar{L}(q;\; \mu, K) 
      & = \frac{1}{(1 + \mu q)^K} \, \frac{1}{(1 + \mu q)^{K_0}}
         = \frac{1}{(1 + \mu q)^{K_0+K}} 
 \end{split}\nonumber\\
  \begin{split}
     \label{eq-semi-group-time-caus-equal-time-const-Lapl-transform}
     & = \bar{U}(q;\; \mu, K_0 + K).
  \end{split}
\end{align}
By differentiating the temporal scale-space representation
(\ref{eq-temp-scsp-peak-uni-distr}) with respect to time $t$ we
obtain
\begin{align}
  \begin{split}
     \label{eq-temp1der-scsp-peak-uni-distr}
      L_t(t;\; \mu, K) 
     & = \frac{(\mu  (K+K_0-1)-t)}{\mu  t} \, L(t;\; \mu, K) 
  \end{split}\\
  \begin{split}
      L_{tt}(t;\; \mu, K) 
      & = \left( 
                 \mu ^2 \left(K^2+K (2 K_0-3)+K_0^2-3 K_0+2\right)
             \right.
  \end{split}\nonumber\\
  \begin{split}
     \label{eq-temp2der-scsp-peak-uni-distr}
         & \phantom{=} \quad
             \left.
                 -2 \mu t (K+K_0-1)+t^2
             \right)
            \frac{L(t;\; \mu, K)}{\mu ^2 t^2}
  \end{split}
\end{align}
implying that the maximum point is assumed at
\begin{equation}
  \label{eq-tmax-scsp-peak-uni-distr}
  t_{max} = \mu  (K+K_0-1)
\end{equation}
and the inflection points at
\begin{align}
  \begin{split}
    t_{inflect1} & = \mu  \left(K+K_0-1 -\sqrt{K+K_0-1}\right),
  \end{split}\\
 \begin{split}
    t_{inflect2} & = \mu  \left(K+K_0-1+\sqrt{K+K_0-1}\right).
  \end{split}
\end{align}
This form of the expression for the time of the temporal maximum
implies that the temporal delay of the underlying peak
$t_{max,0} = \mu(K_0 - 1)$ and the temporal delay of the temporal
scale-space kernel $t_{max,U} = \mu(K - 1)$ are not fully additive,
but instead composed according to
\begin{equation}
  t_{max} = t_{max,0} + t_{max,U} + \mu.
\end{equation}
If we define the temporal duration $d$ of the peak as the distance between the
inflection points, if furthermore follows that this temporal duration is related
to the temporal duration $d_0 = 2 \mu \sqrt{K_0-1}$ of the original peak and
the temporal duration $d_U = 2 \mu \sqrt{K-1}$ of the temporal scale-space kernel
according to
\begin{align}
  \begin{split}
     d 
     & = t_{inflect2} - t_{inflect1} = 2 \mu \sqrt{K+K_0-1}
  \end{split}\nonumber\\
  \begin{split}
     = \sqrt{d_0^2 + d_U^2 + 4\mu^2}.
  \end{split}
\end{align}
Notably these expressions are not scale invariant, but instead
strongly dependent on a preferred temporal scale as defined by the time
constant $\mu$ of the primitive first-order integrators that define
the uniform distribution of the temporal scales.

\paragraph{Scale-normalized temporal derivatives.}

When using temporal scale normalization by variance-based
normalization, the first- and second-order scale-normalized
derivatives are given by
\begin{align}
  \begin{split}
     \label{eq-scnorm-temp1der-scsp-peak-uni-distr}
    L_{\zeta}(t;\; \mu, K) 
   & = \sigma^{\gamma} \, L_t(t;\; \mu, K) = (\mu \sqrt{K})^{\gamma} \, L_t(t;\; \mu, K)
  \end{split}\\
  \begin{split}
     \label{eq-scnorm-temp2der-scsp-peak-uni-distr}
   L_{\zeta\zeta}(t;\; \mu, K) 
   & = \sigma^{2\gamma} \, L_{tt}(t;\; \mu, K) = (\mu^2 K)^{\gamma} \, L_{tt}(t;\; \mu, K)
   \end{split}
\end{align}
where $\sigma = \sqrt{\tau}$, $\tau = K \mu^2$ and with 
$L_t(t;\; \mu, K)$ and $L_{tt}(t;\; \mu, K)$ according to
(\ref{eq-temp1der-scsp-peak-uni-distr}) and
     (\ref{eq-temp2der-scsp-peak-uni-distr}).

When using temporal scale normalization by $L_p$-normal\-ization, the
first- and second-order scale-normalized derivatives are on the other
hand given by (Lindeberg \cite[Equation (75)]{Lin16-JMIV})
\begin{align}
  \begin{split}
    L_{\zeta}(t;\; \mu, K) 
   & = \alpha_{1,p}(\mu, K) \, L_t(t;\; \mu, K) 
  \end{split}\\
  \begin{split}
   L_{\zeta\zeta}(t;\; \mu, K) 
   & = \alpha_{2,p}(\mu, K) \, L_{tt}(t;\; \mu, K) 
   \end{split}
\end{align}
with the scale-normalization factors $\alpha_{n,p}(\mu, K)$ determined
such that the $L_p$-norm of the scale-normalized temporal derivative
computation kernel
\begin{equation} 
  L_{\zeta^n}(\cdot;\; \mu, K)  = \alpha_{n,p}(\mu, K) \,
  h_{t^n}(\cdot;\; \mu, K)
\end{equation}
equals the
$L_o$-norm of some other reference kernel, where we here take the
$L_p$-norm of the corresponding Gaussian derivative kernels (Lindeberg \cite[Equation (76)]{Lin16-JMIV})
\begin{align}
  \begin{split}
     \| \alpha_{n,p}(\mu, K) \, h_{t^n}(\cdot;\; \mu, K) \|_p 
     & = \alpha_{n,p}(\mu, K) \, \| h_{t^n}(\cdot;\; \mu, K) \|_p 
  \end{split}\nonumber\\
  \begin{split}
     \label{eq-sc-norm-der-Lp-norm-2}
     & = \| g_{\xi^n}(\cdot;\; \tau) \|_p = G_{n,p}
  \end{split}
\end{align}
for $\tau = \mu^2 K$, thus implying
\begin{align}
  \begin{split}
    L_{\zeta}(t;\; \mu, K) 
   & = \frac{G_{1,p}}{\| U_t(\cdot;\; \mu, K) \|_p} \, L_t(t;\; \mu, K) 
  \end{split}\\
  \begin{split}
   L_{\zeta\zeta}(t;\; \mu, K) 
   & = \frac{G_{2,p}}{\| U_{tt}(\cdot;\; \mu, K) \|_p} \, L_{tt}(t;\; \mu, K) 
   \end{split}
\end{align}
where $G_{1,p}$ and $G_{2,p}$ denote the $L_p$-norms (\ref{eq-Lp-norm-gauss-ders})
of corresponding Gaussian derivative kernels for the value of $\gamma$ at which they
become constant over scales by $L_p$-normalization, and the
$L_p$-norms $\| U_t(\cdot;\; \mu, K) \|_p $ and $\| U_{tt}(\cdot;\;   \mu, K) \|_p$
of the temporal scale-space kernels $U_t$ and $U_{tt}$ 
for the specific case of $p = 1$ are given by (\ref{eq-L1norm-temp1der-uni-distr}) and
(\ref{eq-L1norm-temp2der-uni-distr}).

\paragraph{Temporal scale selection.} 

Let us assume that we want to register that a new object has appeared
by a scale-space extremum of the scale-normalized second-order
derivative response. 

To determine the temporal moment at which the temporal event occurs, we should
formally determine the time where 
$\partial_{\tau}(L_{\zeta\zeta}(t;\; \mu, K)) = 0$, 
which by
our model (\ref{eq-scnorm-temp2der-scsp-peak-uni-distr})
would correspond to
solving a third-order algebraic equation.
To simplify the problem, let us instead approximate the temporal
position of the peak in the second-order derivative by the temporal
position of the peak $t_{max}$ according to
(\ref{eq-tmax-scsp-peak-uni-distr}) in the signal 
and study the evolution properties over scale $K$ of
\begin{equation}
   L_{\zeta\zeta}(t_{max};\; \mu, K) = L_{\zeta\zeta}(\mu  (K+K_0-1);\; \mu, K).
\end{equation}
In the case of variance-based normalization for a general value of
$\gamma$, we have
\begin{align}
  \begin{split}
    & 
    L_{\zeta\zeta}(\mu  (K+K_0-1);\; \mu, K) 
  \end{split}\nonumber\\
  \begin{split}
        \label{eq-Lttnorm-timecausal-peak}
    & = -\frac{K^{\gamma } \mu ^{2 \gamma -3} e^{-K-K_0+1} (K+K_0-1)^{K+K_0-2}}
                     {\Gamma (K+K_0)}
  \end{split}
\end{align}
and in the case of $L_p$-normalization for $p = 1$
\begin{align}
  \begin{split}
     & 
     L_{\zeta\zeta}(\mu  (K+K_0-1);\; \mu, K) =
  \end{split}\nonumber\\
  \begin{split}
     & = -C (K-2)^2 \sqrt{K-1} e^{\sqrt{K-1}-K_0} \Gamma (K) 
             \mu ^{-K-K_0-1} \times
  \end{split}\nonumber\\
  \begin{split}
     & \phantom{=} \quad
        (\mu (K+K_0-1))^{K+K_0} / (K+K_0-1)^2 \Gamma (K+K_0)/
  \end{split}\nonumber\\
  \begin{split}
     & \phantom{=} \quad
         2 \left(
             e^{2 \sqrt{K-1}} \left(K+2 \sqrt{K-1}\right) \left(K-\sqrt{K-1}-1\right)^K
         \right.
  \end{split}\nonumber\\
  \begin{split}
     & \phantom{=} \quad \phantom{2 \left( \vphantom{\sqrt{K-1}^K} \right.}
         \left.
           +\left(K-2 \sqrt{K-1}\right) \left(K+\sqrt{K-1}-1\right)^K
        \right).
  \end{split}
\end{align}
To determine the scale $\hat{K}$ at which the local maximum is
assumed, let us temporarily extend this definition to continuous
values of $K$ and differentiate the corresponding expressions with
respect to $K$. Solving the equation
\begin{equation}
   \partial_K(L_{\zeta\zeta}(\mu  (K+K_0-1);\; \mu, K) = 0
\end{equation}
numerically for different values of $K_0$ then gives the 
dependency on the scale estimate $\hat{K}$ as function of $K_0$ 
shown in Table~\ref{tab-sc-est-temp-peak-2nd-der-uni-distr}
for variance-based normalization with either $\gamma = 3/4$ 
or $\gamma = 1$ and $L_p$-normalization for $p = 1$.

As can be seen from the results in
Table~\ref{tab-sc-est-temp-peak-2nd-der-uni-distr}, 
when using variance-based scale
normalization for $\gamma = 3/4$, the scale estimate $\hat{K}$ closely
follows the scale $K_0$ of the temporal peak and does therefore imply
a good approximate transfer of the scale selection property (\ref{eq-sc-est-temp-peak})
to this temporal scale-space concept.
If one would instead use variance-based normalization for $\gamma = 1$
or $L_p$-normalization for $p = 1$, then that would, however, lead to
substantial overestimates of the temporal duration of the peak.

Furthermore, if we additionally normalize the input signal to having unit contrast,
then the corresponding time-causal correspondence to the
post-normalized magnitude measure
(\protect\ref{eq-sc-sel-post-norm-temp-peak-scale-calib}) 
\begin{align}
  \begin{split}
    & \left. L_{\zeta\zeta,maxmagn,postnorm} \right|_{\gamma=1}
  \end{split}\nonumber\\
  \begin{split}
    & = \tau^{1/4} \left. L_{\zeta\zeta,maxmagn} \right|_{\gamma=3/4}
       =  \frac{K}{K+K_0-1}
  \end{split}
\end{align}
is for scale estimates proportional to the temporal duration of the
underlying temporal peak $\hat{K} \sim K_0$ very close to
constant under variations of the temporal duration of the underlying temporal peak as
determined by the parameter $K_0$,
thus implying a good approximate transfer
of the scale selection property
(\ref{eq-sc-sel-post-norm-temp-peak-const-magn}).

\subsection{Temporal onset ramp}
\label{sec-scsel-onset-ramp-uni-distr}

Consider an input signal defined as a time-causal onset ramp corresponding
to the primitive function of $K_0$ first-order integrators with time
constants $\mu$ coupled in cascade:
\begin{equation}
 \label{eq-ramp-model-uni-distr}
  f(t)
  =  \int_{u = 0}^t
         \frac{u^{K_0-1} \, e^{-u/\mu}}{\mu^{K_0} \, \Gamma(K_0)} \, du
  = \int_{u = 0}^t U(u;\; \mu, K_0) \, du.
\end{equation}
With respect to the application area of vision,
this signal can be seen as an idealized model of a new object with
temporal diffuseness $\tau_0 = K_0 \, \mu^2$ that appears in the field of
view and modelled on a form to be algebraically compatible with the
algebra of the temporal receptive fields.
With respect to the application area of hearing, this signal can be
seen as an idealized model of the onset of a new sound in some
frequency band of the spectrogram, also modelled on a form to be
compatible with the algebra of the temporal receptive fields.

Define the temporal scale-space representation of the signal by
convolution with the temporal scale-space kernel
(\ref{eq-peak-model-uni-distr}) corresponding to $K$ first-order
integrators having the same time constants $\mu$
\begin{align}
   \begin{split}
      L(t;\; \mu, K) 
      & = (U(\cdot;\; \mu, K) * f(\cdot))(t;\; \mu, K)
  \end{split}\nonumber\\
  \begin{split}
    \label{eq-temp-scsp-ramp-uni-distr}
        & = \int_{u = 0}^t U(t;\; \mu, K_0 + K) \, du.
  \end{split}
\end{align}
Then, the first-order temporal derivative is given by
\begin{align}
  \begin{split}
   \label{eq-first-order-temp-der-time-caus-onset-ramp}
  L_t(t;\; \mu, K) 
  = U(t;\; \mu, K_0 + K)
  = \frac{t^{K_0+K-1} \, e^{-t/\mu}}{\mu^{K_0+K} \, \Gamma(K_0+K)}
  \end{split}
\end{align}
which assumes its temporal maximum at $t_{ramp} = \mu (K_0 + K -1)$.

\paragraph{Temporal scale selection.}

Let us assume that we are going to detect a new appearing object from
a local maximum in the first-order derivative over both time and
temporal scales. When using variance-based normalization for a general
value of $\gamma$, the scale-normalized response at the temporal
maximum in the first-order derivative is given by
\begin{align}
  \begin{split}
    L_{\zeta,max} 
    & = L_{\zeta}(\mu (K_0 + K -1);\; \mu, K) 
  \end{split}\nonumber\\
  \begin{split}
    & = \sigma^{\gamma} L_t(\mu (K_0 + K -1);\; \mu, K) 
  \end{split}\nonumber\\
  \begin{split}
    \label{eq-Ltnorm-timecausal-ramp}
    & = \frac{\left(\sqrt{K} \mu \right)^{\gamma } (K+K_0-1)^{K+K_0-1}  e^{-K-K_0+1}}
                  {\mu \, \Gamma (K+K_0)}.
  \end{split}
\end{align}
When using $L_p$-normalization for a general value of $p$, the
corresponding scale-normalized response is
\begin{align}
  \begin{split}
     L_{\zeta,max} 
    & = L_{\zeta}(\mu (K_0 + K -1);\; \mu, K) 
  \end{split}\nonumber\\
  \begin{split}
     & = \frac{G_{1,p}}{\| U_t(\cdot;\; \mu, K) \|_p} L_t(\mu (K_0 + K -1);\; \mu, K) 
  \end{split}
\end{align}
where the $L_p$-norm of the first-order scale-space derivative kernel
can be expressed in terms of exponential functions, the Gamma function
and hypergeometric functions, but is too complex to be written out here.
Extending the definition of these expressions to continuous values
of $K$ and solving the equation
\begin{equation}
   \partial_K(L_{\zeta}(\mu  (K+K_0-1);\; \mu, K) = 0
\end{equation}
numerically for different values of $K_0$ then gives the 
dependency on the scale estimate $\hat{K}$ as function of $K_0$ 
shown in Table~\ref{tab-sc-est-temp-ramp-1st-der-uni-distr}
for variance-based normalization with $\gamma = 1/2$ 
or $L_p$-normalization for $p = 2/3$.

As can be seen from the numerical results, for both variance-based
normalization and $L_p$-normalization with corresponding values of
$\gamma$ and $p$, the numerical scale estimates in terms of $\hat{K}$ 
closely follow the diffuseness scale of the temporal ramp as
parameterized by $K_0$. Thus, for both of these scale normalization models,
the numerical results indicate an approximate transfer of the scale selection property
(\ref{eq-sc-est-temp-peak}) to this temporal scale-space model.
Additionally, the maximum magnitude values according to
(\ref{eq-Ltnorm-timecausal-ramp}) can according to Stirling's formula
$\Gamma(n+1) \approx (n/e)^n \sqrt{2 \pi n}$ be approximated by
\begin{align}
  \begin{split}
    L_{\zeta,max}
     \approx \frac{\sqrt{K}}{\sqrt{2 \pi } \sqrt{K+K_0-1}}
  \end{split}
\end{align}
and are very
stable under variations of the diffuseness scale $K_0$ of the ramp,
and thus implying a good transfer of the scale selection property
(\ref{eq-sc-sel-onset-ramp-const-magn-after-post-norm})
to this temporal scale-space concept.

\begin{figure*}[hbt]
  \begin{center}
    \begin{tabular}{cc}
      {\small\em scale estimate
      $\hat{\sigma}(\lambda_0)\left|_{n = 1, \gamma=1/2} \, \right.$} 
      & {\small\em max magnitude
        $L_{\zeta,ampl,max}(\lambda_0)  \left|_{\gamma=1} \, \right.$}  \\
      \includegraphics[width=0.35\textwidth]{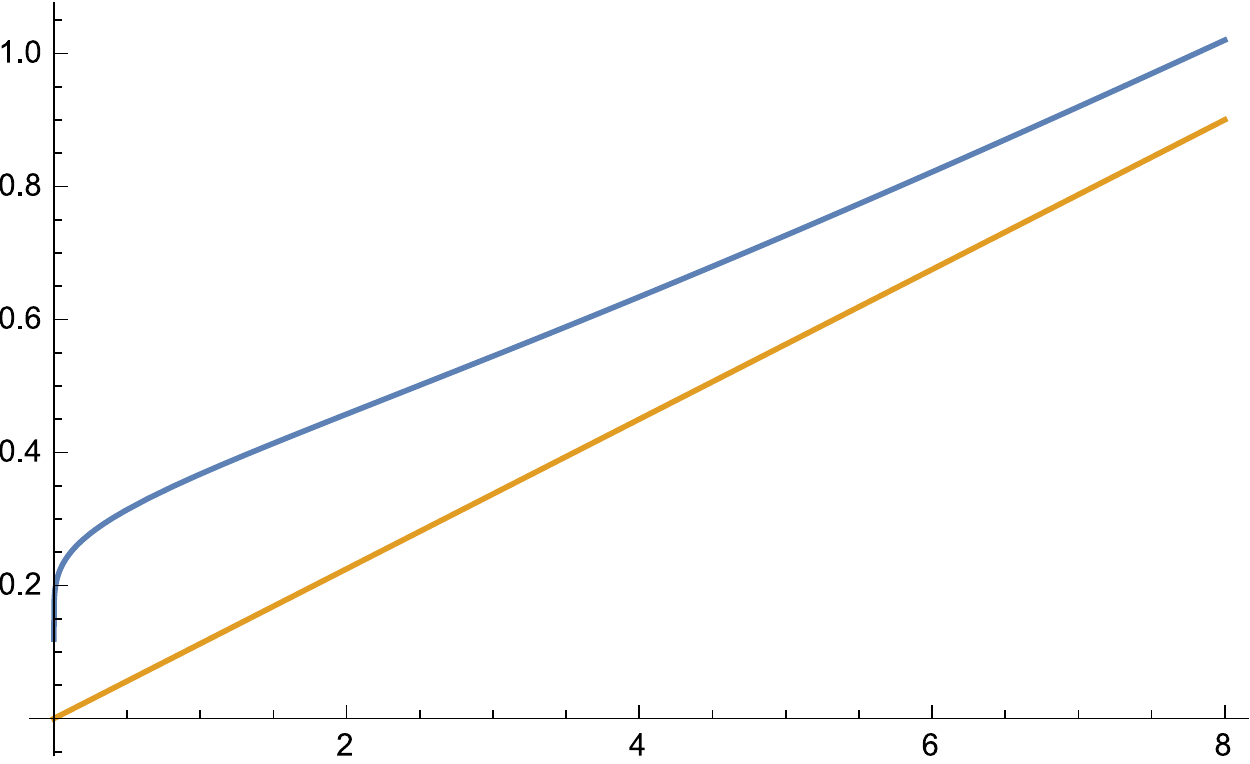} &
      \includegraphics[width=0.35\textwidth]{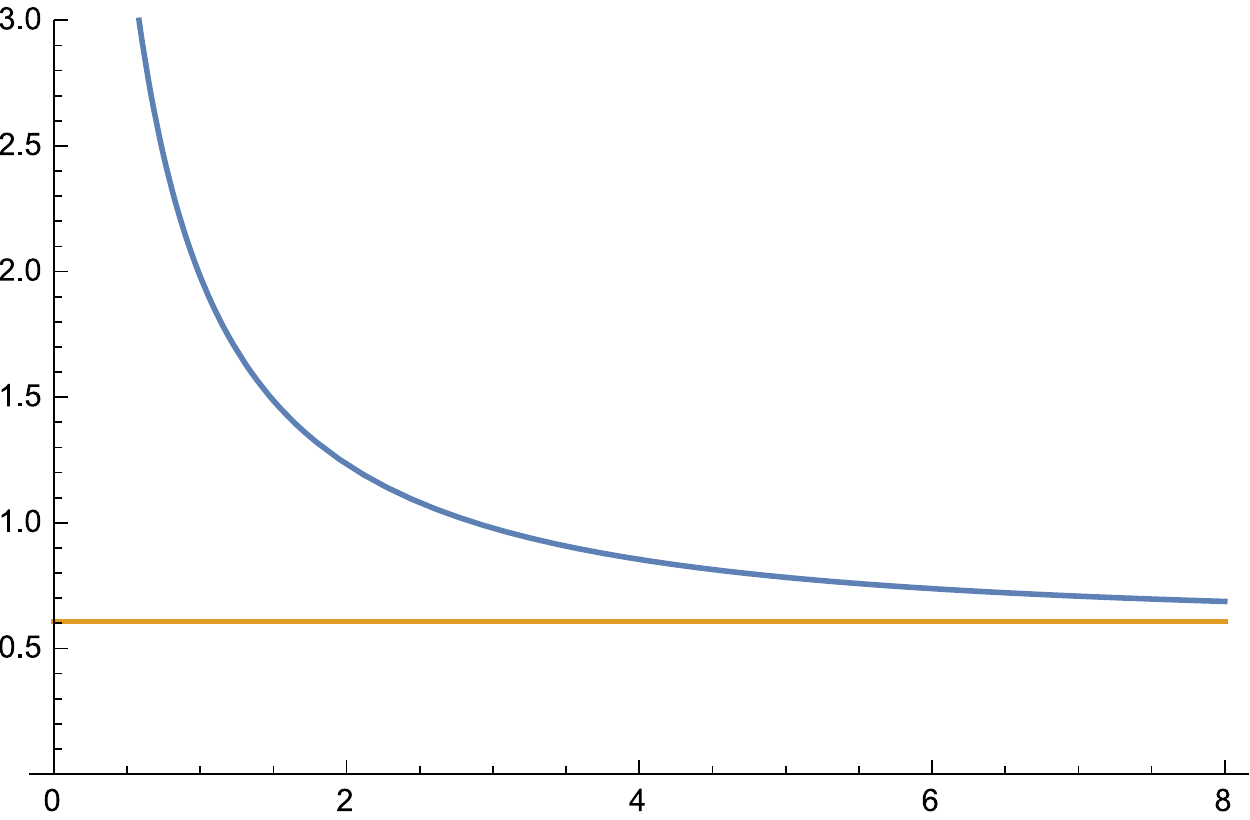} \\
      $\,$\\
      {\small\em scale estimate
      $\hat{\sigma}(\lambda_0)\left|_{n = 2, \gamma=3/4} \, \right.$} 
      & {\small\em max magnitude
        $L_{\zeta\zeta,ampl,max}(\lambda_0)  \left|_{\gamma=1} \, \right.$}  \\
      \includegraphics[width=0.35\textwidth]{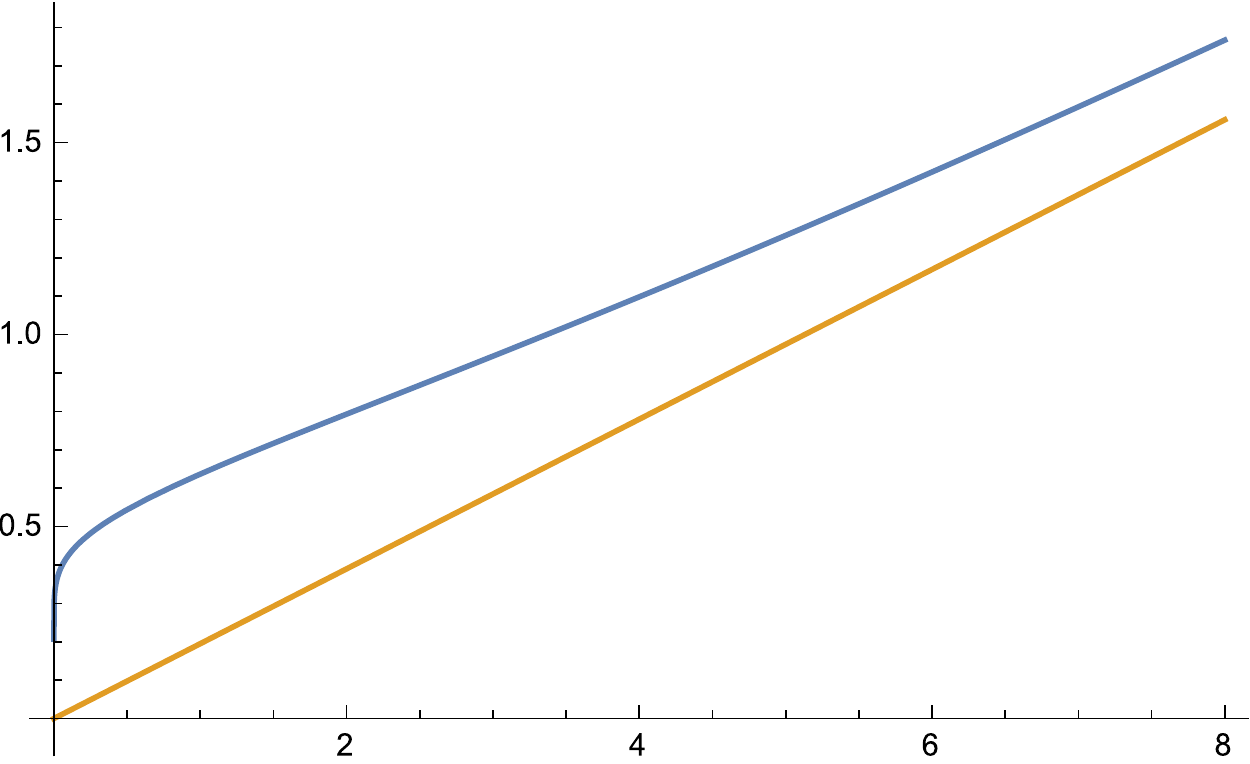} &
      \includegraphics[width=0.35\textwidth]{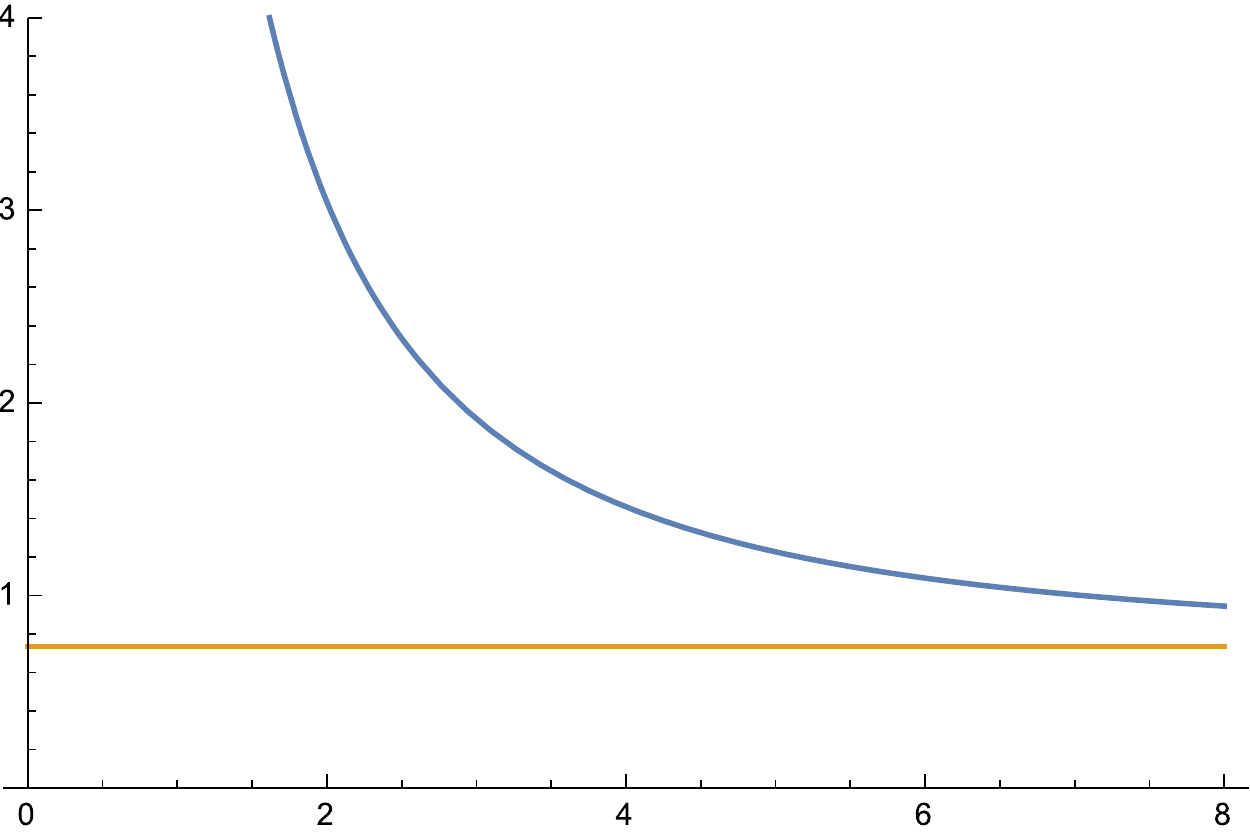} \\
     \end{tabular} 
  \end{center}
  \caption{(left column) Temporal scale estimates $\hat{\sigma}$
    (marked in blue) obtained from a {\em sine wave\/} using the
    maximum value over scale of the amplitude of either (top row) first-order
    scale-normalized temporal derivatives or (bottom row) second-order
    scale-normalized temporal derivatives and as function of
    the wavelength $\lambda_0$ for variance-based temporal scale
    normalization with $\gamma = 1/2$ for first-order derivatives and
    $\gamma = 3/4$ for second-order derivatives. 
    Note that for this temporal scale-space concept based on truncated
    exponential kernels with equal time constants, the temporal scale estimates $\hat{\sigma}$ do not
    constitute a good approximation to the temporal scale estimates
    being proportional to the wavelength $\lambda_0$ of the underlying
    sine wave. Instead, the temporal scale estimates are affected by a
    systematic scale-dependent bias.
    (right column) The maximum scale-normalized magnitude for 
    (top row) first-order scale-normalized temporal derivatives or (bottom row) second-order
    scale-normalized temporal derivatives as function of the
    wavelength $\lambda_0$ using variance-based
    temporal scale normalization with $\gamma = 1$ (marked in blue). 
    The brown curves show corresponding scale estimates and magnitude
    values obtained from the scale-invariant but not time-causal
    Gaussian temporal scale space. Thus, when the temporal scale-space
    concept based on truncated exponential kernels with equal time
    constants coupled in cascade is applied to a sine wave, the maximum magnitude
    measures over temporal scales do not at all constitute a good
    approximation of temporal scale invariance.
    (Horizontal axis: wavelength $\lambda_0$ in units of $\mu$.)}
  \label{fig-sc-est-max-magn-sine-wave-truncexp-uni-distr}
\end{figure*}

\subsection{Temporal sine wave}
\label{sec-temp-sine-wave-temp-scsp-1st-ord-int-eq-time-constants}

Consider a signal defined as a sine wave
\begin{equation}
  \label{eq-sine-model-uni-distr}
  f(t) = \sin \omega_0 t.
\end{equation}
This signal can be seen as a simplified model of a dense temporal
texture with characteristic scale defined as the wavelength 
$\lambda_0 =2\pi/\omega$ of the signal.
In the application area of vision, this can be seen as an idealized
model of watching some oscillating visual phenomena or watching a
dense texture that moves relative the gaze direction.
In the area of hearing, this could be seen as an idealized model of
temporally varying frequencies around some fixed frequency in the
spectrogram corresponding to vibrato.

Define the temporal scale-space representation of the signal by
convolution with the temporal scale-space kernel
(\ref{eq-peak-model-uni-distr}) corresponding to $K$ first-order
integrators with equal time constants $\mu$ coupled in cascade
\begin{align}
   \begin{split}
      L(t;\; \mu, K) 
      & = (U(\cdot;\; \mu, K) * f(\cdot))(t;\; \mu, K)
  \end{split}\nonumber\\
  \begin{split}
    \label{eq-temp-scsp-sine-uni-distr}
         = |\hat{U}(\omega_0;\; \mu, K)| 
            \sin \left( \omega_0 t + \arg \hat{U}(\omega;\; \mu, K) \right)
  \end{split}
\end{align}
where $|\hat{U}(\omega_0;\; \mu, K)|$ and $\arg \hat{U}(\omega;\; \mu, K)$
denote the magnitude and the argument of the Fourier transform $\hat{h}_{composed}(\omega;\; \mu, K)$
of the temporal scale-space kernel $U(\cdot;\; \mu, K)$
according to
\begin{align}
  \begin{split}
    \hat{U}(\omega;\; \mu, K) 
     & = \frac{1}{\left( 1 + i \, \mu \, \omega \right)^K},
 \end{split}\\
  \begin{split}
    |\hat{U}(\omega;\; \mu, K) |
     & = \frac{1}{\left( 1 + \mu^2 \, \omega^2 \right)^{K/2}},
 \end{split}\\
  \begin{split}
    \arg \hat{U}(\omega;\; \mu, K) 
     & = - K \arctan \left( \mu \, \omega \right).
 \end{split}
\end{align}
By differentiating (\ref{eq-temp-scsp-sine-uni-distr}) with respect to
time $t$, it follows that the magnitude of the $n$th order temporal
derivative is given by
\begin{equation}
  L_{t^n,ampl} = \frac{\omega_0^n}{\left( 1 + \mu^2 \, \omega_0^2 \right)^{K/2}}.
\end{equation}

\paragraph{Temporal scale selection.}

Using variance-based temporal scale normalization, the magnitude of
the corresponding scale-normalized temporal derivative is given by
\begin{equation}
  \label{eq-nth-order-temp-der-norm-timecausal-sine}
   L_{\zeta^n,ampl} 
  = \sigma^n L_{t^n,ampl} 
  = \frac{(K \mu^2)^{n \gamma/2} \omega_0^n}{\left( 1 + \mu^2 \, \omega_0^2 \right)^{K/2}}.
\end{equation}
Extending this expression to continuous values of $K$ and
differentiating with respect to $K$ implies that the maximum over
scale is assumed at scale
\begin{equation}
  \hat{K} =  \frac{\gamma  n}{\log \left(1+\mu ^2 \omega_0^2\right)}
\end{equation}
with the following series expansion for small values of $\omega_0$ corresponding
to temporal structures of longer temporal duration
\begin{equation}
\label{eq-sc-est-Khat-sine-wave-truncexp-uni-distr}
  \hat{K} 
  = \frac{\gamma  n}{\mu ^2 \omega_0^2}+\frac{\gamma n}{2}
  -\frac{1}{12} \omega_0^2 \left(\gamma  \mu ^2    n\right)+O\left(\omega_0^4\right).
\end{equation}
Expressing the corresponding scale estimate $\hat{\sigma}$ in terms of
dimension length and parameterized in terms of the wavelength
$\lambda_0 = 2 \pi/\omega_0$ of the sine wave
\begin{align}
  \begin{split}
    \label{eq-sc-est-sine-wave-truncexp-uni-distr}
    \hat{\sigma} 
    & = \sqrt{\hat{\tau}} = \mu \sqrt{\hat{K}} 
      = \mu  \sqrt{\frac{\gamma  n}{\log \left(1+\frac{4 \pi ^2 \mu^2}{\lambda_0^2}\right)}}
  \end{split}\nonumber\\
  \begin{split}
    & = \frac{\sqrt{\gamma  n}}{2 \pi } \lambda_0
          \left(
             1+\frac{\pi ^2 \mu ^2}{\lambda_0^2}+O\left(\left(\frac{\mu}{\lambda_0}\right)^4\right)
          \right)
  \end{split}
\end{align}
we can see that the dominant term $\frac{\sqrt{\gamma  n} \lambda_0 }{2 \pi }$ 
is proportional to the temporal duration of the underlying structures in the
signal and in agreement with the corresponding scale selection
property (\ref{eq-scsel-sine-wave-gauss-temp-scsp-lambda})
of the scale-invariant non-causal Gaussian temporal scale space concept,
whereas the overall expression is not scale invariant.

If the wavelength $\lambda_0$ is much longer than the time constant
$\mu$ of the primitive first-order integrators, then the scale
selection properties in this temporal scale-space model will constitute a 
better approximation of the corresponding scale selection properties in
the scale-invariant non-causal Gaussian temporal scale-space model.

The maximum value over scale is
\begin{equation}
  \label{eq-max-magn-gen-gamma-sine-wave-truncexp-uni-distr}
  L_{\zeta^n,ampl,max} 
  = e^{-\frac{\gamma  n}{2}} \omega_0^n \mu ^{\gamma  n} 
     \left(\frac{\gamma  n}{\log \left(1+\mu ^2 \omega_0^2\right)}\right)^{\frac{\gamma  n}{2}}
\end{equation}
with the following series expansion for large $\lambda_0=2\pi/\omega_0$
and $\gamma = 1$:
\begin{multline}
  \label{eq-max-magn-gamma1-sine-wave-truncexp-uni-distr}
  L_{\zeta^n,ampl,max} 
  = e^{-n/2} n^{n/2} \times
  \\
     \left(
        1+\frac{\pi ^2 \mu ^2 n}{\lambda_0^2}+\frac{\pi ^4 \mu ^4 n (3 n-10)}{6
   \lambda_0^4}+O\left(\left(\frac{\mu}{\lambda_0}\right)^6\right)
     \right).
\end{multline}
Again we can note that the first term agrees with the
corresponding scale selection property
(\ref{eq-scsel-sine-wave-gauss-temp-scsp-max-magn}) for 
the scale-invariant non-causal Gaussian temporal scale space, whereas
the higher order terms are not scale invariant.

Figure~\ref{fig-sc-est-max-magn-sine-wave-truncexp-uni-distr} shows
graphs of the scale estimate $\hat{\sigma}$ according to 
(\ref{eq-sc-est-sine-wave-truncexp-uni-distr}) for $n = 2$ and 
$\gamma = 3/4$ and the maximum response over scale $L_{\zeta^n,ampl,max}$
for $n = 2$ and $\gamma = 1$ as function of the wavelength $\lambda_0$
of the sine wave (marked in blue). For comparison, we also show the corresponding scale
estimates (\ref{eq-scsel-sine-wave-gauss-temp-scsp-lambda})  and
magnitude values (\ref{eq-scsel-sine-wave-gauss-temp-scsp-max-magn})
that would be obtained using temporal scale
selection in the scale-invariant non-causal Gaussian temporal scale space (marked in
brown).

As can be seen from the graphs, both the temporal scale estimate
$\hat{\sigma}(\lambda_0)$ and the maximum magnitude
$L_{\zeta,ampl,max}(\lambda_0)$ obtained from a set of first order integrators
with equal time constants coupled in cascade approach the corresponding results obtained
from the non-causal Gaussian scale space for larger values of
$\lambda_0$ in relation to the time constant $\mu$ of the first-order
integrators. The scale estimate obtained from a set of first-order
integrators with equal time constants is, however, for lower values of
$\lambda_0$ generally significantly higher than the scale estimates obtained from a
non-causal Gaussian temporal scale space. The scale-normalized
magnitude values, which should be constant over scale for a
scale-invariant temporal scale space when $\gamma = 1$
according to the scale selection property
(\ref{eq-scsel-sine-wave-gauss-temp-scsp-max-magn}), 
are for lower values of
$\lambda_0$ much higher than the scale-invariant limit value when
performing scale selection in the temporal scale-space concept
obtained by coupling a set of first-order integrators with equal time
constants in cascade. 
The scale selection 
properties (\ref{eq-scsel-sine-wave-gauss-temp-scsp-lambda}) 
and (\ref{eq-scsel-sine-wave-gauss-temp-scsp-max-magn}) are consequently not transferred
to this temporal scale-space concept for the sine wave model,
which demonstrates the need
for using a scale-invariant temporal scale-space concept when
formulating mechanisms for temporal scale selection.
The scale selection properties of such a scale-invariant time-causal
temporal concepts will be analysed in
Section~\ref{sec-scsel-prop-time-caus-trunc-exp-log-distr}, and
showing that it is possible to obtain temporal scale estimates for a
dense sine wave that are truly proportional to the wavelength of the
signal, {\em i.e.\/}, a characteristic estimate of the temporal
duration of the temporal structures in the signal.

Concerning this theoretical analysis, it should be noted that we have
here for the expressions
(\ref{eq-sc-est-sine-wave-truncexp-uni-distr}),
(\ref{eq-max-magn-gen-gamma-sine-wave-truncexp-uni-distr}) and
(\ref{eq-max-magn-gamma1-sine-wave-truncexp-uni-distr})
disregarded the rounding of the continuous value $\hat{K}$ in 
(\ref{eq-sc-est-Khat-sine-wave-truncexp-uni-distr}) to the nearest
integer upwards or downwards where it assumes its maximum value over
temporal scales.
Thereby, the graphs in
Figure~\ref{fig-sc-est-max-magn-sine-wave-truncexp-uni-distr} may
appear somewhat different if such quantization effects because of
discrete temporal scale levels are also included.
The lack of true temporal scale invariance will, however, still
prevail.

Concerning the motivation to the theoretical analysis in this section, while the purpose
of this analysis has been to investigate how the temporal scale
estimates depend on the frequency or the wavelength of the signal, it
should be emphasized that the primary purpose has not been to develop a
method for only estimating the frequency or the wavelength of a sine
wave. Instead, the primary purpose has been to carry out a closed-form
theoretical analysis of the properties of temporal scale selection
when applied to a model signal for which such closed-form
theoretical analysis can be carried out.
Compared to using {\em e.g.\/} a Fourier transform for estimating
the local frequency content in a signal, it should be noted that the computation of a
Fourier transform requires a complementary parameter --- a window
scale over which the Fourier transform is to be computed. The frequency
  estimate will then be an average of the frequency content over the
  entire interval as defined by the window scale parameter. 
Using the the proposed temporal scale selection methodology
  it is on the other hand possible to estimate the temporal scale
  without using any complementary window scale parameter. Additionally,
  the temporal scale estimate will be instantaneous and not an average over
  multiple cycles of {\em e.g.\/} a periodic signal, see also the later
  experimental results that will be presented in
  Section~\ref{sec-temp-scsel-1D-temp-signal} in particular
  Figure~\ref{fig-expsine-tempscspextr-recfiltlogscvar-discgaussvar}. 

\section{Scale selection properties for the time-causal temporal scale
  space concept based on
  the scale-invariant time-causal limit kernel}
\label{sec-scsel-prop-time-caus-trunc-exp-log-distr}

In this section, we will analyse the scale selection properties
for the time-causal scale-space concept based on convolution with the
scale-invariant time-causal limit kernel.

The analysis starts with a detailed study of a sine wave, for which closed-form
theoretical analysis is possible and showing that the selected
temporal scale level $\hat{\sigma}$ measured in units of 
dimension time $[\mbox{time}]$ according to 
$\hat{\sigma} = \sqrt{\hat{\tau}}$ will be proportional to the wavelength
of the signal, in accordance with true scale invariance. We also show
that despite the discrete nature of the temporal scale levels in this
temporal scale-space concept, local extrema over scale will
nevertheless be preserved under scaling
transformations of the form $\lambda_1 = c^j \lambda_0$, with $c$
denoting the distribution parameter of the time-causal limit kernel.

Then, we present a general result about temporal scale invariance that
holds for temporal derivatives of any order and for any input signal,
showing that under a temporal scaling transformation of the form 
$t' = c^j t$, local extrema over scales are preserved under such
temporal scaling transformations with the temporal scale estimates
transforming according to $\tau' = c^j \tau$. 
We also show that if the scale normalization power $\gamma = 1$
corresponding to $p = 1$, the scale-normalized magnitude responses
will be preserved in accordance with true temporal scale invariance.

\subsection{Time-causal temporal scale space based on
  the scale-invariant time-causal limit kernel}

Given the temporal scale-space model based on truncated exponential
kernels (\ref{eq-trunc-exp-kern-prim}) coupled in cascade
\begin{equation}
  \label{eq-comp-trunc-exp-cascade}
  h_{composed}(\cdot;\; \mu) 
  = *_{k=1}^{K} h_{exp}(\cdot;\; \mu_k),
\end{equation}
having a composed Fourier transform of the form
  \begin{align}
    \begin{split}
       H_{composed}(q;\; \mu) 
       & = \int_{t = - \infty}^{\infty} 
              *_{k=1}^{K} h_{exp}(\cdot;\; \mu_k)(t) 
                 \, e^{-qt} \, dt
   \end{split}\nonumber\\
   \begin{split}
       \label{eq-expr-comp-kern-trunc-exp-filters}
        & =  \prod_{k=1}^{K} \frac{1}{1 + \mu_k q},
    \end{split}
  \end{align}
and as arises from the assumptions of
(i)~linearity, (ii)~temporal shift invariance, (iii)~temporal
causality and (iv)~non-creation of new local extrema or equivalently
zero-crossings with increasing scale,
it is more natural to distribute the temporal scale
levels logarithmically over temporal scales
\begin{equation}
  \label{eq-distr-tau-values}
  \tau_k = c^{2(k-K)} \tau_{max} \quad\quad (1 \leq k \leq K)
\end{equation}
so that the distribution in terms of effective temporal scale
$\tau_{eff} = \log \tau$ (Lindeberg \cite{Lin92-PAMI}) becomes uniform.
This implies that time
constants of the individual first-order integrators should for some 
$c > 1$ be given by (Lindeberg \cite{Lin15-SSVM,Lin16-JMIV})
\begin{align}
  \begin{split}
     \label{eq-mu1-log-distr}
     \mu_1 & = c^{1-K} \sqrt{\tau_{max}},
  \end{split}\\
  \begin{split}
     \label{eq-muk-log-distr}
     \mu_k & = \sqrt{\tau_k - \tau_{k-1}} = c^{k-K-1} \sqrt{c^2-1} \sqrt{\tau_{max}} \quad (2 \leq k \leq K).
  \end{split}
\end{align}
Specifically, if one lets the number of temporal scale levels tend to
infinity with the density of temporal scale levels becoming infinitely
dense towards $\tau \rightarrow 0$, it can be shown that this leads to
a {\em scale-invariant time-causal limit kernel\/} having a Fourier transform of the form
(Lindeberg \cite[Section~5]{Lin16-JMIV})
\begin{align}
  \begin{split}
     \hat{\Psi}(\omega;\; \tau, c) 
     & = \lim_{K \rightarrow \infty} \hat{h}_{exp}(\omega;\; \tau, c, K) 
  \end{split}\nonumber\\
  \begin{split}
     \label{eq-FT-comp-kern-log-distr-limit}
     & = \prod_{k=1}^{\infty} \frac{1}{1 + i \, c^{-k} \sqrt{c^2-1} \sqrt{\tau} \, \omega}.
  \end{split}
\end{align}

\begin{figure*}[hbtp]
  \begin{center}
    \begin{tabular}{cc}
      {\small $n = 1$, $c = \sqrt{2}$} 
      &  {\small $n = 2$, $c = \sqrt{2}$} \\
      \includegraphics[width=0.35\textwidth]{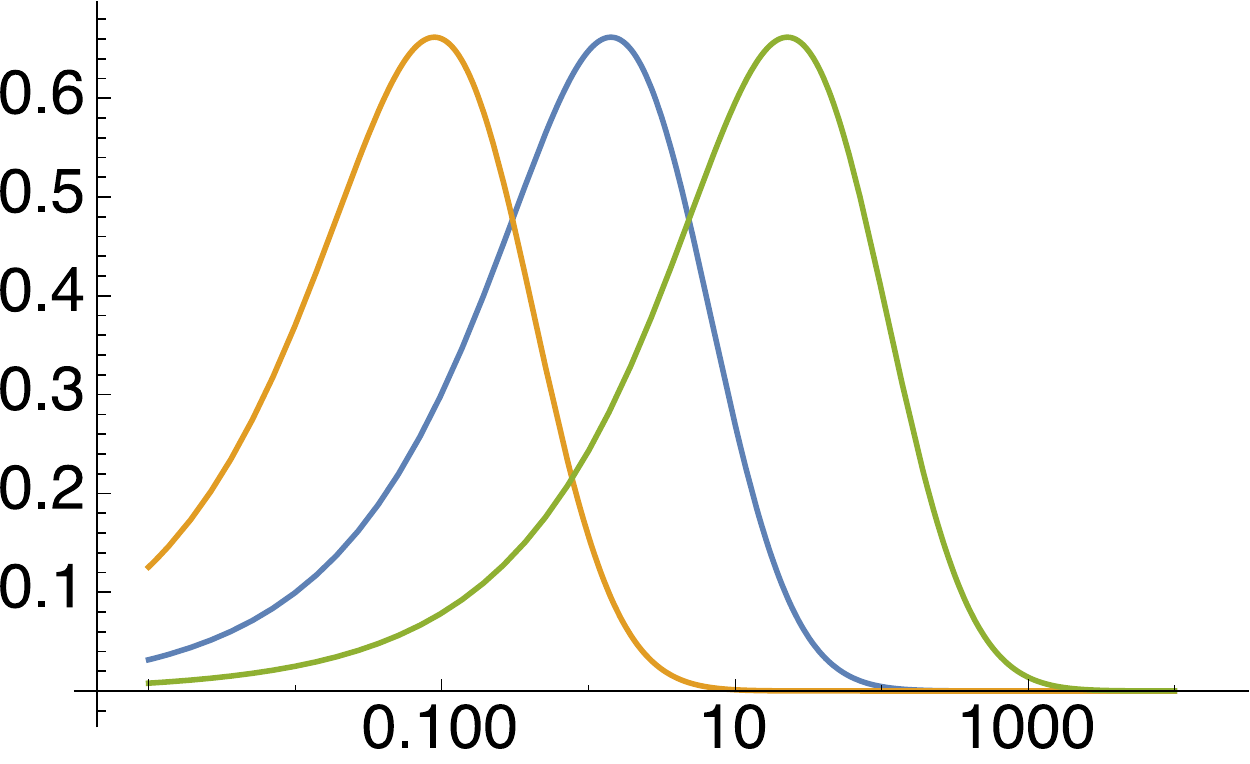} &
      \includegraphics[width=0.35\textwidth]{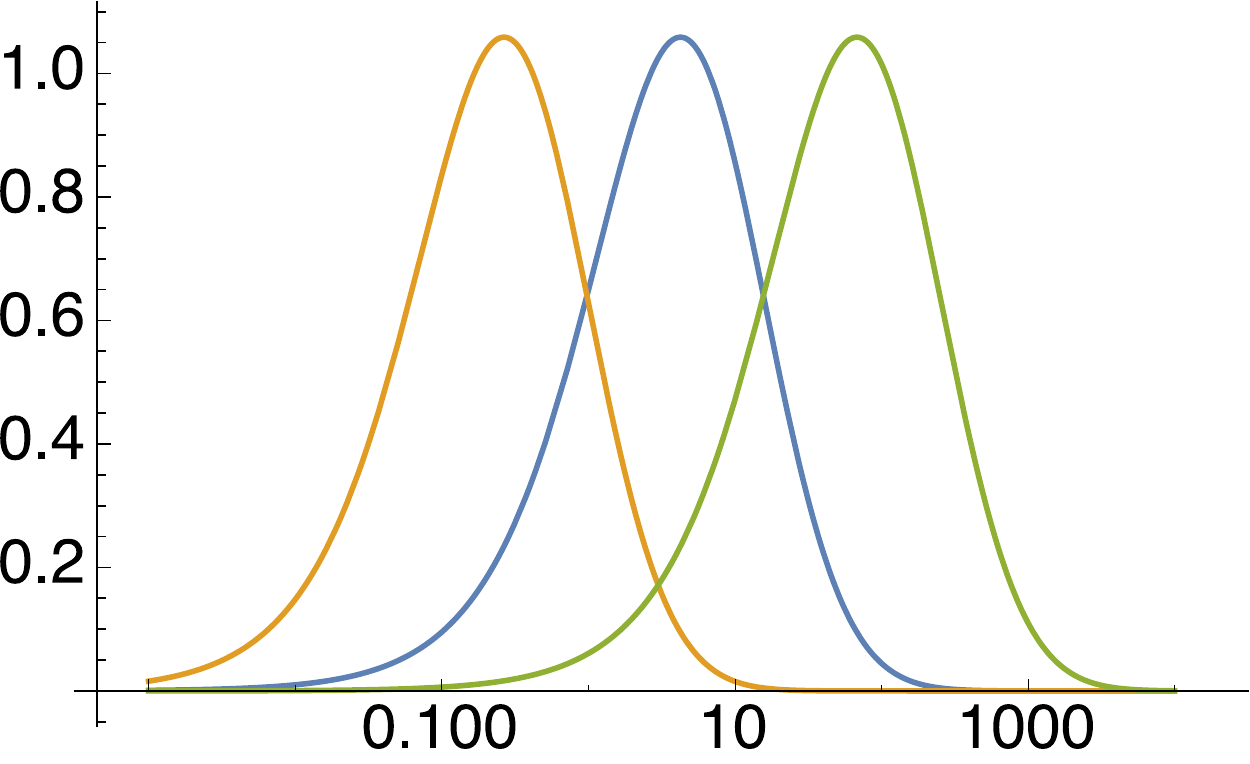} \\
  \\
      {\small $n = 1$, $c = 2$} 
      &  {\small $n = 2$, $c = 2$} \\
      \includegraphics[width=0.35\textwidth]{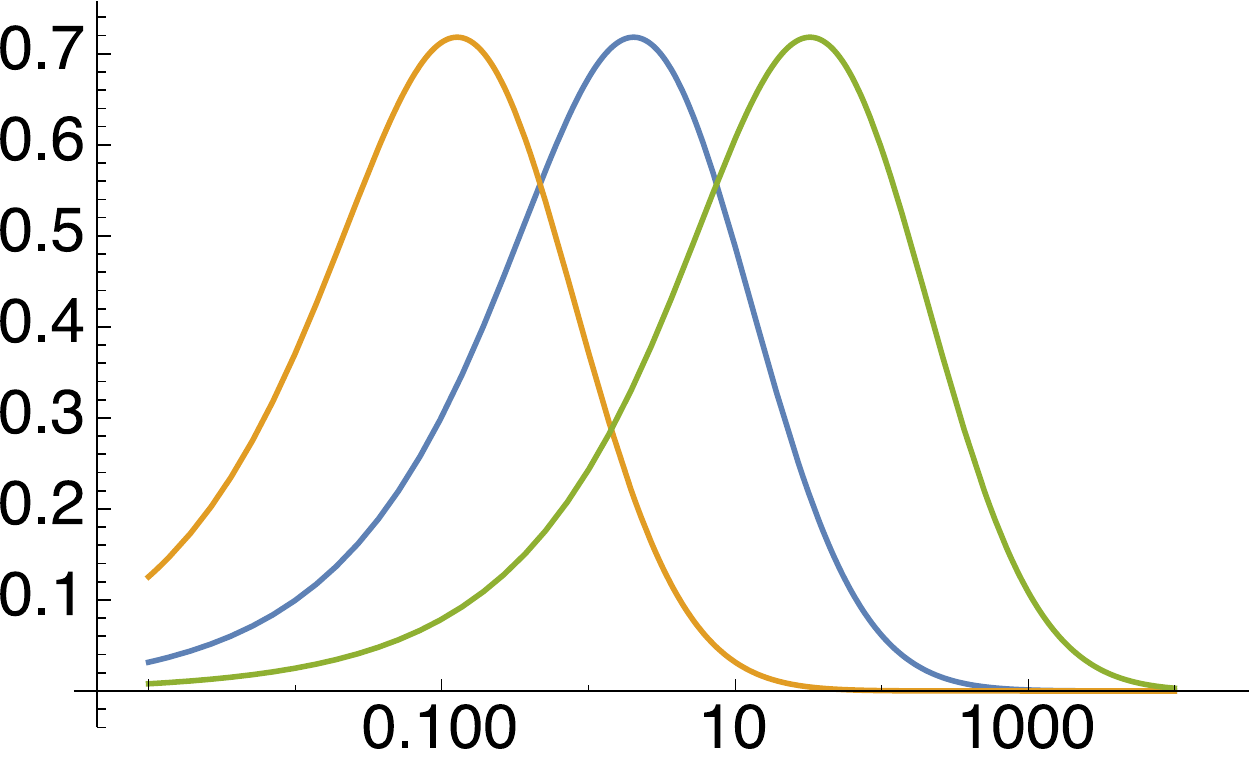} &
      \includegraphics[width=0.35\textwidth]{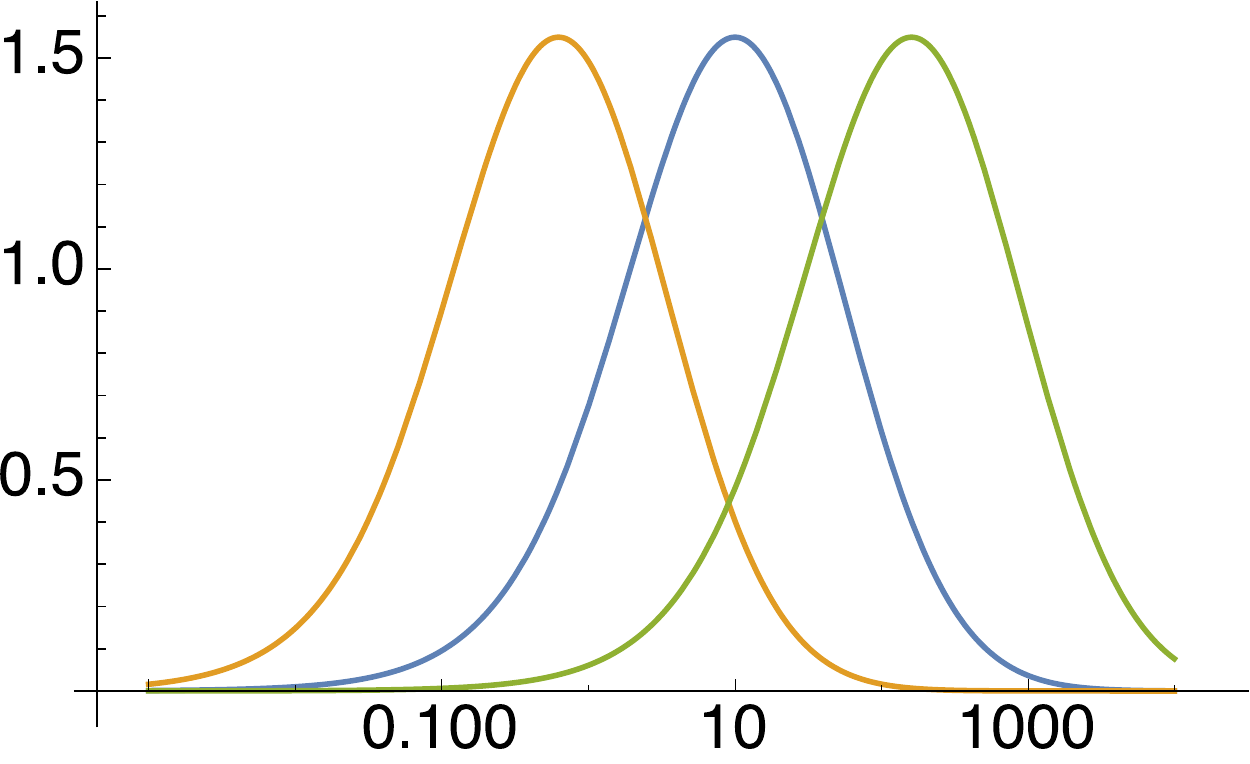} \\
     \end{tabular} 
  \end{center}
   \caption{Scale-space signatures showing the variation over scales
     of the amplitude of scale-normalized temporal derivatives in the
     scale-space representation of {\em sine waves\/} with angular frequencies
   $\omega_0 = 1/4$ (green curves), $\omega_0 = 1$ (blue curves) and 
   $\omega_0 = 4$ (brown curves) under
   convolution with the time-causal limit kernel approximated by
   the slowest $K = 32$ temporal smoothing steps and for $\gamma = 1$
   corresponding to $p = 1$. 
   (left column) First-order temporal derivatives $n = 1$.
   (right column) Second-order temporal derivatives $n = 2$.
   (top row) Distribution parameter $c = \sqrt{2}$.
   (bottom row) Distribution parameter $c = 2$.
   Note how these graphs reflect temporal scale invariance in the
   sense that (i)~a variation in the angular frequency of the underlying
   signal corresponds to a mere shift of the scale-space signature to
   either finer or coarser temporal scales and (ii)~the maximum
   magnitude response is independent of the frequency of the
   underlying signal.
   (Horizontal axis: Temporal scale $\tau$ on a logarithmic axis.)}
  \label{fig-scspsign-sine-waves-limit-kern}

  \begin{center}
    \begin{tabular}{cc}
      {\small $n = 1$, $c = \sqrt{2}$} 
      &  {\small $n = 2$, $c = \sqrt{2}$} \\
      \includegraphics[width=0.35\textwidth]{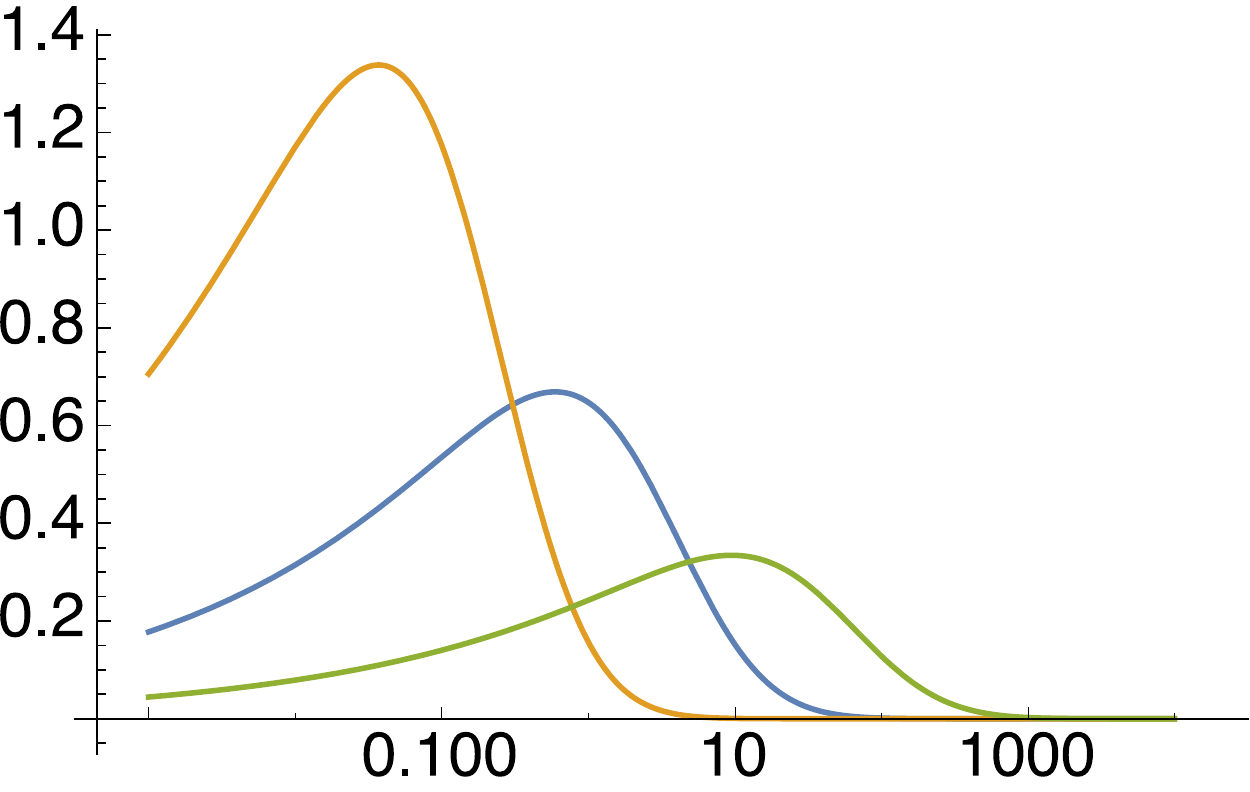} &
      \includegraphics[width=0.35\textwidth]{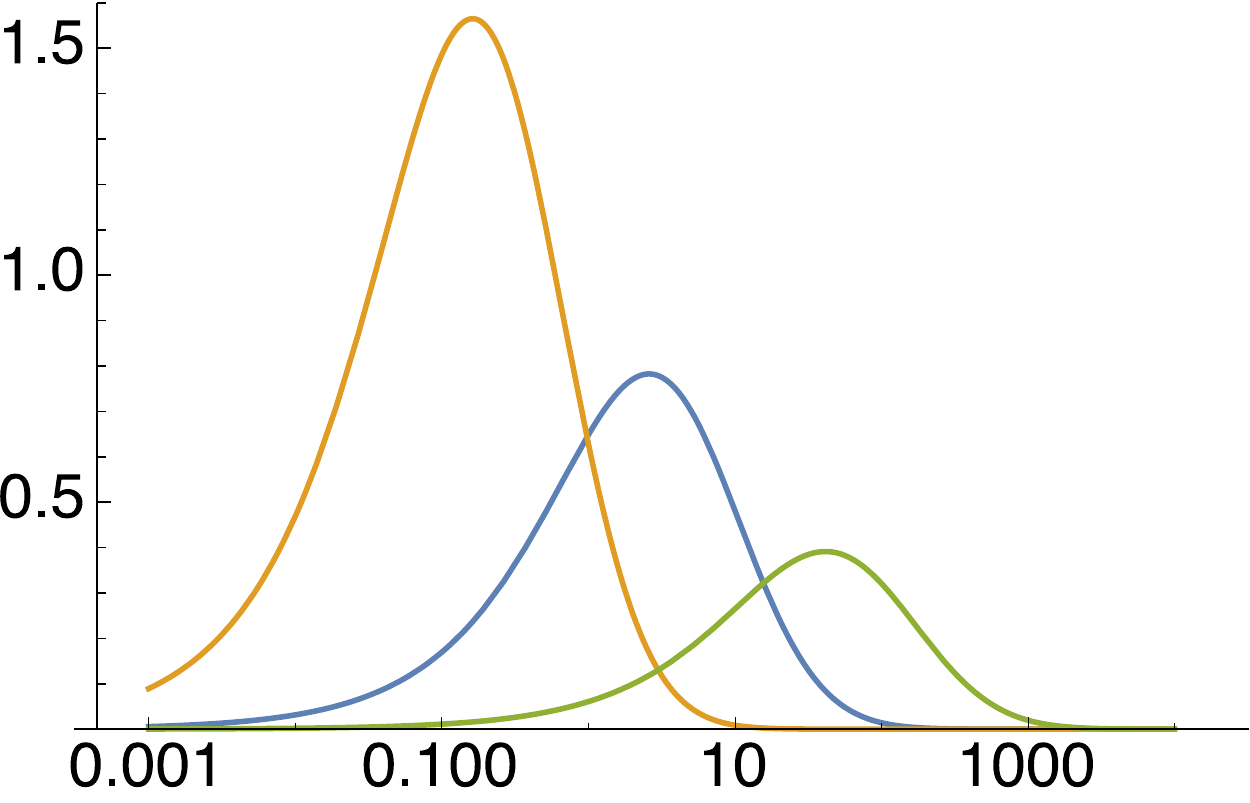} \\
  \\
      {\small $n = 1$, $c = 2$} 
      &  {\small $n = 2$, $c = 2$} \\
      \includegraphics[width=0.35\textwidth]{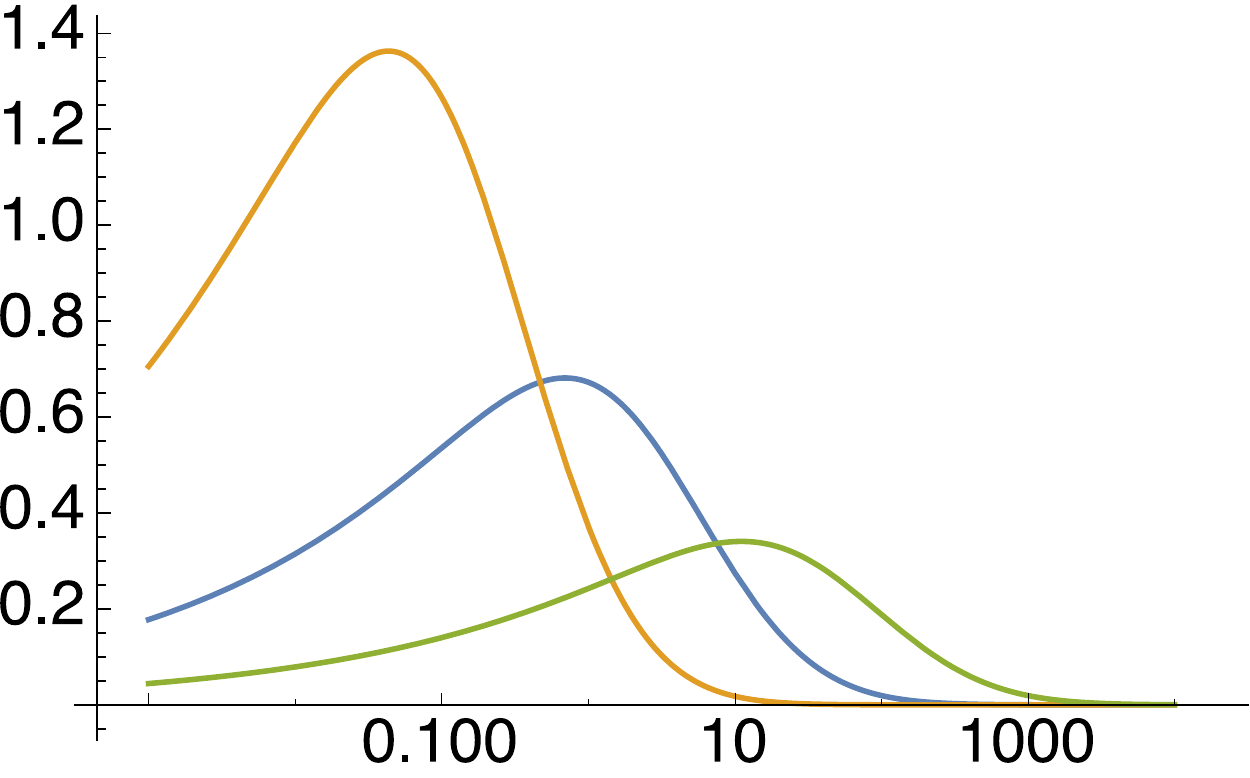} &
      \includegraphics[width=0.35\textwidth]{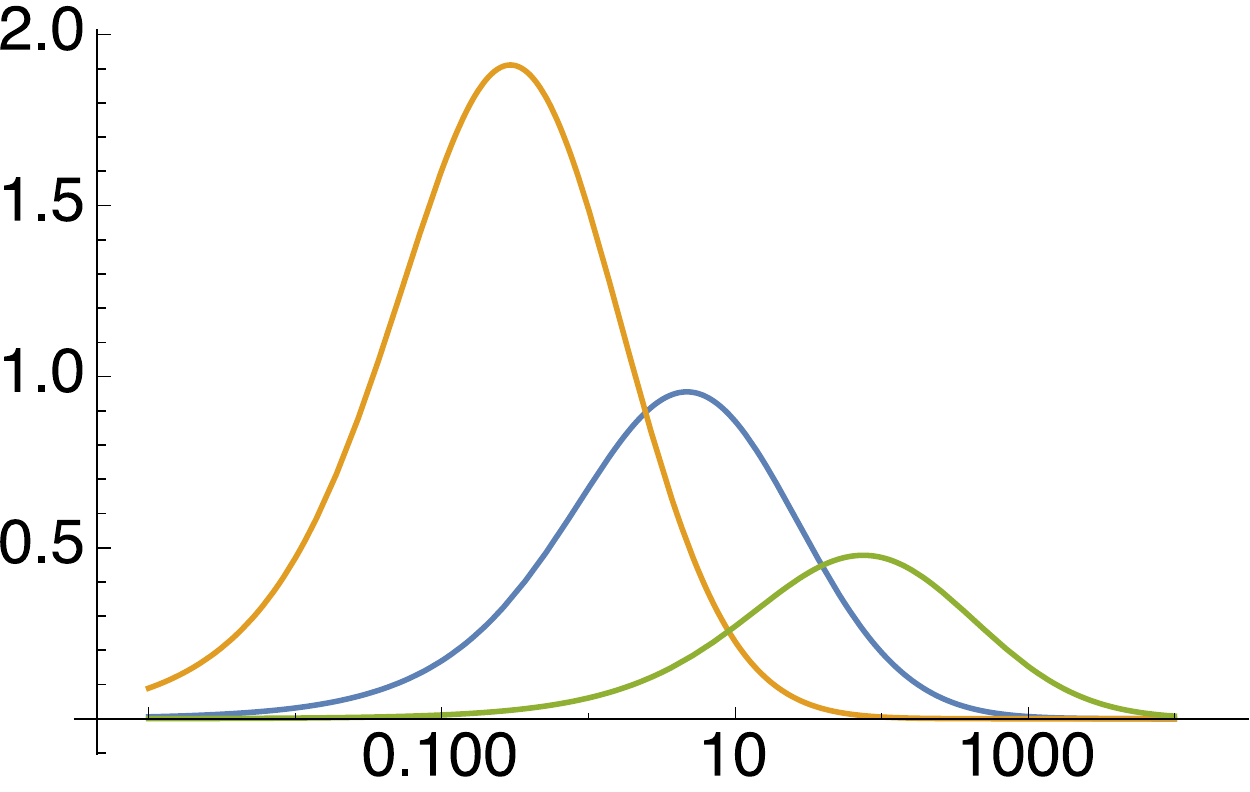} \\
     \end{tabular} 
  \end{center}
   \caption{Corresponding results as in
     Figure~\protect\ref{fig-scspsign-sine-waves-limit-kern} above but with
    (left column) $\gamma = 1/2$ for first-order derivatives and
    (right column) $\gamma = 3/4$ for second-order derivatives.
    Note that the use of scale normalization powers $\gamma < 1$
    implies that the maxima over temporal scales are moved to finer
    temporal scales and that the (uncompensated) maximum magnitude
    responses are no longer scale invariant.}
  \label{fig-scspsign-sine-waves-limit-kern-gamma-0p5-0p75}
\end{figure*}

\begin{figure*}[hbtp]
  \begin{center}
    \begin{tabular}{cc}
      {\small $\hat{\sigma}(\lambda_0)$ for $c = \sqrt{2}$} 
     & {\small $\hat{\sigma}(\lambda_0)$ for $c = 2$} \\
      \includegraphics[width=0.40\textwidth]{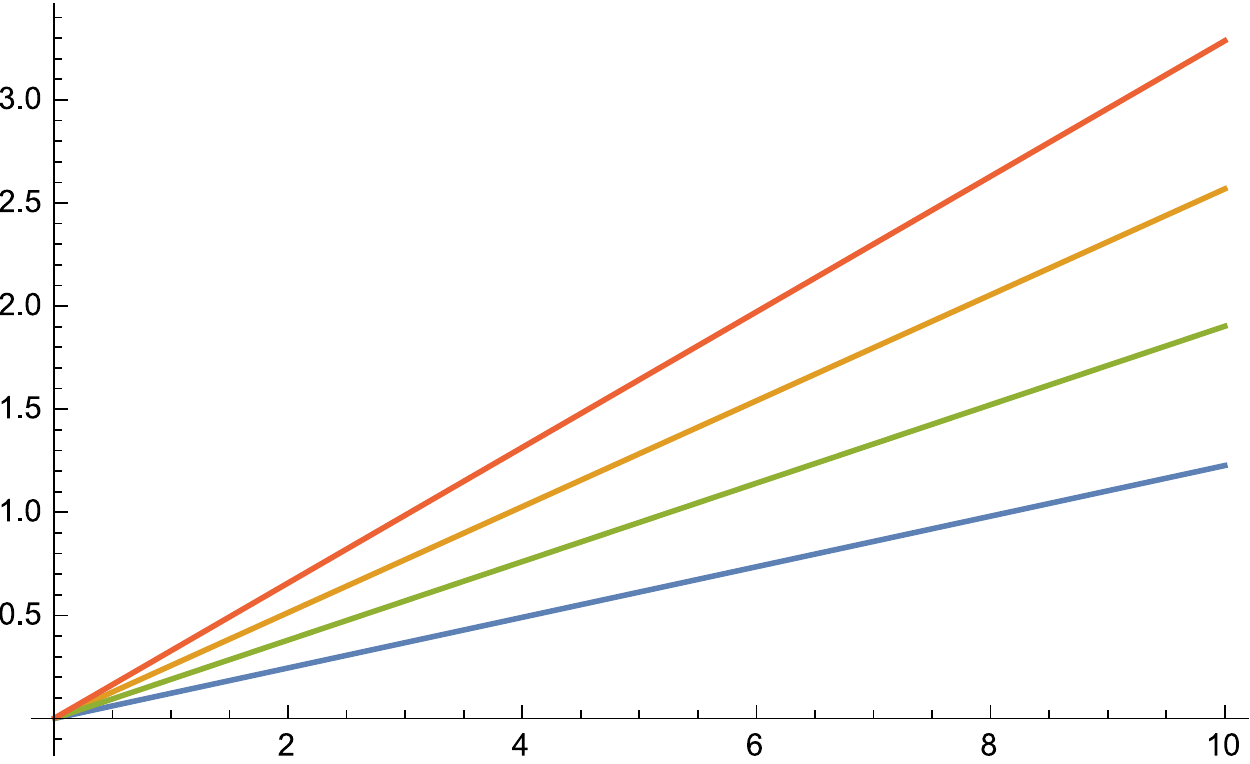}
      & \includegraphics[width=0.40\textwidth]{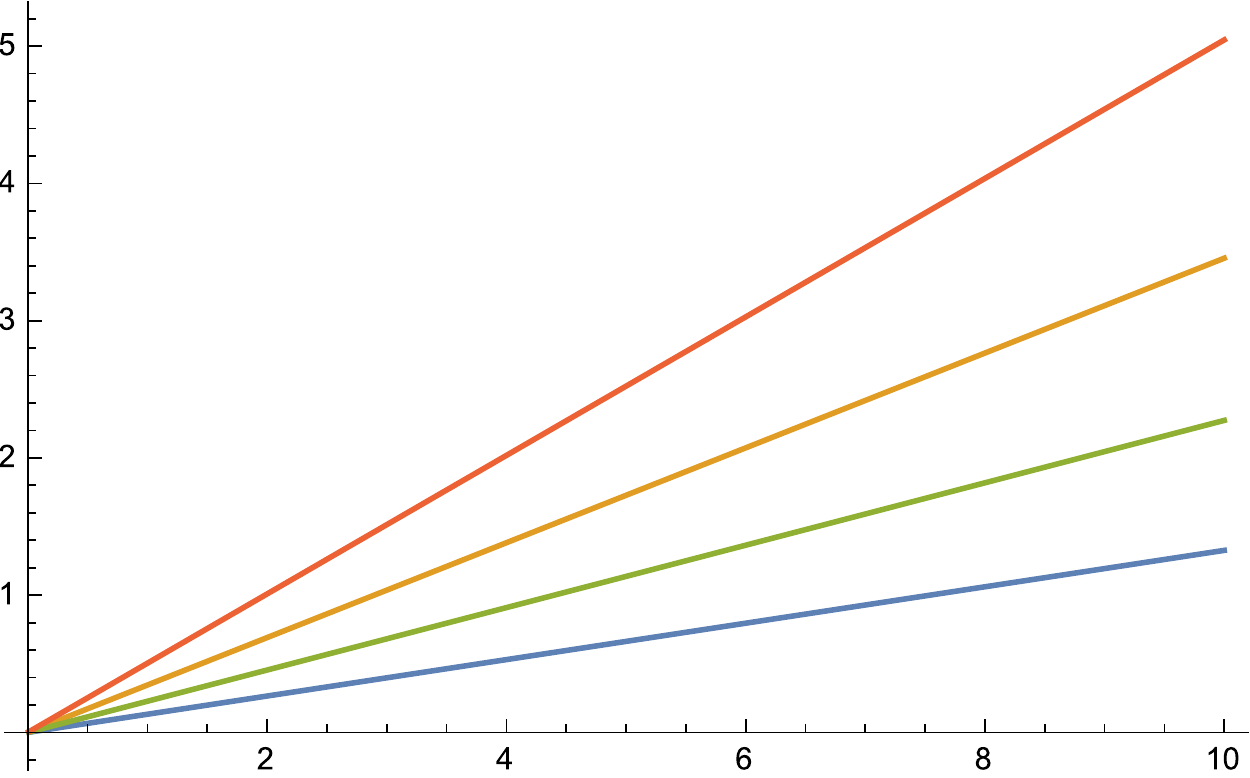}
      \\
    \end{tabular}
  \end{center}
   \caption{Scale estimates $\hat{\sigma} = \sqrt{\hat{\tau}}$ as
     function of the wavelength $\lambda_0$ for {\em sine waves\/} of
     different angular frequencies $\omega_0 = 2 \pi/\lambda_0$ and
     different orders $n$ of temporal differentiation and different
     values of $\gamma$: (blue curves) $n = 1$ and $\gamma = 1/2$,
    (brown curves) $n = 2$ and $\gamma = 3/4$,
    (green curves) $n = 1$ and $\gamma = 1$,
    (red curves) $n = 2$ and $\gamma = 1$. Note how the scale estimates
    expressed in dimensions of time $\sigma = \sqrt{\tau}$ are
    directly proportional to the 
    wavelength $\lambda_0$ of the sine wave and in
    agreement with temporal scale invariance.
    (Horizontal axis: wavelength $\lambda_0$.)
   (For $c = \sqrt{2}$ the time-causal limit kernel has been approximated by the
    slowest $K = 32$ temporal smoothing stages and for $c = 2$ by the
    slowest $K = 12$ temporal smoothing stages.)}
  \label{fig-sc-est-sine-waves-limit-kern}
\end{figure*}

\subsection{Temporal sine wave}
\label{sec-temp-sine-wave-temp-scsp-limit-kern}

Consider an input signal defined as a sine wave
\begin{equation}
  \label{eq-sine-model-limit-kern}
  f(t) = \sin \omega_0 t,
\end{equation}
and taken as an idealized model of a oscillating signal with temporal structures
having characteristic temporal duration $\lambda_0 = 2 \pi/\omega_0$.

For the time-causal temporal scale-space 
defined by
convolution with the time-causal semi-group $\Psi(t;\; \tau, c)$,
the temporal scale-space representation $L(t;\; \tau, c)$ is given by
\begin{equation}
  L(t;\; \tau, c)
  = \left| \hat{\Psi}(\omega_0;\; \tau, c) \right|
      \sin
      \left(
         \omega_0 t + \arg \hat{\Psi}(\omega_0;\; \tau, c)
      \right)
\end{equation}
where the magnitude $|\hat{\Psi}(\omega;\; \tau, c) |$ and the 
argument $\arg \hat{\Psi}(\omega;\; \tau, c) $ of the Fourier transform 
$\hat{\Psi}(\omega;\; \tau, c)$ of the time-causal limit kernel
are given by
\begin{align}
  \begin{split}
     |\hat{\Psi}(\omega;\; \tau, c) |
     & = \prod_{k=1}^{\infty} \frac{1}{\sqrt{1 + c^{-2k} (c^2-1) \tau \, \omega^2}},
 \end{split}\\
  \begin{split}
    \arg \hat{\Psi}(\omega;\; \tau, c) 
    & = \sum_{k=1}^{\infty} \arctan \left( c^{-k} \sqrt{c^2-1} \sqrt{\tau} \, \omega \right).
 \end{split}
\end{align}
Thus, the magnitude on the $n$th order temporal derivative is 
given by
\begin{equation}
  L_{t^n,ampl} = \omega^n \, |\hat{\Psi}(\omega;\; \tau, c) |
\end{equation}
and for the $n$th order scale-normalized derivate based on
variance-based scale normalization
the amplitude as function of scale is 
\begin{align}
  \begin{split}
     L_{\zeta^n,ampl} 
     & = \tau^{n \gamma/2} \omega^n \, |\hat{\Psi}(\omega;\; \tau, c) |
 \end{split}\nonumber\\
  \begin{split}
     \label{eq-magn-sc-norm-ders-sine-wave-limit-kern}
     & = \tau^{n \gamma/2} \omega^n
            \prod_{k=1}^{\infty} \frac{1}{\sqrt{1 + c^{-2k} (c^2-1) \tau \, \omega^2}}.
 \end{split}
\end{align}
Figure~\ref{fig-scspsign-sine-waves-limit-kern} shows graphs of the
variation of this entity as function of temporal scale for different
angular frequencies $\omega_0$, orders of temporal differentiation $n$ and
the distribution parameter $c$.
As can be seen from the graphs, the maxima over scales are assumed at
coarser scales with increasing wavelength of the sine wave.
The maxima over temporal scale are also assumed at coarser scales for
second-order derivatives than for first-order derivatives.

Specifically, when $\gamma = 1$ the magnitude values at the local
extrema over scale are constant over scale, which implies a transfer
of the scale selection property
(\ref{eq-sc-inv-temp-scaling-gauss-temp-scsp}) to this temporal
scale-space model.
Notably, this situation is in clear contrast to the situation for the temporal scale-space
generated by first-order integrators with equal time constants coupled
in cascade. In
Figure~\ref{fig-sc-est-max-magn-sine-wave-truncexp-uni-distr} it was
shown that because of the lack of true temporal scale invariance of
that temoral scale-space concept, the maximum magnitude values are not
constant over scales for $\gamma = 1$ as they should be according to
the scale-invariant scale selection property (\ref{eq-sc-inv-temp-scaling-gauss-temp-scsp}).

When choosing lower values of $\gamma$ as motivated from the
determination of the parameter $\gamma$ for scale selection in a
Gaussian scale space to make the scale estimate reflect the width 
of a Gaussian peak for second-order derivatives or reflect the 
width of a diffuse ramp for first-order derivatives, which leads
to $\gamma = 1/2$ for first-order derivatives
(\ref{eq-gamma-0p5-1st-der-gauss-temp-scsp}) and 
$\gamma = 3/4$ for second-order derivatives 
(\ref{eq-gamma-0p75-2nd-der-gauss-temp-scsp}), the local extrema over
scale are moved to finer scales
(see Figure~\ref{fig-scspsign-sine-waves-limit-kern-gamma-0p5-0p75}).
Then, however, the maximum magnitude values
are no longer the same for sine waves of different frequencies,
implying that a complementary magnitude normalization step is
necessary (see Section~\ref{sec-gen-sc-result-limit-kernel-sine-wave}
for additional details).

\paragraph{Local extrema over temporal scale.}

Taking the logarithm of the expression
(\ref{eq-magn-sc-norm-ders-sine-wave-limit-kern}) gives
\begin{multline}
  \label{eq-log-magn-sc-norm-ders-sine-wave-limit-kern}
  \log L_{\zeta^n,ampl}
  = \frac{n \gamma}{2} \log \tau + n \log \omega_0 \\
     - \frac{1}{2} \sum_{k=1}^{\infty} \log \left( 1 + c^{-2k} (c^2-1) \tau \, \omega_0^2 \right).
\end{multline}
If we treat $\tau$ as a continuous variable and differentiate with
respect to $\tau$ we obtain
\begin{equation}
  \label{eq-der-wrt-temp-scale-ampl-sine-wave-limit-kernel}
  \partial_{\tau} \left( \log L_{\zeta^n,ampl} \right)
  = \frac{n \gamma}{2 \tau} 
     - \frac{1}{2} \sum_{k=1}^{\infty} \frac{c^{-2k} (c^2-1) \, \omega_0^2}
                                                            {1 + c^{-2k} (c^2-1) \, \tau \, \omega_0^2}.
\end{equation}
Figure~\ref{fig-sc-est-sine-waves-limit-kern} show  graphs of how the
scale estimate $\hat{\sigma}$ obtained by setting the derivative with
respect to temporal scale to zero increases linearly with the wavelength
$\lambda_0$ of the signal, with different slopes of the linear curve
depending on the order of temporal differentiation and the value
of the scale normalization parameter $\gamma$. This overall linear
scaling behaviour can directly be proved by rewriting the expression
$\partial_{\tau} \left( \log L_{\zeta^n,ampl} \right) = 0$ in 
(\ref{eq-der-wrt-temp-scale-ampl-sine-wave-limit-kernel}) into
\begin{equation}
  \frac{n \gamma}{\tau \omega_0^2} 
     - \sum_{k=1}^{\infty} \frac{c^{-2k} (c^2-1) }
                                           {1 + c^{-2k} (c^2-1) \, \tau \, \omega_0^2} = 0.
\end{equation}
Since this expression is a direct function of the dimensionless entity
$\tau \omega_0^2$, it follows that the temporal scale estimates will be of the
form
\begin{equation}
   \hat{\tau} = \frac{\varphi(n\gamma, c)}{\omega_0^2}
\end{equation}
for some function $\varphi(n\gamma, c)$, and thus obeying 
temporal scale invariance in the sense that the scale estimate in
dimension length is proportional to the wavelength of the signal
\begin{equation}
  \label{eq-sc-est-prop-wavelength-limit-kern}
  \hat{\sigma} 
  = \sqrt{\hat{\tau}}
  = \frac{\sqrt{\varphi(n\gamma, c)}}{2 \pi} \, \lambda_0.
\end{equation}
Notice how this situation is in contrast to the results of 
scale selection in the temporal scale-space
concept obtained by coupling first-order integrators with equal time
constants in cascade, where the scale estimate for a sine
wave is not directly proportional to the wavelength of the temporal
signal, but also affected by a wavelength dependent temporal scale bias
(see Equation~(\ref{eq-sc-est-sine-wave-truncexp-uni-distr})
and Figure~\ref{fig-sc-est-max-magn-sine-wave-truncexp-uni-distr} in
Section~\ref{sec-temp-sine-wave-temp-scsp-1st-ord-int-eq-time-constants}).

\begin{table}[!hbt]
  \addtolength{\tabcolsep}{1pt}
  \begin{center}
   \footnotesize
  \begin{tabular}{|c|cc|}
   \hline
   \multicolumn{3}{|c|}{Scale estimates $\hat{\sigma}$ for $c = \sqrt{2}$} \\
   \hline
                                             & $n = 1$   & $n = 2$ \\
  \hline
     $\gamma = 1$                & 1.20  & 2.06 \\
     $\gamma = \gamma_n$ & 0.77 &  1.61 \\
  \hline
  \end{tabular}
  \end{center}
  \begin{center}
  \begin{tabular}{|c|cc|}
  \hline
   \multicolumn{3}{|c|}{Scale estimates $\hat{\sigma}$ for $c = 2$} \\
  \hline
                                             & $n = 1$   & $n = 2$ \\
  \hline
     $\gamma = 1$                & 1.43  & 3.17 \\
     $\gamma = \gamma_n$ & 0.83 &  2.17 \\
  \hline
  \end{tabular}
\end{center}
\caption{Scale estimates $\hat{\sigma}$ computed from local extrema over scale for
  a sine wave with angular
    frequency $\omega_0 = 1$ and different orders of temporal
    differentiation $n$ and different scale normalization parameters $\gamma$ with
  $\gamma_n = 1/2$ for $n = 1$ and $\gamma_n = 3/4$ for $n = 2$, and
  different values of the distribution parameter $c$.
  (For $c = \sqrt{2}$ the time-causal limit kernel has been approximated by the
  slowest $K = 32$ temporal smoothing stages and for $c = 2$ by the
  slowest $K = 12$ temporal smoothing stages.)}
  \label{tab-sc-est-sine-wave-limit-kern}

  \bigskip 

  \addtolength{\tabcolsep}{1pt}
  \begin{center}
   \footnotesize
  \begin{tabular}{|c|cc|}
  \hline
   \multicolumn{3}{|c|}{Ratios $\hat{\sigma}/\sqrt{n\gamma}$ for $c = \sqrt{2}$} \\
   \hline
                                             & $n = 1$   & $n = 2$ \\
  \hline
     $\gamma = 1$                & 1.20  & 1.46 \\
     $\gamma = \gamma_n$ & 1.09 &  1.32 \\
  \hline
  \end{tabular}
  \end{center}
  \begin{center}
  \begin{tabular}{|c|cc|}
  \hline
   \multicolumn{3}{|c|}{Ratios $\hat{\sigma}/\sqrt{n\gamma}$ for $c = 2$} \\
  \hline
                                            & $n = 1$   & $n = 2$ \\
  \hline
     $\gamma = 1$                & 1.43  & 2.24 \\
     $\gamma = \gamma_n$ & 1.18 &  1.77 \\
  \hline
  \end{tabular}
\end{center}
\caption{Ratios between the scale estimates $\hat{\sigma}$ in
  Table~\ref{tab-sc-est-sine-wave-limit-kern} and
   $\sqrt{n \gamma}$ computed from local extrema over scale for
  a sine wave with angular
    frequency $\omega_0 = 1$. For the corresponding entities obtained
    from scale selection for a sine wave in a non-causal Gaussian scale space, this
    ratio is equal to one for all combinations of $n$ and $\gamma$. These scale selection entities thus reveal
    a larger deviation from a Gaussian behaviour both for larger
    values of $c$ and for temporal derivatives of higher order.}
  \label{tab-sc-est-sine-wave-limit-kern-ratios}
\end{table}

Table~\ref{tab-sc-est-sine-wave-limit-kern} shows numerical values of
the differences between the results for different orders of
differentiation $n$ and different values of $\gamma$.
These numerical entities become particularly illuminating by
forming the ratio $\hat{\sigma}/\sqrt{n\gamma}$ as shown in 
Table~\ref{tab-sc-est-sine-wave-limit-kern-ratios}.
For a non-causal Gaussian scale space, this ratio should be equal to
one for all combinations of $n$ and $\gamma$
(see Equation~(\ref{eq-scsel-sine-wave-gauss-temp-scsp-omega})).
For this non-causal temporal scale space, we can, however, note that
the deviation from one increases both with larger values of the
distribution parameter%
\footnote{The reason why the these ratios depend on the distribution
  parameter $c$
can specifically be explained by observing that the temporal duration of
the temporal derivatives of the time-causal limit kernel
$\Psi_{t^{\alpha}}(t;\; \tau, c)$   will depend
on the distribution parameter $c$ -- see
Appendix~\ref{sec-scaletime-approx-limit-kernel} for a derivation
and explicit estimates of the width of the temporal derivatives of the
time-causal limit kernel.}
 $c$ and with increasing order of temporal
differentiation $n$, which both lead to larger degrees of temporal
asymmetry due to the non-causal temporal dimension.

In the essential proportionality of the scale estimate $\hat{\sigma}$ to the
wavelength $\lambda_0$ of the signal according to 
(\ref{eq-sc-est-prop-wavelength-limit-kern}), the main component of
the scale selection property
(\ref{eq-scsel-sine-wave-gauss-temp-scsp-lambda}) is thereby
transferred to this temporal scale-space concept, although the
proportionality constant has to be modified depending on the value of the
temporal scale distribution parameter $c$, the order of temporal differentiation $n$
and the scale normalization parameter $\gamma$.


\paragraph{Preservation of local extrema over temporal scale under
  temporal scaling transformations.}

Let us assume that the continuous magnitude function
(\ref{eq-log-magn-sc-norm-ders-sine-wave-limit-kern}) assumes a maximum over
temporal scales for some pair $(\omega_0, \tau_0)$ and that the 
derivative with respect to temporal scale is thereby zero
\begin{multline}
 \left.
    \partial_{\tau} \left( \log L_{\zeta^n,ampl} \right)
 \right|_{(\omega_0, \tau_0)}
  = \\
  \frac{n \gamma}{2 \tau_0} 
   - \frac{1}{2} \sum_{k=1}^{\infty} \frac{c^{-2k} (c^2-1) \, \omega_0^2}
                                                           {1 + c^{-2k} (c^2-1) \, \tau_0 \, \omega_0^2} = 0.
\end{multline}
Let us next assume that we feed in a different sine wave with
wavelength $\lambda_1 = c^j \lambda_0$ for some integer $j$ 
(for the same value of $c$ as used in the definition of the time-causal
limit kernel) and
corresponding to $\omega_1 = c^{-j} \omega_0$ with its matching
scale $\tau_1 = c^{2j} \tau_0$.
Then, it holds that
\begin{align}
  \begin{split}
 &
  \left. 
    \partial_{\tau} \left( \log L_{\zeta^n,ampl} \right)
 \right|_{(\omega_1, \tau_1)}
 \end{split}\nonumber\\
  \begin{split}
  & =
      \frac{n \gamma}{2 \tau_1} 
       - \frac{1}{2} \sum_{k=1}^{\infty} \frac{c^{-2k} (c^2-1) \, \omega_1^2}
                                                              {1 + c^{-2k} (c^2-1) \, \tau_1 \, \omega_1^2} 
  \end{split}\nonumber\\
  \begin{split}
  & =
     c^{-2j}
     \left( 
      \frac{n \gamma}{2 \tau_0} 
       - \frac{1}{2} \sum_{k=1}^{\infty} \frac{c^{-2k} (c^2-1) \, \omega_0^2}
                                                              {1 + c^{-2k} (c^2-1) \, \tau_0 \, \omega_0^2} 
     \right)
 \end{split}\nonumber\\
 \begin{split}
  \label{eq-scale-cov-temp-scsel-sine-wave-limit-kern}
   & =  c^{-2j} 
    \left.
      \partial_{\tau} \left( \log L_{\zeta^n,ampl} \right)
    \right|_{(\omega_0, \tau_0)}.
 \end{split}
\end{align}
This result implies that the sign of the derivate with respect to temporal
scale is preserved between matching angular frequencies
 and scales $(\omega_0, \tau_0)$ and $(\omega_1, \tau_1)$.
Specifically, local extrema over temporal
scales are preserved under uniform scaling transformations of the temporal domain
$t' = c^j t$, implying scale covariance of the temporal scales that
are selected from local extrema over scales of scale-normalized
temporal derivatives. 

Note also that although the analysis in
Equation~(\ref{eq-scale-cov-temp-scsel-sine-wave-limit-kern}) is
performed based on a temporary extension of $\tau$ into a continuous
variable, the scale covariance still holds when the continuous
function is sampled into a discrete set of temporal scale levels.%
\footnote{A formal proof of the transfer of this preservation property
  of local extrema over temporal scales from a temporary extension 
  of the temporal scale parameter $\tau$ into a
  continuous variable back into a restricted discrete set of temporal scale levels can
  be stated as follows: From
  Equation~(\ref{eq-scale-cov-temp-scsel-sine-wave-limit-kern})
  it follows that the continuous temporal scale-space signatures for the two sine
waves of wavelengths $\lambda_0$ and $\lambda_1 = c^j \lambda_0$ will
increase and decrease respectively at corresponding matching temporal scale
levels $\tau_0$ and $\tau_1 = c^{2j} \tau_0$. If we next sample these
scale-space signatures at some discrete set of temporal scale levels 
$\tau_{0,k} = c^{2k}$ and $\tau_{1,k} = c^{2j} c^{2k}$, then it
follows that the discrete maxima over temporal scales will also be
related according to $\tau_{1,max} = c^{2j} \tau_{0,max}$.
This preservation property of local extrema does, however, only hold for
temporal scaling factors $S$ that are integer powers of the
distribution parameter $c$, i.e., only $S = c^j$.
Alternatively, this preservation property can also be derived from the more
general scale invariance property under temporal scaling
transformations (\ref{eq-transf-prop-sc-norm-temp-ders-limit-kern})
that is stated in next section.}

\subsection{General scale invariance property under temporal scaling transformations}
\label{sec-gen-sc-result-limit-kernel-sine-wave}

In Lindeberg \cite[Appendix~3]{Lin16-JMIV} it is shown that
for two temporal signals $f$ and $f'$
that are related by a temporal scaling transform 
$f'(t') = f(t)$ for $t' = c^{j'-j} t$ with the corresponding
transformation between corresponding temporal scale levels $\tau' = c^{2(j'-j)} \tau$,
the scale-normalized temporal derivatives defined by either
$L_p$-normalization or variance-based normalization in 
the scale-space representation defined by
convolution with the time-causal limit kernel $\Psi(t;\; \tau, c)$
are {\em for any temporal input signal\/} $f$ related according to
\begin{align}
  \begin{split}
     L'_{\zeta'^n}(t';\, \tau', c) 
     & = c^{(j'-j)n (\gamma-1)} \, L_{\zeta^n}(t;\, \tau, c)
  \end{split}\nonumber\\
  \begin{split}
     \label{eq-transf-prop-sc-norm-temp-ders-limit-kern}
     & = c^{(j'-j) (1 - 1/p)} \, L_{\zeta^n}(t;\, \tau, c).
  \end{split}
\end{align}
This result specifically implies that the scale-space signatures,
which are the graphs that show the variation in the strength of
scale-normalized derivatives over scale, will be rescaled copies of
each other for signals that are related by a uniform scaling
transformation of the temporal domain. 

Specifically, local temporal
scale estimates $\hat{\tau}$ and $\hat{\tau}'$ as determined from local extrema 
over temporal scales in the two temporal domains will be assumed at corresponding temporal scale
levels and will thus be transformed in a scale-covariant way for any
temporal scaling transformation of the form $t' = c^{j'-j} t$. 
In units of the temporal variance, it holds that
\begin{equation}
  \hat{\tau}' = c^{2(j'-j)} \hat{\tau}
\end{equation}
and in units of the temporal standard deviation
\begin{equation}
  \hat{\sigma}' = c^{j'-j} \hat{\sigma}.
\end{equation}
If $\gamma = 1$ corresponding to $p = 1$, the magnitude values at
corresponding temporal scale levels will be equal. If $\gamma \neq 1$ 
corresponding to $p \neq 1$, the magnitude values will be related
according to (\ref{eq-transf-prop-sc-norm-temp-ders-limit-kern}).
Thereby, this expression provides a way to normalize maximum strength measures
between local extrema over scales assumed at different temporal scales
as obtained {\em e.g.\/} in the scale-space signatures shown in 
Figure~\ref{fig-scspsign-sine-waves-limit-kern-gamma-0p5-0p75}.

Note that by this construction we have been able to transfer the
   temporal scale invariance property (9) and (10) that holds for a
   non-causal Gaussian temporal scale-space concept to also hold for a time-causal
   temporal scale-space concept, which is a novel type of theoretical
   construction.
   This property is, however, restricted
   to the temporal scale-space concept based on convolution with the
   time-causal limit kernel and does, for example, not hold for the
   time-causal temporal scale-space concept based on convolution with
   a cascade of truncated exponential kernels having equal time
   constants and corresponding to a uniform distribution of the
   temporal scale levels in units of the composed temporal variance.

\begin{figure*}[hbtp]
  \begin{center}
    \begin{tabular}{cc}
        {\small\em Temporal peak for $K_0 = 5$}
        & {\small\em Scale-space signature $M_{peak,uni}(K)$} \\
        \includegraphics[width=0.35\textwidth]{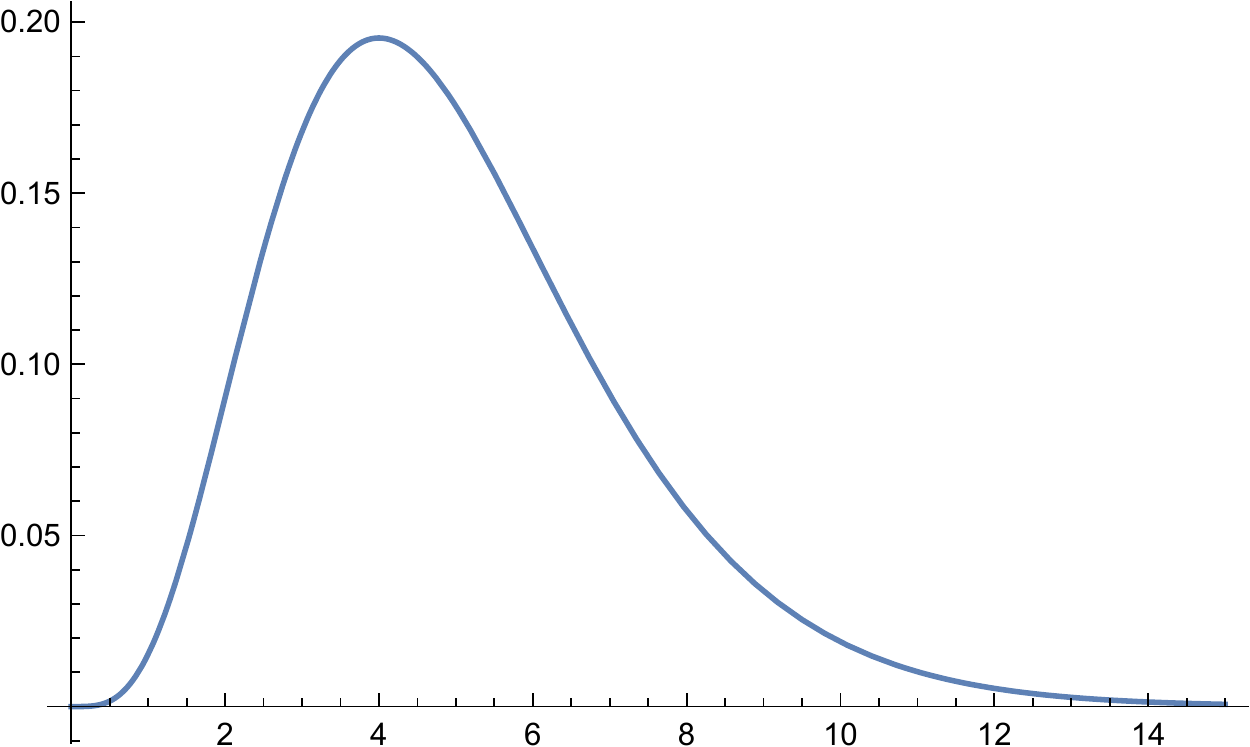} 
        & \includegraphics[width=0.35\textwidth]{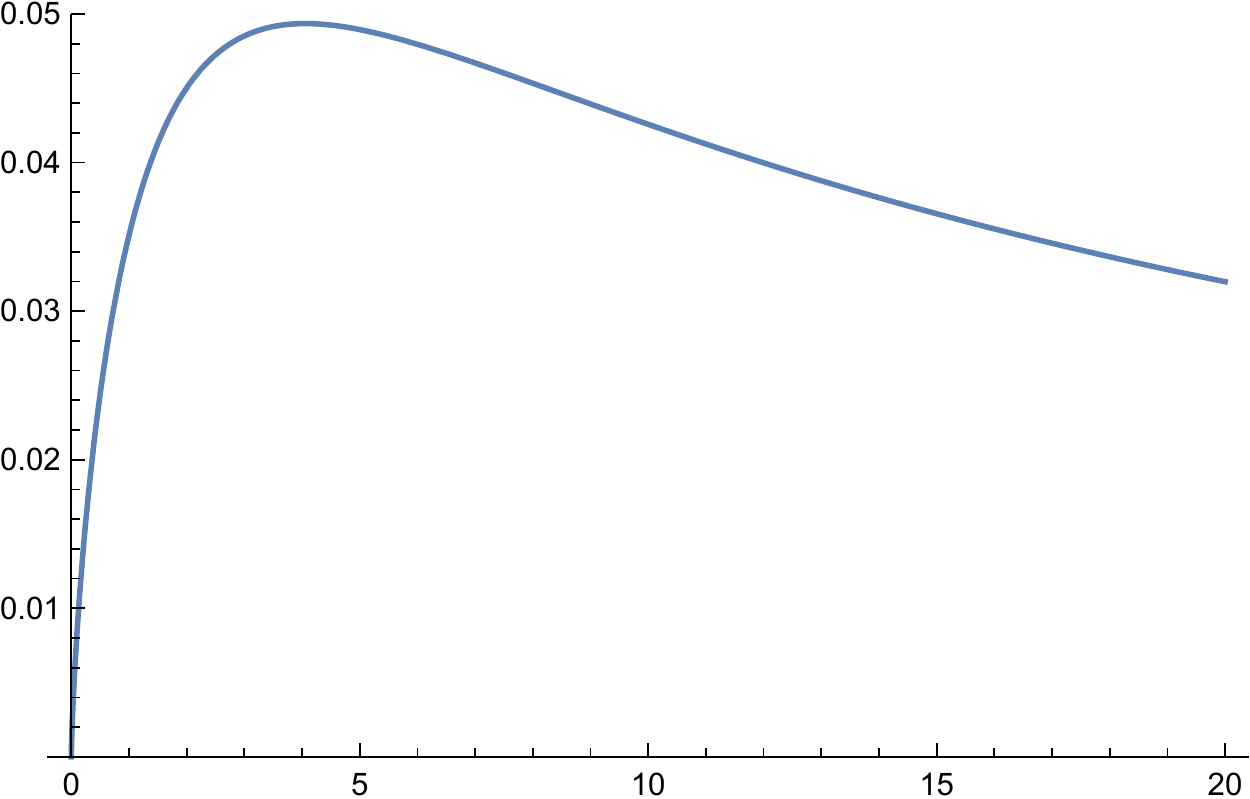}
      \\ $\,$ \\
        {\small\em Temporal onset ramp for $K_0 = 5$}
        & {\small\em Scale-space signature $M_{onset,uni}(K)$} \\
        \includegraphics[width=0.35\textwidth]{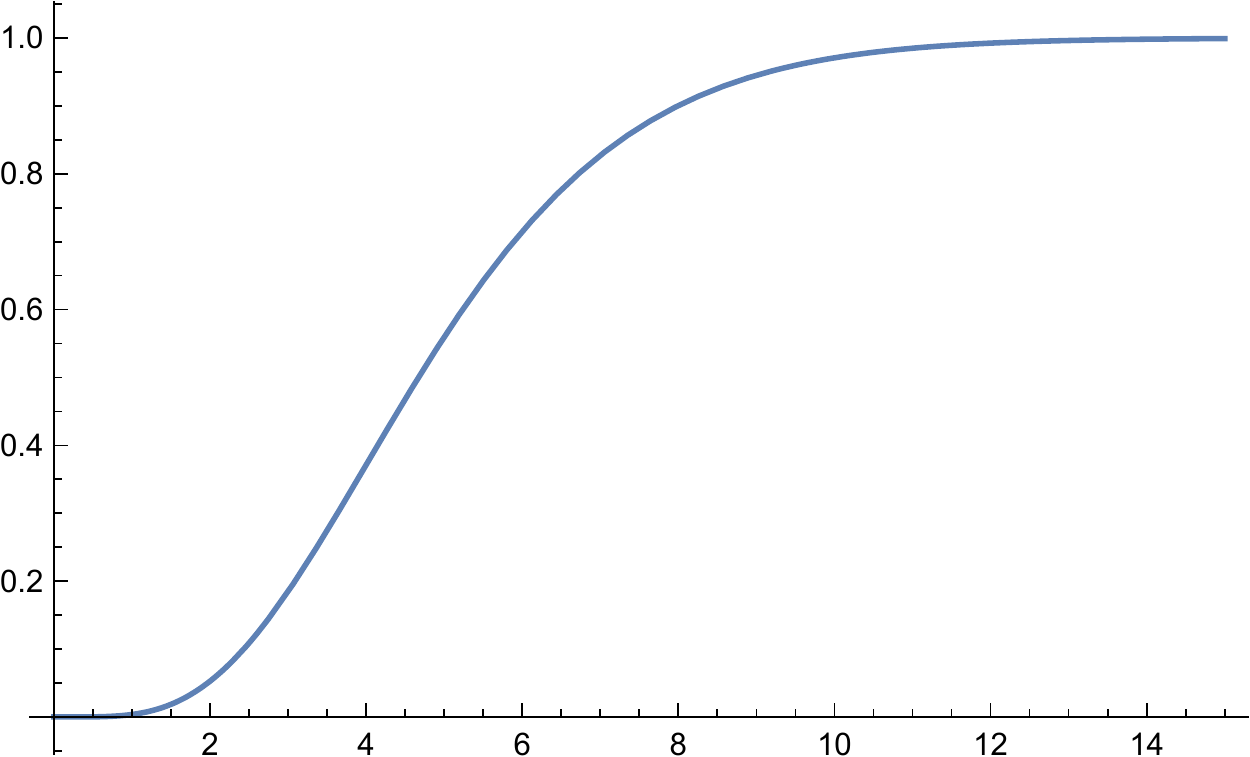} 
        & \includegraphics[width=0.35\textwidth]{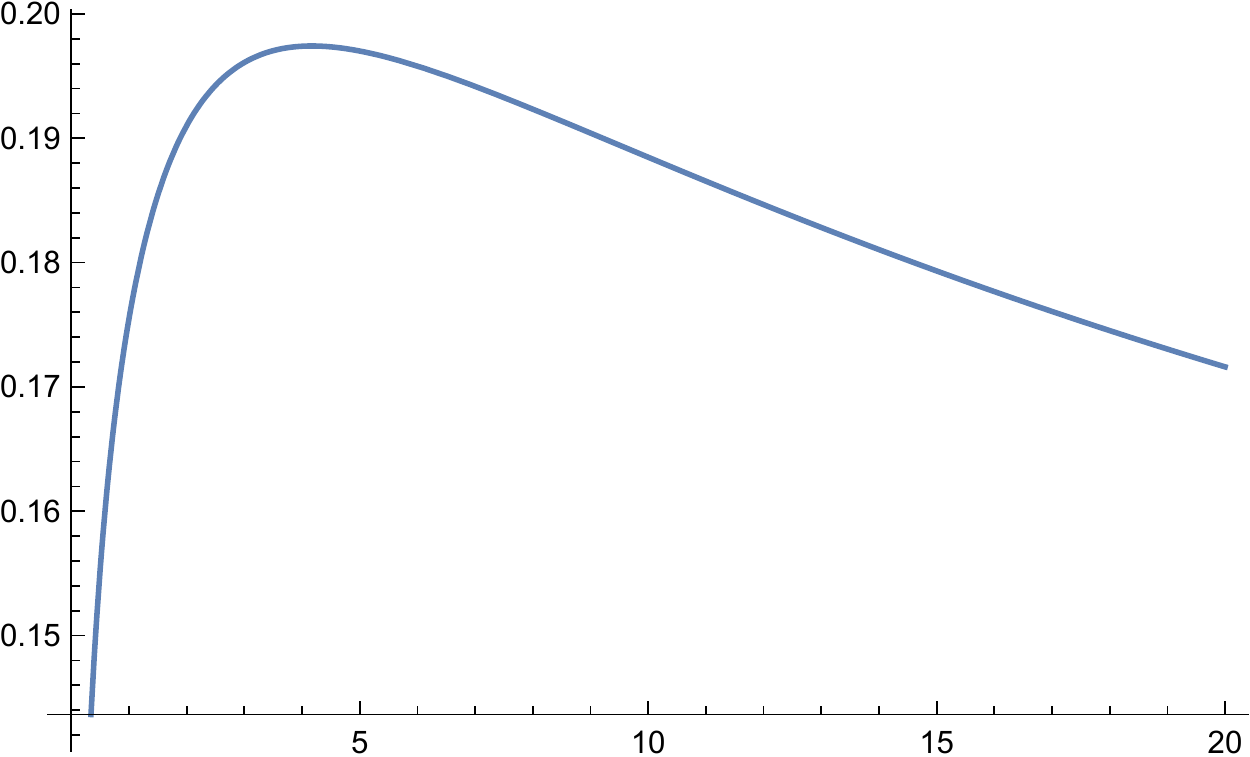}
      \\
$\,$ \\
        {\small\em Sine wave for $\omega_0 = 1$}
        & {\small\em Scale-space signature $M_{sine,uni}(K)$} \\
        \includegraphics[width=0.35\textwidth]{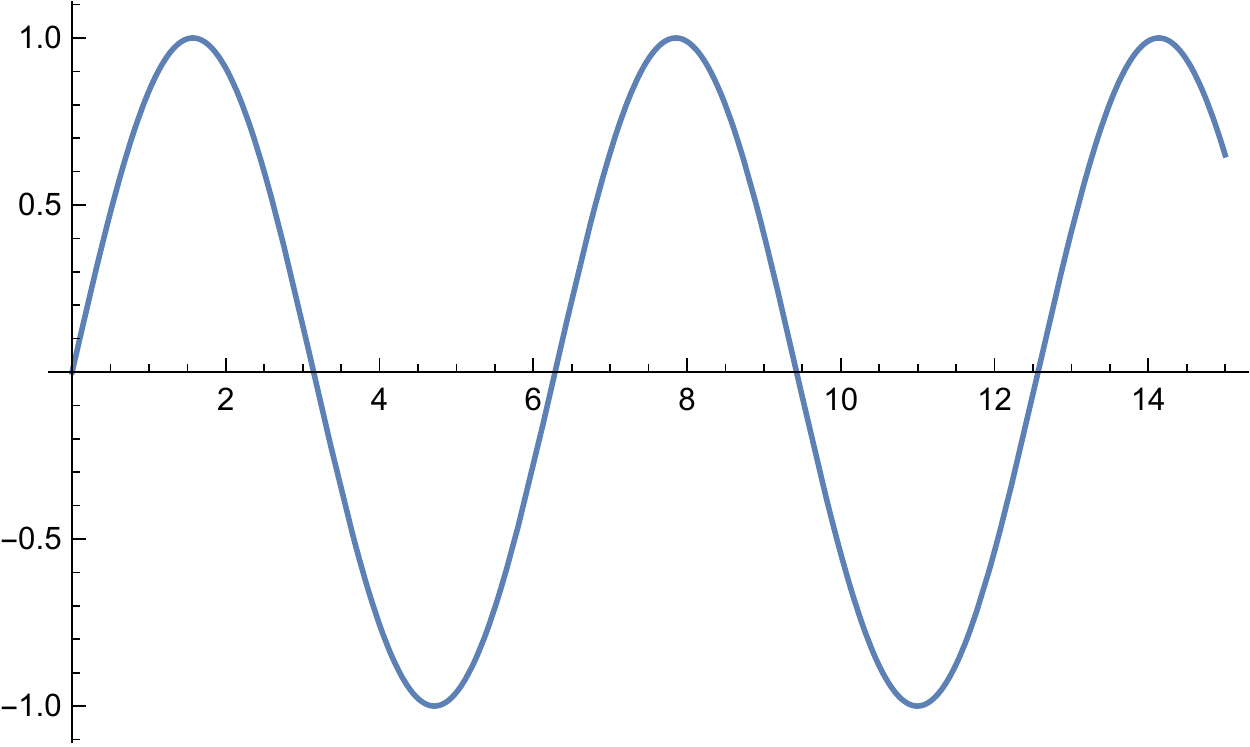} 
        & \includegraphics[width=0.35\textwidth]{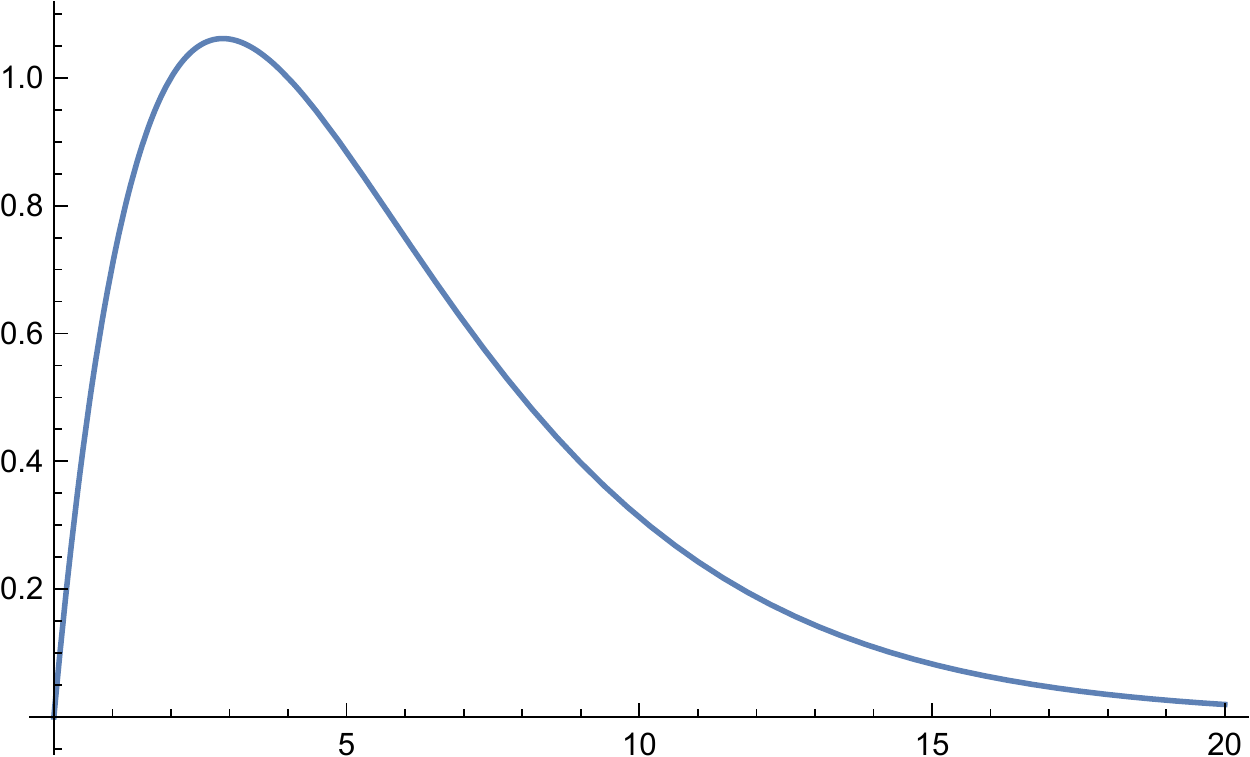}
      \\ 
$\,$ \\
        {\small\em Sine wave for $\omega_0 = 1$}
        & {\small\em Scale-space signature $M_{sine,limit}(K)$} \\
        \includegraphics[width=0.35\textwidth]{sine-unidistr-graph-omega0-1-tmax-15-eps-converted-to} 
        & \includegraphics[width=0.35\textwidth]{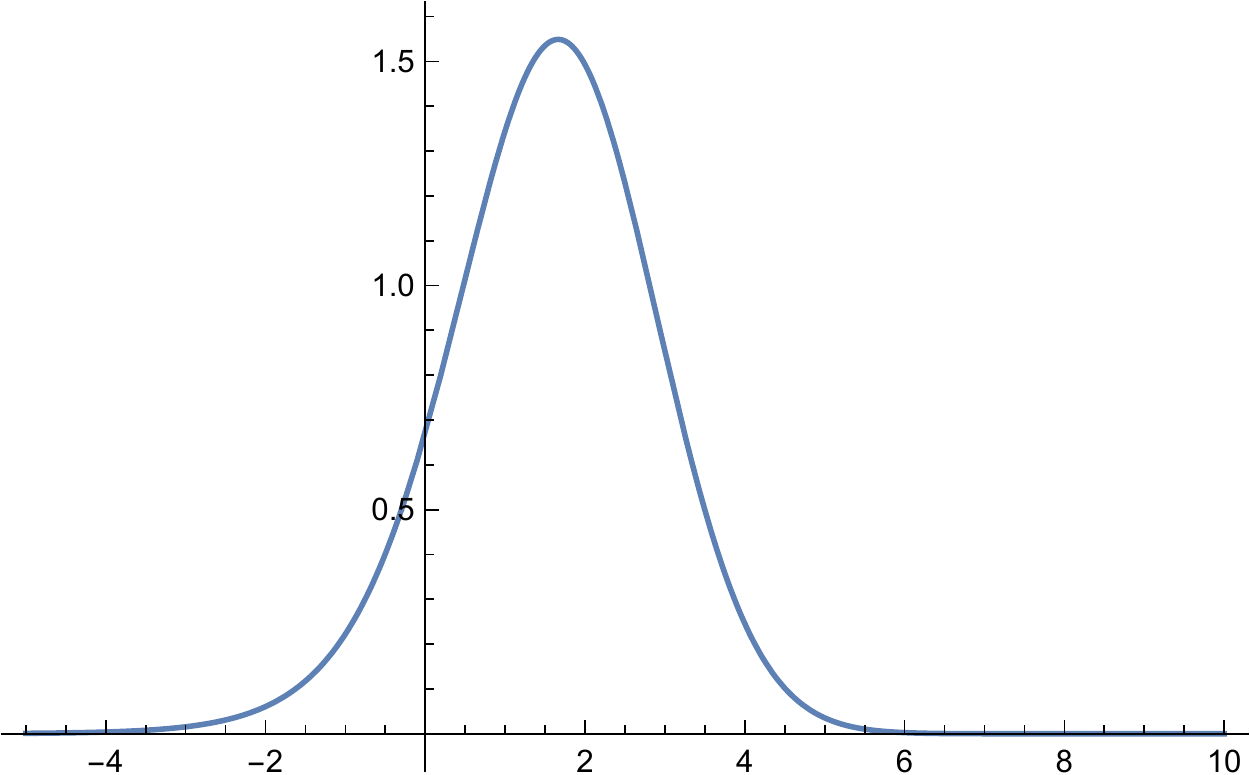}
      \\
    \end{tabular}
  \end{center}
  \caption{Graphs of the three types of model signals for which
    closed-form expressions can be computed for how the
    scale-normalized derivative-based magnitude responses vary as
    function of the scale parameter:
    (top row) Time-causal temporal peak model according to
    (\ref{eq-peak-model-uni-distr}) with temporal duration determined
    by $K_0 = 5$ for which the second-order
    scale-normalized temporal derivative response at the peak varies
    according to (\ref{eq-M-peak-uni}) here with $\gamma = 3/4$,
    (second row) Time-causal onset ramp model according to
    (\ref{eq-ramp-model-uni-distr}) with temporal duration
    determined by $K_0 = 5$ for which the first-order
    scale-normalized temporal derivative response at the ramp varies
    according to (\ref{eq-M-onset-uni}) here with $\gamma = 1/2$,
    (third row) Temporal sine wave according to
    (\ref{eq-sine-model-uni-distr}) with
    angular frequency $\omega_0 = 1$ for which the amplitude of the
    second-order temporal derivative response varies according to
    (\ref{eq-M-sine-uni}) here with $\gamma = 1$  and
   (bottom row) Temporal sine wave according to
   (\ref{eq-sine-model-limit-kern}) for which the
   amplitude of the second-order temporal response measure varies
   according to (\ref{eq-M-sine-limit}) here with $\gamma = 1$ .
   The scale-space signatures in the first three rows have been
   computed using the time-causal temporal scale-space concept based on truncated
   exponential kernels with equal time constants $\mu = 1$ coupled in cascade,
   whereas the scale-space signature in the bottom row has been
   computed computed using the time-causal temporal scale-space
   concept obtained by convolution with the scale-invariant limit
   kernel with distribution parameter $c = 2$. Note specifically that the shape of the scale-space
   signature in the third row is different from the shape in the
   fourth row because of the uniform distribution of temporal scale
   levels $K$ in the third row and the logarithmic distribution in the
   fourth row.
   (Horizontal axis for the figures in the left column: time $t$.)
   (Horizontal axis for the figures in the right column: temporal
   scale level $K$.)}
  \label{fig-model-signals-and-scsp-signatures}
\end{figure*}

\section{Influence of discrete temporal scale levels on the
  theoretical analysis}
\label{sec-infl-disc-temp-scale-levels}

In the theoretical analysis of scale selection properties of
(i)~a temporal peak in Section~\ref{sec-scsel-temp-blob-uni-distr},
(ii)~a temporal onset ramp in Section~\ref{sec-scsel-onset-ramp-uni-distr}
and (ii)~a temporal sine wave in
Section~\ref{sec-temp-sine-wave-temp-scsp-1st-ord-int-eq-time-constants}
and
Section~\ref{sec-temp-sine-wave-temp-scsp-limit-kern},
we did first compute closed-form expressions for how the
scale-normalized temporal magnitude measures
depend upon the temporal scale
levels $\tau_K = K \mu^2$ according to (\ref{eq-Lttnorm-timecausal-peak}),
(\ref{eq-Ltnorm-timecausal-ramp}) and
(\ref{eq-nth-order-temp-der-norm-timecausal-sine}) for the time-causal
scale-space concept based on truncated exponential kernels with equal
time constant coupled in cascade or how the scale-normalized magnitude
measure depends on the temporal
scale level $\tau_K = c^{2K} \tau_0$ according to
(\ref{eq-magn-sc-norm-ders-sine-wave-limit-kern}) for the time-causal temporal
scale-space concept based on the scale-invariant limit kernel:
\begin{align}
  \begin{split}
     & 
    M_{peak,uni}(K) = - L_{\zeta\zeta}(\mu  (K+K_0-1);\; \mu, K) 
  \end{split}\nonumber\\
  \begin{split}
    \label{eq-M-peak-uni}
    & = \frac{K^{\gamma } \mu ^{2 \gamma -3} e^{-K-K_0+1} (K+K_0-1)^{K+K_0-2}}
                   {\Gamma (K+K_0)},
  \end{split}\\
  \begin{split}
     M_{onset,uni}(K) = L_{\zeta}(\mu (K_0 + K -1);\; \mu, K) 
  \end{split}\nonumber\\
  \begin{split}
    \label{eq-M-onset-uni}
    & = \frac{\left(\sqrt{K} \mu \right)^{\gamma } (K+K_0-1)^{K+K_0-1}  e^{-K-K_0+1}}
                  {\mu \, \Gamma (K+K_0)},
  \end{split}\\
  \begin{split}
     & M_{sine,uni}(K) = L_{\zeta^n,ampl}(K)
  \end{split}\nonumber\\
  \begin{split}
    \label{eq-M-sine-uni}
   & = \frac{(K \mu^2)^{n \gamma/2} \omega_0^n}{\left( 1 + \mu^2 \, \omega_0^2 \right)^{K/2}},
  \end{split}
\end{align}
\begin{align}
  \begin{split}
     & M_{sine,limit}(K) = L_{\zeta^n,ampl}(c^{2K} \tau_0)
  \end{split}\nonumber\\
  \begin{split}
    \label{eq-M-sine-limit}
     &  = (c^{2K} \tau_0)^{n \gamma/2} \omega^n
            \prod_{k=1}^{\infty} \frac{1}{\sqrt{1 + c^{-2k} (c^2-1) c^{2K} \tau_0 \, \omega^2}}.
  \end{split}
\end{align}
Then, to compute the temporal scale levels at which the
scale-normalized derivative responses assumed their maximum values
over temporal scales, we temporarily extended these magnitude measures
from being defined over discrete integer temporal scale levels $K$ to
a continuum over $K$, to be able to differentiate the closed-form
expressions with respect to the temporal scale level.

A general question that could be raised in this context therefore
concerns how good approximation the results from the continuous
approximation of local extrema over scales are with respect to a setting
where the temporal scale levels are required to be discrete.
A common property of the four types of scale-space signatures
according to equations~(\ref{eq-M-peak-uni})--(\ref{eq-M-sine-uni}) and
shown in Figure~\ref{fig-model-signals-and-scsp-signatures} is that they are unimodal, {\em i.e.\/},
they assume a single maximum over temporal scales and do first
increase and then decrease.
Thereby, when the continuous variable $K_c$ with its associated maximum
over temporal scales $\hat{K}_c$ obtained from a continuous analysis is
in a second stage restricted to be discrete, it follows that the
discrete maximum over discrete temporal scales $\hat{K}_d$ is guaranteed to be
assumed at either the nearest lower or the nearest higher
integer. Thus, we obtain the discrete temporal scale estimate by
rounding the continuous scale estimate $\hat{K}_c$ to either the nearest
lower or the nearest higher integer. 

Whether the value should be
rounded upwards or downwards depends on how close the continuous
estimate $\hat{K}_c$ is to the nearest downwards {\em vs.\/} upwards
integers and on the local degree of asymmetry of the scale-space
signature around the maximum over temporal scales.

When implementing and executing a temporal scale selection algorithm
in practice, the situation can on the other hand be reverse. Given a set of discrete
temporal scale levels, we may detect a local maximum over temporal
scales at some discrete temporal scale level $\hat{K}_d$.
If we would like to use this temporal scale estimate for estimating
the temporal duration of the underlying temporal structure that gave
rise to the response, {\em e.g.\/}\ according to the methodology outlined
in appendix~\ref{sec-scaletime-approx-limit-kernel}, we may on the
other hand would like to compute a better continuous estimate $\hat{K}_c$ of the temporal
scale level than as restricted by the discrete temporal scale levels.

A straightforward way of computing a more accurate temporal scale estimate in
such a situation is by interpolating a parabola over the measurements
over the temporal scale levels in an analogous way as subresolution
spatial scale estimates can be obtained over a spatial scale-space
representation \cite{Lin97-IJCV,LinBre03-ScSp,Low04-IJCV}.
Let $(x_0, y_0)$ denote the scale level and the magnitude measure at the
discrete maximum and let $(x_0-1, y_{-1})$ and $(x_0+1, y_{1})$
denote the corresponding scale level and magnitude measure at the nearest
lower and upper temporal scales, respectively.
Assuming the following form of the interpolating function
\begin{equation}
  \label{eq-parabol-interpol}
  y(x) = a \, \frac{(x - x_0)^2}{2} + b \, (x - x_0) + c,
\end{equation}
the interpolation coefficients become
\begin{align}
  \begin{split}
     a & = y_1 - 2 y_0 + y_{-1}
  \end{split}\\
  \begin{split}
     b & = (y_1 -  y_{-1})/2
  \end{split}\\
  \begin{split}
     c & = y_0 
  \end{split}
\end{align}
with the corresponding subresolution estimate of the maximum over
scales
\begin{equation}
  \label{eq-parabol-interpol-subresolution-estimate}
  \hat{x} = x_0 - \frac{b}{a} = x_0 - \frac{y_1 -  y_{-1}}{2(y_1 - 2 y_0 + y_{-1})}.
\end{equation}
Note that the correction offset $\Delta x = -b/a$ is restricted to the
interval $\Delta x \in [-1/2, 1/2]$ implying that the location $x_0$
of the discrete maximum is guaranteed to be on the sampling grid point $x_i$
nearest to the subresolution estimate $\hat{x}$.

\section{Temporal scale selection for 1-D temporal signals}
\label{sec-temp-scsel-1D-temp-signal}

To illustrate the derived scale selection properties, we will in this
section show the result of applying temporal scale selection to
different types of purely temporal signals.

\begin{figure*}
  \begin{center}
    \begin{tabular}{cc}
       {\small\em Scale-space extrema from the time-causal limit kernel}
        & {\small\em Scale-space extrema from the non-causal Gaussian kernel} \\
\includegraphics[width=0.48\textwidth]{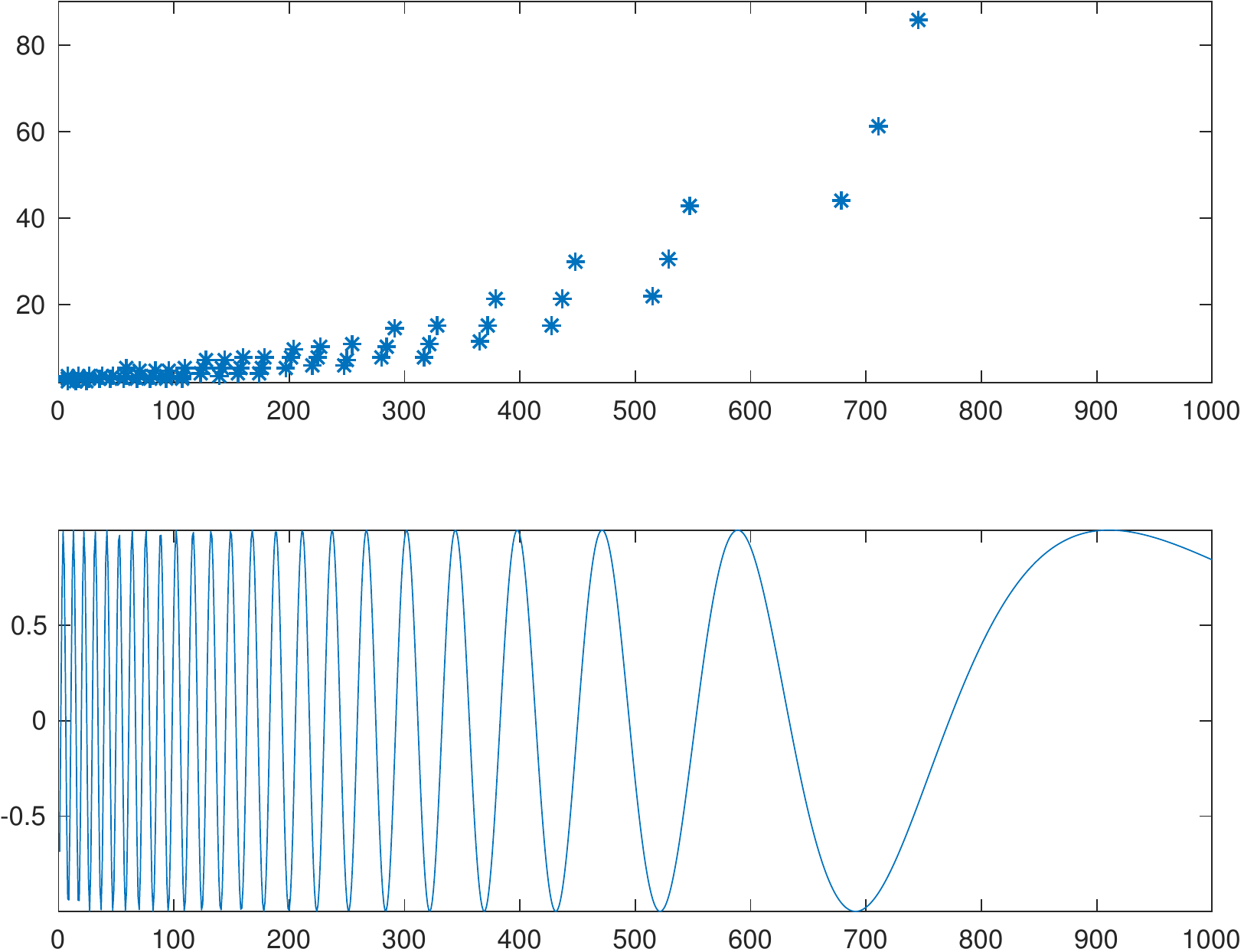} 
        & \includegraphics[width=0.48\textwidth]{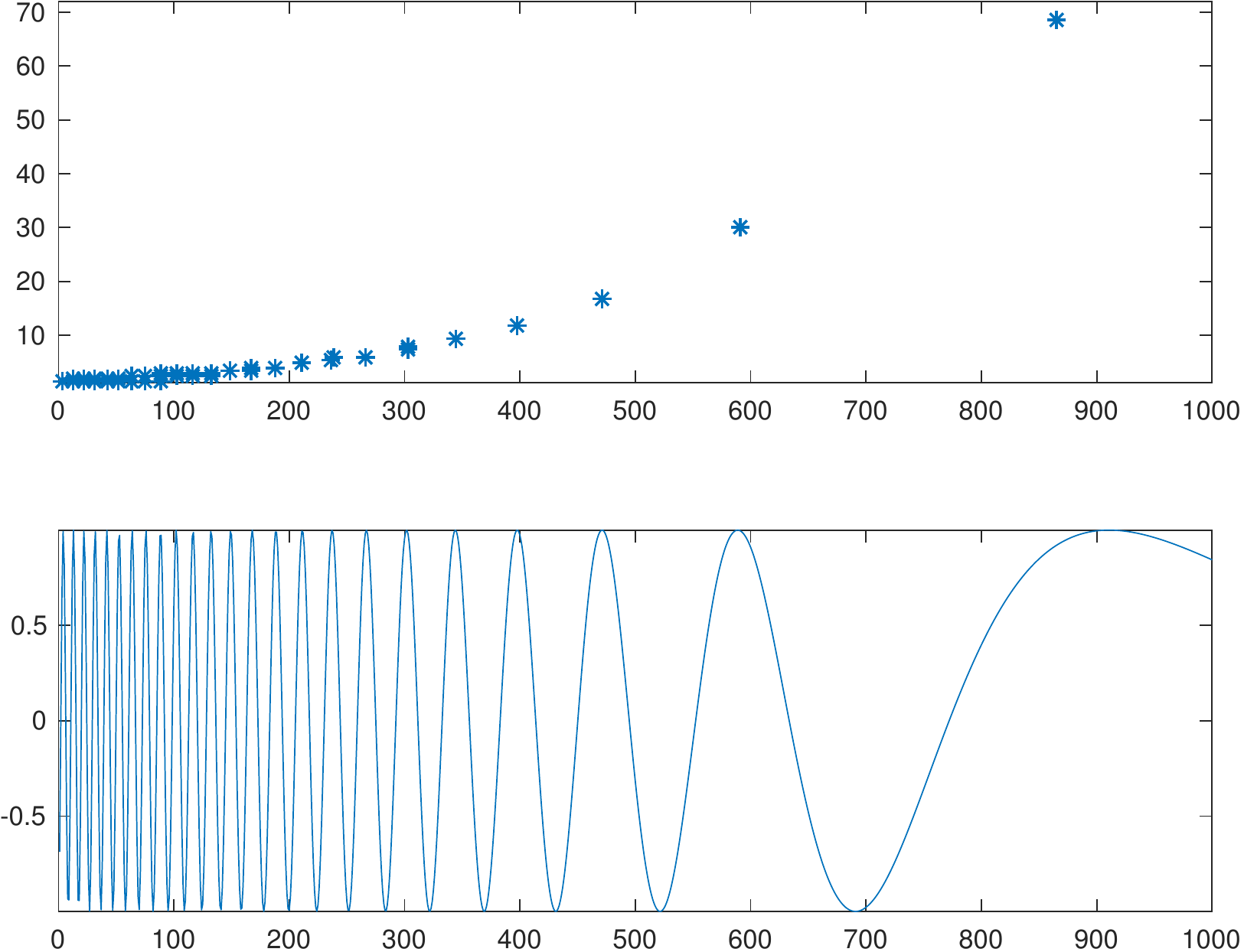}
   \end{tabular}
  \end{center}
  \caption{Temporal scale selection by scale-space extrema detection
    applied to a synthetic sine
    wave signal $f(t) = \sin(\exp((b-t)/a))$ for $a = 200$ and $b =
    1000$ with temporally
    varying frequency so that the wavelength increases with time $t$.
    (top left) Temporal scale-space maxima of the scale-normalized
    second-order temporal derivative $-L_{\zeta\zeta}$ detected using the
    time-causal temporal scale-space concept corresponding to
    convolution with the time-causal limit kernel for $c = \sqrt{2}$
    and with each scale-space maximum marked at
    the point $(\hat{t}, \hat{\sigma})$ with $\hat{\sigma} = \sqrt{\hat{\tau}}$ 
    at which the scale-space maximum is assumed.
    (top right) Temporal scale-space maxima detected using the
    non-causal Gaussian temporal scale-space concept using
    5 temporal scale levels per scale octave.
    For both temporal scale-space concepts, the scale-normalized
    temporal derivatives have been defined using $l_p$-normalization
    for $p = 2/3$ and corresponding to $\gamma = 3/4$ for second-order
    temporal derivatives. (Horizontal axis: time $t$.) (Vertical axis
    in top row: Temporal scale estimates in units of $\sigma =
    \sqrt{\tau}$.) (Vertical axis in bottom row: Signal strength $f(t)$.)}
  \label{fig-expsine-tempscspextr-recfiltlogscvar-discgaussvar}

  \bigskip

  \begin{center}
    \begin{tabular}{cc}
        {\small\em $-L_{\zeta\zeta}$ computed with time-causal limit kernel}
        & {\small\em $-L_{\zeta\zeta}$ computed with non-causal Gaussian kernel} \\
\includegraphics[width=0.35\textwidth]{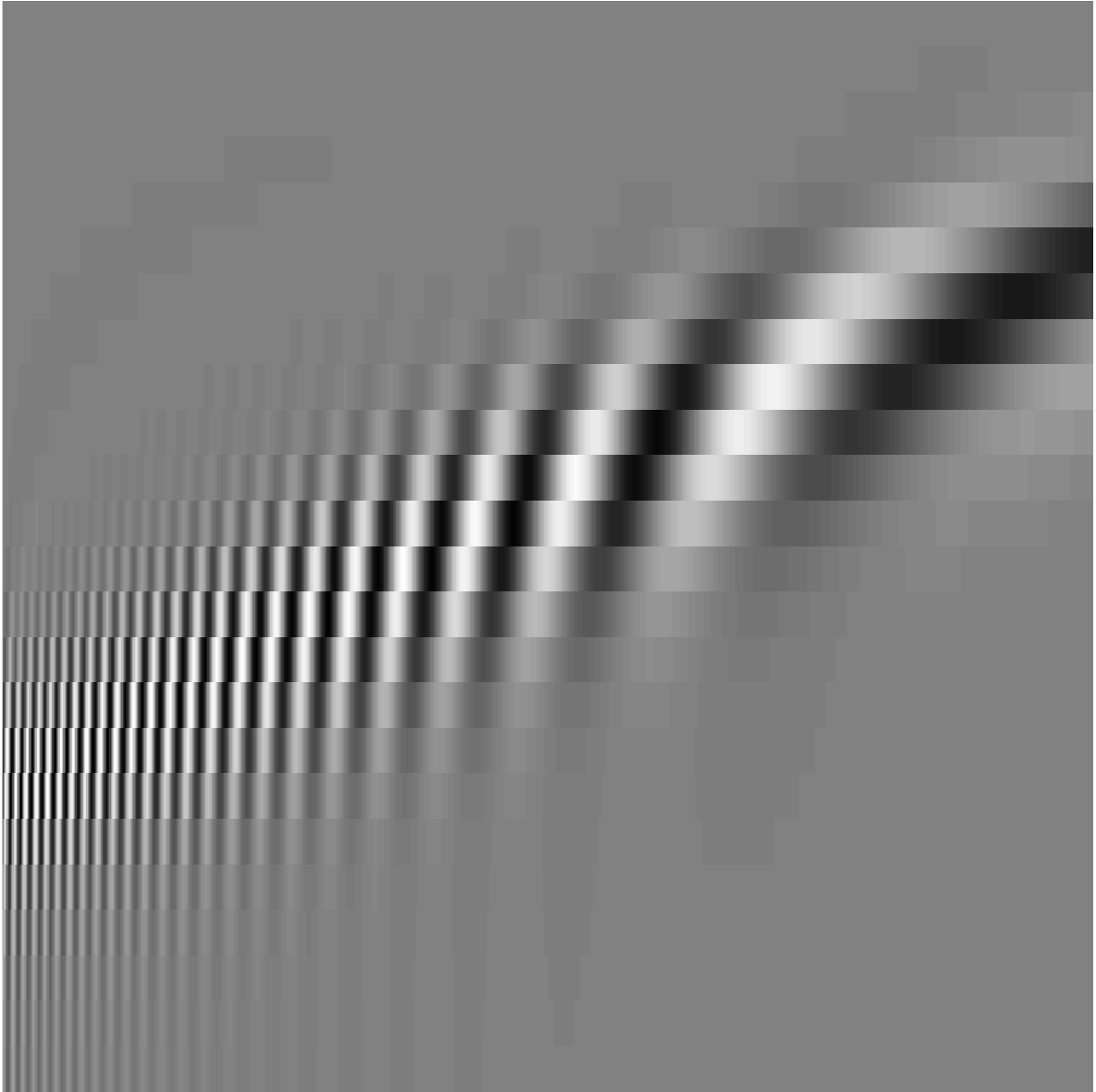} 
        & \includegraphics[width=0.35\textwidth]{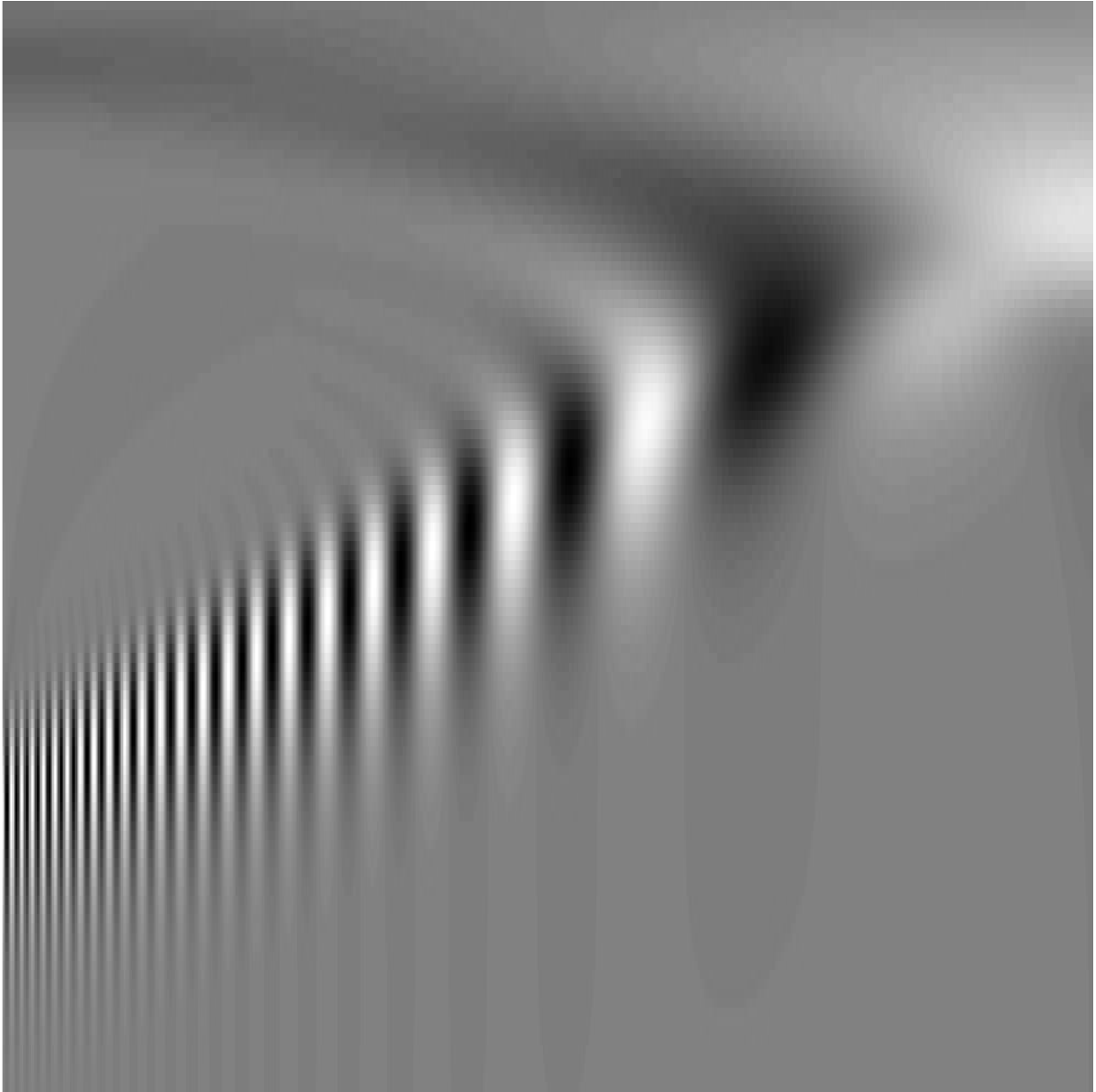}
   \end{tabular}
  \end{center}
  \caption{Temporal scale-space representation of the scale-normalized
  second-order temporal derivative $-L_{\zeta\zeta}$ computed from a synthetic sine
    wave signal $f(t) = \sin(\exp((b-t)/a))$ for $a = 200$ and $b =
    1000$ with temporally
    varying frequency so that the wavelength increases with time $t$.
    (left) Using the time-causal temporal scale-space concept corresponding to
    convolution with the time-causal limit kernel for $c = \sqrt{2}$.
    (right) Using the non-causal Gaussian temporal scale-space
    concept.
    For both temporal scale-space concepts, the scale-normalized
    temporal derivatives have been defined using $l_p$-normalization
    for $p = 1$ and corresponding to $\gamma = 1$. Note how the notion
    of scale-normalized temporal derivatives implies that stronger responses
    are obtained at temporal scale levels proportional to the duration
    of the underlying structures in the temporal signal.
    (Horizontal axis: time $t$.) (Vertical axis: effective temporal
    scale $\log \tau$.) 
    }
  \label{fig-expsine-negLtt-recfiltlogscvar-discgaussvar}
\end{figure*}

The bottom rows in
Figure~\ref{fig-expsine-tempscspextr-recfiltlogscvar-discgaussvar}
shows a one-dimensional model signal having a temporally varying frequency of
the form
\begin{equation}
  f(t) = \sin\left(\exp\left(\frac{b-t}{a}\right)\right)
\end{equation}
defined such that the local wavelength increases with time $t$.
Figure~\ref{fig-expsine-negLtt-recfiltlogscvar-discgaussvar} shows the
temporal scale-space representation of the scale-normalized 
second-order temporal $-L_{\zeta\zeta}$ as function of scale for the
maximally scale-invariant choice of $\gamma = 1$ corresponding to $p = 1$.
Note how structures in the signal of longer temporal duration give
rise responses at coarser temporal scales in agreement with the
derived theoretical properties of the scale-covariant scale-space
concepts over a time-causal {vs.\/}\ a non-causal temporal domain.

In the top rows in
Figure~\ref{fig-expsine-tempscspextr-recfiltlogscvar-discgaussvar} we
show the results of applying local temporal scale selection by detecting local maxima over
both time and scale of the scale-normalized second-order temporal derivative
$-L_{\zeta\zeta}$ for the scale-calibrated choice of $\gamma = 3/4$
corresponding to $p = 2/3$. Each detected scale-space extremum has
been marked by a
star at the point in scale-space $(\hat{t}, \hat{\sigma}) = (\hat{t}, \sqrt{\hat{\tau}})$
at which the local maximum over temporal scale was assumed.
Such temporal scale selection has been performed using two types of
temporal scale-space concepts:
(i)~based on the time-causal scale-space representation corresponding
to convolution with the scale-invar\-iant limit kernel with distribution 
parameter $c = \sqrt{2}$ approximated by a finite number of the at 
least $K = 8$ slowest primitive smoothing steps at the finest level of scale or
(ii)~based on the non-causal Gaussian kernel using 5 temporal scale
levels per scale octave.

The discrete implementation of the time-causal temporal scale-space
representation corresponding to convolution with the scale-invariant
limit kernel has been based on recursive filters over time
according to the methodology described in Lindeberg
\cite[section~6.2]{Lin16-JMIV} whereas the discrete implementation
of the non-causal Gaussian temporal scale-space concept has been based
on the discrete analogue of the Gaussian kernel described in Lindeberg
\cite{Lin90-PAMI,Lin93-JMIV}.
Discrete implementation of scale-normal\-ized temporal derivatives has
in turn been based on discrete $l_p$-normalization according to the
methodology outlined in \cite[section~7]{Lin16-JMIV}.

As can be seen from the results in the top rows in
Figure~\ref{fig-expsine-tempscspextr-recfiltlogscvar-discgaussvar},
the temporal scale estimates $\hat{\sigma}$ increase proportional to
the local wavelength $\lambda$ according to the derived scale-invariant
properties of the temporal scale-space concepts based on convolution with
the time-causal limit kernel or the non-causal Gaussian kernel.
The scale estimates obtained using the time-causal {\em vs.\/} the
non-causal temporal scale space concepts are, however, not equal in units of the
standard deviation $\hat{\sigma}$ of the temporal scale-space kernel
--- see also the theoretical analysis in
section~\ref{sec-scsel-prop-time-caus-trunc-exp-log-distr} with
specifically the numerical comparisons in
Table~\ref{tab-sc-est-sine-wave-limit-kern-ratios}.
Thus, different scale calibration factors are needed to transform the
temporal scale estimates in units of the temporal standard deviation
to units of the temporal duration of the signal for the time-causal
{\em vs.\/} the non-causal Gaussian temporal scale-space concepts ---
compare with the theoretical analysis in
appendix~\ref{sec-scaletime-approx-limit-kernel}.

\begin{figure}
  \begin{center}
    \begin{tabular}{c}
       {\small\em Post-filtered scale-space extrema from the time-causal limit kernel} \\
 \includegraphics[width=0.45\textwidth]{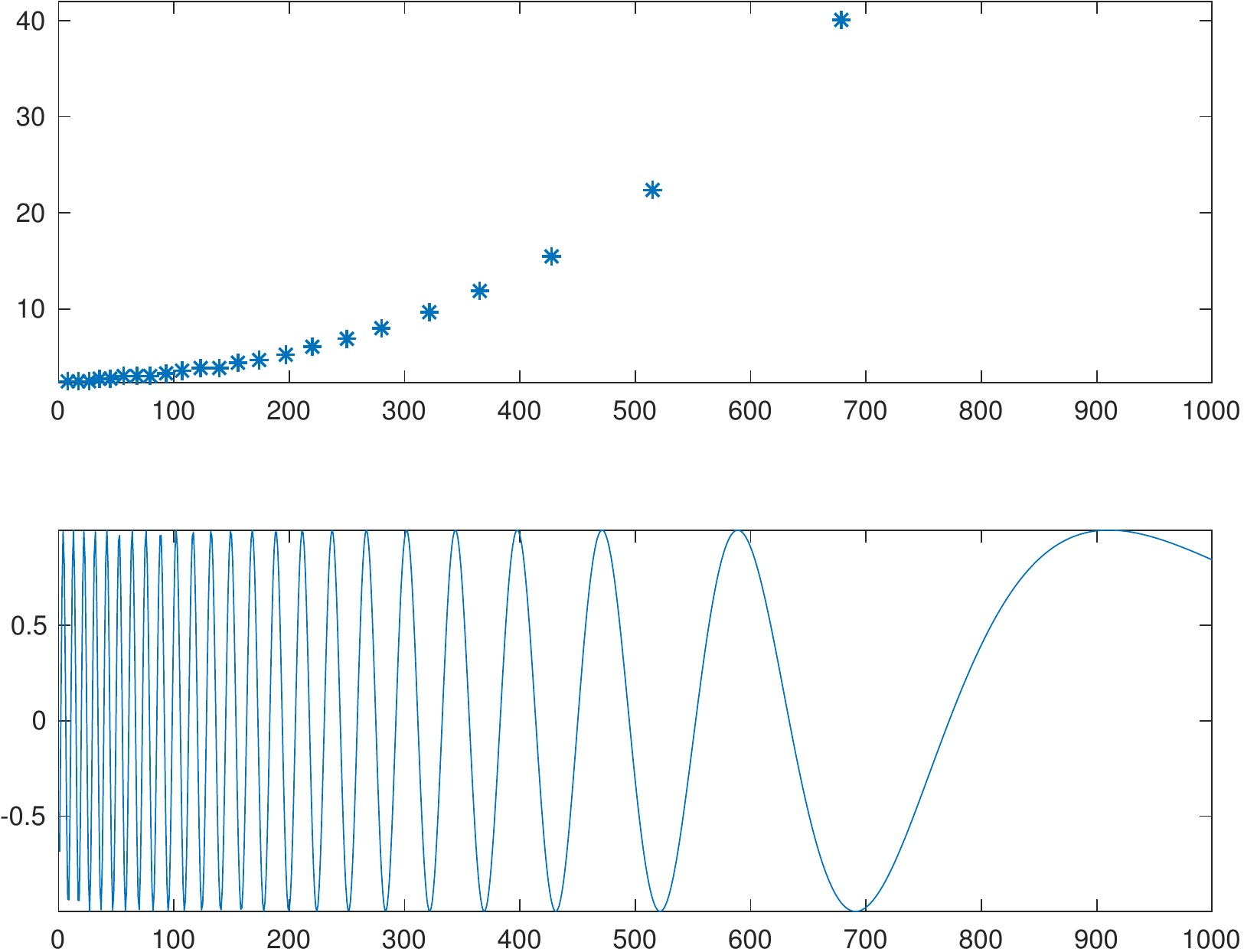} 
   \end{tabular}
  \end{center}
  \caption{The result of post-filtering the scale-space extrema shown
    in
    Figure~\ref{fig-expsine-tempscspextr-recfiltlogscvar-discgaussvar}(left)
    to result in a single extremum over temporal scales for each
    underlying temporal structure.}
  \label{fig-expsine-tempscspextr-recfiltlogscvar-postfilt}

   \medskip

  \begin{center}
    \begin{tabular}{c}
       {\small\em Delay compensated scale-space extrema from the time-causal limit kernel} \\
 \includegraphics[width=0.45\textwidth]{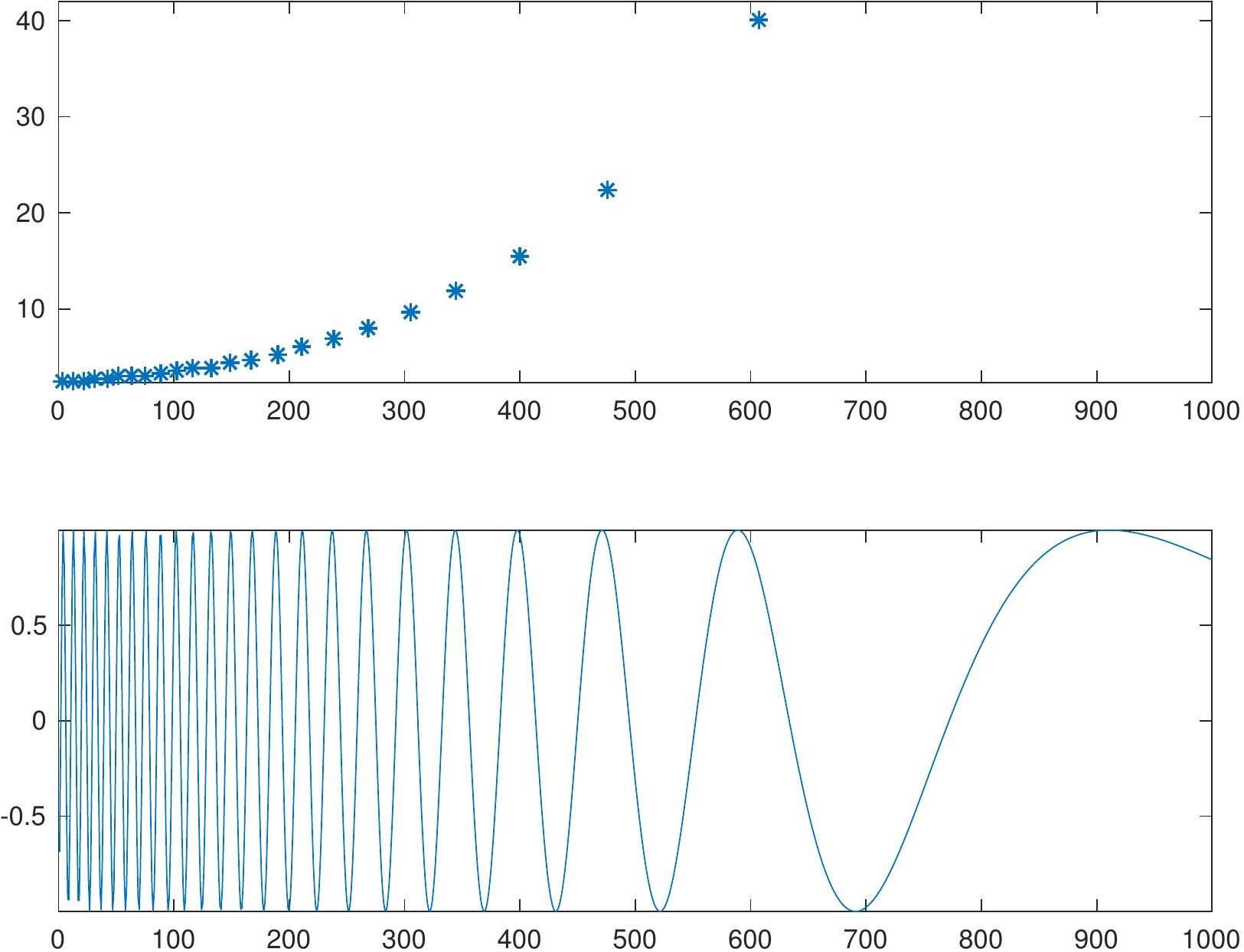} 
   \end{tabular}
  \end{center}
  \caption{The result of performing temporal delay compensation of the scale-space extrema shown
    in
    Figure~\ref{fig-expsine-tempscspextr-recfiltlogscvar-postfilt} by
    correcting the temporal moment $\hat{t}$ of the scale-space
    extremum by an estimate of the temporal delay computed from the
    position of the temporal maximum of the temporal scale-space
    kernel according to \cite[Section~4]{Lin16-JMIV}. Note that
    although the computations by necessity have to be associated with
    a temporal delay, we can nevertheless compute a good estimate of
    when the underlying event occurred that gave rise to the feature response.}
  \label{fig-expsine-tempscspextr-recfiltlogscvar-postfilt-delaycomp}
\end{figure}

\begin{figure}
  \begin{center}
    \begin{tabular}{cc}
       {\small\em Time-causal peak $\sigma_0 = 32$} 
      & {\small\em Non-causal peak $\sigma_0 = 32$} \\
 \includegraphics[width=0.22\textwidth]{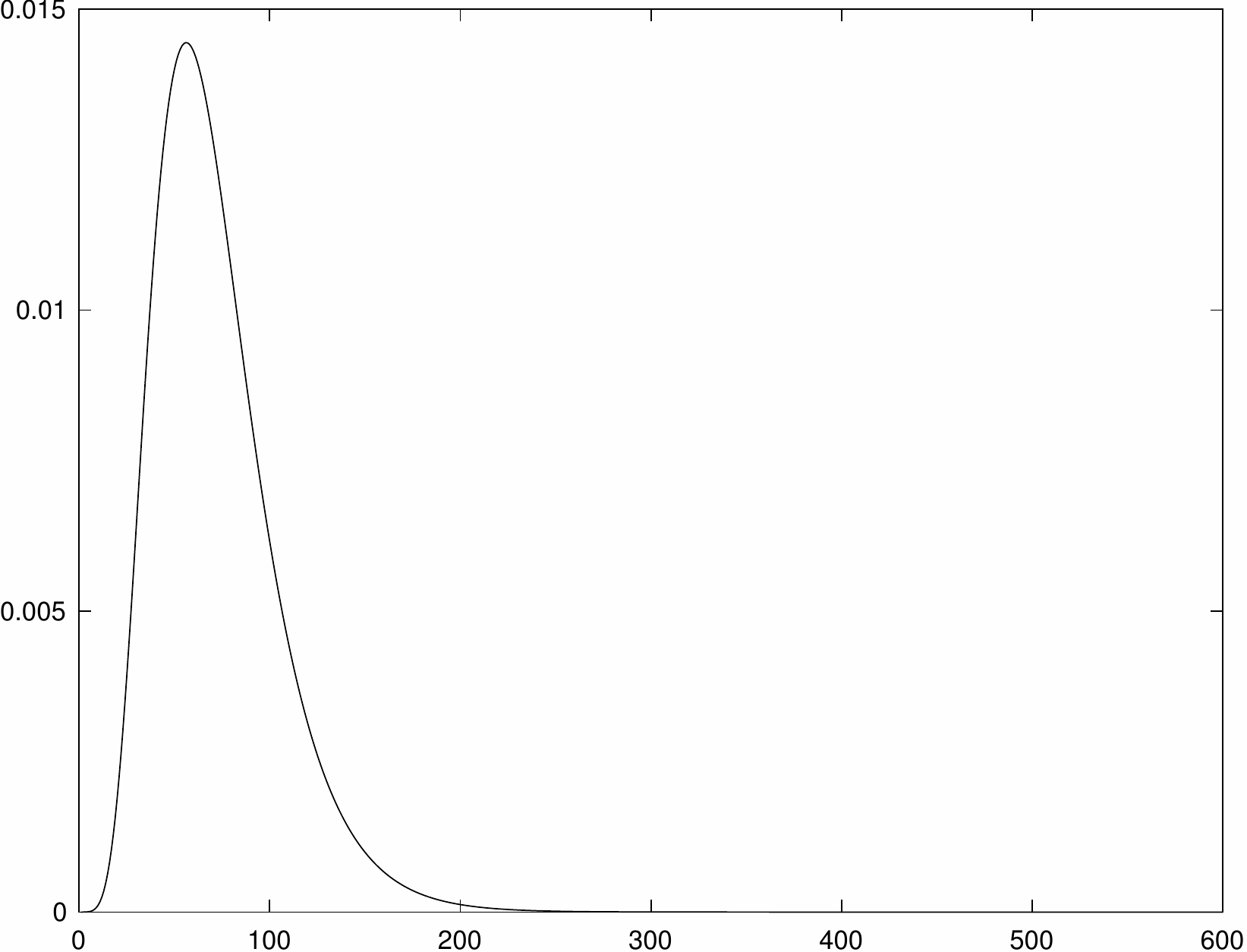} &
 \includegraphics[width=0.22\textwidth]{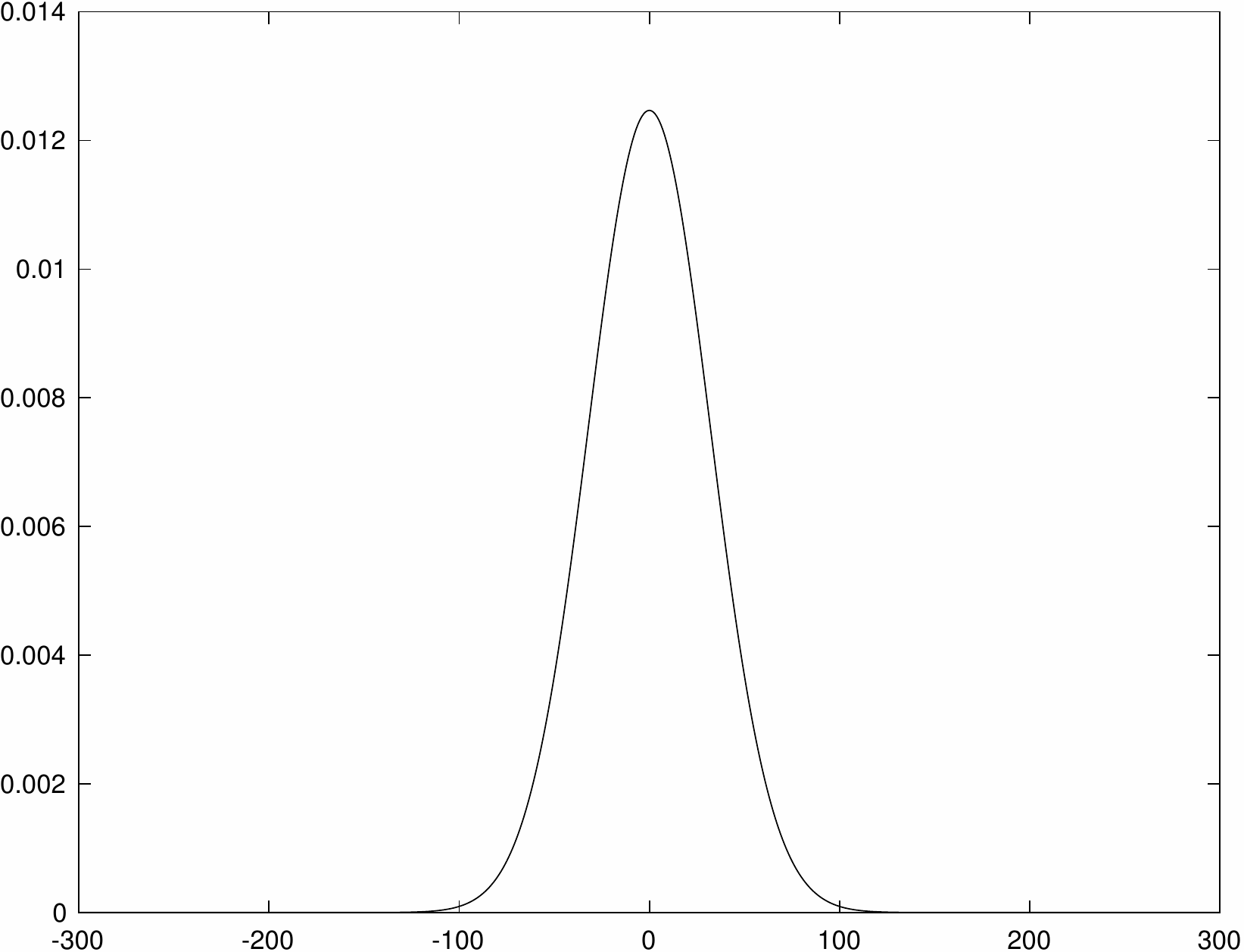}
      \\
      $\,$ \\
       {\small\em Time-causal scale-space $-L_{\zeta\zeta}$} 
      & {\small\em Non-causal scale-space $-L_{\zeta\zeta}$} \\
 \includegraphics[width=0.20\textwidth]{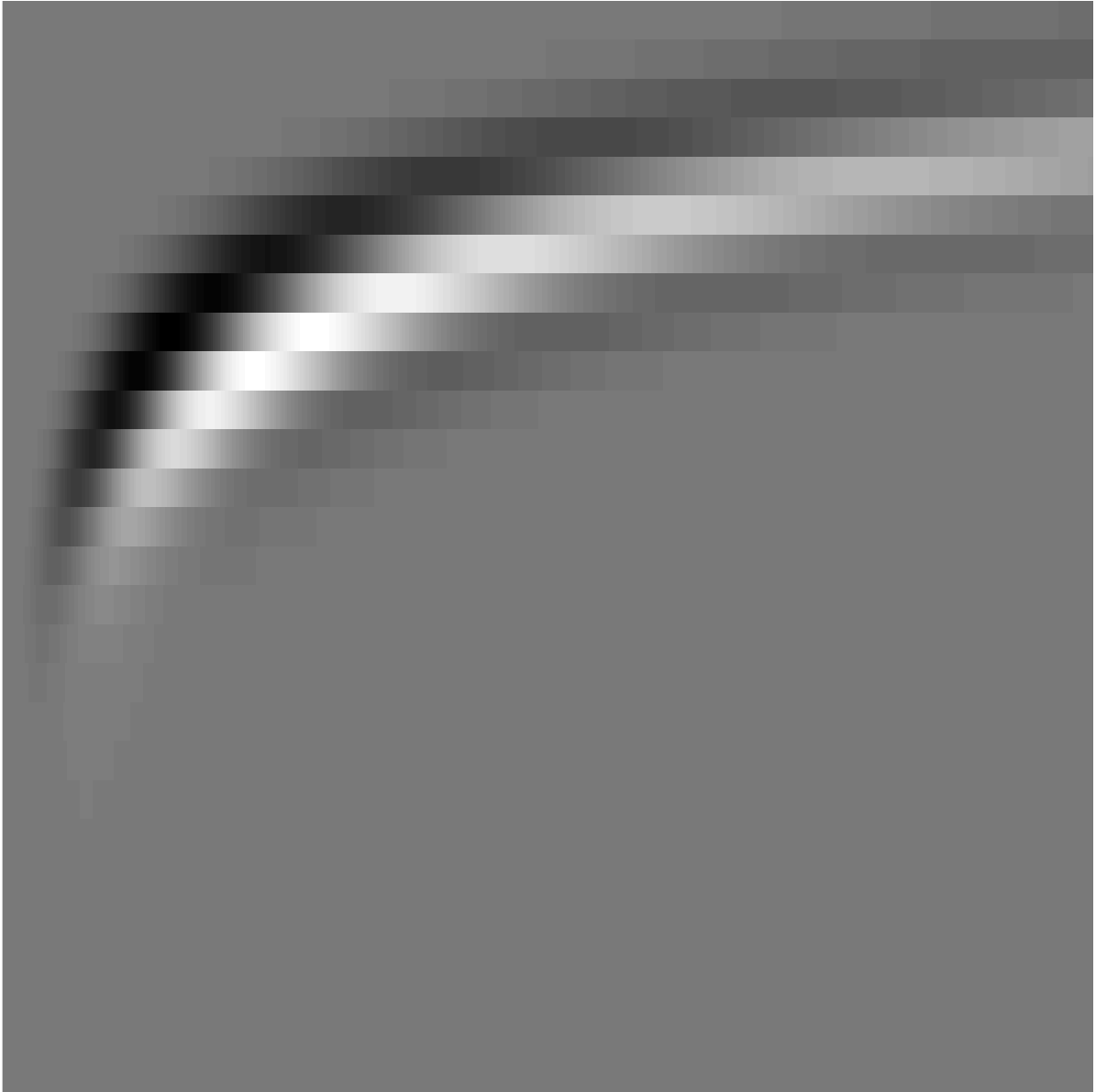} &
 \includegraphics[width=0.20\textwidth]{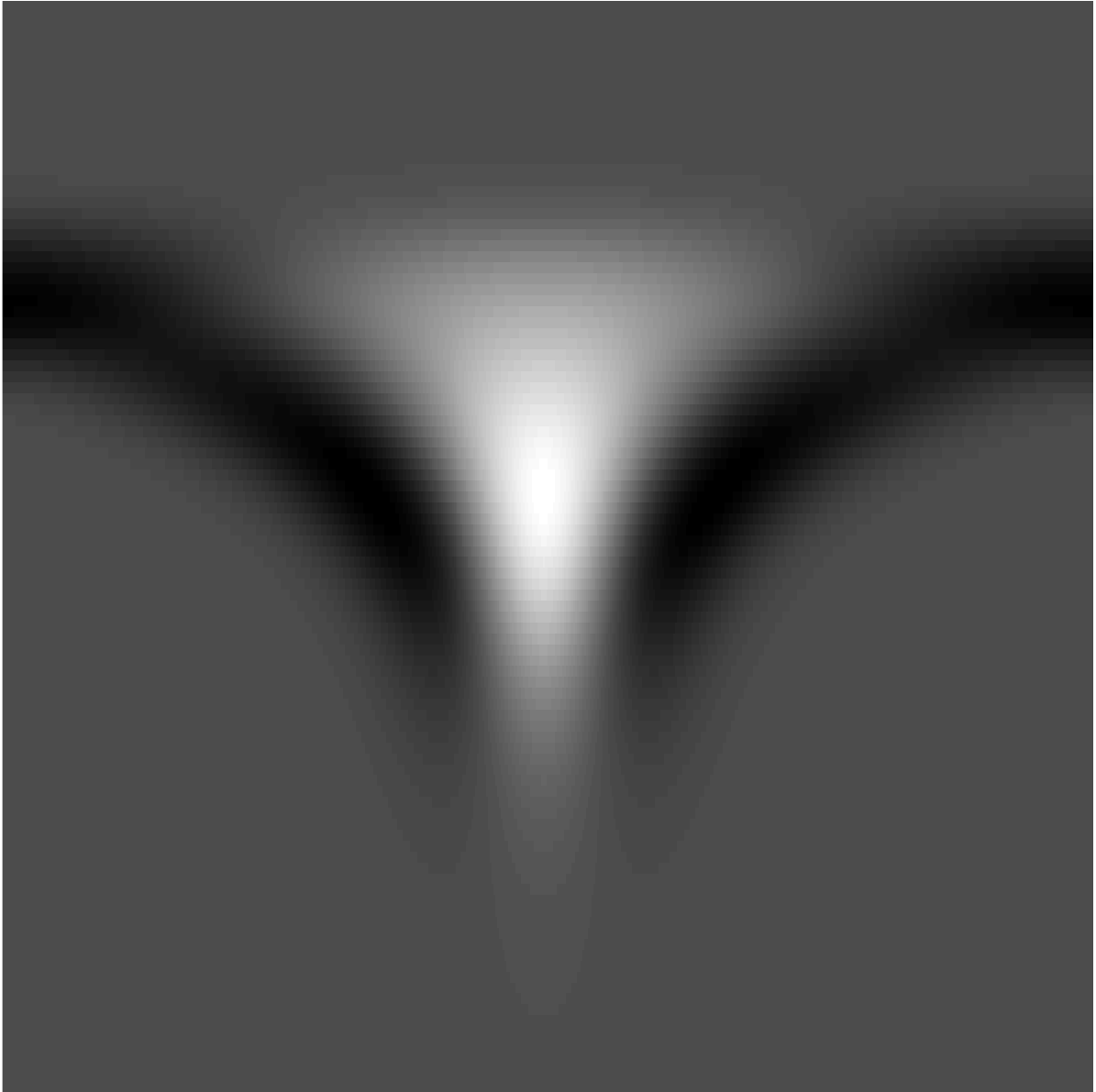} \\
   \end{tabular}
  \end{center}
  \caption{Illustration of the temporal dynamics over multiple
    temporal scales that arises when a temporal scale-space representation
    responds to a temporal peak with temporal standard deviation
    $\sigma_0 = 32$ for (left column) the time-causal
    representation corresponding to convolution with the time-causal
    limit kernel and (right column) the non-causal Gaussian temporal
    scale space. 
    (top row) Graphs of the input kernels for the cases of a
    time-causal peak and a non-causal peak, respectively. 
    (bottom row) Scale-space maps of the scale-normalized second-order
  temporal derivative $-L_{\zeta\zeta}$.
    Note that if we slice the time-causal scale-space map vertically
    at a temporal moment before the full development of the temporal
    scale-space maximum at the temporal scale corresponding to the
    temporal scale of the peak, we will get earlier temporal responses
  at finer temporal scales, whereas if we slice the time-causal
  scale-space map at a temporal moment after the full development of
  the temporal scale-space maximum, we will get later temporal
  responses at coarser temporal scales. When handling multiple
  temporal scale levels in a time-causal real-time situation, it it
  therefore natural to include explicit mechanisms for tracking and
  handling how the
  temporal structures evolve over temporal scales. (Horizontal axis:
  time $t$.) (Vertical axis in bottom row: effective temporal scale
  $\log \tau$.)}
  \label{fig-temp-dyn-time-caus-peak}
\end{figure}

\begin{figure*}
  \begin{center}
    \begin{tabular}{cc}
       {\small\em Scale-space extrema from the time-causal limit kernel} 
      & {\small\em Scale-space extrema from the non-causal Gaussian kernel} \\
 \includegraphics[width=0.48\textwidth]{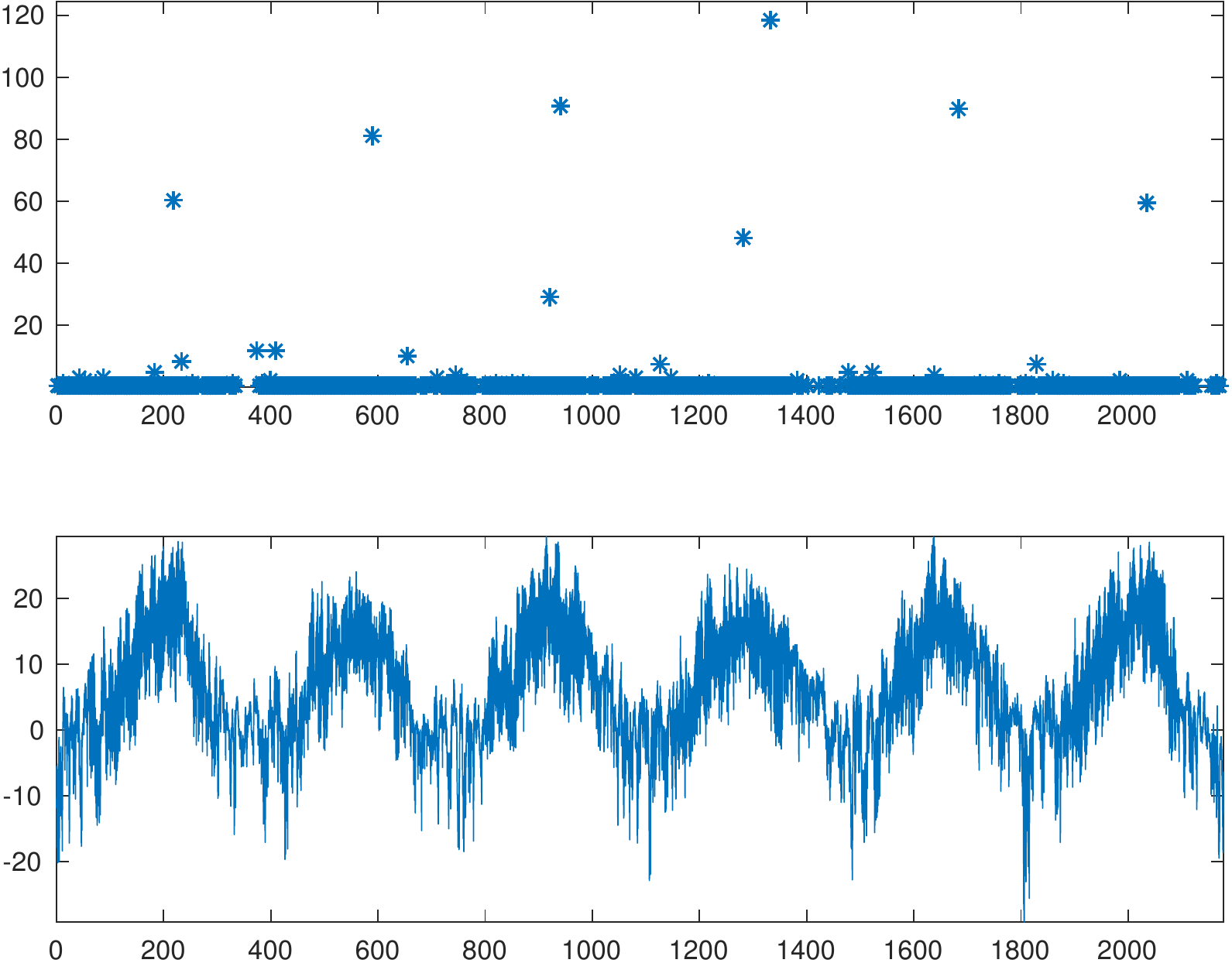} &
 \includegraphics[width=0.48\textwidth]{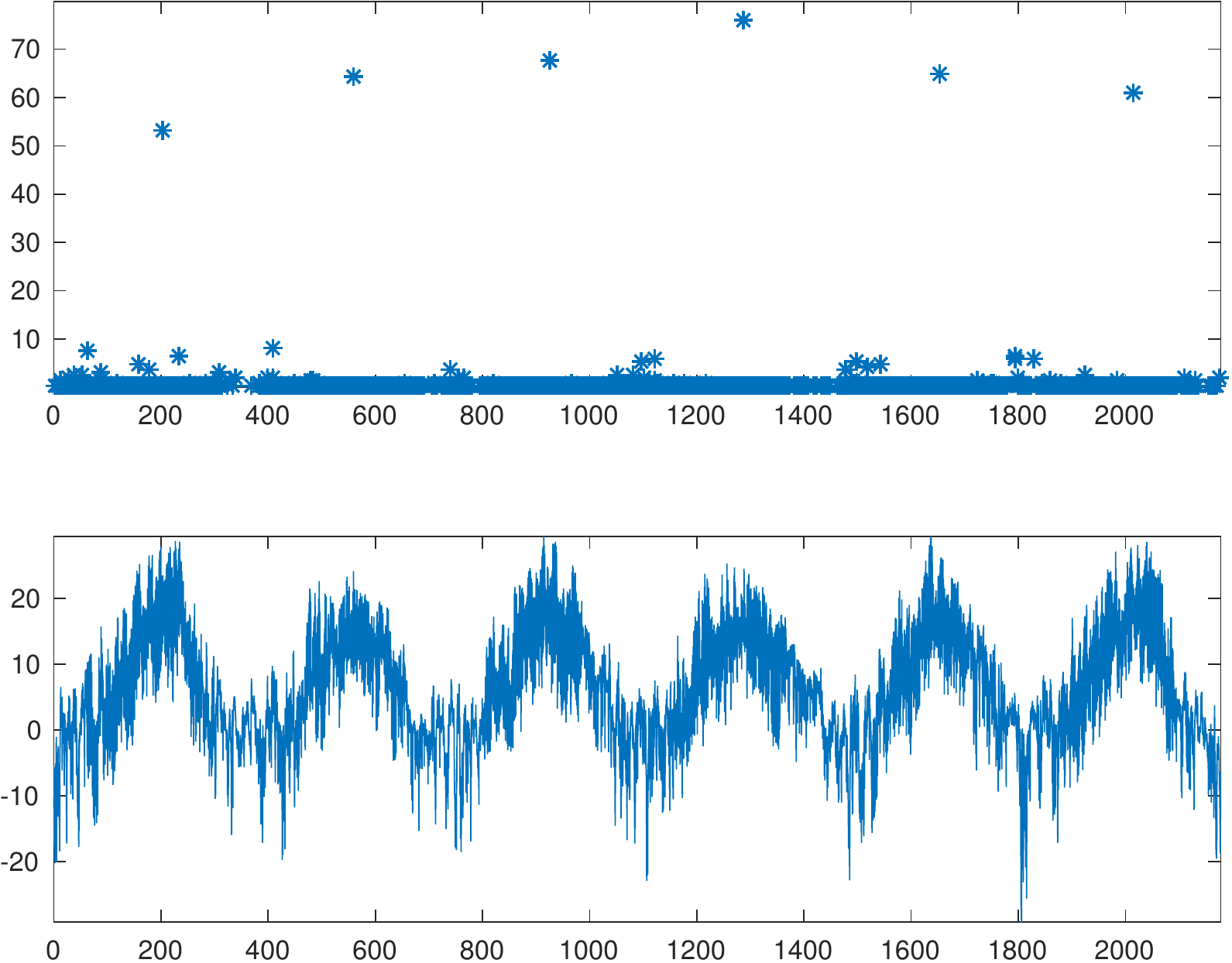} \\
   \end{tabular}
  \end{center}
  \caption{Temporal scale selection by scale-space extrema detection
    applied to a measured
    temperature signal (showing hourly temperature measurements at
    Tullinge outside Stockholm during the years 1997-2002, open data courtesy of the
    Swedish Meteorological and Hydrological Institute, SMHI, in Sweden). 
(top left) Temporal scale-space maxima of the scale-normalized
    second-order temporal derivative $-L_{\zeta\zeta}$ detected using the
    time-causal temporal scale-space concept corresponding to
    convolution with the time-causal limit kernel for $c = \sqrt{2}$
    and with each scale-space maximum marked at
    the delay-compensated point $(\hat{t}-\delta, \hat{\sigma})$ with $\hat{\sigma} = \sqrt{\hat{\tau}}$ 
    at which the scale-space maximum is assumed with the temporal
    delay estimate $\delta$ determined from the temporal location of
    the peak of the corresponding temporal scale-space kernel at the
    given scale.
    (top right) Temporal scale-space maxima detected using the
    non-causal Gaussian temporal scale-space concept using
    5 temporal scale levels per scale octave.
    For both temporal scale-space concepts, the scale-normalized
    temporal derivatives have been defined using discrete $l_p$-normalization
    for $p = 2/3$ corresponding to $\gamma = 3/4$ for second-order temporal derivatives.
Note how the temporal scale selection method is able to extract the coarse
  scale temporal phenomena although the signal contains very strong 
  variations at finer temporal scales. (Horizontal axis: time $t$ in units of
  days.) (Vertical axis in upper figure: temporal scale estimates in
  units of days.)}
  \label{fig-tullinge-temp-tempscspextr-recfiltlogscvar-postfilt}

\bigskip

  \begin{center}
    \begin{tabular}{cc}
        {\small\em Quasi quadrature $\sqrt{{\cal Q}_t L}$ computed with time-causal limit kernel}
        & {\small\em Quasi quadrature $\sqrt{{\cal Q}_t L}$ computed with non-causal Gaussian kernel} \\
\includegraphics[width=0.35\textwidth]{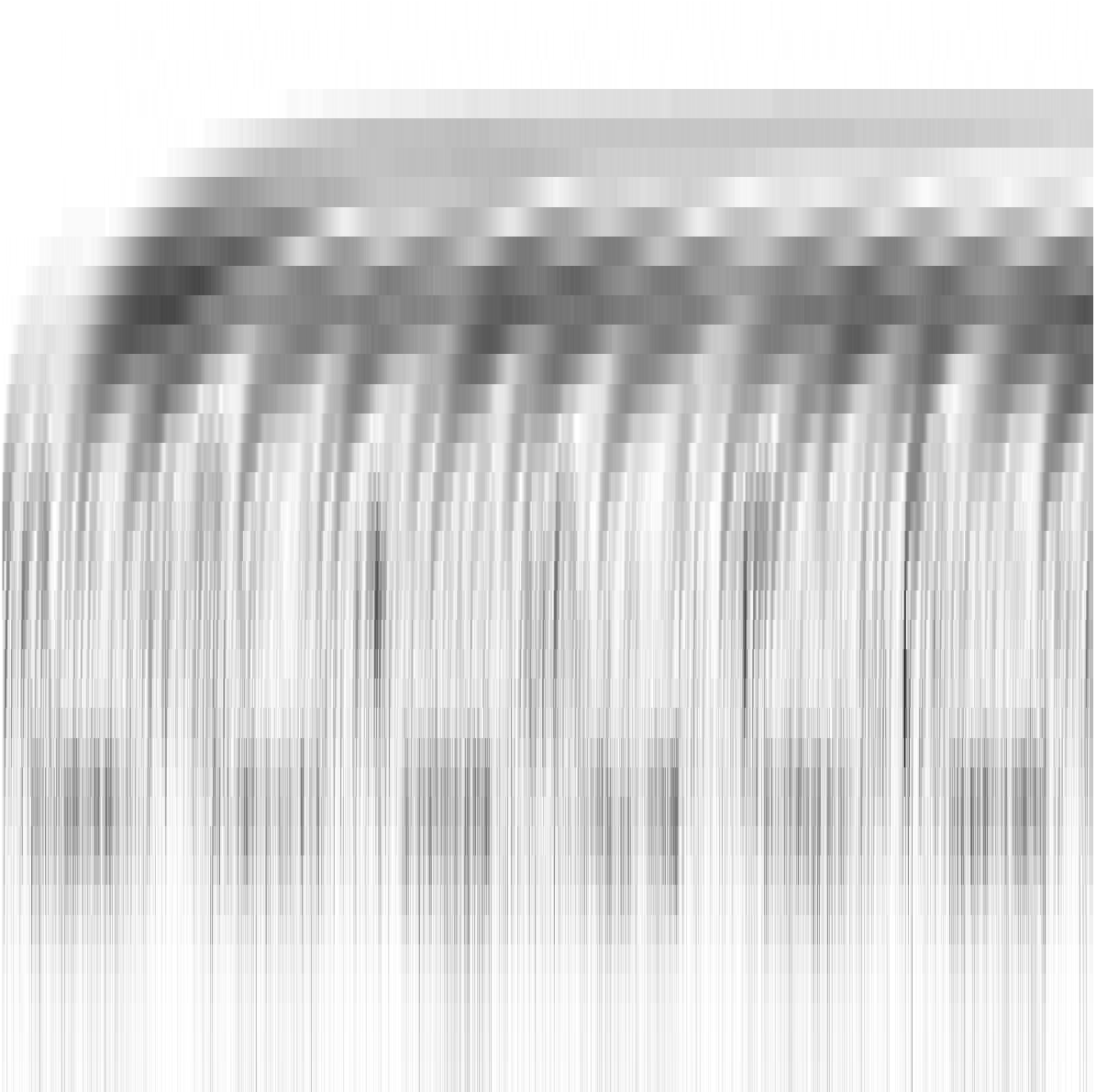} 
        & \includegraphics[width=0.35\textwidth]{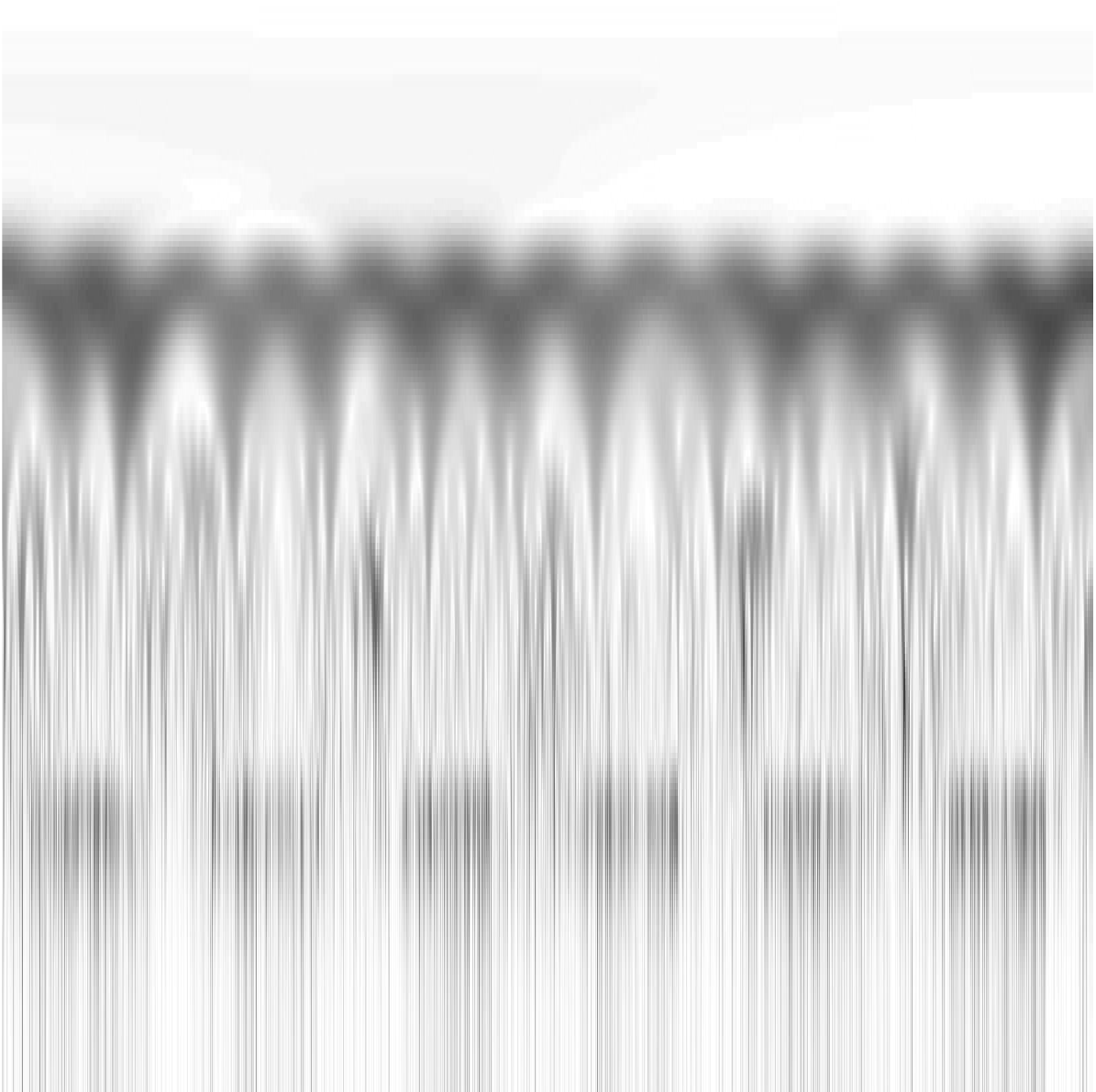}
   \end{tabular}
  \end{center}
  \caption{Temporal scale-space representation of the scale-normalized
  quasi quadrature measure ${\cal Q}_t L = (L_{\zeta}^2 + C L_{\zeta\zeta}^2)/t^{\Gamma}$ 
    for $C = 1/\sqrt{(1 - \Gamma))(2 - \Gamma)}$, $\gamma = 1$ and
    $\Gamma = 0$ computed from the temperature signal in
    Figure~\ref{fig-tullinge-temp-tempscspextr-recfiltlogscvar-postfilt}.
    (left) Using the time-causal temporal scale-space concept corresponding to
    convolution with the time-causal limit kernel for $c = \sqrt{2}$.
    (right) Using the non-causal Gaussian temporal scale-space
    concept.
    For both temporal scale-space concepts, the scale-normalized
    temporal derivatives have been defined using discrete $l_p$-normalization.
    Note how this operator responds to different types of temporal
    structures at different temporal scales with a particularly strong
    response due to temporal variations caused by the annual temperature cycle.
    (Horizontal axis: time $t$.) (Vertical axis: effective temporal
    scale $\log \tau$.)}
    \label{fig-tullinge-temp-quasi-recfiltlogscvar-discgaussvar}
\end{figure*}

Another qualitative difference that can be noted with regard to
temporal scale selection in a time-causal temporal scale-space
representation {\em vs.\/} a non-causal Gaussian temporal scale-space
representation is that any measurement performed in a time-causal
temporal scale-space concept is associated with an inherent
temporal delay $\delta$, whereas the temporal delay can be defined to be zero
for a non-causal Gaussian temporal scale-space. Thus, the local
extrema over scales will be assumed later with increasing temporal
scale levels as induced by temporal structures of having longer
temporal duration. 
Varying the parameter $q$ in the scale calibration criteria
(\ref{eq-sc-est-temp-peak}) and (\ref{eq-sc-est-onset-ramp}) to values
of $q < 1$ provides a straightforward way of enforcing responses to be
obtained at finer temporal scales and thereby implying shorter
temporal delays, at the potential cost of a larger
likelihood of false positive responses by not detecting the underlying
temporal structures at the same temporal scales as they occur.

Yet another side effect of the longer temporal delays at coarser
temporal scales is that
multiple local responses over scales may be obtained with respect to
the same underlying temporal structure, with first responses obtained
at finer temporal scales followed by later responses at coarser temporal scales. 
Because of the temporal shift caused by different temporal delays
between adjacent temporal scale levels, scale-space extrema detection
over a local $3 \times 3$ neighbourhood may not detect a single
extremum over temporal scales as for the non-causal Gaussian temporal
scale-space concept. Therefore, explicit handling of different
temporal delays at different temporal scale levels is needed when
performing temporal scale selection in a time-causal temporal
scale-space representation.
The presented theory of temporal scale selection properties is
intended to be generally applicable with respect to different such
strategies for handling the temporal delays in specific algorithms.

\subsection{Post-filtering of responses at adjacent temporal scales}

In view of the behaviour of image structures over temporal scales
illustrated in
Figure~\ref{fig-expsine-negLtt-recfiltlogscvar-discgaussvar}, 
one way of suppressing multiple responses to the same underlying
structure at different temporal scales is by performing an additional
search around each scale-space extremum as follows: If a point
$(\hat{t}, \hat{\tau})$ at temporal scale level with temporal scale index $k$ is a
scale-space maximum, perform an additional search at the nearest finer
temporal scale level $k-1$ to previous temporal
moments $t-j$ as long as the scale-normalized values monotonically increase.
When the monotone increase stops and a local temporal maximum has been found,
then compare if the temporal maximum value at the nearest finer temporal scale is
greater than the temporal maximum value at the current scale. If so, suppress
the scale-space maximum at the current scale.
In a corresponding manner, perform a search at the next coarser
temporal scale $k+1$ to the following temporal moments $t+j$ as long 
as the scale-normalized values monotonically increase.
When the mono\-tone increase stops and a local temporal maximum has been found,
then compare if the temporal maximum value at the next coarser temporal scale
is greater than the temporal maximum value at the current scale. If so,
suppress the scale-space maximum at the current scale.
By this type of straightforward scale-space tracking over adjacent temporal scales, 
a single response will be obtained to the same underlying
structure as illustrated in
Figure~\ref{fig-expsine-tempscspextr-recfiltlogscvar-postfilt}. 
Additionally, a more accurate temporal scale estimate can be computed by performing
the parabolic interpolation according to (\ref{eq-parabol-interpol})
and (\ref{eq-parabol-interpol-subresolution-estimate}) over the
nearest backward and forward temporal maxima at the adjacent finer and coarser
temporal scales as opposed to an interpolation over temporal scales at
the same temporal moment as at which the
temporal scale-space maximum was assumed.

Note that in a real-time situation, the necessary information needed
to perform a search to the past can be stored by a process that
records the value of temporal maxima at each temporal scale.
In a corresponding manner, later deletion or subresolution
interpolation of the scale estimate of a scale-space maximum can only
be performed when time has passed to the location of the next
temporal maximum at the nearest coarser temporal scale.
If a preliminary feature response has been obtained at any temporal
scale while a potential response at the nearest coarser temporal scale
is not available yet because of its longer temporal scale, a real-time
system operating over multiple scales should preferably be designed
with the ability to correct or adjust preliminary measurements when
more information at coarser scales becomes available --- see
also Figure~\ref{fig-temp-dyn-time-caus-peak} for an illustration of
the underlying temporal dynamics that arises when processing signals
at multiple scales using a time-causal temporal scale-space representation.

While these descriptions have been given regarding scale-space maxima
and local maxima, the procedure for handling scale-space minima and local minima is analogous with the polarity
of the signal reversed.

\subsection{Temporal delay compensation}
 
In Figure~\ref{fig-expsine-tempscspextr-recfiltlogscvar-postfilt-delaycomp}
we have additionally adjusted the temporal location of every temporal
scale-space maximum by an estimate of the temporal delay $\delta$ computed from
the location of the temporal maximum of the underlying temporal
scale-space kernel according to \cite[Section~4]{Lin16-JMIV}. Note that although any temporal event detected at
a coarser temporal scale will by necessity be associated with non-zero temporal
delay, we can nevertheless retrospectively compute a good estimate of when the
underlying event occurred that gave rise to the registered feature response.

\subsection{Temporal scale selection for a real measurement signal}

While a main purpose of this article is to develop a theory of
temporal scale selection to be used in conjunction with a
spatio-temporal scale-space concept for video analysis or a
spectro-temporal scale-space concept for audio analysis, we argue that
this theory is applicable to much larger classes of time-dependent
measurement signals. For the
purpose of isolating the effect to a purely one-dimensional
measurement signal, we do in 
Figure~\ref{fig-tullinge-temp-tempscspextr-recfiltlogscvar-postfilt}
show the result of applying corresponding temporal scale selection to a
real measurement signal showing hourly measurements of the temperature
at a weather station. Note how the temporal scale selection method
based on scale-space extrema is
able to extract the coarse scale temperature peaks although the signal
contains substantial high-amplitude variations at finer scales.

For the purpose of having the local feature responses being less
dependent on the local phase of the signal than for either first- or
second-order temporal derivatives $L_{\zeta}$ or $L_{\zeta\zeta}$,
we do in
Figure~\ref{fig-tullinge-temp-quasi-recfiltlogscvar-discgaussvar} show
the result of computing a quasi quadrature measure
\begin{equation}
   {\cal Q}_t L = \frac{L_{\zeta}^2 + C L_{\zeta\zeta}^2}{t^{\Gamma}}
\end{equation}
derived in \cite{Lin16-spattempscsel} to for 
\begin{equation}
  C = \frac{1}{\sqrt{(1 - \Gamma))(2 - \Gamma)}}
\end{equation}
constitute an improved version of an earlier proposed quasi quadrature
measure in Lindeberg \cite{Lin97-AFPAC}.

\begin{figure}
  \begin{center}
    \begin{tabular}{cc}
        {\small\em $\sum_t {\cal Q}_t L$ time-causal} 
      & {\small\em $\sum_t {\cal Q}_t L$  non-causal} \\
\includegraphics[width=0.22\textwidth]{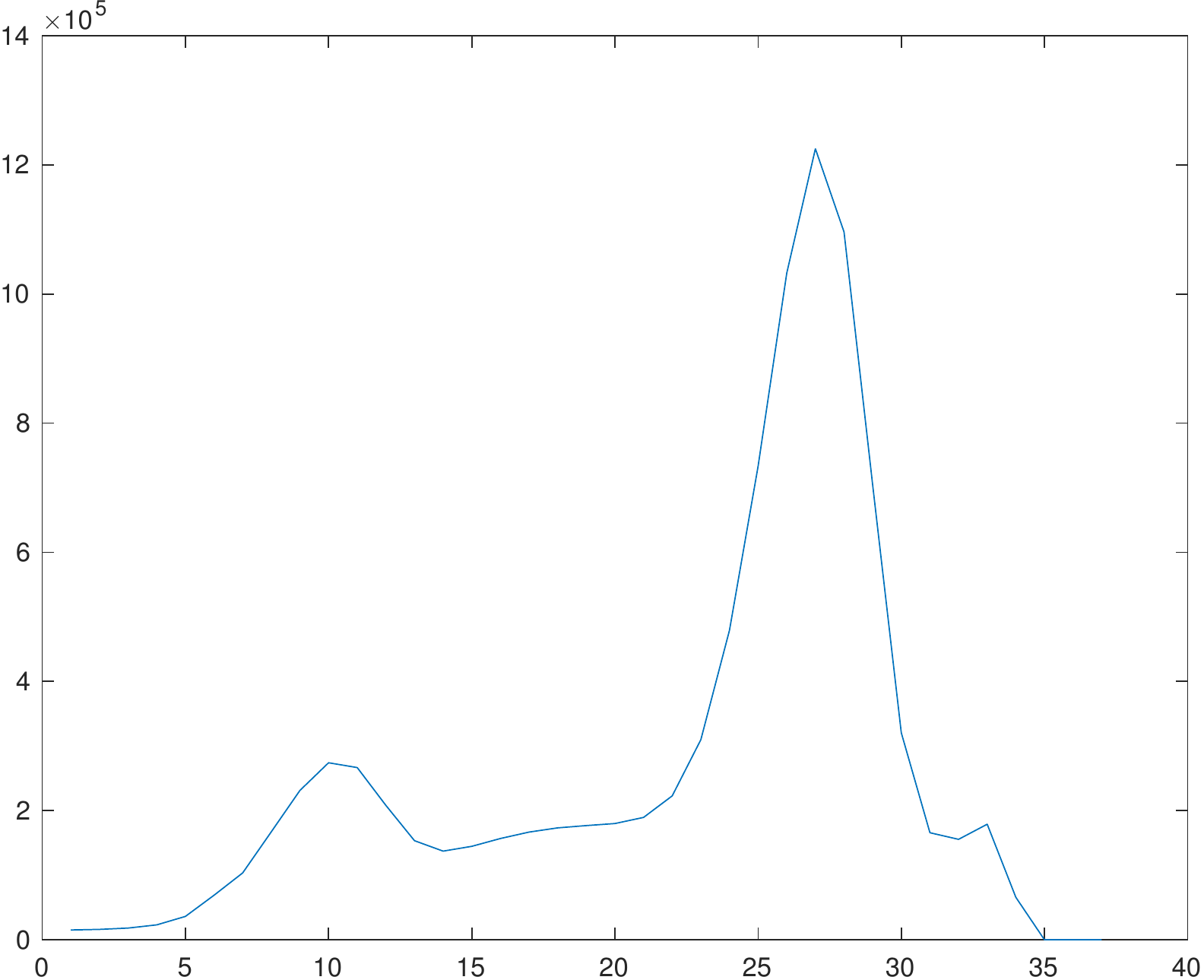} &
 \includegraphics[width=0.22\textwidth]{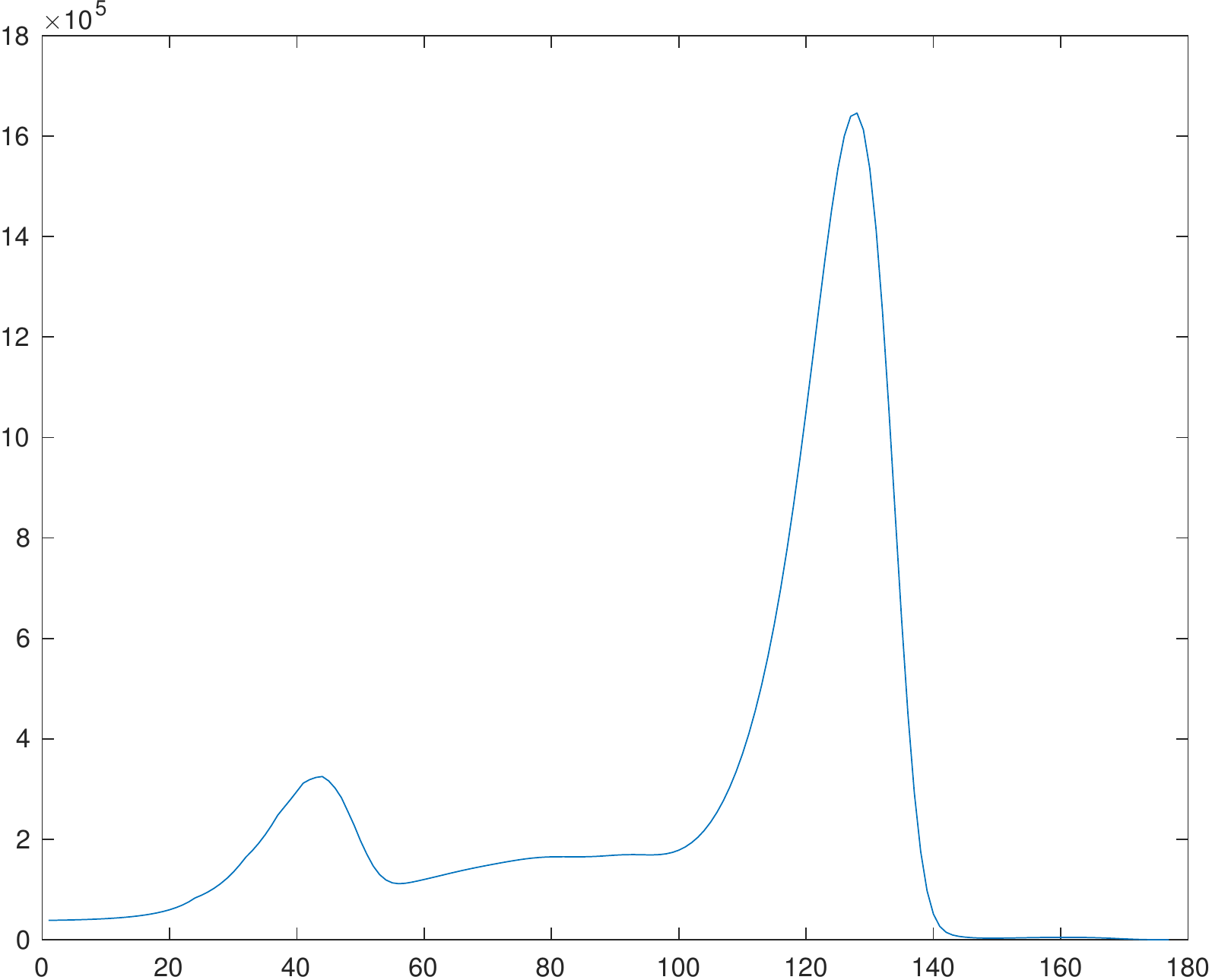} \\
   \end{tabular}
  \end{center}
  \caption{Graphs that for each temporal scale level $\sigma_k$ show
    the sum of the quasi quadrature
    responses in
    Figure~\ref{fig-tullinge-temp-quasi-recfiltlogscvar-discgaussvar}
    over time $t$. With a complementary stationarity assumption, the
    peaks over temporal scales reflect the temporal scale levels
    $\sigma_k$ at which there is the largest scale-normalized
    variability in terms of first- and second-order temporal
    derivative responses. For the time-causal model to the left the
    local maxima over temporal scale are assumed at temporal scale levels
    $\hat{\sigma}_1 = 0.207$~days and $\hat{\sigma}_2 = 71.5$ days,
    respectively. For the non-causal model to the right, the local
    maxima over temporal scales are assumed at temporal scale levels
    $\hat{\sigma}_1 = 0.236$~days and $\hat{\sigma}_2 = 85.2$ days.
    The ratios between these temporal scale levels are 332 for the
    time-causal model and 345 for the non-causal model in good
    qualitative agreement with the finer scale peaks corresponding to
    the daily cycle and the coarser scale peaks corresponding to the
    yearly cycle of 365 days. (Horizontal axis: temporal scale index
    $k$ corresponding to a uniform sampling in terms of effective
    temporal scale $\log \tau$.)}
   \label{fig-tullinge-temp-quasi-recfiltlogscvar-discgaussvar-sumovert}
\end{figure}

\begin{table}[!hbtp]
  \addtolength{\tabcolsep}{2pt}
  \begin{center}
    \footnotesize
    \begin{tabular}{cccccc}
    \hline
       \multicolumn{6}{c}{Temporal scale estimates from temporal peaks} \\
    \hline
       \multicolumn{3}{c}{$c = \sqrt{2}$} & \multicolumn{3}{c}{$c = 2$} \\
  \hline
       $\sigma_0$ & $\left.\hat{\sigma_0}\right|_{L_p}$ & $\left.\hat{\sigma_0}\right|_{var}$
       & $\sigma_0$ & $\left.\hat{\sigma_0}\right|_{L_p}$ & $\left.\hat{\sigma_0}\right|_{var}$ \\
  \hline
       2 &   2.26 &   2.14 &   2 &   1.58 &   2.16 \\
       4 &   4.11 &   4.05 &   4 &   4.57 &   4.06 \\
       8 &   7.76 &   8.06 &   8 &   8.02 &   7.96 \\
     16 & 15.90 & 16.03 & 16 & 15.60 & 15.86 \\
     32 & 31.97 & 32.06 & 32 & 31.35 & 31.66 \\
     64 & 64.12 & 64.13 & 64 & 63.15 & 63.31 \\
  \hline
 \end{tabular}
\end{center}
\caption{Experimental results of the accuracy of the temporal scale
  estimates when detecting temporal scale-space extrema of the
  scale-normalized derivative $-L_{\zeta\zeta}$ in a temporal
  peak defined as a time-causal limit
  kernel with temporal standard deviation $\sigma_0 = \sqrt{\tau_0}$
  for distribution parameter $c$ and then detecting the strongest
  temporal scale-space maximum with temporal scale estimate 
  $\hat{\sigma} = \sqrt{\hat{\tau}}$ also using the same value of the
  distribution parameter $c$ and for two ways of defining temporal
  scale-normalized derivatives by either (i)~discrete $l_p$-normalization or
  (ii)~variance-based normalization. 
}
  \label{tab-temp-sc-est-temp-peak}

  \bigskip

  \begin{center}
    \footnotesize
    \begin{tabular}{cccccc}
    \hline
       \multicolumn{6}{c}{Temporal scale estimates from temporal onset
      ramps} \\
    \hline
       \multicolumn{3}{c}{$c = \sqrt{2}$} & \multicolumn{3}{c}{$c = 2$} \\
  \hline
       $\sigma_0$ & $\left.\hat{\sigma_0}\right|_{L_p}$ & $\left.\hat{\sigma_0}\right|_{var}$
       & $\sigma_0$ & $\left.\hat{\sigma_0}\right|_{L_p}$ & $\left.\hat{\sigma_0}\right|_{var}$ \\
  \hline
       2 &   2.01 &   1.96 &   2 &   2.04 &   1.97 \\
       4 &   3.66 &   3.90 &   4 &   3.68 &   3.98 \\
       8 &   7.71 &   8.02 &   8 &   7.60 &   7.94 \\
     16 & 15.68 & 16.02 & 16 & 15.42 & 15.90 \\
     32 & 32.00 & 32.04 & 32 & 31.48 & 31.77 \\
     64 & 64.08 & 64.08 & 64 & 63.44 & 63.54 \\
  \hline
 \end{tabular}
\end{center}
\caption{Experimental results of the accuracy of the temporal scale
  estimates when detecting temporal scale-space extrema of the
  scale-normalized derivative $L_{\zeta}$ in a temporal onset ramp
  defined as the primitive function of a time-causal limit
  kernel with temporal standard deviation $\sigma_0 = \sqrt{\tau_0}$
  for distribution parameter $c$ and then detecting the strongest
  temporal scale-space maximum with temporal scale estimate 
  $\hat{\sigma} = \sqrt{\hat{\tau}}$ also using the same value of the
  distribution parameter $c$ and for two ways of defining temporal
  scale-normalized derivatives by either (i)~discrete $l_p$-normalization or
  (ii)~variance-based normalization. 
  }
  \label{tab-temp-sc-est-temp-onset-ramp}

  \bigskip
  \begin{center}
    \footnotesize
    \begin{tabular}{cccccc}
    \hline
       \multicolumn{6}{c}{Temporal scale estimates from non-causal
      Gaussian scale space} \\
    \hline
       \multicolumn{3}{c}{temporal peak} & \multicolumn{3}{c}{temporal
                                           onset ramp} \\
  \hline
       $\sigma_0$ & $\left.\hat{\sigma_0}\right|_{L_p}$ & $\left.\hat{\sigma_0}\right|_{var}$
       & $\sigma_0$ & $\left.\hat{\sigma_0}\right|_{L_p}$ & $\left.\hat{\sigma_0}\right|_{var}$ \\
  \hline
       2 &   2.95 &   1.92 &   2 &   1.59 &   2.07 \\
       4 &   4.04 &   3.97 &   4 &   3.86 &   4.03 \\
       8 &   8.10 &   7.98 &   8 &   7.92 &   8.02 \\
     16 & 15.89 & 15.99 & 16 & 15.94 & 16.01 \\
     32 & 32.06 & 32.00 & 32 & 31.97 & 32.00 \\
     64 & 64.02 & 64.00 & 64 & 63.97 & 64.00 \\
  \hline
 \end{tabular}
\end{center}
\caption{Experimental results of the accuracy of the temporal scale
  estimates for the non-causal Gaussian temporal scale-space concept
  when applied to the detection temporal scale-space maxima of the
  scale-normalized derivative $-L_{\zeta\zeta}$ for a temporal peak
  and to the scale-normalized derivative $L_{\zeta}$ for onset ramp
  detection for two ways of defining temporal
  scale-normalized derivatives by either (i)~discrete $l_p$-normalization or
  (ii)~variance-based normalization. 
 }
  \label{tab-temp-sc-est-temp-onset-ramp-discgauss}
\end{table}

Note how this operator responds to different types of temporal
structures at different temporal scales with a particularly strong
response due to the temporal variations caused by the daily and annual
temperature cycles. This effect becomes more immediately noticeable if
we complement the above temporal scale selection method with a
complementary assumption about stationarity of the signal and sum up
the scale-normalized feature responses of the quasi quadrature measure
over time for every temporal scale level. In the graphs shown in 
Figure~\ref{fig-tullinge-temp-quasi-recfiltlogscvar-discgaussvar-sumovert}
we do then obtain two major peaks over temporal scales, with the finer scale peak
corresponding to the daily temperature cycle and the coarser scale
peak to the annual temperature cycle. 

Note that the purpose of
this experiment is not primarily to develop an algorithm for detecting
periodic variations in a signal but to illustrate that the temporal
scale selection mechanism produces intuitively reasonable temporal
scale estimates for a real-world 1-D signal with known properties.

\subsection{Numerical accuracy of the temporal scale estimates}

To investigate how well the temporal scale estimates generated by the
resulting temporal scale selection mechanism reflect the temporal
scale in the underlying temporal signal, we generated temporal model
signals defined as either (i)~a temporal peak modelled as time-causal
limit kernel with temporal scale $\tau_0$ and distribution parameter
$c$ for different values of $\tau_0$ and $c$ and (ii)~a temporal onset 
ramp modelled as the primitive function of the temporal peak model.
Then, we detected scale-space extrema of the second-order
scale-normalized temporal derivative $-L_{\zeta\zeta}$ for peak detection or
scale-space extrema of the first-order scale-normalized temporal 
derivative $L_{\zeta}$ for onset detection, using the time-causal
temporal scale-space concept corresponding to convolution with the
time-causal limit kernel for the same value of $c$ and resulting in temporal
scale estimates $\hat{\sigma} = \sqrt{\hat{\tau}}$.

Table~\ref{tab-temp-sc-est-temp-peak} and
Table~\ref{tab-temp-sc-est-temp-onset-ramp} 
show results from this
experiment for the time-causal temporal peak and the temporal onset ramp models,
respectively. Note how well the temporal scale estimates
derived for truncated exponential kernels with a uniform distribution
of the temporal scale levels
in section~\ref{sec-scsel-temp-blob-uni-distr} and 
section~\ref{sec-scsel-onset-ramp-uni-distr} do also generalize to
truncated exponential kernels with a logarithmic distribution of the
temporal scale levels.
Table~\ref{tab-temp-sc-est-temp-onset-ramp-discgauss} shows
corresponding results for the non-causal Gaussian scale-space concept
applied to non-causal temporal peak and temporal onset
ramp models, respectively, and again with a very good agreement
between the local scale estimates $\hat{\sigma}$ in relation to the 
inherent temporal scale $\sigma_0$ in the signal.

For this experiment, we can specifically note that variance-based temporal scale
normalization does on average lead to slightly more accurate temporal
scale estimates compared to discrete $l_p$-normalization, which can be contrasted
to previous results regarding spatial scale selection in hybrid
pyramids by Lindeberg and Bretzner \cite{LinBre03-ScSp}, where
$l_p$-normalization lead to much more accurate scale selection results
compared to variance-based normalization.

\section{Temporal scale selection in spatio-temporal video data}
\label{sec-sc-sel-video-data}

In this section, we will develop a basic proof of concept of applying
the proposed theory and methodology for selecting local temporal
scales in video data based on a small set of specific spatio-temporal feature
detectors formulated in terms of spatio-temporal differential
invariants. A more detailed treatment of this topic with examples for more general
families of differential expressions for spatio-temporal scale
selection with associated spatio-temporal image models for scale calibration
is presented in a companion paper \cite{Lin16-spattempscsel}.

\subsection{Spatio-temporal receptive field model}

For applying the proposed framework for temporal scale selection to
spatio-temporal video data, we follow the approach with idealized
models of spatio-temporal receptive fields of the form
\begin{equation}
      \label{eq-spat-temp-RF-model}
       T(x_1, x_2, t;\; s, \tau;\; v, \Sigma)  
       = g(x_1 - v_1 t, x_2 - v_2 t;\; s, \Sigma) \, h(t;\; \tau)
\end{equation}
as previously derived, proposed and studied in Lindeberg \cite{Lin10-JMIV,Lin13-BICY,Lin16-JMIV}
where
  \begin{itemize}
  \item
     $x = (x_1, x_2)^T$ denotes the image coordinates,
  \item
     $t$ denotes time,
  \item
    $s$ denotes the spatial scale,
\item
    $\tau$ denotes the temporal scale,
  \item
   $v = (v_1, v_2)^T$ denotes a local image velocity,
  \item
    $\Sigma$ denotes a spatial covariance matrix determining the
    spatial shape of an affine Gaussian kernel
   $g(x;\; s, \Sigma)  = \frac{1}{2 \pi s \sqrt{\det\Sigma}} e^{-x^T \Sigma^{-1} x/2s}$,
  \item
     $g(x_1 - v_1 t, x_2 - v_2 t;\; s, \Sigma)$ denotes a spatial affine Gaussian kernel
     that moves with image velocity $v = (v_1, v_2)$ in space-time and
\item
   $h(t;\; \tau)$ is a temporal smoothing kernel over time.
\end{itemize}
and we specifically here choose as temporal smoothing kernel over time
either (i)~the time-causal temporal scale-space kernel corresponding to a set of
first-order integrators with equal time constants coupled in cascade
(\ref{eq-temp-scsp-kernel-uni-distr})
\begin{equation}
 \label{eq-temp-scsp-kernel-uni-distr-spat-temp-scsel}
  h(t;\; \tau) = U(t;\; \mu, K) = \frac{t^{K-1} \, e^{-t/\mu}}{\mu^K \, \Gamma(K)}  
\end{equation}
with $\tau = K \mu^2$ or (ii)~the time-causal limit kernel 
\begin{equation}
 \label{eq-temp-scsp-kernel-log-distr}
  h(t;\; \tau) = \Psi(t;\; \tau, c)
\end{equation}
defined via its Fourier transform of the form
(\ref{eq-FT-comp-kern-log-distr-limit}).

For simplicity, we shall in this treatment restrict ourselves to
space-time separable receptive fields obtained by setting the image
velocity to zero $v = (v_1, v_2) = (0, 0)$ and to receptive fields
that are rotationally symmetric over the spatial domain as obtained by
setting the spatial covariance matrix to a unit matrix $\Sigma = I$.
The resulting spatio-temporal receptive fields that we then obtain correspond to
complementing our time-causal temporal scale-space concepts studied in 
Sections~\ref{sec-scsel-prop-time-caus-trunc-exp-uni-distr}--\ref{sec-scsel-prop-time-caus-trunc-exp-log-distr}
with a rotationally symmetric spatial Gaussian scale-space concept
over the spatial domain.

Specifically, the natural way of expressing spatio-temporal scale
selection mechanisms within this space-time separable spatio-temporal
scale-space concept
\begin{equation}
  \label{eq-sep-spat-temp-scsp-def}
  L(x_1, x_2, t;\; s, \tau) 
  = \left( T(\cdot, \cdot, \cdot;\; s, \tau) * f(\cdot, \cdot, \cdot) \right)(x_1, x_2, t;\; s, \tau) 
\end{equation}
is by studying scale-normalized partial derivates of the form
(Lindeberg \cite[Section~8.5, Equation~(108)]{Lin16-JMIV}
\begin{equation}
  \label{eq-sc-norm-part-der}
  L_{x_1^{m_1} x_2^{m_2} t^n,norm} = s^{(m_1 + m_2) \gamma_s/2} \, \alpha_n(\tau) \, L_{x_1^{m_1} x_2^{m_2} t^n}.
\end{equation}
where the factor $s^{(m_1 + m_2) \gamma_s/2}$ transforms the regular partial
spatial derivatives to corresponding scale-normalized spatial
derivatives with $\gamma_s$ denoting the spatial scale normalization parameter
(Lindeberg \cite{Lin97-IJCV}) and the factor
$\alpha_n(\tau)$ is the scale normalization factor for
scale-normalized temporal derivatives according to either
variance-based normalization (Lindeberg \cite[Section~7.2, Equation~(74)]{Lin16-JMIV})
\begin{equation}
  \alpha_n(\tau) = \tau^{n \gamma_{\tau}/2}
\end{equation}
with $\gamma_{\tau}$ denoting the temporal scale normalization parameter
or scale-normalized temporal derivatives according to 
$L_p$-normalization (Lindeberg \cite[Section~7.2, Equation~(76)]{Lin16-JMIV})
\begin{equation}
  \alpha_n(\tau) 
  = \frac{\| g_{\xi^n}(\cdot;\; \tau) \|_p}{\| h_{t^n}(\cdot;\; \tau) \|_p}
  = \frac{G_{n,\gamma_{\tau}}}{\| h_{t^n}(\cdot;\; \tau) \|_p}.
\end{equation}
and $G_{n,\gamma_{\tau}}$ denotes the $L_p$-norm of the non-causal
temporal Gaussian derivative kernel for the $\gamma_{\tau}$-value for
which this $L_P$-norm becomes constant over temporal scales.

\subsection{Differential entities for spatio-temporal scale selection}

Inspired by the way neurons in the lateral geniculate nucleus (LGN)
respond to visual input (DeAngelis et al \cite{DeAngOhzFre95-TINS,deAngAnz04-VisNeuroSci}),
which for many LGN cells can be modelled by idealized operations of
the form (Lindeberg \cite[Equation~(108)]{Lin13-BICY})
\begin{equation}
  \label{eq-lgn-model-2}
  h_{LGN}(x, y, t;\; s, \tau) 
  = \pm (\partial_{xx} + \partial_{yy}) \, g(x, y;\; s) \, \partial_{t^n} \, h(t;\; \tau),
\end{equation}
let us consider the following differential entities 
(Lindeberg \cite[Section~8.4, Equations~(95)--(96)]{Lin16-JMIV})
\begin{align}
  \begin{split}
    \partial_t (\nabla_{(x,y)}^2 L) & = L_{xxt} + L_{yyt}
  \end{split}\\
  \begin{split}
    \partial_{tt} (\nabla_{(x,y)}^2 L) & = L_{xxtt} + L_{yytt}
  \end{split}
\end{align}
which correspond to first- and second-order temporal derivatives of
the spatial Laplacian operator and study the corresponding
scale-normalized spatio-temporal derivative expressions for 
$\gamma_s = 1$:
\begin{align}
  \begin{split}
    \partial_{t,norm} (\nabla_{(x,y),norm}^2 L) 
     & = s \, \alpha_1(\tau) \, (L_{xxt} + L_{yyt})
  \end{split}\nonumber\\
  \begin{split}
      &  = s \, \alpha_1(\tau) \, \partial_{t} (\nabla_{(x,y)}^2 L),
  \end{split}\\
  \begin{split}
    \partial_{tt,norm} (\nabla_{(x,y),norm}^2 L) 
     & = s \, \alpha_2(\tau) \, (L_{xxtt} + L_{yytt}) 
  \end{split}\nonumber\\
  \begin{split}
      & = s \, \alpha_2(\tau) \, \partial_{tt} (\nabla_{(x,y)}^2 L).
  \end{split}
\end{align}
Notably, we do not focus on extending the previously
established use of the spatial Laplacian operator for spatial scale
selection to a spatio-temporal Laplacian operator for spatio-temporal
scale selection, since the most
straightforward way of defining such an operator 
$\nabla_{(x, y, t)}^2 L = L_{xx} + L_{yy} + \kappa^2 L_{tt}$ for some $\kappa$
is not covariant under independent rescaling of the spatial and temporal
coordinates as occurs if observing the same scene with cameras having
independently different spatial and temporal sampling rates.
The differential entities $\partial_{t,norm} (\nabla_{(x,y),norm}^2 L)$ and
$\partial_{tt,norm} (\nabla_{(x,y),norm}^2 L)$ are on the other hand
truly covariant under independent rescalings of the spatial and
temporal dimensions and therefore better candidates to be used as
primitives in spatio-temporal scale selection algorithms.


\subsection{Spatio-temporal scale selection properties}

\subsubsection{Response to a localized Gaussian blink}

Consider a local idealized spatio-temporal image pattern defined as
the combination of a rotationally symmetric Gaussian blob $g(x, y;\; s_0)$
over the spatial domain and a time-causal temporal peak $U(t;\; \mu, K_0)$
of the form (\ref{eq-peak-model-uni-distr}) over the temporal domain
\begin{equation}
  \label{eq-model-gauss-blink}
   f(x, y, t) = g(x, y;\; s_0) \, U(t;\; \mu, K_0).
\end{equation}
If we define the spatio-temporal scale-space representation of this
spatio-temporal image pattern of the form
(\ref{eq-sep-spat-temp-scsp-def}) with the temporal scale-space kernel
chosen as the composed kernel $U(t;\; \mu, K)$ corresponding to a cascade of
first-order integrators with equal time constants coupled in cascade 
(\ref{eq-temp-scsp-kernel-uni-distr-spat-temp-scsel}),
then it follows from the semi-group property of the spatial Gaussian
kernel and the semi-group property
(\ref{eq-semi-group-time-caus-equal-time-const-Lapl-transform}) of the
time-causal temporal kernel $U(t;\; \mu, K)$ that the spatio-temporal
scale-space representation will be of the form
\begin{equation}
  \label{eq-spat-temp-scsp-time-caus-blink}
   L(x, y, t;\; s, \tau) = g(x, y;\; s_0 + s) \, U(t;\; \mu, K_0 + K).
\end{equation}

\paragraph{The second-order temporal derivative of the spatial Laplacian of
  the Gaussian.}

Specifically, the scale-normalized differential entity $\partial_{tt,norm} (\nabla_{(x,y),norm}^2 L)$
constituting an idealized model of a ``lagged'' LGN cell \cite[Figure~3(right)]{Lin16-JMIV}
will by a combination of the Laplacian response of Gaussian
\begin{equation}
\nabla^2 g(x, y;\; s) = (x^2 + y^2 -2 s)/s^2 \, g(x, y;\; s), 
\end{equation}
the second-order temporal derivative of $L_{tt}(t;\; \mu, K)$ of a
time-causal peak in Equation~(\ref{eq-temp2der-scsp-peak-uni-distr}) 
and the temporal scale normalization operation 
\begin{equation} 
  L_{\zeta\zeta}(t;\; \mu, K) = (\mu^2 K)^{\gamma_{\tau}} \,
  L_{tt}(t;\; \mu, K)
\end{equation}
in Equation~(\ref{eq-scnorm-temp2der-scsp-peak-uni-distr}) assume the form
\begin{align}
  \begin{split}
      & \partial_{tt,norm} (\nabla_{(x,y),norm}^2 L)
  \end{split}\nonumber\\
  \begin{split}
      & = \frac{(\mu^2 K)^{\gamma_{\tau}}}{\mu ^2 t^2} 
             \left( 
                  \mu ^2 \left(K^2+K (2 K_0-3)+K_0^2-3 K_0+2\right)
             \right.
 \end{split}\nonumber\\
  \begin{split}
       & \phantom{ \frac{(\mu^2 K)^{\gamma_{\tau}}}{\mu ^2 t^2} \left( \right.}
              \quad
              \left.
               -2 \mu t (K+K_0-1)+t^2
              \right) \times
  \end{split}\nonumber\\
  \begin{split}
       & \phantom{=} \quad \frac{(x^2 + y^2 -2 (s_0 + s))}{(s_0 + s)^2} g(x, y;\; s_0 + s) \, U(t;\; \mu, K_0 + K).
  \end{split}
\end{align}
Specifically, based on previously established scale selection
properties of the spatial Laplacian of the Gaussian (Lindeberg
\cite{Lin97-IJCV,Lin12-JMIV})
and the second-order scale-normalized temporal derivatives of the
time-causal scale space (Section~\ref{sec-scsel-temp-blob-uni-distr}),
it follows that this spatio-temporal differential entity will for
spatial and temporal scale normalization powers $\gamma_s = 1$ and
$\gamma_{\tau} = 3/4$, respectively, assume
its local extremum over both spatial and temporal scales at spatial
scale 
\begin{equation}
  \hat{s} = s_0
\end{equation}
and at a temporal scale that is a good
approximation of the temporal scale of the temporal peak
$\hat{K} \approx K_0$ corresponding to 
(see Table~\ref{tab-sc-est-temp-peak-2nd-der-uni-distr})
\begin{equation}
  \hat{\tau} \approx \tau_0.
\end{equation}
Thus, simultaneous spatio-temporal scale selection using the
differential entity $\partial_{tt,norm} (\nabla_{(x,y),norm}^2 L)$
applied to the model signal (\ref{eq-model-gauss-blink}) will estimate
both the spatial extent and the temporal duration of the Gaussian
blink.

\paragraph{The determinant of the spatio-temporal Hessian.}

For general values of the spatial and temporal scale normalization
parameters $\gamma_s$ and $\gamma_{\tau}$, the determinant of the
spatio-temporal Hessian is given by
\begin{multline}
    \label{eq-spat-temp-detHess-norm}
    \det {\cal H}_{(x,y,t),norm} L 
    = \, s^{2 \gamma_s} \tau^{\gamma_{\tau}} 
        \left( L_{xx} L_{yy} L_{tt} + 2 L_{xy} L_{xt} L_{yt} \right. \\
       \left. - L_{xx} L_{yt}^2 - L_{yy} L_{xt}^2 - L_{tt} L_{xy}^2 \right).
\end{multline}
In the specific case when the spatio-temporal scale-space
representation of a time-causal Gaussian blink is of the form
(\ref{eq-spat-temp-scsp-time-caus-blink}), if we restrict the
analysis to the spatial origin $(x, y) = (0, 0)$ where 
$g_x = g_y = 0$ and to the temporal maximum point 
$t_{max}$ where $h_t = 0$, implying that $L_{xy} = L_{xt} = L_{yt} = 0$,
 it follows that the determinant of
spatio-temporal Hessian at the spatio-temporal maximum reduces to the
form
\begin{multline}
    \left. \det {\cal H}_{(x,y,t),norm} L \right|_{(x, y) = (0, 0), t = t_{max}}
    = \\ \left. \left(  s^{2 \gamma_s}  g \, g_{xx} \, g_{yy} \right) \right|_{(x, y) = (0, 0)}
       \left.  \left( (\mu^2 K)^{\gamma_{\tau}}  \, U^2 \, U_{tt} \right) \right|_{t = t_{max}},
\end{multline}
we can observe that the spatial and temporal scale selection
properties of the determinant of the spatio-temporal Hessian will be
different from the scale selection properties of the second-order
temporal derivative of the spatial Laplacian.
In a companion paper (Lindeberg \cite{Lin16-spattempscsel}), it is
shown that for a corresponding non-causal Gaussian temporal
scale-space concept, spatial and temporal scale normalization
parameters equal to $\gamma_s = 5/4$ and $\gamma_{\tau} = 5/4$
lead to scale estimates $\hat{s}$ and $\hat{\tau}$ corresponding 
to the spatial and temporal extents $\hat{s} = s_0$ and 
$\hat{\tau} = \tau_0$ of a Gaussian blink.

\begin{table*}[!hbt]
  \addtolength{\tabcolsep}{1pt}
  \begin{center}
   \footnotesize
  \begin{tabular}{ccc}
  \hline
   \multicolumn{3}{c}{Scale estimate $\hat{K}$ and max magnitude
    $\theta_{postnorm} $ from temporal peak (uniform distr)} \\
  \hline
    $K_0$ 
       & $\hat{K}$ (var, $\gamma=5/4$) 
       & $\left. \theta_{postnorm} \right|_{\gamma=1}$ (var, $\gamma=5/4$) \\
  \hline
     4 &   3.1 & 0.508 \\
     8 &   7.1 & 0.503 \\
   16 & 15.1 & 0.501 \\
   32 & 31.1 & 0.502 \\
   64 & 63.1 & 0.501 \\
\hline
  \end{tabular}
\end{center}
\caption{Numerical estimates of the value of $\hat{K}$ at which the
  temporal component of the determinant of the spatio-temporal Hessian
  assumes its maximum over temporal scale (with the discrete
  expression over discrete temporal scales extended to a continuous
  variation) as function of $K_0$ and for either 
  (i)~variance-based normalization with $\gamma = 5/4$,
  (iii)~variance-based normalization with $\gamma = 1$ and
 (iv)~$L_p$-normalization with $p = 1$.
 For the case of variance-based normalization with $\gamma = 5/4$, (ii)~the
post-normalized magnitude measure $\left. L_{\zeta\zeta,maxmagn,norm} \right|_{\gamma=1}$ according to
(\protect\ref{eq-sc-sel-post-norm-temp-peak-scale-calib}) and at the corresponding
scale (i) is also shown.
  Note that for $\gamma = 5/4$ the temporal scale estimate $\hat{K}$ constitutes a good
  approximation to the temporal scale estimate being proportional to
  the temporal scale of the underlying temporal peak and that
  the maximum magnitude estimate $\left. \theta_{postnorm} \right|_{\gamma=1}$ 
  constitutes a good approximation to the maximum magnitude measure
  being constant under variations of the temporal duration of the
  underlying spatio-temporal image structure.}
  \label{tab-sc-est-temp-peak-tempofspattemdethess-uni-distr}
\end{table*}

By performing a corresponding study of the temporal scale selection
properties of the purely temporal component of this expression
\begin{equation}
  \theta_{norm} = \left.  \left( (\mu^2 K)^{\gamma_{\tau}}  \, U^2 \, U_{tt} \right) \right|_{t = t_{max}}
\end{equation}
for the specific case of time-causal temporal scale-space representation based on a
uniform distribution of the intermediate temporal scale levels
as previously done for the temporal scale selection properties of the
second-order temporal derivative of a temporal peak in
Table~\ref{tab-sc-est-temp-peak-2nd-der-uni-distr}, 
we obtain the results shown in
Table~\ref{tab-sc-est-temp-peak-tempofspattemdethess-uni-distr}.
The column labelled $\hat{K}$ shows that the maximum over temporal
scales is obtained at a temporal scale level near the temporal scale
of the underlying temporal peak
\begin{equation}
  \hat{\tau} \approx \tau_0,
\end{equation}
whereas the column labelled $\theta_{postnorm}$ shows that if we
normalize the input signal $f(t) = U(t;\; \mu, K_0)$ to having unit
contrast, then the corresponding post-normalized differential entity
\begin{equation}
  \left. \theta_{postnorm} \right|_{\gamma=1} 
  = \frac{1}{\tau^{1/4}} \left. \theta_{norm} \right|_{\gamma=5/4} 
  = \frac{K}{K+K_0-1}
\end{equation}
is approximately constant for temporal peaks with different temporal
duration as determined by the parameter $K_0$.
In these respects, this time-causal scale selection method implies a
good approximate transfer of the scale selection property of treating
similar temporal structures of different temporal duration in a uniform manner.

\subsubsection{Response to a localized Gaussian onset blob}

Consider a local idealized spatio-temporal image pattern defined as
the combination of a rotationally symmetric Gaussian blob $g(x, y;\; s_0)$
over the spatial domain and a time-causal onset ramp $\int_{u = 0}^t U(u;\; \mu, K_0) \, du$
of the form (\ref{eq-ramp-model-uni-distr}) over the temporal domain
\begin{equation}
  \label{eq-model-gauss-onset-ramp}
   f(x, y, t) = g(x, y;\; s_0) \, \int_{u = 0}^t U(u;\; \mu, K_0) \, du.
\end{equation}
Again defining the temporal scale-space representation of this
spatio-temporal image pattern of the form (\ref{eq-sep-spat-temp-scsp-def}) with the temporal scale-space kernel
chosen as the composed kernel $U(t;\; \mu, K)$ corresponding to a cascade of
first-order integrators with equal time constants coupled in cascade 
(\ref{eq-temp-scsp-kernel-uni-distr-spat-temp-scsel}),
it follows from the semi-group property of the spatial Gaussian kernel and the semi-group property
(\ref{eq-semi-group-time-caus-equal-time-const-Lapl-transform}) of the
time-causal temporal kernel $U(t;\; \mu, K)$ that the spatio-temporal
scale-space representation will be of the form
\begin{equation}
   L(x, y, t;\; s, \tau) = g(x, y;\; s_0 + s) \, \int_{u = 0}^t U(u;\; \mu, K_0) \, du.
\end{equation}

\paragraph{The first-order temporal derivative of the spatial Laplacian of
  the Gaussian.}

For the scale-normalized differential entity $\partial_{t,norm} (\nabla_{(x,y),norm}^2 L)$
constituting an idealized model of a ``non-lagged'' LGN cell \cite[Figure~3(left)]{Lin16-JMIV}
will by a combination of the Laplacian response of Gaussian
$\nabla^2 g(x, y;\; s) = (x^2 + y^2 -2 s)/s^2 \, g(x, y;\; s)$,
the first-order temporal derivative $L_t(t;\; \mu, K)$ of a
time-causal onset ramp in
Equation~(\ref{eq-first-order-temp-der-time-caus-onset-ramp})
and the temporal scale normalization operation 
$L_{\zeta}(t;\; \mu, K) = (\mu \sqrt{K})^{\gamma_{\tau}} \, L_{t}(t;\; \mu, K)$
in Equation~(\ref{eq-scnorm-temp1der-scsp-peak-uni-distr})  assume
the form
\begin{align}
  \begin{split}
      & \partial_{t,norm} (\nabla_{(x,y),norm}^2 L)
 \end{split}\nonumber\\
 \begin{split}
      & =  \frac{(x^2 + y^2 -2 (s_0 + s))}{(s_0 + s)^2} \,
              g(x, y;\; s_0 + s) \times
 \end{split}\nonumber\\
 \begin{split}
      & \phantom{=}
          \quad (\mu \sqrt{K})^{\gamma_{\tau}} \, U(t;\; \mu, K_0 + K).
  \end{split}
\end{align}
Specifically, based on previously established scale selection
properties of the spatial Laplacian of the Gaussian (Lindeberg
\cite{Lin97-IJCV,Lin12-JMIV})
and the first-order scale-normalized temporal derivatives of the
time-causal scale space (Section~\ref{sec-scsel-onset-ramp-uni-distr}),
it follows that this spatio-temporal differential entity will for
spatial and temporal scale normalization powers $\gamma_s = 1$ and
$\gamma_{\tau} = 1/2$, respectively, assume
its local extremum over both spatial and temporal scales at spatial
scale 
\begin{equation}
  \hat{s} = s_0
\end{equation} and at a temporal scale that is a good
approximation of the temporal scale of the temporal onset ramp
$\hat{K} \approx K_0$ corresponding to 
(see Table~\ref{tab-sc-est-temp-ramp-1st-der-uni-distr})
\begin{equation}
  \hat{\tau} \approx \tau_0.
\end{equation}
Thus, simultaneous spatio-temporal scale selection using the
differential entity $\partial_{t,norm} (\nabla_{(x,y),norm}^2 L)$
applied to the model signal (\ref{eq-model-gauss-onset-ramp}) will estimate
both the spatial extent and the temporal duration of a Gaussian
onset blob.

\subsection{General scale selection property for temporal modelling
  and time-causal scale space based on the scale-invariant time-causal limit kernel}

If we instead model the Gaussian blink (\ref{eq-model-gauss-blink}) and 
the onset Gaussian blob (\ref{eq-model-gauss-onset-ramp}) by the
scale-invariant time-causal limit kernel $\Psi(t;\; \tau, c)$ over the temporal domain
\begin{align}
  \begin{split}
     \label{eq-model-gauss-blink-limitkern}
     f(x, y, t) & = g(x, y;\; s_0) \, \Psi(t;\; \tau, c),
  \end{split}\\
  \begin{split}
     \label{eq-model-gauss-onset-ramp-limitkern}
      f(x, y, t) & = g(x, y;\; s_0) \, \int_{u = 0}^t \Psi(u;\; \tau, c) \, du,
  \end{split}
\end{align}
then by the general transformation property of the time-causal limit kernel under
temporal scaling transformations by a temporal scaling factor $S =c^j$ 
that is an integer power of the distribution parameter $c$ of the time-causal
limit kernel (Lindeberg \cite[Equation~(44)]{Lin16-JMIV})
\begin{equation}
  \label{eq-sc-transf-limit-kernel}
   S \, \Psi(S \, t;\; S^2 \tau, c) = \Psi(t;\; \tau, c),
\end{equation}
it holds that the corresponding spatio-temporal scale-space representations
of two temporally scaled video sequences $f'(x', y', t) = f(x, y, t)$
for $(x', y', t') = (x, y, S t)$ are related according to
(Lindeberg \cite[Equation~(46)]{Lin16-JMIV})
\begin{equation}
  L'(x', y';, t';\; s', \tau', c) = L(x, y, t;\; s, \tau, c) 
\end{equation}
for $(s', \tau')  = (s, S^2 \tau)$ if  $S = c^j$.
The corresponding scale-normalized temporal derivatives are in turn related
according to (\ref{eq-transf-prop-sc-norm-temp-ders-limit-kern})
\begin{equation}
  \label{eq-transf-prop-sc-norm-temp-ders-limit-kern-S}
  L'_{\zeta'^n}(x', y', t';\, s', \tau', c) 
  = S^{n (\gamma_{\tau}-1)} \, L_{\zeta^n}(x, y, t;\, s, \tau, c).
\end{equation}
If the scale-normalized temporal derivative $L_{\zeta^n}(x, y, t;\, s,\tau, c)$ 
computed from the original video sequence $f$ 
assumes a local extremum over temporal scales at
$(x, y, t;\; s, \tau) = (x_0, y_0, t_0;\; s_0, \tau_0$), 
then by the general scale-invariance property of
temporal scale selection in the temporal scale-space concept
based on the time-causal limit kernel, which is described in
Section~\ref{sec-gen-sc-result-limit-kernel-sine-wave}, 
it follows that the scale-normalized temporal derivative $L_{\zeta'^n}(x', y', t';\, s',\tau', c)$ 
computed from the temporally scaled video sequence $f'$ will assume
a local extremum over temporal scales at 
\begin{equation}
  (x_0', y_0', t_0';\; s_0', \tau_0') 
  = (x_0, y_0, S t_0;\; s_0, S^2 \tau_0).
\end{equation}
This scale-invariant property can also be extended to spatio-temporal
derivatives $L_{\xi^{m_1}\eta^{m_2}\zeta^n}$ and spatio-temporal
differential invariants ${\cal D}_{norm}$ defined in terms of homogenous polynomials as
well as homogeneous rational expressions of such scale-normalized
spatio-temporal derivatives.
In this way, by performing both the temporal modelling of the
underlying temporal signal in terms of the time-causal limit kernel
and using a temporal scale-space concept based on the time-causal
limit kernels, we can support fully scale-covariant temporal scale
estimates for temporal scale selection in video data defined over a
time-causal spatio-temporal domain.

The only component that remains is to determine how the original
temporal scale estimate $\hat{\tau}$ depends on the distribution
parameter $c$ and the temporal scale normalization parameter
$\gamma_{\tau}$ for some value of $\tau_0$.


\section{Temporal scale selection in spectro-temporal audio data}
\label{sec-sc-sel-audio-data}

For audio signals, corresponding temporal scale selection
methods can be applied to a time-causal spectro-temporal domain,
with the 2-D spatial domain of video data over the spatial dimensions
$(x, y)$ conceptually replaced by a 1-D logspectral domain over the
logspectral dimension $\nu$ in the spectrogram computed at any 
temporal moment using a time-causal receptive field model as proposed in
(Lindeberg and Friberg \cite{LinFri15-PONE,LinFri15-SSVM})
or with the time-causal kernels in that model replaced by the 
time-causal limit kernel (Lindeberg \cite{Lin16-JMIV}).

The analogous operations to the first- and second-order temporal
derivatives of the spatial dimension would then be the first- and
second-order temporal derivatives of the second-order derivative in
the logspectral dimension
\begin{align}
  \begin{split}
    \partial_{t,norm}(L_{\nu\nu,norm}) & = s^{\gamma_s} \, \tau^{\gamma_{\tau}/2} \, L_{\nu\nu t}
  \end{split}\\
  \begin{split}
    \partial_{tt,norm}(L_{\nu\nu,norm}) & = s^{\gamma_s} \, \tau^{\gamma_{\tau}} \, L_{\nu\nu tt}
  \end{split}
\end{align}
where $s$ denotes the logspectral scale and $\tau$ the temporal scale.

By calibrating the logspectral scale normalization
parameter $\gamma_s$ such that the selected temporal
scale should reflect the logspectral width of a spectral band, it
follows that we should use $\gamma_s = 3/4$.
By calibrating the temporal scale normalization parameter
$\gamma_{\tau}$ such that the selected temporal scale 
of $\partial_{t,norm}(L_{\nu,norm})$ should reflect the temporal duration of
an onset, if follows that we should use $\gamma_{\tau} = 1/2$ for this
operator.
By instead calibrating the temporal scale normalization parameter
$\gamma_{\tau}$ such that the selected temporal scale 
of $\partial_{tt,norm}(L_{\nu\nu,norm})$ should reflect the temporal
duration of a beat, if follows that we should use $\gamma_{\tau} = 3/4$ for that
operator.
 
Note that these operations can be expressed both over frequency-time
separable spectro-temporal receptive fields and over glissando-adapted
spectro-temporal receptive fields if we for glissando-adapted
receptive fields also replace the temporal derivative operator $\partial_t$
by the corresponding glissando-adapted temporal derivative operator
$\partial_{\bar t} = \partial_t + v \, \partial_{\nu}$,
where $v$ denotes the glissando parameter.

\section{Summary and conclusions}
\label{sec-sum-disc}

In this treatment, we have proposed a new theoretical framework for
temporal scale selection in a time-causal scale-space representation.
Starting from a general survey of previously proposed temporal
scale-space concepts and a detailed analysis of their relative
advantages and disadvantages, we have focused our efforts on the 
time-causal scale-space concept based on first-order integrators 
coupled in cascade and analysed the extent to which scale-space 
properties that hold for the scale-invariant non-causal Gaussian 
temporal scale-space concept can be transferred to this time-causal 
scale-space concept. Specifically, we have analysed this
time-causal scale-space concept for two specific ways of distributing
the intermediate temporal scale levels using either 
(i)~a uniform distribution over the temporal scales as parameterized
by the variance of the temporal scale-space kernel and corresponding
to convolution with temporal kernels that are Laguerre functions and
in turn
corresponding to temporal derivatives of the Gamma distribution or
(ii)~a logarithmic distribution taken to a recently proposed time-causal limit kernel with an
infinitely dense distribution of temporal scale levels towards zero
temporal scale.

For peak and ramp detection, we have shown that for the time-causal temporal
scale space concept based on first-order integrators with equal time
constant coupled in cascade, we can reasonably well estimate the temporal scale of a
localized temporal peak or a localized onset ramp with corresponding
good approximation of constancy of appropriately post-normalized scale-normalized
magnitude measures of the corresponding feature detectors under
variations in the temporal duration of the underlying temporal peak or
the underlying temporal ramp.
For a non-localized sine wave signal, the lack of temporal scale invariance
is, however, substantial both with regard to a systematic offset in
temporal scale estimates and a lack of corresponding constancy of the
magnitude measures over variations of the wavelength of the underlying
sine wave.

For the time-causal temporal scale-space concept based on convolution
with the time-causal limit kernel with an underlying logarithmic
distribution of the temporal scale levels and taken to the limit of
the time-causal limit kernel with an infinitely dense distribution of
temporal scale levels near temporal scale zero, we have on the other hand
shown that it is possible to achieve perfect temporal invariance in
the respects that (i)~the temporal scale estimates in dimension
$[\mbox{time}]$ are proportional to the wavelength of the underlying
sine wave and (ii)~the magnitude measures remain constant under
variations of the wavelength of the sine wave. 

Additionally, we have shown a general scale invariance result that
holds for any temporal signal and which states that
for temporal scaling transformation with a temporal scaling factor
given as an integer power of the distribution parameter $c$ of the
time-causal limit kernel that is used for generating the temporal
scale space, it holds that:
\begin{itemize}
\item[(i)]
  local extrema over temporal scales of scale-normalized derivatives 
  are preserved under this group of temporal scaling transformations
  with scaling factors of the form $t' = c^j t$ for integer $j$,
\item[(ii)]
  the corresponding scale estimates are transformed in a
  scale-covariant way corresponding to $\hat{\tau}' = c^{2j} \hat{\tau}$ in units of
  the variance of the temporal scale-space kernel,
\item[(iii)]
  if the scale normalization parameter $\gamma$ of variance-based
scale-normalized derivatives is chosen as $\gamma = 1$ and
corresponding to $p = 1$ for $L_p$-normalization, then the magnitude
values of the scale-normalized temporal derivatives are preserved
under scaling transformations with any temporal scaling factor $c^j$
that is an integer power of the distribution parameter $c$ of the
time-causal limit kernel and
\item[(iv)]
 for other values of the scale normalization parameters $\gamma$ or
 $p$, the corresponding scale-normalized derivatives are transformed
 according to a scale covariant power law 
 (\ref{eq-transf-prop-sc-norm-temp-ders-limit-kern}), which is straightforward to compensate
 for by post normalization.
\end{itemize}
In these respects, the proposed framework for temporal scale selection
in the scale-space concept based on the time-causal limit kernel provides the
necessary mechanisms to achieve temporal scale invariance while simultaneously being
expressed over a time-causal and time-recursive temporal domain.
From a theoretical perspective, this is a conceptually novel type of
construction that has not previously been achie\-ved based on any other
type of time-causal temporal scale-space concept.

As experimental confirmation of the derived theoretical results
regarding temporal scale selection properties, we have presented
experimental results of applying two types of more specific temporal
scale selection algorithms to one-dimensional temporal signals, based on either
(i)~sparse scale-space extrema detection by detecting local extrema of
feature responses over both time and temporal scales or (ii)~dense feature maps
over temporal scales here specifically manifested in terms of a temporal quasi
quadrature entity that constitutes an energy measure of the local
strength of scale-normalized first- and second-order temporal
derivates.

We have also described practical details to handle in time-causal
scale selection algorithms in relation to the inherent temporal delays
of time-causal image measurements and proposed specific mechanisms to
handle the differences in temporal delays between time-causal scale-space
representations at different temporal scales.

Experimental results presented for synthetic and real
one-dimensional temporal signals show that it is possible to compute local
estimates of temporal scale levels that in units of the standard deviation of
the underlying temporal scale-space kernel are proportional to the
temporal duration of the underlying structures in the temporal signal
that gave rise to the filter responses.

Beyond these two specific ways of expressing temporal scale selection
mechanisms, we argue that the theoretical results presented in the
paper should also more generally open up for extensions to other
ways of comparing time-dependent filter responses at multiple temporal
scales.

Experimental results obtained by applying this temporal scale
selection theory to video analysis will be reported in a
companion paper \cite{Lin16-spattempscsel}.

The non-causal temporal scale selection theory developed in this paper
as a baseline and reference for time-causal temporal scale selection
can of course also be used for analysing pre-recorded time-dependent signals in 
offline or time-delayed scenarios.

\appendix
\normalsize

\section{Why a semi-group property over temporal scales leads to undesirable
  temporal dynamics in the presence of temporal delays}
\label{app-undesired-temp-dyn-temp-semi-group}

One way of understanding why the assumption about a semi-group
property over temporal scales may lead to undesirable temporal
dynamics for a temporal scale space representation involving temporal
delays can be obtained as follows:

Ideally, for a temporal scale-space concept involving a temporal delay
one would like the
temporal delay $\delta$ to be proportional to the temporal scale
parameter $\sigma$ in terms of dimension $[\mbox{time}]$
\begin{equation}
  \label{eq-prop-req-temp-delay-temp-scale}
  \delta = C \, \sigma
\end{equation}
for some constant $C > 0$. For a temporal scale-space kernel with finite%
\footnote{Regarding the assumption of a finite temporal variance, it
  is interesting to compare the situation with the time-causal
  semi-group kernel $\phi(t;\; \tau) = \frac{1}{\sqrt{2 \pi} \,
    t^{3/2}} \, \tau \, e^{-\tau^2/2t} $
  derived by Fagerstr{\"o}m \cite[Equation~(27)]{Fag05-IJCV} and 
  Lindeberg \cite[Equation~(93)]{Lin10-JMIV}.
  For this kernel, the first- and second-order temporal moments are
  not finite $\int_{t = 0}^{\infty} t \, \phi(t;\; \tau) \, dt
  \rightarrow \infty$ and $\int_{t = 0}^{\infty} t^2 \, \phi(t;\; \tau)
  \, dt \rightarrow \infty$, implying that the analysis in this
  appendix breaks down if applied to the time-causal semi-group, since
this analysis is based on the additive properties of mean values and variances
for non-negative distributions. The fact that the first- and
   second-order temporal moments are infinite for the time-causal semi-group, does on the other hand
   also reflect undesirable temporal dynamics, since temporal smoothing with
   such a kernel leads to slow and smeared out temporal responses
   compared to temporal smoothing with a temporal kernel having finite
   first- and second-order temporal moments.
   If we measure the temporal delay of the time-causal semi-group
   kernel by the position of the temporal maximum $\hat{t} = \tau^2/3$
   (Lindeberg \cite[Equation~(119)]{Lin10-JMIV})
   and its temporal extent from the difference
  between the time instances at which the
   one-dimensional time-causal semi-group kernel 
   assumes half its maximum value $\Delta t \approx 0.900 \, \tau^2$  
   (Lindeberg \cite[Equation~(122)]{Lin10-JMIV}),
   then the temporal delay and the temporal extent of the time-causal
   semi-group kernel are indeed proportional.
   Those measures of the temporal delay and the temporal extent of the
   temporal kernel are, however, not the same as used in the in the
   arguments in this appendix.

The example with the time-causal semi-group therefore demonstrates that at the cost of infinite
   first- and second-order temporal moments it is possible to find a
   a temporal smoothing kernel that both obeys the semi-group property 
   and a proportionality relation between measures of the temporal
   delay and the temporal extent in dimensions of $[\mbox{time}]$.
   Due to the infinite first- and second-order temporal moments, the
   temporal dynamics is, however, undesirable anyway.}
 temporal variance $\tau$, this corresponds to
letting the temporal delay at any temporal scale be proportional to
the square root of the temporal scale parameter $\tau$ according to
\begin{equation}
  \label{eq-rel-temp-delay-temp-var}
  \delta = \phi(\tau) = C \sqrt{\tau}.
\end{equation}
Let us next assume that we have to two temporal scale-space kernels
$h(t;\; \tau_1, \delta_1)$ and $h(t;\; \tau_2, \delta_2)$
with finite temporal variances $\tau_1$ and $\tau_2$ and 
finite temporal means $\delta_1$ and $\delta_2$ from the same family of
temporal kernels $h$.
If the temporal kernels are to obey a semi-group property over
temporal scales, then by the additive property of mean values and
variances under convolution of positive functions, it follows that the
composed temporal scale-space kernel should be given by
\begin{equation}
  h(\cdot;\; \tau_1, \delta_1) * h(t;\; \tau_2, \delta_2)
  = h(t;\; \tau_1 + \tau_2, \delta_1 + \delta_2).
\end{equation}
This property should for example hold for the non-causal Gaussian
temporal scale-space kernels (\ref{eq-gauss-time-delay}) if we require
the kernels to obey a semi-group property over temporal scales.
Combining this property with a fixed relationship between the temporal
delay $\delta$ and the temporal scale $\tau$ according to
$\delta = \phi(\tau)$ does, however, then lead to
\begin{equation}
  \label{eq-counter-arg-delay-temp-scale}
  \delta_1 + \delta_2 = \phi(\tau_1 + \tau_2) = \phi(\tau_1) + \phi(\tau_2).
\end{equation}
This implies that the function $\phi$ must be additive in terms of its
argument $\tau$, implying increasingly longer temporal delays at coarser
temporal scales and thus a violation of the desirable form
of temporal dynamics $\delta = \phi(\tau) = C \sqrt{\tau}$.

We can, however, remedy the situation by replacing the temporal
semi-group property with a weaker cascade smoothing property over
temporal scales
\begin{equation}
  \label{eq-casc-smooth-prop-in-proof-temp-dyn}
  L(\cdot;\; \tau_2, \delta_2) 
  = h(\cdot;\; (\tau_1, \delta_1) \mapsto (\tau_2, \delta_2))
     * L(\cdot;\; \tau_1, \delta_1),
\end{equation}
where the temporal kernels should for any triplets of temporal scale
values and temporal delays $(\tau_1, \delta_1)$, $(\tau_2, \delta_2)$
and $(\tau_3, \delta_3)$ obey the transitive property
\begin{align}
   \begin{split}
       & h(\cdot;\; (\tau_1, \delta_1) \mapsto (\tau_2, \delta_2))
          * h(\cdot;\; (\tau_2, \delta_2) \mapsto (\tau_3, \delta_3)) =
   \end{split}\nonumber\\
   \begin{split}
      &  h(\cdot;\; (\tau_1, \delta_1) \mapsto (\tau_3, \delta_3))
  \end{split}
\end{align}
and we can specifically for the non-causal Gaussian temporal scale-space
concept with time-delayed Gaussian kernels of the form (\ref{eq-gauss-time-delay}) choose
\begin{equation}
  h(\cdot;\; (\tau_1, \delta_1) \mapsto (\tau_2, \delta_2))
  = g(\cdot;\; \tau_2-\tau_1, C(\sqrt{\tau_2} - \sqrt{\tau_1})).
\end{equation}
Based on this form of cascade smoothing property over temporal scale, we
can both (i)~guarantee non-creation of new structures 
in the signal from finer to coarser temporal scales based on the scale-space properties of the temporal
scale-space kernel $g$ and
(ii)~achieve temporal delays that increase linearly with the temporal
scale parameter in terms of dimension $[\mbox{time}]$ such that 
$\delta_1 = C \sqrt{\tau_1}$ and $\delta_2 = C \sqrt{\tau_2}$
in (\ref{eq-casc-smooth-prop-in-proof-temp-dyn}).

In our temporal scale-space concept based on truncated exponential
kernels coupled in cascade (Lindeberg \cite{Lin90-PAMI,Lin15-SSVM,Lin16-JMIV}; 
Lindeberg and Fagerstr{\"o}m \cite{LF96-ECCV}), we can specifically
note that
(i)~the special case when all the time constants are equal implies a
semi-group property over discrete temporal scales and longer temporal
delays at coarser temporal scales (see the second row in
Figure~\ref{fig-temp-kernels-1D}) whereas the 
(ii)~the special case with logarithmically distributed temporal scales
implies that only a weaker cascade smoothing property holds and which
enables much faster temporal response properties  (see the third and fourth rows in
Figure~\ref{fig-temp-kernels-1D}).

Since any time-causal temporal scale-space representation will give
rise to non-zero temporal delays, and we have shown in this section
how the assumption of a semi-group structure over temporal scale leads
to undesirable temporal dynamics in the presence of temporal delays,
we argue that one should not require a semi-group structure over
temporal scales for time-causal scale space and instead require a less
restrictive cascade smoothing property over temporal scales.

\begin{table*}
{\em Regular temporal derivatives of Koenderink's scale-time kernel:}
\begin{align}
  \begin{split}
   \label{eq-scale-time-norm-1-temp-der}
    h_{t,Koe}(t;\; \sigma, \delta)  
    & = -\frac{\log \left(\frac{t}{\delta }\right) e^{-\frac{\log ^2\left(\frac{t}{\delta
   }\right)+\sigma ^4}{2 \sigma ^2}}}{\sqrt{2 \pi } \delta  \sigma ^3 t}
  \end{split}\\
 \begin{split}
   \label{eq-scale-time-norm-2-temp-der}
    h_{tt,Koe}(t;\; \sigma, \delta)  
    & = \frac{e^{-\frac{\log ^2\left(\frac{t}{\delta }\right)+\sigma ^4}{2 \sigma ^2}}
   \left(\sigma ^2 \log \left(\frac{t}{\delta }\right)+\log ^2\left(\frac{t}{\delta
   }\right)-\sigma ^2\right)}{\sqrt{2 \pi } \delta  \sigma ^5 t^2},
  \end{split}\\
 \begin{split}
   \label{eq-scale-time-norm-3-temp-der}
    h_{ttt,Koe}(t;\; \sigma, \delta)  
    & = \frac{e^{-\frac{\log ^2\left(\frac{t}{\delta }\right)+\sigma ^4}{2 \sigma ^2}} \left(-3
   \sigma ^2 \log ^2\left(\frac{t}{\delta }\right)+\left(3 \sigma ^2-2 \sigma ^4\right)
   \log \left(\frac{t}{\delta }\right)-\log ^3\left(\frac{t}{\delta }\right)+3 \sigma
   ^4\right)}{\sqrt{2 \pi } \delta  \sigma ^7 t^3}.
  \end{split}
\end{align}
\caption{Regular first-, second- and third-order temporal derivatives of
  Koenderink's scale-time kernel (renormalized according to
  (\ref{eq-temp-kern-norm-Koe-scale-time})).}
  \label{tab-scale-time-norm-1-2-3-temp-der}

\bigskip

{\em Scale-normalized temporal derivatives of Koenderink's scale-time
  kernel:}

\medskip

\noindent
Using variance-based normalization, the corresponding scale-normalized
temporal derivatives are for a general value of $\gamma$ given by
\begin{align}
  \begin{split}
    \label{eq-sc-norm-1-temp-der-scale-time-var-gen-gamma}
    h_{\zeta,Koe}(t;\; \sigma, \delta)  
    & = \tau^{\gamma/2} h_{t,Koe}(t;\; \sigma, \delta) 
      = \frac{\left(\delta ^2 e^{3 \sigma ^2} \left(e^{\sigma ^2}-1\right)\right)^{\gamma /2}
   \log \left(\frac{\delta }{t}\right) e^{-\frac{\log ^2\left(\frac{t}{\delta
   }\right)+\sigma ^4}{2 \sigma ^2}}}{\sqrt{2 \pi } \delta  \sigma ^3 t},
  \end{split}\\
   \begin{split}
    \label{eq-sc-norm-2-temp-der-scale-time-var-gen-gamma}
    h_{\zeta\zeta,Koe}(t;\; \sigma, \delta)  
    & = \tau^{\gamma} h_{tt,Koe}(t;\; \sigma, \delta) 
      = \frac{\left(\delta ^2 e^{3 \sigma ^2} \left(e^{\sigma ^2}-1\right)\right)^{\gamma }
   e^{-\frac{\log ^2\left(\frac{t}{\delta }\right)+\sigma ^4}{2 \sigma ^2}} \left(\log
   \left(\frac{t}{\delta }\right) \left(\log \left(\frac{t}{\delta }\right)+\sigma
   ^2\right)-\sigma ^2\right)}{\sqrt{2 \pi } \delta  \sigma ^5 t^2},
  \end{split}
\end{align}
which for the specific value of $\gamma = 1$ reduce to
\begin{align}
  \begin{split}
    \label{eq-sc-norm-1-temp-der-scale-time-var-gamma1}
    h_{\zeta,Koe}(t;\; \sigma, \delta)  
    & = \sqrt{\tau} \, h_{t,Koe}(t;\; \sigma, \delta) 
      = \frac{\sqrt{e^{\sigma ^2}-1} \log \left(\frac{\delta }{t}\right) e^{\sigma ^2-\frac{\log
   ^2\left(\frac{\delta }{t}\right)}{2 \sigma ^2}}}{\sqrt{2 \pi } \sigma ^3 t},
  \end{split}\\
   \begin{split}
    \label{eq-sc-norm-2-temp-der-scale-time-var-gamma1}
    h_{\zeta\zeta,Koe}(t;\; \sigma, \delta)  
    & = \tau \,  h_{tt,Koe}(t;\; \sigma, \delta) 
      = \frac{\delta  \left(e^{\sigma ^2}-1\right) e^{-\frac{\log ^2\left(\frac{t}{\delta
   }\right)-5 \sigma ^4}{2 \sigma ^2}} \left(\log \left(\frac{t}{\delta }\right)
   \left(\log \left(\frac{t}{\delta }\right)+\sigma ^2\right)-\sigma ^2\right)}{\sqrt{2
   \pi } \sigma ^5 t^2},
  \end{split}
\end{align}
or when using $L_p$-normalization for $p = 1$:
\begin{align}
  \begin{split}
    \label{eq-sc-norm-1-temp-der-scale-time-Lp-norm-p1}
    h_{\zeta,Koe}(t;\; \sigma, \delta)  
    & = \frac{G_{1,1}}{\| h_{t,Koe}(\cdot;\; \sigma, \delta) \|_1} \,
            h_{t,Koe}(t;\; \sigma, \delta) 
      = -\frac{\log \left(\frac{t}{\delta }\right) e^{-\frac{\log ^2\left(\frac{t}{\delta
   }\right)}{2 \sigma ^2}}}{\sqrt{2 \pi } \sigma ^2 t},
  \end{split}\\
   \begin{split}
    \label{eq-sc-norm-2-temp-der-scale-time-Lp-norm-p1}
    h_{\zeta\zeta,Koe}(t;\; \sigma, \delta)  
    & = \frac{G_{2,1}}{\| h_{tt,Koe}(\cdot;\; \sigma, \delta) \|_1} \,
            h_{tt,Koe}(t;\; \sigma, \delta) 
      = \frac{\sqrt{\frac{2}{\pi }} \delta  e^{-\frac{2 \log ^2\left(\frac{t}{\delta
   }\right)+\sigma ^4}{4 \sigma ^2}} \left(\log \left(\frac{t}{\delta }\right) \left(\log
   \left(\frac{t}{\delta }\right)+\sigma ^2\right)-\sigma ^2\right)}{\sigma ^3 t^2
   \left(\sigma  \sinh \left(\frac{1}{4} \sigma  \sqrt{\sigma ^2+4}\right)+\sqrt{\sigma
   ^2+4} \cosh \left(\frac{1}{4} \sigma  \sqrt{\sigma ^2+4}\right)\right)}.
  \end{split}
\end{align}
\caption{Scale-normalized first- and second-order temporal derivatives of
  Koenderink's scale-time kernel using either variance-based
  normalization for a general value of $\gamma$, variance-based
  normalization for $\gamma = 1$ or $L_P$-normalization for $p = 1$.}
  \label{tab-scale-time-1-2-3-scale-norm-temp-der}
\end{table*}

\section{Scale normalization of temporal derivatives in Koenderink's scale-time model}
\label{sec-sc-norm-koe-scale-time}

In his scale-time model, Koenderink \cite{Koe88-BC} proposed to perform a
logarithmic mapping of the past via a time delay and then
applied Gaussian smoothing in the transformed temporal domain.
Following the slight modification of this model proposed in 
(Lindeberg \cite[Appendix~2]{Lin16-JMIV}) to
have the temporal kernels normalized to unit $L_1$-norm such
that a constant signal should remain unchanged under temporal
smoothing, these kernels can be written on the form
\begin{equation}
  \label{eq-temp-kern-norm-Koe-scale-time}
  h_{Koe}(t;\; \sigma, \delta) 
  =\frac{1}{\sqrt{2 \pi } \sigma \,\delta}
  e^{-\frac{\log ^2\left(\frac{t}{\delta }\right)}{2 \sigma ^2} -\frac{\sigma^2}{2}}.
 \end{equation}
where $\delta$ represents the temporal delay and $\sigma$ is a
dimensionless temporal scale parameter relative to the logarithmically
transformed temporal domain.
The temporal mean of this kernel is (Lindeberg \cite[Appendix~2, Equation~(152)]{Lin16-JMIV})
\begin{equation}
  \bar{t} = \int_{t=-\infty}^{\infty} t \, h_{Koe}(t;\; \sigma, \delta) \, dt
  = \delta \, e^{\frac{3 \sigma ^2}{2}} 
 \end{equation}
and the temporal variance (Lindeberg \cite[Appendix~2, Equation~(153)]{Lin16-JMIV})
\begin{align}
  \begin{split}
     \tau
     & = \int_{t=-\infty}^{\infty} (t - \bar{t})^2 \, h_{Koe}(t;\; \sigma, \delta) \, dt
       = \delta ^2 e^{3 \sigma ^2} \left(e^{\sigma ^2}-1\right).
  \end{split}
\end{align}
In relation to the proportionality requirement (\ref{eq-rel-temp-delay-temp-var})
between the temporal
delay and the temporal scale parameter in terms of dimension $[\mbox{time}]$ used for the
theoretical arguments in Appendix~\ref{app-undesired-temp-dyn-temp-semi-group}
\begin{equation}
  \bar{t} = C \sqrt{\tau}
\end{equation}
it follows that that this relation is satisfied if and only if
\begin{equation}
  C = \frac{1}{\sqrt{e^{\sigma ^2}-1}}
\end{equation}
is held constant between the temporal scale-time representations at
different temporal scales, in other words only if a one-parameter family of
scale-time representations is generated by keeping the dimensionless temporal
scale parameter $\sigma$ constant while varying only the temporal delay
parameter $\delta$. If proportionality between the temporal delay and
the temporal scale parameter of dimension $[\mbox{time}]$ is required,
then the dimensionless scale parameter $\sigma$ should
therefore be determined from the proportionality constant $C$ according to
\begin{equation}
  \sigma = \sqrt{\log \left( 1 + \frac{1}{C^2} \right)}.
\end{equation}

Differentiating the kernel (\ref{eq-temp-kern-norm-Koe-scale-time})
with respect to time gives the expressions for the first-, second- and
third-order temporal derivatives in equations
(\ref{eq-scale-time-norm-1-temp-der}),
(\ref{eq-scale-time-norm-2-temp-der}) and
(\ref{eq-scale-time-norm-3-temp-der}) in Table~\ref{tab-scale-time-norm-1-2-3-temp-der}.
The first-order temporal derivative has its zero-crossing at
\begin{equation}
  \label{eq-tmax-scale-time}
  t_{max} = \delta,
\end{equation}
the second-order temporal derivative has its zero-crossings at
\begin{align}
  \begin{split}
    t_{inflect1} & = \delta  \, e^{-\frac{1}{2} \sigma  \left(\sqrt{\sigma ^2+4}+\sigma \right)},
  \end{split}\\
 \begin{split}
    t_{inflect2} & = \delta  \, e^{-\frac{1}{2} \sigma  \left(\sigma -\sqrt{\sigma ^2+4}\right)}.
  \end{split}
\end{align}
and its peaks at the zero-crossings of the third-order derivative
\begin{align}
  \begin{split}
    t_{3,1} & = e^{-\sigma  \left(\sqrt{\sigma ^2+3}+\sigma \right)},
  \end{split}\\
  \begin{split}
    t_{3,2} & = \delta  e^{-\sigma ^2},
  \end{split}\\
  \begin{split}
   \label{eq-t-3-3-scale-time}
    t_{3,3} & = \delta  e^{\sigma  \left(\sqrt{\sigma ^2+3}-\sigma \right)}.
  \end{split}
\end{align}
The $L_1$-norms of the first- and second-order temporal scale-space
kernels are thereby given by
\begin{align}
  \begin{split}
    & \| h_{t,Koe}(\cdot;\; \sigma, \delta) \|_1
     = 2 h_{Koe}(t_{max};\; \sigma, \delta) 
       = \frac{\sqrt{\frac{2}{\pi }} e^{-\frac{\sigma ^2}{2}}}{\delta  \sigma },
  \end{split}\\
  \begin{split}
    & \| h_{tt,Koe}(\cdot;\; \sigma, \delta) \|_1
   \end{split}\nonumber\\
  \begin{split}
    & = 2 \left( h_{t,Koe}(t_{inflect1};\; \sigma, \delta) -  h_{t,Koe}(t_{inflect2};\; \sigma, \delta) \right)
  \end{split}\nonumber\\
  \begin{split}
    & =
       \frac{\sqrt{\frac{2}{\pi}} e^{-\frac{\sigma
             ^2}{4}-\frac{1}{2}}}{\delta ^2 \sigma ^2} 
       \left(
          \sigma \sinh \left(\frac{1}{4} \sigma  \sqrt{\sigma ^2+4}\right)
      \right)
 \end{split}\nonumber\\
  \begin{split}
     & \phantom{= \frac{\sqrt{\frac{2}{\pi}} e^{-\frac{\sigma^2}{4}-\frac{1}{2}}}{\delta ^2 \sigma ^2} } \quad
       \left.
           +\sqrt{\sigma ^2+4} \cosh \left(\frac{1}{4} \sigma  \sqrt{\sigma ^2+4}\right)
        \right).
  \end{split}
\end{align}
Based on these characteristics, we can define scale-normalized
temporal derivatives of Koenderink's scale-time kernel according to
Equations~(\ref{eq-sc-norm-1-temp-der-scale-time-var-gen-gamma})--(\ref{eq-sc-norm-2-temp-der-scale-time-Lp-norm-p1}) in
Table~\ref{tab-scale-time-1-2-3-scale-norm-temp-der}.

\section{Estimating the temporal duration of underlying temporal structures from scale-time approximations of temporal derivatives of the time-causal limit kernel}
\label{sec-scaletime-approx-limit-kernel}

In (Lindeberg \cite[Appendix~2]{Lin16-JMIV}) the following
transformation between the parameters in Koenderink's scale-time
kernels and the time-causal limit kernel (\ref{eq-FT-comp-kern-log-distr-limit}) is derived
\begin{equation}
  \label{eq-var-transf-par-tau-c-sigma-delta-explogdistr-scaletime-models}
  \left\{
    \begin{array}{l}
        \tau = \delta^2 \, e^{3 \sigma ^2} \left(e^{\sigma ^2}-1\right)  \\
      c = \frac{e^{\sigma ^2}}{2-e^{\sigma ^2}}
    \end{array}
  \right.
  \quad\quad
\left\{
    \begin{array}{l}
      \sigma = \sqrt{\log \left(\frac{2 c}{c+1}\right)}  \\
      \delta = \frac{(c+1)^2 \sqrt{\tau}}{2 \sqrt{2} \sqrt{(c-1) c^3}}
    \end{array}
  \right.
\end{equation}
under the conditions  $c > 1$ and $\sigma < \sqrt{\log 2} \approx 0.832$
by requiring the first- and second-order temporal moments of the
kernels in the two families to be equal.

Given this approximate mapping between the time-causal limit kernel
and the temporal kernels in Koenderink's scale-time model, we can
approximate the positions of the temporal peak, the peaks in the
first- and second-order temporal derivatives of the time-causal limit
kernel based on our previously derived expressions for the maximum
point $t_{max}$, the inflection points $t_{inflect1}$ and
$t_{inflect2}$ as well as the zero-crossings of the third-order derivative
$t_{3,1}$, $t_{3,2}$ and $t_{3,2}$ according to
(\ref{eq-tmax-scale-time})--(\ref{eq-t-3-3-scale-time})
in Appendix~\ref{sec-sc-norm-koe-scale-time}:
\begin{align}
  \begin{split}
     t_{max} & \approx \delta = \frac{(c+1)^2}{2 \sqrt{2} \sqrt{\frac{(c-1) c^3}{\tau }}},
  \end{split}\\
  \begin{split}
    t_{inflect1} & \approx \delta  \, e^{-\frac{1}{2} \sigma  \left(\sqrt{\sigma ^2+4}+\sigma \right)}
  \end{split}\nonumber\\
 \begin{split}
                   & = \frac{(c+1)^{5/2} \sqrt{\frac{\tau }{c-1}} e^{-\frac{1}{2} \sqrt{\log
   \left(\frac{2 c}{c+1}\right) \left(\log \left(\frac{2
   c}{c+1}\right)+4\right)}}}{4 c^2},
  \end{split}\\
 \begin{split}
    t_{inflect2} & \approx \delta  \, e^{-\frac{1}{2} \sigma  \left(\sigma -\sqrt{\sigma ^2+4}\right)}
  \end{split}\nonumber\\
 \begin{split}
                   & = \frac{(c+1)^{5/2} \sqrt{\frac{\tau }{c-1}} e^{\frac{1}{2} \sqrt{\log \left(\frac{2
   c}{c+1}\right) \left(\log \left(\frac{2 c}{c+1}\right)+4\right)}}}{4 c^2},
  \end{split}\\
  \begin{split}
    t_{3,1} & \approx e^{-\sigma  \left(\sqrt{\sigma ^2+3}+\sigma \right)}
  \end{split}\nonumber\\
 \begin{split}
                   & = \frac{(c+1)^3 \sqrt{\frac{\tau }{c-1}} e^{-\sqrt{\log \left(\frac{2 c}{c+1}\right)
   \left(\log \left(\frac{2 c}{c+1}\right)+3\right)}}}{4 \sqrt{2} c^{5/2}},
  \end{split}
\end{align}
\begin{align}
  \begin{split}
    t_{3,2} & \approx \delta  e^{-\sigma ^2}
  \end{split}\nonumber\\
 \begin{split}
                   & = \frac{(c+1)^3}{4 \sqrt{2} \sqrt{\frac{(c-1) c^5}{\tau }}},
  \end{split}\\
  \begin{split}
    t_{3,3} & \approx \delta  e^{\sigma  \left(\sqrt{\sigma ^2+3}-\sigma \right)}
  \end{split}\nonumber\\
 \begin{split}
                   & = \frac{(c+1)^3 e^{\sqrt{\log \left(\frac{2 c}{c+1}\right) \left(\log \left(\frac{2
   c}{c+1}\right)+3\right)}}}{4 \sqrt{2} \sqrt{\frac{(c-1) c^5}{\tau }}}
  \end{split}
\end{align}
Specifically, this leads to the following estimates of how the
temporal width of the first- and second-order temporal derivatives
depend on the distribution parameter $c$
\begin{align}
 \begin{split}
     d_1 & = t_{inflect2} - t_{inflect1} 
               \approx 2 \delta  e^{-\sigma^2/2} \sinh \left(\frac{\sigma}{2} \sqrt{\sigma ^2+4}\right)
  \end{split}\nonumber\\
 \begin{split}
             & = \frac{(c+1)^{5/2}}{2 c^2 \sqrt{c-1}} \times
  \end{split}\nonumber\\
 \begin{split}
    \label{eq-width-1st-temp-der-limit-kern-scale-time-approx}
    & \phantom{=} \quad \sinh \left(\frac{1}{2} \sqrt{\log
   \left(\frac{2 c}{c+1}\right) \left(\log \left(\frac{2
   c}{c+1}\right)+4\right)}\right) \sqrt{\tau}
  \end{split}\\
 \begin{split}
     d_2 & = t_{3,3} - t_{3,2} 
               \approx 2 \delta  e^{-\sigma ^2} \sinh \left(\sigma  \sqrt{\sigma ^2+3}\right)
  \end{split}\nonumber\\
 \begin{split}
            & = \frac{(c+1)^3}{2 \sqrt{2} \, c^2 \sqrt{c \, (c-1)}} \times
  \end{split}\nonumber\\
 \begin{split}
    \label{eq-width-2nd-temp-der-limit-kern-scale-time-approx}
     & \phantom{=} \quad \sinh \left(\sqrt{\log \left(\frac{2 c}{c+1}\right) \left(\log
   \left(\frac{2 c}{c+1}\right)+3\right)}\right) \sqrt{\tau}
  \end{split}
\end{align}
which can be compared to the corresponding width measures for the non-causal
Gaussian kernel
\begin{align}
 \begin{split}
    \label{eq-width-1st-temp-der-gauss-kern}
     d_1 & = t_{inflect2} - t_{inflect1} 
                = (\delta + \sqrt{\tau}) - (\delta - \sqrt{\tau}) = 2 \sqrt{\tau},
\end{split}\\
 \begin{split}
    \label{eq-width-2nd-temp-der-gauss-kern}
     d_2 & = t_{3,3} - t_{3,1} 
                = (\delta + \sqrt{3 \tau}) - (\delta - \sqrt{3 \tau})
                = 2 \sqrt{3} \sqrt{\tau}.
\end{split}
\end{align}

\begin{figure*}[hbtp]
  \begin{center}
    \begin{tabular}{cc}
      {\small $d_1(\tau, c)$ for $\tau=1$} 
     & {\small $d_2(\tau, c)$ for $\tau=1$} \\
      \includegraphics[width=0.45\textwidth]{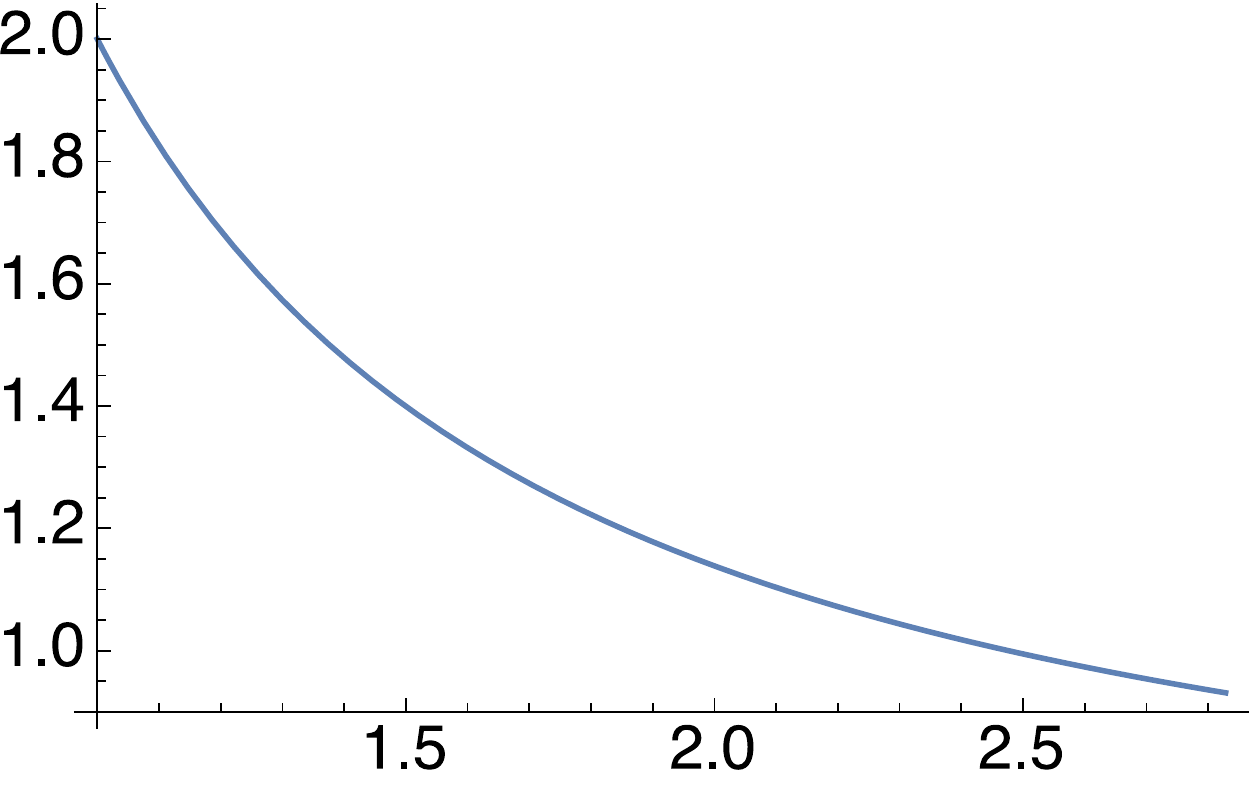}
      & \includegraphics[width=0.45\textwidth]{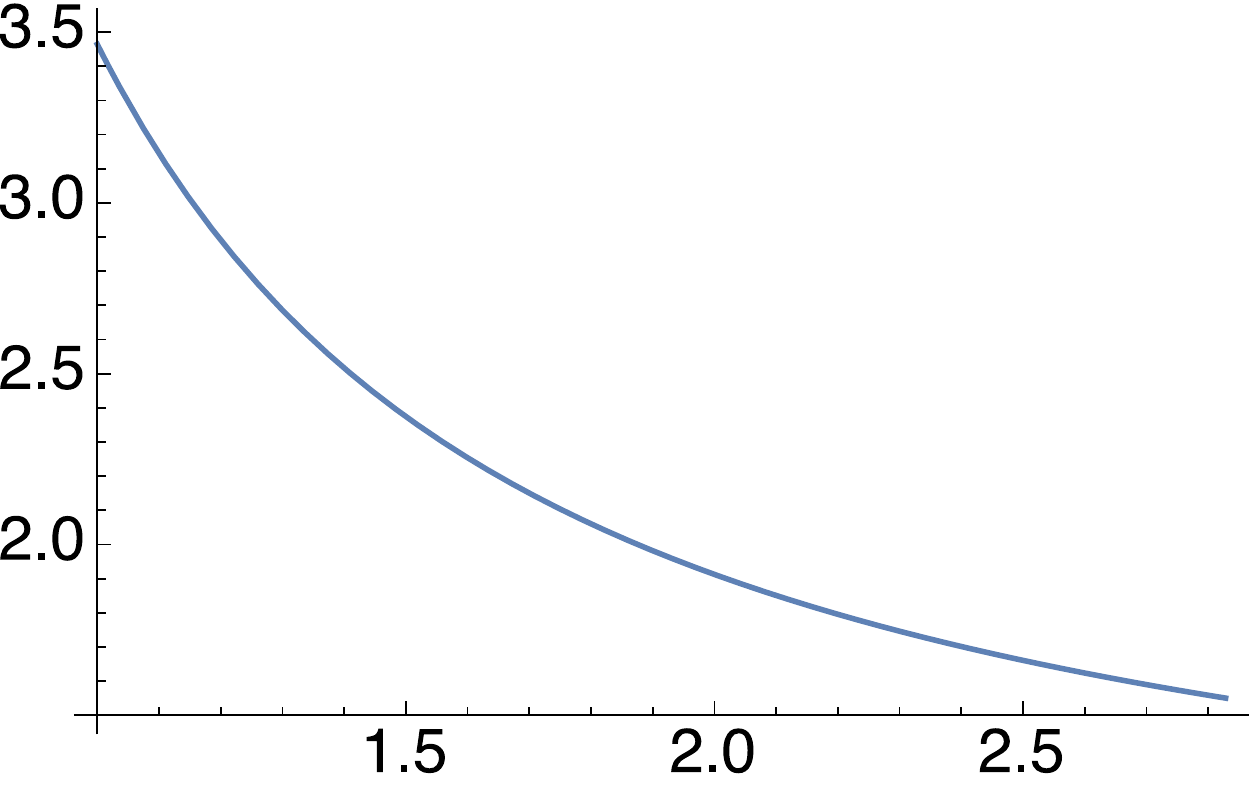}
      \\
    \end{tabular}
  \end{center}
   \caption{Scale-time approximations of the temporal durations $d_1(\tau, c)$
     and $d_2(\tau, c)$ of the first- and second-order temporal
     derivatives of the time-causal limit kernel according to
     (\protect\ref{eq-width-1st-temp-der-limit-kern-scale-time-approx})
     and
     (\protect\ref{eq-width-2nd-temp-der-limit-kern-scale-time-approx})
     as function of the distribution parameter $c$ for $\tau = 1$.}
  \label{fig-temp-widths-w1-w2-scale-time-approx}
\end{figure*}

Figure~\ref{fig-temp-widths-w1-w2-scale-time-approx} shows graphs of
the width measures
(\ref{eq-width-1st-temp-der-limit-kern-scale-time-approx})
and (\ref{eq-width-2nd-temp-der-limit-kern-scale-time-approx})
of the first- and
second-order temporal derivatives time-causal limit kernel obtained as
obtained from a scale-time approximation.
As can be seen from the graphs, the width measures vary by about 30~\%
when the distribution parameter is varied between $c = \sqrt{2}$ and
$c = 2$.
Thus, the value of the distribution parameter $c$ must be
taken into explicit account when transferring the temporal scale
parameter $\tau$ to a characteristic length estimate on the temporal axis.
Notably when the distribution parameter tends to $c \rightarrow 1$,
the temporal width estimates approach the corresponding width estimates
(\ref{eq-width-1st-temp-der-gauss-kern}) and
(\ref{eq-width-2nd-temp-der-gauss-kern}) of the Gaussian kernel.

Given that a temporal feature has been detected from a local maximum over
temporal scales in either the first- or second-order temporal
derivative of the time-causal limit kernel, if we use the behaviour of
the Gaussian temporal scale-space model
(\ref{eq-width-1st-temp-der-gauss-kern}) and
(\ref{eq-width-2nd-temp-der-gauss-kern}) for additional calibration of
the proportionality constant of the scale estimate,
we do then obtain the following
estimates $\hat{d}$ of the temporal duration of the corresponding
temporal feature as function of the temporal scale estimate
$\hat{\tau}$ and the distribution parameter $c$:
\begin{align}
  \begin{split}
     \hat{d}_1 
     & = \frac{d_1(\hat{\tau}, c)}{2}
  \end{split}\nonumber\\
  \begin{split}
     & = \frac{(c+1)^{5/2}}{4 c^2 \sqrt{c-1}} \times
  \end{split}\nonumber\\
  \begin{split}
   & \phantom{=} \quad \sinh \left(\frac{1}{2} \sqrt{\log
   \left(\frac{2 c}{c+1}\right) \left(\log \left(\frac{2
   c}{c+1}\right)+4\right)}\right) \sqrt{\hat{\tau}},
  \end{split}\\
  \begin{split}
     \hat{d}_2 
     & = \frac{d_2(\hat{\tau}, c)}{2\sqrt{3}}
  \end{split}\nonumber\\
 \begin{split}
     & = \frac{(c+1)^3}{4 \sqrt{6} \, c^2 \sqrt{c \, (c-1)}} \times
  \end{split}\nonumber\\
  \begin{split}
   & \phantom{=} \quad \sinh \left(\sqrt{\log \left(\frac{2 c}{c+1}\right) \left(\log
   \left(\frac{2 c}{c+1}\right)+3\right)}\right) \sqrt{\hat{\tau}}.
  \end{split}
\end{align}

\bibliographystyle{spmpsci}

{\footnotesize
\bibliography{defs,tlmac}

\begin{thebibliography}{100}
\providecommand{\url}[1]{{#1}}
\providecommand{\urlprefix}{URL }
\expandafter\ifx\csname urlstyle\endcsname\relax
  \providecommand{\doi}[1]{DOI~\discretionary{}{}{}#1}\else
  \providecommand{\doi}{DOI~\discretionary{}{}{}\begingroup
  \urlstyle{rm}\Url}\fi

\bibitem{AdeBer85-JOSA}
Adelson, E., Bergen, J.: Spatiotemporal energy models for the perception of
  motion.
\newblock Journal of Optical Society of America \textbf{A~2}, 284--299 (1985)

\bibitem{AerJoh81-BICY}
Aertsen, A.M.H.J., Johannesma, P.I.M.: The spectro-temporal receptive field: A
  functional characterization of auditory neurons.
\newblock Biological Cybernetics \textbf{42}(2), 133--143 (1981)

\bibitem{AgaSnaSimSeiSze09-ICCV}
Agarwal, S., Snavely, N., Simon, I., Seitz, S.M., Szeliski, R.: Building {R}ome
  in a day.
\newblock In: Proc.\ International Conference on Computer Vision (ICCV 2009),
  pp. 72--79 (2009)

\bibitem{AliSocJoaSev16-ApplSci}
Al{\'\i}as, F., Socor{\'o}, J.C., Sevillano, X.: A review of physical and
  perceptual feature extraction techniques for speech, music and environmental
  sounds.
\newblock Applied Sciences \textbf{6}(5), 143 (2016)

\bibitem{BayEssTuyGoo08-CVIU}
Bay, H., Ess, A., Tuytelaars, T., van Gool, L.: Speeded up robust features
  {(SURF)}.
\newblock Computer Vision and Image Understanding \textbf{110}(3), 346--359
  (2008)

\bibitem{BerReyRid14-SensMEMSElOptSyst}
van~der Berg, E.S., Reyneke, P.V., de~Ridder, C.: Rotational image correlation
  in the {G}auss-{L}aguerre domain.
\newblock In: Third SPIE Conference on Sensors, MEMS and Electro-Optic Systems:
  Proc.\ of SPIE, vol. 9257, pp. 92,570F--1--92,570F--17 (2014)

\bibitem{BicLagGroTis06-CVPRW}
Bicego, M., Lagorio, A., Grosso, E., Tistarelli, M.: On the use of {SIFT}
  features for face authentication.
\newblock In: Proc.\ Computer Vision and Pattern Recognition Workshop (CVPRW
  2006), p.~35 (2006)

\bibitem{BosZisMun07-ICCV}
Bosch, A., Zisserman, A., Munoz, X.: Image classification using random forests
  and ferns.
\newblock In: Proc.\ International Conference on Computer Vision (ICCV 2007),
  pp. 1--8. Rio de Janeiro, Brazil (2007)

\bibitem{BreLapLin02-FG}
Bretzner, L., Laptev, I., Lindeberg, T.: Hand-gesture recognition using
  multi-scale colour features, hierarchical features and particle filtering.
\newblock In: Proc. Face and Gesture, pp. 63--74. Washington D.C., USA (2002)

\bibitem{BL97-CVIU}
Bretzner, L., Lindeberg, T.: Feature tracking with automatic selection of
  spatial scales.
\newblock Computer Vision and Image Understanding \textbf{71}(3), 385--392
  (1998)

\bibitem{BroLow05-3DIM}
Brown, M., Lowe, D.G.: Unsupervised 3d object recognition and reconstruction in
  unordered datasets.
\newblock In: Proc.\ 3-D Digital Imaging and Modeling (3DIM 2005), pp. 56--63
  (2005)

\bibitem{BroLow07-IJCV}
Brown, M., Lowe, D.G.: Automatic panoramic image stitching using invariant
  features.
\newblock International Journal of Computer Vision \textbf{74}(1), 59--73
  (2007)

\bibitem{ChoVerHalCro00-ECCV}
Chomat, O., de~Verdiere, V., Hall, D., Crowley, J.: Local scale selection for
  {G}aussian based description techniques.
\newblock In: Proc.\ European Conf. on Computer Vision (ECCV 2000),
  \emph{Lecture Notes in Computer Science}, vol. 1842, pp. I:117--133.
  Springer-Verlag, Dublin, Ireland (2000)

\bibitem{DatJosLiWan08-CompSurv}
Datta, R., Joshi, D., Li, J., Wang, J.Z.: Image retrieval: {I}deas, influences,
  and trends of the new age.
\newblock ACM Computing Surveys \textbf{40}(2), 5 (2008)

\bibitem{deAngAnz04-VisNeuroSci}
DeAngelis, G.C., Anzai, A.: A modern view of the classical receptive field:
  Linear and non-linear spatio-temporal processing by {V1} neurons.
\newblock In: L.M. Chalupa, J.S. Werner (eds.) The Visual Neurosciences,
  vol.~1, pp. 704--719. MIT Press (2004)

\bibitem{DeAngOhzFre95-TINS}
DeAngelis, G.C., Ohzawa, I., Freeman, R.D.: Receptive field dynamics in the
  central visual pathways.
\newblock Trends in Neuroscience \textbf{18}(10), 451--457 (1995)

\bibitem{DerWil12-PAMI}
Derpanis, K.G., Wildes, R.P.: Spacetime texture representation and recognition
  based on a spatiotemporal orientation analysis.
\newblock IEEE Transactions on Pattern Analysis and Machine Intelligence
  \textbf{34}(6), 1193--1205 (2012)

\bibitem{EldZuc98-PAMI}
Elder, J., Zucker, S.: Local scale control for edge detection and blur
  estimation.
\newblock IEEE Trans. Pattern Analysis and Machine Intell. \textbf{20}(7),
  699--716 (1998)

\bibitem{EzzBouPog07-InterSpeech}
Ezzat, T., Bouvrie, J.V., Poggio, T.: Spectro-temporal analysis of speech using
  2-{D} {G}abor filters.
\newblock In: INTERSPEECH, pp. 506--509 (2007)

\bibitem{Fag05-IJCV}
Fagerstr{\"o}m, D.: Temporal scale-spaces.
\newblock International Journal of Computer Vision \textbf{2--3}, 97--106
  (2005)

\bibitem{FleLan95-PAMI}
Fleet, D.J., Langley, K.: Recursive filters for optical flow.
\newblock IEEE Trans. Pattern Analysis and Machine Intell. \textbf{17}(1),
  61--67 (1995)

\bibitem{Flo97-book}
Florack, L.M.J.: Image Structure.
\newblock Series in Mathematical Imaging and Vision. Springer (1997)

\bibitem{FraNieHooWalVie00-MED}
Frangi, A.F., J., N.W., Hoogeveen, R.M., van Walsum, T., Viergever, M.A.:
  Model-based quantitation of 3{D} magnetic resonance angiographic images.
\newblock IEEE Trans.\ on Medical Imaging \textbf{18}(10), 946--956 (2000)

\bibitem{GL94-IJCV}
G{\aa}rding, J., Lindeberg, T.: Direct computation of shape cues using
  scale-adapted spatial derivative operators.
\newblock International Journal of Computer Vision \textbf{17}(2), 163--191
  (1996)

\bibitem{Gui98-TIP}
Guichard, F.: A morphological, affine, and {G}alilean invariant scale-space for
  movies.
\newblock IEEE Trans.\ Image Processing \textbf{7}(3), 444--456 (1998)

\bibitem{Haa04-book}
ter Haar~Romeny, B.: Front-End Vision and Multi-Scale Image Analysis.
\newblock Springer (2003)

\bibitem{RomFloNie01-SCSP}
ter Haar~Romeny, B., Florack, L., Nielsen, M.: Scale-time kernels and models.
\newblock In: Proc.\ Int.\ Conf.\ Scale-Space and Morphology in Computer Vision
  (Scale-Space'01), Springer Lecture Notes in Computer Science. Springer,
  Vancouver, Canada (2001)

\bibitem{HalVerCro00-ECCV}
Hall, D., de~Verdiere, V., Crowley, J.: Object recognition using coloured
  receptive fields.
\newblock In: Proc.\ European Conf. on Computer Vision (ECCV 2000),
  \emph{Springer Lecture Notes in Computer Science}, vol. 1842, pp. I:164--177.
  Springer, Dublin, Ireland (2000)

\bibitem{HanXuZhu15-JMIV}
Han, Z., Xu, Z., Zhu, S.C.: Video primal sketch: {A} unified middle-level
  representation for video.
\newblock Journal of Mathematical Imaging and Vision \textbf{53}(2), 151--170
  (2015)

\bibitem{HarZis04-Book}
Hartley, R., Zisserman, A.: Multiple View Geometry in Computer Vision.
\newblock Cambridge University Press (2004).
\newblock Second Edition

\bibitem{HasMayZel12-CVPR}
Hassner, T., Mayzels, V., Zelnik-Manor, L.: On {SIFT}s and their scales.
\newblock In: Proc.\ Computer Vision and Pattern Recognition (CVPR 2012), pp.
  1522--1528. Providence, Rhode Island (2012)

\bibitem{HecDomJouGoe11-SpeechComm}
Heckmann, M., Domont, X., Joublin, F., Goerick, C.: A hierarchical framework
  for spectro-temporal feature extraction.
\newblock Speech Communication \textbf{53}(5), 736--752 (2011)

\bibitem{HubWie59-Phys}
Hubel, D.H., Wiesel, T.N.: Receptive fields of single neurones in the cat's
  striate cortex.
\newblock J Physiol \textbf{147}, 226--238 (1959)

\bibitem{HubWie05-book}
Hubel, D.H., Wiesel, T.N.: Brain and Visual Perception: {T}he Story of a
  25-Year Collaboration.
\newblock Oxford University Press (2005)

\bibitem{Iij62}
Iijima, T.: Observation theory of two-dimensional visual patterns.
\newblock Tech. rep., Papers of Technical Group on Automata and Automatic
  Control, IECE, Japan (1962)

\bibitem{MutLow08-IJCV}
J.~Mutch, J., Lowe, D.G.: Object class recognition and localization using
  sparse features with limited receptive fields.
\newblock International Journal of Computer Vision \textbf{80}(1), 45--57
  (2008)

\bibitem{JacPle08-CircSystVidTech}
Jacobs, N., Pless, R.: Time scales in video surveillance.
\newblock IEEE Transactions on Circuits and Systems for Video Technology
  \textbf{18}(8), 1106--1113 (2008)

\bibitem{JaiSeb07-CVIU}
Jaimes, A., Sebe, N.: Multimodal human--computer interaction: {A} survey.
\newblock Computer Vision and Image Understanding \textbf{108}(1), 116--134
  (2007)

\bibitem{JhuSerWolPog07-ICCV}
Jhuang, H., Serre, T., Wolf, L., Poggio, T.: A biologically inspired system for
  action recognition.
\newblock In: International Conference on Computer Vision (ICCV'07), pp. 1--8
  (2007)

\bibitem{KadBra01-IJCV}
Kadir, T., Brady, M.: Saliency, scale and image description.
\newblock International Journal of Computer Vision \textbf{45}(2), 83--105
  (2001)

\bibitem{KanMorNag05-ScSp}
Kang, Y., Morooka, K., Nagahashi, H.: Scale invariant texture analysis using
  multi-scale local autocorrelation features.
\newblock In: Proc.\ Scale Space and PDE Methods in Computer Vision
  (Scale-Space'05), \emph{Springer Lecture Notes in Computer Science}, vol.
  3459, pp. 363--373. Springer (2005)

\bibitem{Kar68}
Karlin, S.: Total Positivity.
\newblock Stanford Univ. Press (1968)

\bibitem{KlaMarSch08-BMVC}
Kl{\"a}ser, A., Marszalek, M., Schmid, C.: A spatio-temporal descriptor based
  on {3D}-gradients.
\newblock In: Proc. British Machine Vision Conf. Leeds, U.K. (2008)

\bibitem{Kle02-ActAcust}
Kleinschmidt, M.: Methods for capturing spectro-temporal modulations in
  automatic speech recognition.
\newblock Acta Acustica united with Acustica \textbf{88}(3), 416--422 (2002)

\bibitem{Koe84-BC}
Koenderink, J.J.: The structure of images.
\newblock Biological Cybernetics \textbf{50}, 363--370 (1984)

\bibitem{Koe88-BC}
Koenderink, J.J.: Scale-time.
\newblock Biological Cybernetics \textbf{58}, 159--162 (1988)

\bibitem{KoeDoo92-PAMI}
Koenderink, J.J., {van Doorn}, A.J.: Generic neighborhood operators.
\newblock IEEE Trans. Pattern Analysis and Machine Intell. \textbf{14}(6),
  597--605 (1992)

\bibitem{KriMalAyaValTro00-CVIU}
Krissian, K., Malandain, G., Ayache, N., Vaillant, R., Trousset, Y.:
  Model-based detection of tubular structures in 3{D} images.
\newblock Computer Vision and Image Understanding \textbf{80}(2), 130--171
  (2000)

\bibitem{LapCapSchLin07-CVIU}
Laptev, I., Caputo, B., Schuldt, C., Lindeberg, T.: Local velocity-adapted
  motion events for spatio-temporal recognition.
\newblock Computer Vision and Image Understanding \textbf{108}, 207--229 (2007)

\bibitem{LapLin03-ICCV}
Laptev, I., Lindeberg, T.: Space-time interest points.
\newblock In: Proc.\ Int.\ Conf.\ on Computer Vision (ICCV 2003), pp. 432--439.
  Nice, France (2003)

\bibitem{LapLin04-ECCVWS}
Laptev, I., Lindeberg, T.: Local descriptors for spatio-temporal recognition.
\newblock In: Proc.\ ECCV'04 Workshop on Spatial Coherence for Visual Motion
  Analysis, \emph{Springer Lecture Notes in Computer Science}, vol. 3667, pp.
  91--103. Prague, Czech Republic (2004)

\bibitem{LarDarDahPed12-ECCV}
Larsen, A.B.L., Darkner, S., Dahl, A.L., Pedersen, K.S.: Jet-based local image
  descriptors.
\newblock In: Proc.\ European Conference on Computer Vision (ECCV 2012),
  \emph{Springer Lecture Notes in Computer Science}, vol. 7574, pp.
  III:638--650. Springer (2012)

\bibitem{LazSchPon05-PAMI}
Lazebnik, S., Schmid, C., Ponce, J.: A sparse texture representation using
  local affine regions.
\newblock IEEE Trans. Pattern Analysis and Machine Intell. \textbf{27}(8),
  1265--1278 (2005)

\bibitem{LewSebDjeJai06-ACM-Multi}
Lew, M.S., Sebe, N., Djeraba, C., Jain, R.: Content-based multimedia
  information retrieval: {S}tate of the art and challenges.
\newblock ACM Trans.\ on Multimedia Computing, Communications, and Applications
  \textbf{2}(1), 1--19 (2006)

\bibitem{Li09-EncBiometr}
Li, S.Z. (ed.): Encyclopedia of Biometrics.
\newblock Springer Science \& Business Media (2009)

\bibitem{LiTaxLoo11-ScSp}
Li, Y., Tax, D.M.J., Loog, M.: Supervised scale-invariant segmentation (and
  detection).
\newblock In: Proc.\ Scale Space and Variational Methods in Computer Vision
  (SSVM 2011), \emph{Springer Lecture Notes in Computer Science}, vol. 6667,
  pp. 350--361. Springer, Ein Gedi, Israel (2012)

\bibitem{Lin90-PAMI}
Lindeberg, T.: Scale-space for discrete signals.
\newblock IEEE Trans. Pattern Analysis and Machine Intell. \textbf{12}(3),
  234--254 (1990)

\bibitem{Lin93-JMIV}
Lindeberg, T.: Discrete derivative approximations with scale-space properties:
  A basis for low-level feature extraction.
\newblock Journal of Mathematical Imaging and Vision \textbf{3}(4), 349--376
  (1993)

\bibitem{Lin92-PAMI}
Lindeberg, T.: Effective scale: {A} natural unit for measuring scale-space
  lifetime.
\newblock IEEE Trans. Pattern Analysis and Machine Intell. \textbf{15}(10),
  1068--1074 (1993)

\bibitem{Lin93-SCIA}
Lindeberg, T.: On scale selection for differential operators.
\newblock In: Proc.\ 8th Scandinavian Conf.\ on Image Analysis (SCIA'93), pp.
  857--866. Norwegian Society for Image Processing and Pattern Recognition,
  Troms{\o}, Norway (1993)

\bibitem{Lin93-Dis}
Lindeberg, T.: Scale-Space Theory in Computer Vision.
\newblock Springer (1993)

\bibitem{Lin94-SI}
Lindeberg, T.: Scale-space theory: {A} basic tool for analysing structures at
  different scales.
\newblock Journal of Applied Statistics \textbf{21}(2), 225--270 (1994).
\newblock Also available from
  http://www.csc.kth.se/$\sim$tony/abstracts/Lin94-SI-abstract.html

\bibitem{Lin97-ICSSTCV}
Lindeberg, T.: Linear spatio-temporal scale-space.
\newblock In: B.M. ter Haar~Romeny, L.M.J. Florack, J.J. Koenderink, M.A.
  Viergever (eds.) Proc.\ International Conference on Scale-Space Theory in
  Computer Vision (Scale-Space'97), \emph{Springer Lecture Notes in Computer
  Science}, vol. 1252, pp. 113--127. Springer, Utrecht, The Netherlands (1997)

\bibitem{Lin97-AFPAC}
Lindeberg, T.: On automatic selection of temporal scales in time-casual
  scale-space.
\newblock In: G.~Sommer, J.J. Koenderink (eds.) Proc.\ AFPAC'97: Algebraic
  Frames for the Perception-Action Cycle, \emph{Springer Lecture Notes in
  Computer Science}, vol. 1315, pp. 94--113. Kiel, Germany (1997)

\bibitem{Lin98-IJCV}
Lindeberg, T.: Edge detection and ridge detection with automatic scale
  selection.
\newblock International Journal of Computer Vision \textbf{30}(2), 117--154
  (1998)

\bibitem{Lin97-IJCV}
Lindeberg, T.: Feature detection with automatic scale selection.
\newblock International Journal of Computer Vision \textbf{30}(2), 77--116
  (1998)

\bibitem{Lin97-IVC}
Lindeberg, T.: A scale selection principle for estimating image deformations.
\newblock Image and Vision Computing \textbf{16}(14), 961--977 (1998)

\bibitem{Lin99-CVHB}
Lindeberg, T.: Principles for automatic scale selection.
\newblock In: Handbook on Computer Vision and Applications, pp. 239--274.
  Academic Press, Boston, USA (1999).
\newblock Also available from http://www.csc.kth.se/cvap/abstracts/cvap222.html

\bibitem{CVAP257}
Lindeberg, T.: Linear spatio-temporal scale-space.
\newblock Tech. Rep. ISRN KTH/NA/P-{}-01/22-{}-SE, Dept. of Numerical Analysis
  and Computer Science, KTH (2001).
\newblock Available from http://www.csc.kth.se/cvap/abstracts/cvap257.html

\bibitem{Lin10-JMIV}
Lindeberg, T.: Generalized {G}aussian scale-space axiomatics comprising linear
  scale-space, affine scale-space and spatio-temporal scale-space.
\newblock Journal of Mathematical Imaging and Vision \textbf{40}(1), 36--81
  (2011)

\bibitem{Lin12-Scholarpedia}
Lindeberg, T.: Scale invariant feature transform.
\newblock Scholarpedia \textbf{7}(5), 10,491 (2012)

\bibitem{Lin13-BICY}
Lindeberg, T.: A computational theory of visual receptive fields.
\newblock Biological Cybernetics \textbf{107}(6), 589--635 (2013)

\bibitem{Lin13-PONE}
Lindeberg, T.: Invariance of visual operations at the level of receptive
  fields.
\newblock {PLOS ONE} \textbf{8}(7), e66,990 (2013)

\bibitem{Lin12-JMIV}
Lindeberg, T.: Scale selection properties of generalized scale-space interest
  point detectors.
\newblock Journal of Mathematical Imaging and Vision \textbf{46}(2), 177--210
  (2013)

\bibitem{Lin14-EncCompVis}
Lindeberg, T.: Scale selection.
\newblock In: K.~Ikeuchi (ed.) Computer Vision: A Reference Guide, pp.
  701--713. Springer (2014)

\bibitem{Lin15-JMIV}
Lindeberg, T.: Image matching using generalized scale-space interest points.
\newblock Journal of Mathematical Imaging and Vision \textbf{52}(1), 3--36
  (2015)

\bibitem{Lin15-SSVM}
Lindeberg, T.: Separable time-causal and time-recursive spatio-temporal
  receptive fields.
\newblock In: Proc.\ Scale-Space and Variational Methods for Computer Vision
  (SSVM 2015), \emph{Lecture Notes in Computer Science}, vol. 9087, pp.
  90--102. Springer (2015)

\bibitem{Lin16-JMIV}
Lindeberg, T.: Time-causal and time-recursive spatio-temporal receptive fields.
\newblock Journal of Mathematical Imaging and Vision \textbf{55}(1), 50--88
  (2016)

\bibitem{Lin16-spattempscsel}
Lindeberg, T.: Spatio-temporal scale selection in video data.
\newblock In preparation  (2017)

\bibitem{LinBre03-ScSp}
Lindeberg, T., Bretzner, L.: Real-time scale selection in hybrid multi-scale
  representations.
\newblock In: L.~Griffin, M.~Lillholm (eds.) Proc.\ Scale-Space Methods in
  Computer Vision (Scale-Space'03), \emph{Springer Lecture Notes in Computer
  Science}, vol. 2695, pp. 148--163. Springer, Isle of Skye, Scotland (2003)

\bibitem{LF96-ECCV}
Lindeberg, T., Fagerstr{\"o}m, D.: Scale-space with causal time direction.
\newblock In: Proc.\ European Conf. on Computer Vision (ECCV'96),
  \emph{Springer Lecture Notes in Computer Science}, vol. 1064, pp. 229--240.
  Cambridge, UK (1996)

\bibitem{LinFri15-PONE}
Lindeberg, T., Friberg, A.: Idealized computational models of auditory
  receptive fields.
\newblock PLOS ONE \textbf{10}(3), e0119,032:1--58 (2015)

\bibitem{LinFri15-SSVM}
Lindeberg, T., Friberg, A.: Scale-space theory for auditory signals.
\newblock In: Proc.\ Scale-Space and Variational Methods for Computer Vision
  (SSVM 2015), \emph{Lecture Notes in Computer Science}, vol. 9087, pp. 3--15.
  Springer (2015)

\bibitem{LG93-ICCV}
Lindeberg, T., G{\aa}rding, J.: Shape from texture from a multi-scale
  perspective.
\newblock In: T.S.H. H.-H.~Nagel, Y.~Shirai (eds.) Proc.\ Int.\ Conf. on
  Computer Vision (ICCV'93), pp. 683--691. IEEE Computer Society Press, Berlin,
  Germany (1993)

\bibitem{LG96-IVC}
Lindeberg, T., G{\aa}rding, J.: Shape-adapted smoothing in estimation of 3-{D}
  depth cues from affine distortions of local 2-{D} structure.
\newblock Image and Vision Computing \textbf{15}, 415--434 (1997)

\bibitem{LiuYueTor11-PAMI}
Liu, C., Yuen, J., Torralba, A.: {SIFT} flow: {D}ense correspondence across
  scenes and its applications.
\newblock IEEE Transactions on Pattern Analysis and Machine Intelligence
  \textbf{33}(5), 978--994 (2011)

\bibitem{LiuWanYaoZha12-CVPR}
Liu, X.M., Wang, C., Yao, H., Zhang, L.: The scale of edges.
\newblock In: Proc.\ Computer Vision and Pattern Recognition (CVPR 2012), pp.
  462--469 (2012)

\bibitem{LooLiTax09-LNCS}
Loog, M., Li, Y., Tax, D.: Maximum membership scale selection.
\newblock In: Multiple Classifier Systems, \emph{Springer Lecture Notes in
  Computer Science}, vol. 5519, pp. 468--477. Springer (2009)

\bibitem{Low04-IJCV}
Lowe, D.G.: Distinctive image features from scale-invariant keypoints.
\newblock International Journal of Computer Vision \textbf{60}(2), 91--110
  (2004)

\bibitem{Mah16-JMIV}
Mahmoodi, S.: Linear neural circuitry model for visual receptive fields.
\newblock Journal of Mathematical Imaging and Vision \textbf{54}(2), 1--24
  (2016)

\bibitem{MeyKol08-InterSpeech}
Meyer, B.T., Kollmeier, B.: Optimization and evaluation of {G}abor feature sets
  for {ASR}.
\newblock In: INTERSPEECH, pp. 906--909 (2008)

\bibitem{MikSch04-IJCV}
Mikolajczyk, K., Schmid, C.: Scale and affine invariant interest point
  detectors.
\newblock International Journal of Computer Vision \textbf{60}(1), 63--86
  (2004)

\bibitem{MikTuySchZisMatSchKadGoo05-IJCV}
Mikolajczyk, K., Tuytelaars, T., Schmid, C., Zisserman, A., Matas, J.,
  Schaffalitzky, F., Kadir, T., van Gool, L.: A comparison of affine region
  detectors.
\newblock International Journal of Computer Vision \textbf{65}(1--2), 43--72
  (2005)

\bibitem{MilEscReaSch01-JNeuroPhys}
Miller, L.M., Escabi, N.A., Read, H.L., Schreiner, C.: Spectrotemporal
  receptive fields in the lemniscal auditory thalamus and cortex.
\newblock Journal of Neurophysiology \textbf{87}(1), 516--527 (2001)

\bibitem{MraNav03-IJCV}
Mr{\'a}zek, P., Navara, M.: Selection of optimal stopping time for nonlinear
  diffusion filtering.
\newblock International Journal of Computer Vision \textbf{52}(2--3), 189--203
  (2003)

\bibitem{NegBraCroLau08-ExpRob}
Negre, A., Braillon, C., Crowley, J.L., Laugier, C.: Real-time
  time-to-collision from variation of intrinsic scale.
\newblock Experimental Robotics \textbf{39}, 75--84 (2008)

\bibitem{NieWanFei08-IJCV}
Niebles, J.C., Wang, H., Fei-Fei, L.: Unsupervised learning of human action
  categories using spatial-temporal words.
\newblock International Journal of Computer Vision \textbf{79}(3), 299--318
  (2008)

\bibitem{Par08-ECCV}
Paris, S.: Edge-preserving smoothing and mean-shift segmentation of video
  streams.
\newblock In: Proc.\ European Conf. on Computer Vision (ECCV 2008), Springer
  Lecture Notes in Computer Science, pp. 460--473. Springer, Marseille, France
  (2008)

\bibitem{PatAllGig95-JASA}
Patterson, R.D., Allerhand, M.H., Giguere, C.: Time-domain modeling of
  peripheral auditory processing: {A} modular architecture and a software
  platform.
\newblock The Journal of the Acoustical Society of America \textbf{98}(4),
  1890--1894 (1995)

\bibitem{PatRobHolMcKeoZhaAll92-AudPhysPerc}
Patterson, R.D., Robinson, K., Holdsworth, J., McKeown, D., Zhang, C.,
  Allerhand, M.: Complex sounds and auditory images.
\newblock Auditory Physiology and Perception \textbf{83}, 429--446 (1992)

\bibitem{Pop09-IVC}
Poppe, R.: A survey on vision-based human action recognition.
\newblock Image and Vision Computing \textbf{28}(6), 976--990 (2010)

\bibitem{Por02-HumCompStud}
Porta, M.: Vision-based user interfaces: Methods and applications.
\newblock International Journal of Human-Computer Studies \textbf{57}, 27--73
  (2002)

\bibitem{RivBre04-ImAnalRec}
Rivero-Moreno, C.J., Bres, S.: Spatio-temporal primitive extraction using
  {H}ermite and {L}aguerre filters for early vision video indexing.
\newblock In: Image Analysis and Recognition, \emph{Springer Lecture Notes in
  Computer Science}, vol. 3211, pp. 825--832 (2004)

\bibitem{RotLazSchPon06-IJCV}
Rothganger, F., Lazebnik, S., Schmid, C., Ponce, J.: {3D} object modeling and
  recognition using local affine-invariant image descriptors and multi-view
  spatial constraints.
\newblock International Journal of Computer Vision \textbf{66}(3), 231--259
  (2006)

\bibitem{SanGevSno10-PAMI}
van~de Sande, K.E.A., Gevers, T., Snoek, C.G.M.: Evaluating color descriptors
  for object and scene recognition.
\newblock IEEE Trans. Pattern Analysis and Machine Intell. \textbf{32}(9),
  1582--1596 (2010)

\bibitem{SatNakShiAtsYouKolGerKik98-MIA}
Sato, Y., Nakajima, S., Shiraga, N., Atsumi, H., Yoshida, S., Koller, T.,
  Gerig, G., Kikinis, R.: {3D} multi-scale line filter for segmentation and
  visualization of curvilinear structures in medical images.
\newblock Medical Image Analysis \textbf{2}(2), 143--168 (1998)

\bibitem{SchBezWagNey07-ICASSP}
Schlute, R., Bezrukov, L., Wagner, H., Ney, H.: Gammatone features and feature
  combination for large vocabulary speech recognition.
\newblock In: IEEE Int.\ Conf.\ on Acoustics, Speech and Signal Processing
  (ICASSP'07), vol.~IV, pp. 649--652 (2007)

\bibitem{Sch50}
Schoenberg, I.J.: On {P}\`olya frequency functions. ii. {V}ariation-diminishing
  integral operators of the convolution type.
\newblock Acta Sci. Math. (Szeged) \textbf{12}, 97--106 (1950)

\bibitem{Sch88-book}
Schoenberg, I.J.: I. J. Schoenberg Selected Papers, vol.~2.
\newblock Springer (1988).
\newblock Edited by C. de Boor

\bibitem{SeLowLit05-TROB}
Se, S., Lowe, D.G., Little, J.J.: Vision-based global localization and mapping
  for mobile robots.
\newblock IEEE Transactions on Robotics \textbf{21}(3), 364--375 (2005)

\bibitem{ShaClaZel12-BMVC}
Shabani, A.H., Clausi, D.A., Zelek, J.S.: Improved spatio-temporal salient
  feature detection for action recognition.
\newblock In: British Machine Vision Conference (BMVC'11), pp. 1--12. Dundee,
  U.K. (2011)

\bibitem{ShaMatt10-CIVR}
Shao, L., Mattivi, R.: Feature detector and descriptor evaluation in human
  action recognition.
\newblock In: Proc.\ ACM Int.\ Conf.\ on Image and Video Retrieval (CIVR'10),
  pp. 477--484. Xian, China (2010)

\bibitem{SicKha08-HandBookRob}
Siciliano, B., Khatib, O. (eds.): Springer Handbook of Robotics.
\newblock Springer Science \& Business Media (2008)

\bibitem{SpoCoilTra00-ICIP}
Sporring, J., Colios, C.J., Trahanias, P.E.: Generalized scale selection.
\newblock In: Proc. Int.\ Conf.\ on Image Processing (ICIP'00), pp. 920--923.
  Vancouver, Canada (2000)

\bibitem{SurVorPelJosSeePal15-JMIV}
Surya, P.V.B., Vorotnikov, D., Pelapur, R., Jose, S., Seetharaman, G.,
  Palaniappan, K.: Multiscale {T}ikhonov-total variation image restoration
  using spatially varying edge coherence exponent.
\newblock IEEE Transactions on Image Processing \textbf{24}(12), 5220--5235
  (2015)

\bibitem{TuyGoo04-IJCV}
Tuytelaars, T., van Gool, L.: Matching widely separated views based on affine
  invariant regions.
\newblock International Journal of Computer Vision \textbf{59}(1), 61--85
  (2004)

\bibitem{TuyMik08-Book}
Tuytelaars, T., Mikolajczyk, K.: A Survey on Local Invariant Features,
  \emph{Foundations and Trends in Computer Graphics and Vision}, vol. 3(3).
\newblock Now Publishers (2008)

\bibitem{WanUllKlaLapSch09-BMVC}
Wang, H., Ullah, M.M., Kl{\"a}ser, A., Laptev, I., Schmid, C.: Evaluation of
  local spatio-temporal features for action recognition.
\newblock In: Proc.\ British Machine Vision Conference (BMVC 2009). London,
  U.K. (2009)

\bibitem{WanQiaTan15-CVPR}
Wang, L., Qiao, Y., Tang, X.: Action recognition with trajectory-pooled
  deep-convolutional descriptors.
\newblock In: IEEE Conference on Computer Vision and Pattern Recognition (CVPR
  2015), pp. 4305--4314 (2015)

\bibitem{WeiRonBoy11-CVIU}
Weinland, D., Ronfard, R., Boyer, E.: A survey of vision-based methods for
  action representation, segmentation and recognition.
\newblock Computer Vision and Image Understanding \textbf{115}(2), 224--241
  (2011)

\bibitem{WilTuyGoo08-ECCV}
Willems, G., Tuytelaars, T., van Gool, L.: An efficient dense and
  scale-invariant spatio-temporal interest point detector.
\newblock In: Proc.\ European Conf. on Computer Vision (ECCV 2008),
  \emph{Springer Lecture Notes in Computer Science}, vol. 5303, pp. 650--663.
  Marseille, France (2008)

\bibitem{Wit83}
Witkin, A.P.: Scale-space filtering.
\newblock In: Proc. 8th Int. Joint Conf. Art. Intell., pp. 1019--1022.
  Karlsruhe, Germany (1983)

\bibitem{WuZhaShi11-ASLP}
Wu, Q., Zhang, L., Shi, G.: Robust multifactor speech feature extraction based
  on {G}abor analysis.
\newblock IEEE Trans.\ on Audio, Speech, and Language Processing
  \textbf{19}(4), 927--936 (2011)

\bibitem{ZelIra01-CVPR}
Zelnik-Manor, L., Irani, M.: Event-based analysis of video.
\newblock In: Proc.\ Computer Vision and Pattern Recognition (CVPR'01), pp.
  II:123--130 (2001)

\end{thebibliography}
}

\medskip

\begin{figure}[!h]
\begin{center}
\includegraphics[width=0.30\textwidth]{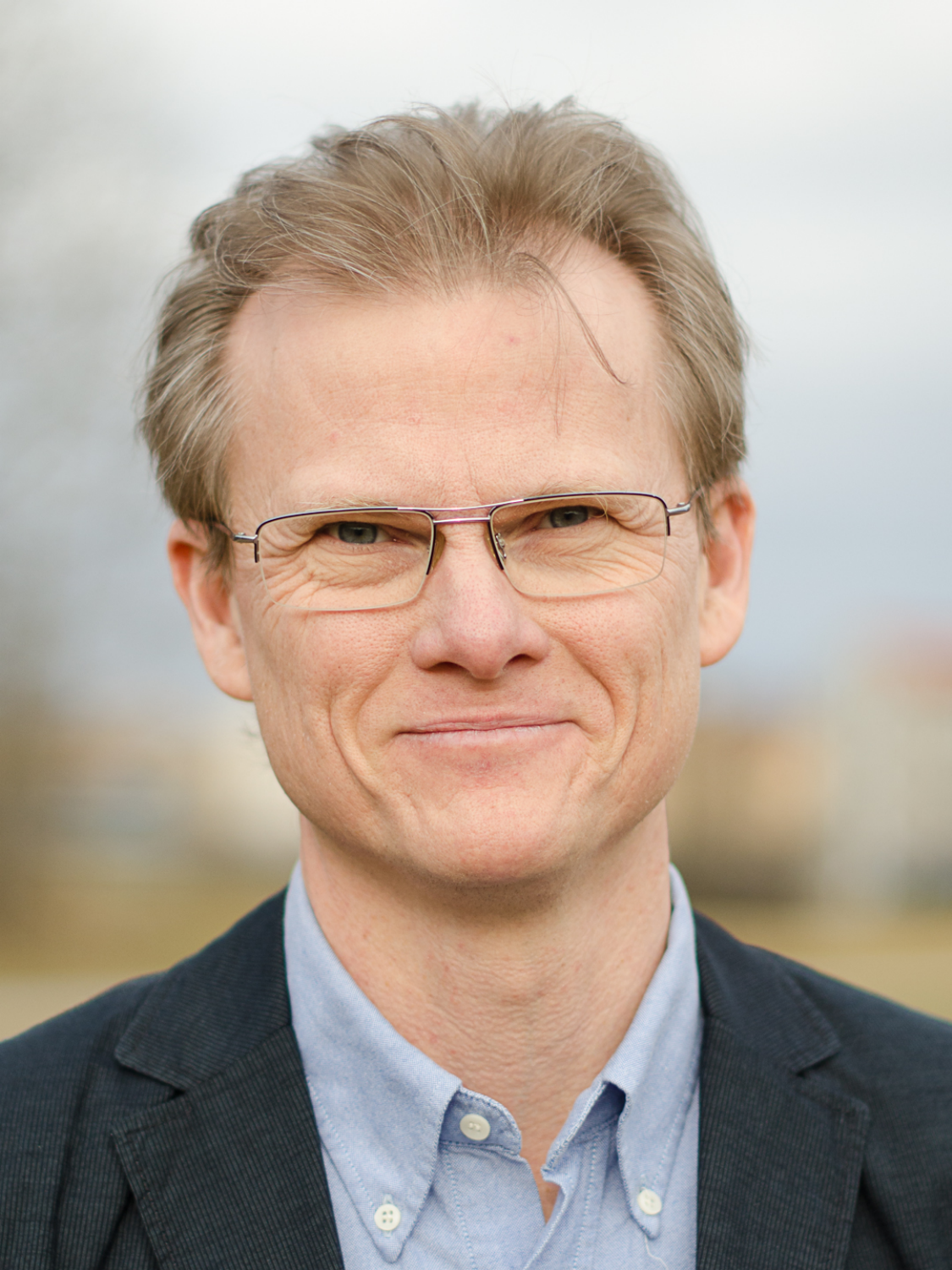}
\end{center}
\end{figure}

\noindent
{\bf Tony Lindeberg} is a Professor of Computer Science at KTH Royal Institute of Technology in Stockholm, Sweden. He was born in Stockholm in 1964, received his MSc degree in 1987, his PhD degree in 1991, became docent in 1996, and was appointed professor in 2000. He was a Research Fellow at the Royal Swedish Academy of Sciences between 2000 and 2010.
 
His research interests in computer vision relate to scale-space representation, image features, object recognition, spatio-temporal recognition, focus-of-attention and computational modelling of biological vision. He has developed theories and methodologies for continuous and discrete scale-space representation, visual and auditory receptive fields, detection of salient image structures, automatic scale selection, scale-invariant image features, affine invariant features, affine and Galilean normalization, temporal, spatio-temporal and spectro-temporal scale-space concepts as well as spatial and spatio-temporal image descriptors for image-based recognition. He has also worked on topics in medical image analysis and gesture recognition. He is author of the book Scale-Space Theory in Computer Vision. 

\end{document}